\pdfoutput=1

\documentclass[11pt]{article}

\usepackage[preprint]{acl}

\usepackage{times}
\usepackage{latexsym}
\usepackage{amsmath}
\usepackage{amsfonts}
\usepackage{graphicx}
\usepackage{makecell}
\usepackage{multirow}
\usepackage{booktabs}
\usepackage{nicematrix}
\usepackage{hhline}
\usepackage[dvipsnames, table]{xcolor}
\usepackage{CJKutf8}

\usepackage[T1]{fontenc}
\usepackage[utf8]{inputenc}

\usepackage{microtype}

\usepackage{inconsolata}

\usepackage{graphicx}
\DeclareMathOperator*{\argmax}{arg\,max}

\title{Overfitting Mitigation via Singular Value Decomposition \\ in Minimum Bayes Risk Decoding}

\author{
 \textbf{Riza Setiawan Soetedjo}\textsuperscript{1},
 \textbf{Yusuke Sakai}\textsuperscript{1},
 \textbf{Hidetaka Kamigaito}\textsuperscript{1},
\\
 \textbf{Katsuhiko Hayashi}\textsuperscript{1,2},
 \textbf{Taro Watanabe}\textsuperscript{1}
\\
 \textsuperscript{1}Nara Institute of Science and Technology (NAIST)
 \textsuperscript{2}The University of Tokyo
\\
 \texttt{riza.setiawan\_soetedjo.rs6@naist.ac.jp}
\\
 \texttt{\{sakai.yusuke.sr9, kamigaito.h, taro.watanabe\}@is.naist.jp}
\\
 \texttt{katsuhiko-hayashi@g.ecc.u-tokyo.ac.jp}
}

\begin{document}
\maketitle
\begin{abstract}
Minimum Bayes Risk (MBR) decoding enables high-quality text generation by selecting the hypothesis that maximizes a utility metric over sampled pseudo-references. However, it is highly susceptible to metric overfitting: it can irregularly inflate the chosen utility metric at the direct expense of other unoptimized evaluation metrics. To mitigate this, we introduce SVD-MBR, which frames the pairwise utility matrix as a noisy information signal. By computing a low-rank approximation via Singular Value Decomposition (SVD) and retaining only the top-$k$ components, we effectively decouple true consensus from metric noise. Experiments demonstrate that SVD-MBR successfully regularizes decoding, yielding substantial gains across a range of generalized metrics. Furthermore, we reveal that this denoising is metric-dependent: neural metrics encode a robust low-rank consensus ideal for SVD, whereas surface-level metrics struggle to separate signal from metric noise.
Our code is available at \href{https://github.com/naist-nlp/mbr-svd}{https://github.com/naist-nlp/mbr-svd}.

\end{abstract}

\section{Introduction}

Text generation remains a fundamental yet challenging task. Typically, models rely on maximum a posteriori (MAP) decoding, which generates a high-probability sequence conditioned on the preceding context. However, recent studies have demonstrated superior performance using sampling-based generation methods~\cite{eikema-aziz-2020-map, suzgun-etal-2023-follow}, including Minimum Bayes Risk (MBR) decoding~\cite{kumar-byrne-2004-minimum}.

\begin{table}[ht!]
\centering
\small
\begin{tabular}{@{}p{0.2\linewidth} p{0.73\linewidth}@{}}
\toprule
\textbf{Source} & The committee decided to postpone the meeting until next week. \\
\textbf{Reference} & Der Ausschuss beschloss, die Sitzung auf nächste Woche zu verschieben. \\
\midrule
\textbf{MAP} & Der Ausschuss beschloss, die Sitzung auf nächste Woche zu verschieben. \\
\addlinespace
\textbf{MBR} & Das Komitee hat entschieden das Treffen bis zur nächsten Woche zu postponieren. \\
\midrule
\textbf{Utility Function} & chrF~\cite{popovic-2015-chrf} \\
\midrule
\textbf{Evaluation} & 
    \vspace{-6mm}
    \begin{tabular}[t]{@{}l c c@{}}
    \\
    \textbf{Metric} & \textbf{MAP} & \textbf{MBR} \\
    \addlinespace
    \textbf{chrF} & $52.8$ & \textbf{54.2} (\textcolor{green}{$\uparrow$}) \\
    COMET & \textbf{77.3} & $76.3$ (\textcolor{red}{$\downarrow$}) \\
    BLEURT & \textbf{56.5} & $56.4$ (\textcolor{red}{$\downarrow$}) \\
    \end{tabular} \\
\bottomrule
\end{tabular}
\caption{Illustration of metric overfitting in MBR decoding. While MBR decoding improves performance on the utility function metric, i.e., chrF~\cite{popovic-2015-chrf}, it degrades performance on other evaluation metrics.}
\label{tab:method_illustration}
\end{table}

MBR decoding chooses the optimal hypothesis out of the samples based on two assumptions: (1) the model's distribution adequately approximates the true distribution; (2) the selected text exhibits the highest consensus with the other samples.
This process requires a candidate pool, a set of pseudo-references to act as guides, and a utility function to evaluate them. Ideally, pseudo-references reliably reflect the true distribution, and the utility function is resistant to noise. In practice, however, pseudo-references are noisy, and metrics struggle to discern subtle variations. 
We formally define this ``noise'' as the metric bias that causes MBR decoding to overfit when it excessively maximizes the utility function.
Consequently, the calculation of expected scores is highly susceptible to this overfitting, where outlier scores in the pairwise matrix cause the selection to inherit specific biases of the utility metric~\cite{freitag-etal-2023-epsilon}.

Previous studies have empirically reported this overfitting problem. \citet{freitag-etal-2022-high} evaluated multiple metrics as utility functions given the observation that while scores for the specific metric used as the utility function increased compared to beam search, unoptimized metrics stagnated or degraded. We illustrated this issue in Table~\ref{tab:method_illustration}. \citet{deguchi-etal-2024-mbrs} observed a similar tendency, showing that across multiple MBR variants, the selected text consistently improves the most with the target utility metric at the expense of other metrics. 

In this study, we address this gap by a novel, intrinsic approach to mitigate metric overfitting. We hypothesize that metric overfitting is driven by noise within the internal $N \times M$ pairwise utility matrix. To counter this, we introduce SVD-MBR, which utilizes Singular Value Decomposition (SVD) to denoise the matrix prior to hypothesis selection. We demonstrate that metric overfitting is an inherent limitation rather than a byproduct of specific datasets, MBR variants, or utility metrics. Our empirical evaluation for SVD-MBR under MT and Summarization tasks shows that applying a low-rank approximation successfully filters out metric hallucinations, reinforcing our proposed method to denoise the pairwise matrix using SVD as an effective solution to the overfitting problem.

\section{Background and Related Work}
\label{sec:background_related}

\paragraph{MBR decoding.}

The goal of MBR decoding is to select the hypothesis that maximizes the expected utility over pseudo-references.
Given a candidate set of hypotheses $\mathcal{H} = \{\boldsymbol{h}_1, \dots, \boldsymbol{h}_{|\mathcal{H}|}\}$ and a set of pseudo-references $\mathcal{Y} = \{\boldsymbol{y}_1, \dots, \boldsymbol{y}_{|\mathcal{Y}|}\}$ sampled from a generation model $\theta$ conditioned on a source input $x$, we rely on a Monte Carlo (MC) estimate, as the exact expected risk over the full model distribution is intractable.
We define an evaluation metric serves as the utility function $u(\boldsymbol{h}, \boldsymbol{y})$ to evaluate each candidate against each pseudo-reference, yielding a pairwise utility matrix $\mathbf{M} \in \mathbb{R}^{|\mathcal{H}| \times |\mathcal{Y}|}$ where $m_{i,j} = u(\boldsymbol{h}_i, \boldsymbol{y}_j)$, expressed as:
\begin{equation}
\mathbf{M} = \begin{bmatrix} 
m_{1,1} & m_{1,2} & \cdots & m_{1,|\mathcal{Y}|} \\ 
m_{2,1} & m_{2,2} & \cdots & m_{2,|\mathcal{Y}|} \\ 
\vdots & \vdots & \ddots & \vdots \\ 
m_{|\mathcal{H}|,1} & m_{|\mathcal{H}|,2} & \cdots & m_{|\mathcal{H}|,|\mathcal{Y}|} 
\end{bmatrix}.
\end{equation}
The decision rule selects the optimal hypothesis $\boldsymbol{h}^*$ that maximizes this expected utility:
\begin{equation}
\label{eq:mbr_expected_score}
\boldsymbol{h}^* = \argmax_{\boldsymbol{h}_i \in \mathcal{H}} \frac{1}{|\mathcal{Y}|}\sum_{j=1}^{|\mathcal{Y}|} m_{i,j}.
\end{equation}
Ideally, the pseudo-reference set $\mathcal{Y}$ would perfectly capture the true distribution, allowing $\mathbf{M}$ to yield an unbiased estimate of the expected utility. In practice, a finite sample cannot fully represent this distribution, inherently introducing noise into the calculation. Previous studies~\cite{freitag-etal-2023-epsilon, kamigaito-etal-2025-diversity} have demonstrated that this MC estimation is highly sensitive to the chosen sampling method and utility function. Crucially, these components dictate the topology of the pairwise matrix $\mathbf{M}$, and consequently, its susceptibility to the metric overfitting problem.

\paragraph{Overfitting in MBR decoding.}

MBR has resurfaced as a strong alternative to MAP decoding, largely because MAP's reliance on maximizing sequence probability often fails to reflect the true distribution learned by the model~\cite{eikema-aziz-2020-map}. 
However, recent literature highlights that MBR decoding tends to overfitting toward the utility metric. \citet{muller-sennrich-2021-understanding} demonstrated that MBR inherits metric biases, and \citet{freitag-etal-2023-epsilon} confirmed this problem persists when examining various sampling methods.
While \citet{kovacs-etal-2024-mitigating} have attempted to mitigate this by ensembling Quality Estimation (QE) with MBR, such approaches act as external workarounds to metric bias rather than addressing the internal noise of the pairwise utility matrix itself.

\paragraph{MBR matrix is low-rank.}

Standard MBR decoding is computationally intensive, requiring utility calculations between every hypothesis and pseudo-reference ($\mathcal{O}(N\times M)$). This bottleneck has prompted multiple studies to focus on decoding efficiency. For instance, \citet{eikema-aziz-2022-sampling} reduced the computational cost to $\mathcal{O}(N\times k)$ by filtering the number of pseudo-references. More fundamentally, \citet{trabelsi-2024-efficient-mbr} observed that the pairwise matrix exhibits an inherently low-rank structure, justifying a method to subsample candidates and approximate the matrix in fewer iterations. However, prior literature has exclusively exploited this low-rank property for computational speedups. Although matrix factorization techniques such as SVD and Non-Negative Matrix Factorization (NMF) are foundational for denoising and filtering in fields like recommender systems~\cite{wang-etal-2025-svd, corso-2016-nmf}, their application to denoising the MBR matrix has the potential to mitigate metric overfitting.

\section{Preliminary Analysis: Overfitting in MBR Decoding}
\label{sec:mbr_overfit}

While prior work has identified metric overfitting as a tendency in MBR decoding (\S\ref{sec:background_related}), its exact behavior across different decoding configurations remains largely under-explored. 
In this section, we empirically quantify this overfitting problem by systematically evaluating multiple MBR variants, utility functions, and reference pool sizes.

\subsection{Experimental Setup}

\begin{table*}[ht!]
\centering
\resizebox{\textwidth}{!}{
\begin{tabular}{p{0.2\linewidth} p{0.15\linewidth} l l l l l l l l}
\toprule
\textbf{Decoding Method} & \textbf{Util. Function} & \textbf{$|\mathcal{Y}|$} & \textbf{BLEU} & \textbf{chrF} & \textbf{BLEURT} & \textbf{COMET} & \textbf{BERTScore} & \textbf{COMETKiwi} & \textbf{$\bar{Z}_{\text{other}}$} \\
\midrule
MAP$\epsilon$ & N/A & - & 23.165 & 50.448 & 53.758 & 75.112 & 78.042 & 74.082 & 0.018 \\
\cline{1-10} \cline{2-10}
\multirow[t]{15}{*}{MBR} & \multirow[t]{3}{*}{BLEU} & 4 & \cellcolor{red!17} 21.472 (-1.693) & \cellcolor{red!19} 48.987 (-1.461) & \cellcolor{red!11} 52.098 (-1.660) & \cellcolor{red!14} 72.736 (-2.375) & \cellcolor{green!5} 78.249 (+0.207) & \cellcolor{red!21} 71.689 (-2.392) & \cellcolor{red!20} -0.708 (-0.726) \\
 &  & 32 & \cellcolor{green!3} 23.462 (+0.297) & \cellcolor{green!1} 50.591 (+0.143) & \cellcolor{red!3} 53.204 (-0.554) & \cellcolor{red!3} 74.598 (-0.514) & \cellcolor{green!16} 78.668 (+0.627) & \cellcolor{red!7} 73.199 (-0.882) & \cellcolor{red!3} -0.096 (-0.114) \\
 &  & 256 & \cellcolor{green!9} 24.065 (+0.900) & \cellcolor{green!7} 51.024 (+0.575) & 53.643 (-0.115) & \cellcolor{red!1} 74.926 (-0.186) & \cellcolor{green!20} 78.819 (+0.777) & \cellcolor{red!4} 73.576 (-0.505) & \cellcolor{green!1} 0.078 (+0.060) \\
\cline{2-10}
 & \multirow[t]{3}{*}{chrF} & 4 & \cellcolor{red!20} 21.209 (-1.957) & \cellcolor{red!1} 50.364 (-0.084) & \cellcolor{red!2} 53.334 (-0.424) & \cellcolor{red!9} 73.625 (-1.486) & \cellcolor{green!12} 78.541 (+0.499) & \cellcolor{red!13} 72.626 (-1.456) & \cellcolor{red!10} -0.352 (-0.370) \\
 &  & 32 & \cellcolor{red!2} 22.958 (-0.207) & \cellcolor{green!24} 52.235 (+1.786) & \cellcolor{green!4} 54.484 (+0.726) & \cellcolor{green!1} 75.415 (+0.304) & \cellcolor{green!27} 79.096 (+1.054) & \cellcolor{green!3} 74.434 (+0.353) & \cellcolor{green!8} 0.336 (+0.318) \\
 &  & 256 & 23.254 (+0.089) & \cellcolor{green!27} 52.494 (+2.045) & \cellcolor{green!6} 54.726 (+0.968) & \cellcolor{green!3} 75.715 (+0.603) & \cellcolor{green!29} 79.169 (+1.127) & \cellcolor{green!4} 74.635 (+0.554) & \cellcolor{green!11} 0.442 (+0.424) \\
\cline{2-10}
 & \multirow[t]{3}{*}{BLEURT} & 4 & \cellcolor{red!27} 20.516 (-2.649) & \cellcolor{red!18} 49.069 (-1.380) & \cellcolor{green!36} 59.185 (+5.427) & \cellcolor{green!7} 76.338 (+1.226) & \cellcolor{green!12} 78.508 (+0.466) & \cellcolor{green!2} 74.383 (+0.301) & \cellcolor{red!6} -0.203 (-0.221) \\
 &  & 32 & \cellcolor{red!16} 21.640 (-1.525) & \cellcolor{red!3} 50.192 (-0.257) & \cellcolor{green!47} 60.921 (+7.163) & \cellcolor{green!12} 77.057 (+1.946) & \cellcolor{green!21} 78.873 (+0.831) & \cellcolor{green!10} 75.283 (+1.202) & \cellcolor{green!7} 0.299 (+0.281) \\
 &  & 256 & \cellcolor{red!14} 21.786 (-1.380) & \cellcolor{red!1} 50.364 (-0.085) & \cellcolor{green!49} 61.119 (+7.361) & \cellcolor{green!12} 77.152 (+2.041) & \cellcolor{green!22} 78.906 (+0.864) & \cellcolor{green!11} 75.311 (+1.230) & \cellcolor{green!9} 0.357 (+0.339) \\
\cline{2-10}
 & \multirow[t]{3}{*}{COMET} & 4 & \cellcolor{red!13} 21.876 (-1.289) & \cellcolor{red!3} 50.220 (-0.228) & \cellcolor{green!18} 56.463 (+2.705) & \cellcolor{green!44} 82.338 (+7.226) & \cellcolor{green!19} 78.802 (+0.761) & \cellcolor{green!39} 78.414 (+4.333) & \cellcolor{green!14} 0.549 (+0.531) \\
 &  & 32 & \cellcolor{red!7} 22.444 (-0.721) & \cellcolor{green!6} 50.939 (+0.490) & \cellcolor{green!22} 57.109 (+3.351) & \cellcolor{green!49} 83.075 (+7.963) & \cellcolor{green!23} 78.956 (+0.914) & \cellcolor{green!42} 78.839 (+4.757) & \cellcolor{green!22} 0.834 (+0.816) \\
 &  & 256 & \cellcolor{red!5} 22.665 (-0.500) & \cellcolor{green!7} 50.993 (+0.545) & \cellcolor{green!22} 57.101 (+3.343) & \cellcolor{green!49} 83.116 (+8.004) & \cellcolor{green!24} 78.981 (+0.940) & \cellcolor{green!43} 78.945 (+4.863) & \cellcolor{green!24} 0.884 (+0.866) \\
\cline{2-10}
 & \multirow[t]{3}{*}{BERTScore} & 4 & \cellcolor{red!21} 21.156 (-2.009) & \cellcolor{red!10} 49.651 (-0.797) & \cellcolor{red!4} 53.021 (-0.737) & \cellcolor{red!11} 73.227 (-1.884) & \cellcolor{green!26} 79.066 (+1.024) & \cellcolor{red!16} 72.225 (-1.856) & \cellcolor{red!12} -0.428 (-0.446) \\
 &  & 32 & \cellcolor{red!2} 22.947 (-0.218) & \cellcolor{green!11} 51.276 (+0.828) & \cellcolor{green!4} 54.457 (+0.699) & \cellcolor{red!2} 74.745 (-0.367) & \cellcolor{green!46} 79.841 (+1.799) & \cellcolor{red!3} 73.697 (-0.385) & \cellcolor{green!7} 0.291 (+0.273) \\
 &  & 256 & 23.200 (+0.035) & \cellcolor{green!13} 51.475 (+1.027) & \cellcolor{green!6} 54.685 (+0.927) & 75.051 (-0.060) & \cellcolor{green!50} 79.963 (+1.921) & \cellcolor{red!1} 73.964 (-0.118) & \cellcolor{green!10} 0.399 (+0.382) \\
\cline{1-10} \cline{2-10}
\multirow[t]{15}{*}{Probabilistic MBR} & \multirow[t]{3}{*}{BLEU} & 4 & \cellcolor{red!32} 20.116 (-3.049) & \cellcolor{red!36} 47.730 (-2.718) & \cellcolor{red!18} 51.033 (-2.725) & \cellcolor{red!28} 70.590 (-4.521) & \cellcolor{red!9} 77.688 (-0.354) & \cellcolor{red!42} 69.419 (-4.663) & \cellcolor{red!39} -1.383 (-1.401) \\
 &  & 32 & \cellcolor{red!16} 21.606 (-1.559) & \cellcolor{red!22} 48.792 (-1.657) & \cellcolor{red!14} 51.635 (-2.123) & \cellcolor{red!21} 71.691 (-3.421) & 78.058 (+0.016) & \cellcolor{red!30} 70.690 (-3.392) & \cellcolor{red!27} -0.947 (-0.965) \\
 &  & 256 & 23.182 (+0.017) & \cellcolor{red!3} 50.189 (-0.259) & \cellcolor{red!8} 52.543 (-1.216) & \cellcolor{red!13} 72.919 (-2.192) & \cellcolor{green!13} 78.557 (+0.515) & \cellcolor{red!22} 71.635 (-2.447) & \cellcolor{red!12} -0.424 (-0.441) \\
\cline{2-10}
 & \multirow[t]{3}{*}{chrF} & 4 & \cellcolor{red!39} 19.417 (-3.748) & \cellcolor{red!31} 48.145 (-2.304) & \cellcolor{red!18} 51.052 (-2.706) & \cellcolor{red!29} 70.402 (-4.710) & \cellcolor{red!7} 77.749 (-0.293) & \cellcolor{red!43} 69.246 (-4.835) & \cellcolor{red!39} -1.376 (-1.394) \\
 &  & 32 & \cellcolor{red!28} 20.422 (-2.743) & \cellcolor{red!13} 49.419 (-1.029) & \cellcolor{red!12} 51.880 (-1.878) & \cellcolor{red!21} 71.636 (-3.476) & \cellcolor{green!2} 78.143 (+0.102) & \cellcolor{red!30} 70.652 (-3.429) & \cellcolor{red!25} -0.910 (-0.928) \\
 &  & 256 & \cellcolor{red!20} 21.228 (-1.937) & \cellcolor{red!2} 50.275 (-0.173) & \cellcolor{red!7} 52.635 (-1.123) & \cellcolor{red!16} 72.534 (-2.578) & \cellcolor{green!9} 78.398 (+0.356) & \cellcolor{red!23} 71.508 (-2.573) & \cellcolor{red!16} -0.574 (-0.592) \\
\cline{2-10}
 & \multirow[t]{3}{*}{BLEURT} & 4 & \cellcolor{red!48} 18.610 (-4.556) & \cellcolor{red!44} 47.140 (-3.309) & 53.821 (+0.063) & \cellcolor{red!20} 71.771 (-3.340) & \cellcolor{red!13} 77.540 (-0.502) & \cellcolor{red!30} 70.692 (-3.389) & \cellcolor{red!42} -1.491 (-1.509) \\
 &  & 32 & \cellcolor{red!30} 20.244 (-2.921) & \cellcolor{red!22} 48.763 (-1.685) & \cellcolor{green!25} 57.546 (+3.788) & 75.063 (-0.048) & \cellcolor{green!6} 78.274 (+0.233) & \cellcolor{red!2} 73.812 (-0.270) & \cellcolor{red!13} -0.456 (-0.474) \\
 &  & 256 & \cellcolor{red!14} 21.764 (-1.401) & 50.382 (-0.067) & \cellcolor{green!45} 60.621 (+6.863) & \cellcolor{green!12} 77.114 (+2.002) & \cellcolor{green!23} 78.948 (+0.907) & \cellcolor{green!10} 75.236 (+1.154) & \cellcolor{green!9} 0.362 (+0.344) \\
\cline{2-10}
 & \multirow[t]{3}{*}{COMET} & 4 & \cellcolor{red!50} 18.422 (-4.743) & \cellcolor{red!49} 46.782 (-3.666) & \cellcolor{red!14} 51.579 (-2.179) & \cellcolor{red!12} 73.141 (-1.971) & \cellcolor{red!18} 77.349 (-0.692) & \cellcolor{red!26} 71.096 (-2.985) & \cellcolor{red!45} -1.613 (-1.631) \\
 &  & 32 & \cellcolor{red!35} 19.774 (-3.391) & \cellcolor{red!27} 48.443 (-2.005) & \cellcolor{green!3} 54.324 (+0.566) & \cellcolor{green!23} 78.953 (+3.841) & \cellcolor{green!1} 78.093 (+0.052) & \cellcolor{green!16} 75.909 (+1.827) & \cellcolor{red!14} -0.482 (-0.500) \\
 &  & 256 & \cellcolor{red!9} 22.292 (-0.873) & \cellcolor{green!4} 50.756 (+0.307) & \cellcolor{green!21} 56.967 (+3.209) & \cellcolor{green!47} 82.694 (+7.582) & \cellcolor{green!24} 78.990 (+0.948) & \cellcolor{green!41} 78.687 (+4.605) & \cellcolor{green!21} 0.780 (+0.763) \\
\cline{2-10}
 & \multirow[t]{3}{*}{BERTScore} & 4 & \cellcolor{red!48} 18.550 (-4.615) & \cellcolor{red!50} 46.751 (-3.697) & \cellcolor{red!22} 50.458 (-3.300) & \cellcolor{red!35} 69.418 (-5.693) & \cellcolor{red!12} 77.578 (-0.464) & \cellcolor{red!50} 68.539 (-5.543) & \cellcolor{red!50} -1.768 (-1.786) \\
 &  & 32 & \cellcolor{red!37} 19.591 (-3.574) & \cellcolor{red!32} 48.034 (-2.414) & \cellcolor{red!14} 51.578 (-2.180) & \cellcolor{red!26} 70.842 (-4.270) & \cellcolor{green!4} 78.214 (+0.172) & \cellcolor{red!38} 69.763 (-4.318) & \cellcolor{red!35} -1.233 (-1.250) \\
 &  & 256 & \cellcolor{red!5} 22.613 (-0.552) & \cellcolor{green!5} 50.861 (+0.413) & \cellcolor{green!2} 54.123 (+0.364) & \cellcolor{red!4} 74.409 (-0.702) & \cellcolor{green!42} 79.686 (+1.644) & \cellcolor{red!7} 73.284 (-0.797) & \cellcolor{green!3} 0.126 (+0.108) \\
\cline{1-10} \cline{2-10}
\multirow[t]{5}{*}{Model-based MBR} & BLEU & 256 & \cellcolor{green!8} 23.972 (+0.807) & \cellcolor{green!7} 50.988 (+0.540) & 53.624 (-0.134) & \cellcolor{red!1} 74.840 (-0.272) & \cellcolor{green!19} 78.779 (+0.737) & \cellcolor{red!6} 73.414 (-0.667) & 0.044 (+0.026) \\
\cline{2-10}
 & chrF & 256 & \cellcolor{red!2} 22.955 (-0.210) & \cellcolor{green!26} 52.433 (+1.984) & \cellcolor{green!6} 54.752 (+0.994) & \cellcolor{green!3} 75.662 (+0.551) & \cellcolor{green!30} 79.210 (+1.168) & \cellcolor{green!4} 74.623 (+0.541) & \cellcolor{green!11} 0.414 (+0.396) \\
\cline{2-10}
 & BLEURT & 256 & \cellcolor{red!19} 21.347 (-1.818) & \cellcolor{red!2} 50.257 (-0.191) & \cellcolor{green!50} 61.244 (+7.486) & \cellcolor{green!12} 77.086 (+1.974) & \cellcolor{green!21} 78.884 (+0.842) & \cellcolor{green!10} 75.262 (+1.180) & \cellcolor{green!7} 0.273 (+0.255) \\
\cline{2-10}
 & COMET & 256 & \cellcolor{red!8} 22.352 (-0.813) & \cellcolor{green!7} 50.978 (+0.529) & \cellcolor{green!22} 57.194 (+3.436) & \cellcolor{green!50} 83.161 (+8.049) & \cellcolor{green!25} 79.015 (+0.974) & \cellcolor{green!44} 79.005 (+4.924) & \cellcolor{green!23} 0.862 (+0.844) \\
\cline{2-10}
 & BERTScore & 256 & 23.105 (-0.060) & \cellcolor{green!13} 51.483 (+1.034) & \cellcolor{green!6} 54.740 (+0.982) & 75.050 (-0.061) & \cellcolor{green!49} 79.963 (+1.921) & \cellcolor{red!1} 73.863 (-0.218) & \cellcolor{green!10} 0.385 (+0.367) \\
\cline{1-10} \cline{2-10}
\textit{Oracle} & N/A & - & \textit{41.875} & \textit{65.930} & \textit{67.723} & \textit{85.800} & \textit{85.050} & \textit{82.773} & \textit{3.628} \\
\bottomrule
\end{tabular}

}
\caption{Performance of decoding variants on WMT22 En$\rightarrow$De ($|\mathcal{H}|=256$). Absolute differences (in parentheses) are relative to the \texttt{MAP$\epsilon$} baseline. Color gradation indicates the relative magnitude of \colorbox{green!15}{improvement} or \colorbox{red!15}{degradation}.}
\label{tab:mbr_overfitting_wmt22_ende}
\end{table*}

\paragraph{Datasets.} 

We conducted our experiments on the WMT-22 English-to-German (En$\rightarrow$De) test set~\cite{kocmi-etal-2022-findings}. We selected this benchmark because it is a standard evaluation dataset widely adopted in recent MBR decoding literature~\cite{trabelsi-2024-efficient-mbr, kamigaito-etal-2025-diversity}. Furthermore, to demonstrate the generalizability of our findings, experiments on additional language directions (De$\rightarrow$En) and alternative text generation tasks (XSum) are detailed in Appendix~\ref{appendix:additional_exp}.

\paragraph{Models and hypotheses generation.}

We generated all hypotheses for the translation tasks using the M2M100 model (418M parameters)~\cite{fan-2020-englishcentricmultilingualmachinetranslation}. To evaluate the scaling behavior of MBR, we generated candidate pools of varying sizes using $\epsilon$-sampling~\cite{hewitt-etal-2022-truncation} with $\epsilon=0.02$. Specifically, we fixed the candidate set size at $|\mathcal{H}| = 256$ and evaluated against pseudo-reference sets of sizes $|\mathcal{Y}| \in \{4, 32, 256\}$. In addition to standard MBR, we experimented with multiple decoding variants, including Model-based MBR~\cite{jinnai-2024-modelmbr} and Probabilistic MBR (PMBR)~\cite{trabelsi-2024-efficient-mbr}. For Model-based MBR, we utilized the normalized log-probabilities of the generation model and fixed both sets at 256 ($|\mathcal{H}| = |\mathcal{Y}| = 256$). Finally, we evaluated the raw $\epsilon$-sampled outputs without MBR as our baseline.

\paragraph{Utility function and evaluation metrics.}

To comprehensively evaluate generation quality, we employed six distinct metrics: BLEU~\cite{papineni-etal-2002-bleu}, chrF~\cite{popovic-2015-chrf}, BLEURT~\cite{sellam-etal-2020-bleurt}, COMET~\cite{rei-etal-2020-comet}, COMETKiwi~\cite{rei-etal-2022-cometkiwi}, and BERTScore~\cite{zhang-2020-bertscore}.
To investigate overfitting behaviors across different metrics, we tested every metric except the reference-free COMETKiwi, as the MBR utility function. We define the evaluation metrics not selected as the utility function as \textbf{\textit{off-target}} metrics.
To quantify the extent of metric overfitting, we standardize evaluation scores across MBR variants and define $\bar{Z}_{other}$ as the average of these standardized scores for all off-target metrics. This isolates unoptimized performance to quantify generalized quality, allowing us to observe the distinct effects of optimizing for surface-level versus neural-based metrics.
Finally, to establish an upper bound for our evaluations, we include an \textit{Oracle} baseline. It is computed independently for each individual evaluation metric.

\subsection{Results}

\paragraph{Observation 1: Overfitting occurs across all utility metrics.}
Table~\ref{tab:mbr_overfitting_wmt22_ende} empirically confirms metric overfitting across all evaluated utility functions. Hypotheses consistently achieve relatively higher scores on the utility metric compared to off-target evaluations. For instance, MBR ($|\mathcal{Y}|=256$) with COMET utility yields a massive +8.004 improvement in COMET relative to the MAP$\epsilon$ baseline, yet simultaneously degrades BLEU by -0.500. Furthermore, this collateral damage systematically divides along metric architectures. Optimizing for neural metrics consistently results in a broad degradation of surface-level scores (e.g., dropping BLEU by up to -2.649 under BLEURT utility). Conversely, optimizing for lexical overlap frequently suppresses or degrades improvements measured by neural metrics like COMETKiwi. This divergence demonstrates that the decoding process forces hypotheses to overfit to the specific lexical or semantic features favored by the evaluator's family, unbalancing overall generation quality.

\paragraph{Observation 2: Overfitting persists across MBR variants.}
Crucially, the results reveal that metric overfitting is an inherent limitation, not exclusive to standard MBR. When observing alternative decoding strategies, namely PMBR and Model-based MBR, the exact same pattern of utility metric inflation and off-target degradation emerges. For example, PMBR utilizing BLEURT as the utility function ($|\mathcal{Y}|=256$) substantially boosts its own evaluation by +6.863, but heavily penalizes BLEU (-1.401) compared to the baseline. Similarly, Model-based MBR optimizing for COMET inflates its score by +8.049 while degrading BLEU by -0.813. This confirms that merely altering the sampling mechanism or expectation estimator is insufficient to prevent overfitting to the utility metric.

\paragraph{Observation 3: Scaling the reference size improves baseline quality.}
Beyond the observed overfitting trends, increasing the pseudo-reference pool ($|\mathcal{Y}|$) consistently leads to improved performance to all evaluation metrics, including both the target utility and off-target scores. For example, in standard MBR utilizing chrF as the utility function, scaling the references from 4 to 256 not only improves chrF from 50.364 to 52.494, but also simultaneously lifts the COMET score from 73.625 to 75.715, and BLEU from 21.209 to 23.254. This trend indicates that providing MBR with a larger reference size universally raises the baseline quality of the selected hypotheses, mitigating the most severe score degradations observed at extremely low reference counts. This aligns with findings from previous studies~\cite{eikema-aziz-2022-sampling, kamigaito-etal-2025-diversity}.

\paragraph{Summary.}
Our preliminary analysis demonstrates that \textbf{\textit{metric overfitting is a persistent limitation in MBR decoding}}. It is not merely an artifact of a specific utility function; rather, it is a bias that persists regardless of the underlying MBR variant employed. Furthermore, while scaling the pseudo-reference pool size generally improves baseline translation quality, generating and evaluating massive candidate pools is computationally prohibitive. More critically, expanding this search space simultaneously provides the decoding algorithm with greater opportunities to exploit the mathematical quirks of the utility metric, thereby exacerbating the overfitting effect. Consequently, a more robust intervention is required to decouple the true consensus from these metric biases.
Moreover, we further investigated this tendency across domains and tasks, e.g., summarization, in Appendix~\ref{appendix:additional_exp}. These results also confirmed that metric overfitting is a universal issue. Regardless of the specific utility functions employed, the result consistently overfit to the designated utility metric. These results definitively demonstrate that the tendency of the decoding algorithm to exploit metric biases is an inherent issue in standard MBR across entirely different domains and metric landscapes.

\section{Proposed Method: SVD-MBR}

\S\ref{sec:mbr_overfit} demonstrated that metric overfitting is a problem in MBR decoding. We hypothesize that this phenomenon occurs because the raw expected scores are highly sensitive to microscopic variances and erratic outliers within the evaluation metrics. To counter this, we introduce Singular Value Decomposition MBR (SVD-MBR) decoding, a novel approach that applies denoising to the pairwise matrix prior to candidate selection. 

Formally, we reframe the pairwise utility matrix $\mathbf{M}$ as an information signal. The true consensus between hypotheses and pseudo-references forms a low-rank pattern, whereas metric biases and outliers manifest as noise. We apply SVD to factorize the matrix $\mathbf{M}$:
\begin{equation}
  \mathbf{M} = \mathbf{U} \mathbf{\Sigma} \mathbf{V}^\top,  
\end{equation}
where $\mathbf{U} \in \mathbb{R}^{|\mathcal{H}| \times |\mathcal{H}|}$ is the orthogonal matrix of left singular vectors, $\mathbf{\Sigma} \in \mathbb{R}^{|\mathcal{H}| \times |\mathcal{Y}|}$ is the diagonal matrix containing the singular values in descending order, and $\mathbf{V} \in \mathbb{R}^{|\mathcal{Y}| \times |\mathcal{Y}|}$ is the orthogonal matrix of right singular vectors. The matrices $\mathbf{U}$ and $\mathbf{V}$ represent the latent structures of the hypotheses and pseudo-references, respectively.

\begin{figure}[!t]
    \centering
    \includegraphics[width=\linewidth]{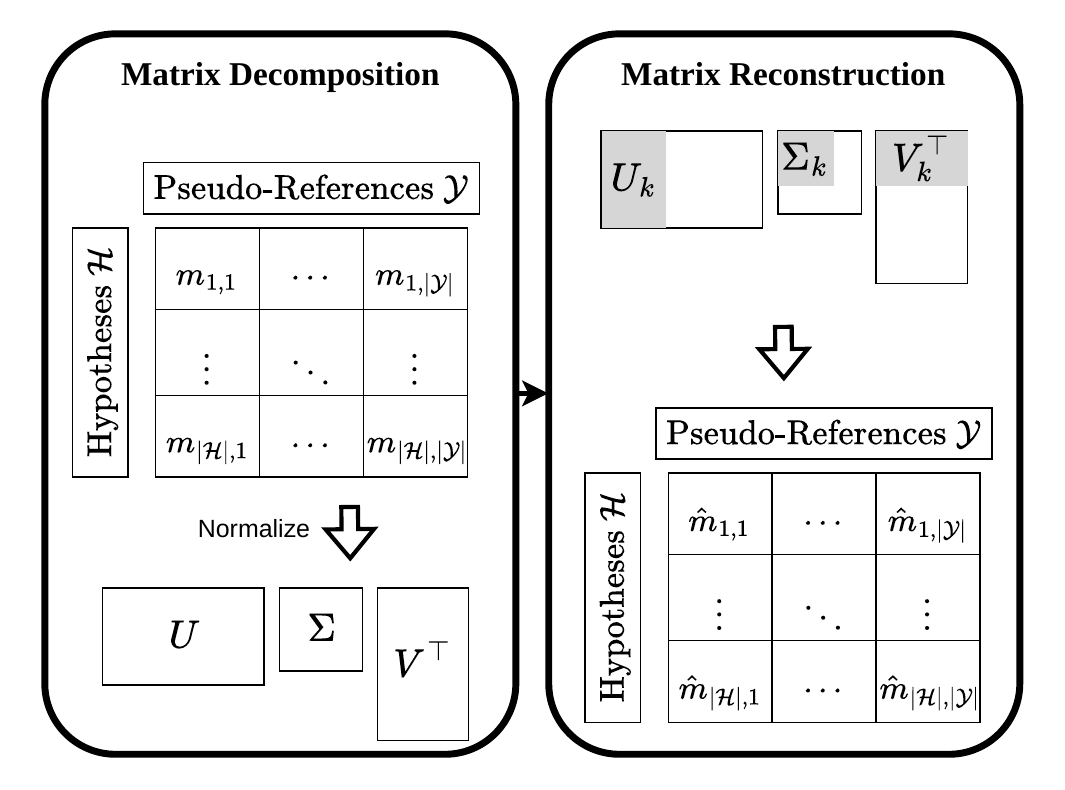}
    \caption{SVD-MBR Methodology}
    \label{fig:methodology_diagram}
\end{figure}

To filter out the noise, we extract only the top-$k$ components by their singular values. We truncate the decomposed matrices to $\mathbf{U}_k \in \mathbb{R}^{|\mathcal{H}| \times k}$, $\mathbf{\Sigma}_k \in \mathbb{R}^{k \times k}$, and $\mathbf{V}_k \in \mathbb{R}^{|\mathcal{Y}| \times k}$. Finally, we reconstruct a low-rank approximation of the original matrix: 

\begin{equation}
    \hat{\mathbf{M}} = \mathbf{U}_k \mathbf{\Sigma}_k \mathbf{V}_k^\top,   
\end{equation}
where $\hat{\mathbf{M}}$ represents the smoothed pairwise utility matrix devoid of the noise. We then substitute this denoised matrix $\hat{\mathbf{M}}$ back into Equation~\ref{eq:mbr_expected_score} to compute the expected scores and extract the optimal hypothesis. Figure~\ref{fig:methodology_diagram} visualizes this entire process.

In our preliminary studies, we applied SVD directly to the raw pairwise utility matrix. However, because evaluation metrics operate on arbitrary numerical scales with high variance, direct decomposition often disproportionately weights mean magnitude over intrinsic consensus. To stabilize the decomposition and center the distribution, we standardize the pairwise matrix by applying z-score normalization prior to factorization. 
We apply this normalization element-wise to prevent it from altering the relative ranking of the expected utility scores. We empirically verify that this transformation does not change the indices in Appendix~\ref{appendix:mbr_normed_mbr}.

\section{Experiments}

\subsection{Experiment Setup}

We retain the experimental configuration detailed in \S\ref{sec:mbr_overfit}. 
We evaluate top-$k$ truncation using three small integers. This design choice is grounded in prior literature~\cite{trabelsi-2024-efficient-mbr}, which establishes that the consensus within an MBR pairwise matrix is inherently a low-rank signal.
In addition, we provide our empirical analysis to support this design choice in Appendix \ref{appendix:mbr_low_rank}. We further employ the previously defined $\bar{Z}_{other}$ score to compare the generalized translation quality of SVD-MBR against standard MBR. Finally, statistical significance is verified via paired bootstrap resampling with 10,000 iterations~\cite{koehn-2004-statistical}.

\begin{table*}[ht!]
\centering
\resizebox{\textwidth}{!}{
\begin{tabular}{llllllllll}
\toprule
\textbf{Decoding Method} & \textbf{Util. Function} & \textbf{$|\mathcal{Y}|$} & \textbf{BLEU} & \textbf{chrF} & \textbf{BLEURT} & \textbf{COMET} & \textbf{BERTScore} & \textbf{COMETKiwi} & \textbf{$\bar{Z}_{\text{other}}$} \\
\midrule
\multirow[t]{15}{*}{MBR} & \multirow[t]{3}{*}{BLEU} & 4 & 21.472 & 48.987 & 52.098 & 72.736 & 78.249 & 71.689 & -1.394 \\
 &  & 32 & 23.462 & 50.591 & 53.204 & 74.598 & 78.668 & 73.199 & -0.547 \\
 &  & 256 & 24.065 & 51.024 & 53.643 & 74.926 & 78.819 & 73.576 & -0.304 \\
\cline{2-10}
 & \multirow[t]{3}{*}{chrF} & 4 & 21.209 & 50.364 & 53.334 & 73.625 & 78.541 & 72.626 & -0.924 \\
 &  & 32 & 22.958 & 52.235 & 54.484 & 75.415 & 79.096 & 74.434 & 0.036 \\
 &  & 256 & 23.254 & 52.494 & 54.726 & 75.715 & 79.169 & 74.635 & 0.183 \\
\cline{2-10}
 & \multirow[t]{3}{*}{BLEURT} & 4 & 20.516 & 49.069 & 59.185 & 76.338 & 78.508 & 74.383 & -0.920 \\
 &  & 32 & 21.640 & 50.192 & 60.921 & 77.057 & 78.873 & 75.283 & -0.171 \\
 &  & 256 & 21.786 & 50.364 & 61.119 & 77.152 & 78.906 & 75.311 & -0.083 \\
\cline{2-10}
 & \multirow[t]{3}{*}{COMET} & 4 & 21.876 & 50.220 & 56.463 & 82.338 & 78.802 & 78.414 & 0.158 \\
 &  & 32 & 22.444 & 50.939 & 57.109 & 83.075 & 78.956 & 78.839 & 0.577 \\
 &  & 256 & 22.665 & 50.993 & 57.101 & 83.116 & 78.981 & 78.945 & 0.653 \\
\cline{2-10}
 & \multirow[t]{3}{*}{BERTScore} & 4 & 21.156 & 49.651 & 53.021 & 73.227 & 79.066 & 72.225 & -1.037 \\
 &  & 32 & 22.947 & 51.276 & 54.457 & 74.745 & 79.841 & 73.697 & -0.018 \\
 &  & 256 & 23.200 & 51.475 & 54.685 & 75.051 & 79.963 & 73.964 & 0.132 \\
\cline{1-10} \cline{2-10}
\multirow[t]{15}{*}{SVD-MBR ($k=1$)} & \multirow[t]{3}{*}{BLEU} & 4 & 21.466 (-0.01) & \cellcolor{red!2} 48.954 (-0.03) & \cellcolor{red!3} 52.013 (-0.08) & \cellcolor{red!41} 72.443 (-0.29) & \cellcolor{red!10} 78.168 (-0.08) & \cellcolor{red!44} 71.316 (-0.37) & \cellcolor{red!22} -1.494 (-0.100) \\
 &  & 32 & \cellcolor{red!7} 23.382 (-0.08) & \cellcolor{red!5} 50.529 (-0.06) & \cellcolor{red!4} 53.094 (-0.11) & \cellcolor{red!32} 74.369 (-0.23) & \cellcolor{red!10} 78.588 (-0.08) & \cellcolor{red!26} 72.978 (-0.22) & \cellcolor{red!20} -0.639 (-0.091) \\
 &  & 256 & \cellcolor{red!13} 23.917 (-0.15) & \cellcolor{red!9} 50.917 (-0.11) & \cellcolor{red!5} 53.492 (-0.15) & \cellcolor{red!31} 74.706 (-0.22) & \cellcolor{red!3} 78.789 (-0.03) & \cellcolor{red!34} 73.285 (-0.29) & \cellcolor{red!18} -0.389 (-0.085) \\
\cline{2-10}
 & \multirow[t]{3}{*}{chrF} & 4 & \cellcolor{red!10} 21.093 (-0.12) & \cellcolor{red!50} 49.807 (-0.56) & \cellcolor{red!21} 52.774 (-0.56) & \cellcolor{red!37} 73.363 (-0.26) & \cellcolor{red!25} 78.339 (-0.20) & \cellcolor{red!37} 72.315 (-0.31) & \cellcolor{red!44} -1.125 (-0.201) \\
 &  & 32 & \cellcolor{green!28} 23.279 (+0.32)$^*$ & \cellcolor{red!45} 51.723 (-0.51) & \cellcolor{red!12} 54.177 (-0.31) & \cellcolor{red!25} 75.233 (-0.18) & \cellcolor{red!13} 78.990 (-0.11) & \cellcolor{red!32} 74.163 (-0.27) & \cellcolor{red!9} -0.005 (-0.041) \\
 &  & 256 & \cellcolor{green!22} 23.507 (+0.25)$^*$ & \cellcolor{red!47} 51.962 (-0.53) & \cellcolor{red!18} 54.256 (-0.47) & \cellcolor{red!44} 75.399 (-0.32) & \cellcolor{red!22} 78.986 (-0.18) & \cellcolor{red!40} 74.294 (-0.34) & \cellcolor{red!25} 0.067 (-0.116) \\
\cline{2-10}
 & \multirow[t]{3}{*}{BLEURT} & 4 & \cellcolor{green!23} 20.780 (+0.26) & 49.069 (+0.00) & \cellcolor{red!50} 57.909 (-1.28) & \cellcolor{red!37} 76.071 (-0.27) & \cellcolor{red!11} 78.412 (-0.10) & \cellcolor{red!17} 74.239 (-0.14) & \cellcolor{red!4} -0.940 (-0.019) \\
 &  & 32 & \cellcolor{green!36} 22.053 (+0.41)$^*$ & \cellcolor{green!3} 50.226 (+0.03) & \cellcolor{red!42} 59.828 (-1.09) & 77.051 (-0.01) & \cellcolor{red!4} 78.839 (-0.03) & \cellcolor{green!6} 75.336 (+0.05) & \cellcolor{green!17} -0.092 (+0.079) \\
 &  & 256 & \cellcolor{green!50} 22.348 (+0.56)$^*$ & \cellcolor{green!18} 50.570 (+0.21) & \cellcolor{red!39} 60.107 (-1.01) & \cellcolor{green!16} 77.266 (+0.11) & \cellcolor{green!8} 78.976 (+0.07) & \cellcolor{green!19} 75.474 (+0.16) & \cellcolor{green!46} 0.127 (+0.210)$^*$ \\
\cline{2-10}
 & \multirow[t]{3}{*}{COMET} & 4 & \cellcolor{red!6} 21.805 (-0.07) & \cellcolor{red!4} 50.169 (-0.05) & \cellcolor{red!6} 56.291 (-0.17) & \cellcolor{red!35} 82.089 (-0.25) & \cellcolor{red!9} 78.729 (-0.07) & \cellcolor{red!11} 78.321 (-0.09) & \cellcolor{red!17} 0.078 (-0.080) \\
 &  & 32 & \cellcolor{green!14} 22.605 (+0.16) & \cellcolor{green!8} 51.032 (+0.09) & \cellcolor{red!3} 57.020 (-0.09) & \cellcolor{red!18} 82.945 (-0.13) & \cellcolor{green!3} 78.986 (+0.03) & \cellcolor{red!6} 78.786 (-0.05) & \cellcolor{green!12} 0.631 (+0.055) \\
 &  & 256 & \cellcolor{red!3} 22.624 (-0.04) & \cellcolor{green!3} 51.034 (+0.04) & \cellcolor{red!1} 57.066 (-0.04) & \cellcolor{red!9} 83.047 (-0.07) & \cellcolor{green!2} 79.004 (+0.02) & \cellcolor{green!3} 78.973 (+0.03) & \cellcolor{green!2} 0.663 (+0.010) \\
\cline{2-10}
 & \multirow[t]{3}{*}{BERTScore} & 4 & \cellcolor{red!41} 20.684 (-0.47) & \cellcolor{red!41} 49.189 (-0.46) & \cellcolor{red!7} 52.836 (-0.19) & \cellcolor{red!22} 73.067 (-0.16) & \cellcolor{red!50} 78.666 (-0.40) & \cellcolor{red!15} 72.095 (-0.13) & \cellcolor{red!50} -1.262 (-0.225) \\
 &  & 32 & \cellcolor{red!9} 22.840 (-0.11) & \cellcolor{red!18} 51.067 (-0.21) & \cellcolor{red!7} 54.254 (-0.20) & \cellcolor{green!18} 74.878 (+0.13) & \cellcolor{red!39} 79.524 (-0.32) & \cellcolor{green!6} 73.748 (+0.05) & \cellcolor{red!14} -0.085 (-0.067) \\
 &  & 256 & \cellcolor{green!17} 23.400 (+0.20) & \cellcolor{green!7} 51.558 (+0.08) & \cellcolor{green!1} 54.713 (+0.03) & \cellcolor{green!27} 75.247 (+0.20) & \cellcolor{red!32} 79.704 (-0.26) & \cellcolor{green!21} 74.141 (+0.18) & \cellcolor{green!19} 0.219 (+0.086)$^*$ \\
\cline{1-10} \cline{2-10}
\multirow[t]{15}{*}{SVD-MBR ($k=2$)} & \multirow[t]{3}{*}{BLEU} & 4 & \cellcolor{red!33} 21.097 (-0.38) & \cellcolor{red!19} 48.772 (-0.22) & \cellcolor{red!7} 51.915 (-0.18) & \cellcolor{red!28} 72.534 (-0.20) & \cellcolor{red!14} 78.131 (-0.12) & \cellcolor{red!34} 71.405 (-0.28) & \cellcolor{red!33} -1.543 (-0.149) \\
 &  & 32 & \cellcolor{red!10} 23.346 (-0.12) & \cellcolor{red!7} 50.513 (-0.08) & \cellcolor{red!3} 53.122 (-0.08) & \cellcolor{red!25} 74.418 (-0.18) & \cellcolor{red!8} 78.604 (-0.06) & \cellcolor{red!36} 72.897 (-0.30) & \cellcolor{red!19} -0.636 (-0.089) \\
 &  & 256 & \cellcolor{red!14} 23.899 (-0.17) & \cellcolor{red!13} 50.878 (-0.15) & \cellcolor{red!7} 53.454 (-0.19) & \cellcolor{red!37} 74.661 (-0.27) & \cellcolor{red!8} 78.752 (-0.07) & \cellcolor{red!38} 73.253 (-0.32) & \cellcolor{red!26} -0.423 (-0.119) \\
\cline{2-10}
 & \multirow[t]{3}{*}{chrF} & 4 & \cellcolor{red!3} 21.164 (-0.04) & \cellcolor{red!8} 50.267 (-0.10) & \cellcolor{red!9} 53.081 (-0.25) & \cellcolor{red!24} 73.453 (-0.17) & \cellcolor{red!5} 78.498 (-0.04) & \cellcolor{red!15} 72.498 (-0.13) & \cellcolor{red!15} -0.993 (-0.069) \\
 &  & 32 & \cellcolor{red!14} 22.794 (-0.16) & \cellcolor{red!11} 52.107 (-0.13) & \cellcolor{green!6} 54.657 (+0.17)$^*$ & \cellcolor{green!15} 75.521 (+0.11) & \cellcolor{red!2} 79.072 (-0.02) & \cellcolor{green!15} 74.561 (+0.13) & \cellcolor{red!3} 0.022 (-0.014) \\
 &  & 256 & \cellcolor{green!7} 23.337 (+0.08) & 52.504 (+0.01) & \cellcolor{green!5} 54.869 (+0.14)$^*$ & \cellcolor{green!20} 75.858 (+0.14) & \cellcolor{green!1} 79.177 (+0.01) & \cellcolor{green!28} 74.872 (+0.24)$^*$ & \cellcolor{green!13} 0.243 (+0.060)$^*$ \\
\cline{2-10}
 & \multirow[t]{3}{*}{BLEURT} & 4 & \cellcolor{green!20} 20.751 (+0.23)$^*$ & \cellcolor{green!11} 49.196 (+0.13) & \cellcolor{red!7} 58.983 (-0.20) & \cellcolor{red!7} 76.282 (-0.06) & \cellcolor{green!1} 78.523 (+0.01) & 74.376 (-0.01) & \cellcolor{green!17} -0.844 (+0.077) \\
 &  & 32 & \cellcolor{green!11} 21.773 (+0.13) & \cellcolor{green!9} 50.299 (+0.11) & \cellcolor{red!6} 60.747 (-0.17) & \cellcolor{green!19} 77.193 (+0.14)$^*$ & \cellcolor{green!3} 78.904 (+0.03) & \cellcolor{green!21} 75.463 (+0.18)$^*$ & \cellcolor{green!19} -0.084 (+0.087)$^*$ \\
 &  & 256 & \cellcolor{green!15} 21.958 (+0.17)$^*$ & \cellcolor{green!15} 50.540 (+0.18)$^*$ & \cellcolor{red!2} 61.055 (-0.06) & \cellcolor{green!17} 77.276 (+0.12)$^*$ & \cellcolor{green!8} 78.972 (+0.07)$^*$ & \cellcolor{green!20} 75.485 (+0.17)$^*$ & \cellcolor{green!27} 0.041 (+0.124)$^*$ \\
\cline{2-10}
 & \multirow[t]{3}{*}{COMET} & 4 & \cellcolor{green!12} 22.014 (+0.14)$^*$ & \cellcolor{green!2} 50.249 (+0.03) & 56.464 (+0.00) & \cellcolor{red!4} 82.308 (-0.03) & 78.809 (+0.01) & \cellcolor{green!1} 78.428 (+0.01) & \cellcolor{green!8} 0.196 (+0.038) \\
 &  & 32 & \cellcolor{green!5} 22.510 (+0.07) & \cellcolor{green!8} 51.030 (+0.09) & \cellcolor{green!5} 57.251 (+0.14)$^*$ & \cellcolor{green!13} 83.171 (+0.10)$^*$ & \cellcolor{green!8} 79.027 (+0.07)$^*$ & \cellcolor{green!8} 78.912 (+0.07)$^*$ & \cellcolor{green!18} 0.658 (+0.082)$^*$ \\
 &  & 256 & \cellcolor{red!4} 22.618 (-0.05) & \cellcolor{green!3} 51.033 (+0.04) & \cellcolor{green!5} 57.229 (+0.13)$^*$ & \cellcolor{green!24} 83.285 (+0.17)$^*$ & \cellcolor{green!3} 79.010 (+0.03) & \cellcolor{green!14} 79.065 (+0.12)$^*$ & \cellcolor{green!7} 0.685 (+0.032)$^*$ \\
\cline{2-10}
 & \multirow[t]{3}{*}{BERTScore} & 4 & \cellcolor{red!16} 20.973 (-0.18) & \cellcolor{red!14} 49.494 (-0.16) & \cellcolor{red!2} 52.963 (-0.06) & \cellcolor{green!15} 73.339 (+0.11) & \cellcolor{red!17} 78.924 (-0.14) & \cellcolor{green!12} 72.331 (+0.11) & \cellcolor{red!12} -1.095 (-0.058) \\
 &  & 32 & \cellcolor{red!3} 22.906 (-0.04) & \cellcolor{red!2} 51.248 (-0.03) & \cellcolor{red!5} 54.315 (-0.14) & \cellcolor{green!19} 74.883 (+0.14) & \cellcolor{red!17} 79.698 (-0.14) & \cellcolor{green!10} 73.789 (+0.09) & \cellcolor{red!1} -0.026 (-0.008) \\
 &  & 256 & \cellcolor{green!2} 23.226 (+0.03) & 51.482 (+0.01) & 54.683 (-0.00) & \cellcolor{green!31} 75.273 (+0.22)$^*$ & \cellcolor{red!13} 79.854 (-0.11) & \cellcolor{green!26} 74.184 (+0.22)$^*$ & \cellcolor{green!8} 0.171 (+0.039)$^*$ \\
\cline{1-10} \cline{2-10}
\multirow[t]{15}{*}{SVD-MBR ($k=3$)} & \multirow[t]{3}{*}{BLEU} & 4 & \cellcolor{red!18} 21.264 (-0.21) & \cellcolor{red!10} 48.871 (-0.12) & \cellcolor{red!3} 51.997 (-0.10) & \cellcolor{red!13} 72.640 (-0.10) & \cellcolor{red!8} 78.178 (-0.07) & \cellcolor{red!13} 71.574 (-0.12) & \cellcolor{red!17} -1.474 (-0.080) \\
 &  & 32 & \cellcolor{red!9} 23.360 (-0.10) & \cellcolor{red!9} 50.490 (-0.10) & \cellcolor{red!7} 53.000 (-0.20) & \cellcolor{red!50} 74.246 (-0.35) & \cellcolor{red!5} 78.622 (-0.05) & \cellcolor{red!50} 72.782 (-0.42) & \cellcolor{red!25} -0.662 (-0.114) \\
 &  & 256 & \cellcolor{red!7} 23.979 (-0.09) & \cellcolor{red!1} 51.003 (-0.02) & \cellcolor{red!4} 53.536 (-0.11) & \cellcolor{red!13} 74.828 (-0.10) & \cellcolor{red!2} 78.797 (-0.02) & \cellcolor{red!18} 73.418 (-0.16) & \cellcolor{red!9} -0.346 (-0.042) \\
\cline{2-10}
 & \multirow[t]{3}{*}{chrF} & 4 & \cellcolor{red!9} 21.097 (-0.11) & \cellcolor{red!13} 50.209 (-0.16) & \cellcolor{red!6} 53.157 (-0.18) & \cellcolor{red!4} 73.593 (-0.03) & \cellcolor{red!7} 78.483 (-0.06) & \cellcolor{red!2} 72.602 (-0.02) & \cellcolor{red!14} -0.990 (-0.067) \\
 &  & 32 & \cellcolor{red!12} 22.817 (-0.14) & \cellcolor{red!7} 52.148 (-0.09) & \cellcolor{green!1} 54.528 (+0.04) & \cellcolor{green!2} 75.430 (+0.02) & \cellcolor{red!5} 79.048 (-0.05) & \cellcolor{red!11} 74.342 (-0.09) & \cellcolor{red!12} -0.019 (-0.055) \\
 &  & 256 & \cellcolor{green!2} 23.285 (+0.03) & \cellcolor{green!5} 52.555 (+0.06) & \cellcolor{green!4} 54.836 (+0.11) & 75.717 (+0.00) & 79.166 (-0.00) & 74.642 (+0.01) & \cellcolor{green!3} 0.197 (+0.014) \\
\cline{2-10}
 & \multirow[t]{3}{*}{BLEURT} & 4 & \cellcolor{green!16} 20.698 (+0.18)$^*$ & \cellcolor{green!9} 49.169 (+0.10)$^*$ & \cellcolor{green!2} 59.253 (+0.07)$^*$ & \cellcolor{green!11} 76.418 (+0.08)$^*$ & \cellcolor{green!4} 78.548 (+0.04)$^*$ & \cellcolor{green!1} 74.393 (+0.01) & \cellcolor{green!18} -0.839 (+0.081)$^*$ \\
 &  & 32 & \cellcolor{green!8} 21.739 (+0.10) & \cellcolor{green!4} 50.244 (+0.05) & \cellcolor{red!1} 60.881 (-0.04) & \cellcolor{green!4} 77.090 (+0.03) & \cellcolor{green!1} 78.889 (+0.02) & \cellcolor{green!1} 75.294 (+0.01) & \cellcolor{green!9} -0.130 (+0.041) \\
 &  & 256 & \cellcolor{green!10} 21.908 (+0.12)$^*$ & \cellcolor{green!11} 50.494 (+0.13)$^*$ & 61.101 (-0.02) & \cellcolor{green!1} 77.162 (+0.01) & \cellcolor{green!3} 78.930 (+0.02) & 75.306 (-0.00) & \cellcolor{green!14} -0.020 (+0.063) \\
\cline{2-10}
 & \multirow[t]{3}{*}{COMET} & 4 & 21.884 (+0.01) & \cellcolor{green!2} 50.252 (+0.03) & 56.467 (+0.00) & \cellcolor{red!2} 82.319 (-0.02) & 78.808 (+0.01) & \cellcolor{green!4} 78.450 (+0.04) & \cellcolor{green!3} 0.172 (+0.014) \\
 &  & 32 & \cellcolor{green!1} 22.466 (+0.02) & \cellcolor{green!4} 50.992 (+0.05) & \cellcolor{green!2} 57.181 (+0.07) & \cellcolor{green!2} 83.094 (+0.02) & \cellcolor{green!5} 79.002 (+0.05)$^*$ & \cellcolor{green!4} 78.874 (+0.04) & \cellcolor{green!10} 0.622 (+0.045)$^*$ \\
 &  & 256 & \cellcolor{red!7} 22.585 (-0.08) & \cellcolor{green!2} 51.020 (+0.03) & \cellcolor{green!2} 57.154 (+0.05) & \cellcolor{green!15} 83.222 (+0.11)$^*$ & 78.988 (+0.01) & \cellcolor{green!1} 78.958 (+0.01) & 0.650 (-0.003) \\
\cline{2-10}
 & \multirow[t]{3}{*}{BERTScore} & 4 & \cellcolor{red!10} 21.042 (-0.11) & \cellcolor{red!2} 49.624 (-0.03) & 53.019 (-0.00) & 73.231 (+0.00) & \cellcolor{red!6} 79.016 (-0.05) & \cellcolor{green!3} 72.255 (+0.03) & \cellcolor{red!5} -1.063 (-0.026) \\
 &  & 32 & \cellcolor{green!1} 22.966 (+0.02) & \cellcolor{red!1} 51.256 (-0.02) & \cellcolor{green!2} 54.528 (+0.07) & \cellcolor{green!33} 74.982 (+0.24)$^*$ & \cellcolor{red!6} 79.793 (-0.05) & \cellcolor{green!22} 73.885 (+0.19)$^*$ & \cellcolor{green!7} 0.017 (+0.035)$^*$ \\
 &  & 256 & \cellcolor{green!9} 23.311 (+0.11)$^*$ & \cellcolor{green!7} 51.564 (+0.09) & \cellcolor{green!1} 54.735 (+0.05) & \cellcolor{green!14} 75.156 (+0.11)$^*$ & \cellcolor{red!2} 79.944 (-0.02) & \cellcolor{green!13} 74.073 (+0.11)$^*$ & \cellcolor{green!13} 0.192 (+0.060)$^*$ \\
\cline{1-10} \cline{2-10}
\textit{Oracle} & N/A & - & \textit{41.875} & \textit{65.930} & \textit{67.723} & \textit{85.800} & \textit{85.050} & \textit{82.773} & \textit{4.927} \\
\cline{1-10} \cline{2-10}
\bottomrule
\end{tabular}

}
\caption{Performance of MBR vs. SVD-MBR. $\bar{Z}_{\text{other}}$ represents the mean z-score of all non-utility function evaluation metrics. Values in parentheses are absolute differences against MBR, with colors indicating relative \colorbox{green!15}{improvement} or \colorbox{red!15}{degradation}. Statistical significance ($p < 0.05$) is marked with $^*$.}
\label{tab:svd_mbr_results_wmt22_ende}
\end{table*}

\subsection{Experiment Results}
\label{sec:svd_mbr_result}

\paragraph{Finding 1: SVD-MBR prevents metric overfitting.}
As shown in Table~\ref{tab:svd_mbr_results_wmt22_ende}, SVD-MBR successfully mitigates the metric overfitting observed in the MBR baseline. Specifically, when employing BLEURT as the utility function, SVD-MBR~($k=1$) significantly improves generalized translation quality, achieving a substantial gain of $+0.210$ in the aggregate $\bar{Z}_{other}$ score. At rank $k=2$, we observe significant improvements across almost all utility functions. With the exception of BLEU, optimizing for any other metric yields a statistically significant increase in both the off-target metrics and the $\bar{Z}_{other}$ score, particularly when BLEURT or COMET are used as the utility function. This demonstrates that our overfitting mitigation approach provides a robust solution, successfully preventing the utility function from hijacking the selection process and yielding genuine improvements in overall generation quality.

\paragraph{Finding 2: Excessive $k$ reintroduces metric overfitting.}
Our results empirically justify the decision to keep $k$ extremely small. By tracking the performance trajectory from $k=1$ through $k=3$, it becomes evident that retaining higher singular components simply reintroduces the noise responsible for overfitting. For instance, when optimizing for BLEURT~($|\mathcal{Y}|=256$), the improvement for $\bar{Z}_{other}$ steadily diminishes as $k$ increases, dropping from $+0.210$ at $k=1$, down to $+0.124$ at $k=2$, and finally to $+0.063$ at $k=3$. However, the widespread degradation observed at $k=1$, particularly for utility functions like chrF and BERTScore, suggests that $k=1$ can sometimes be overly aggressive, discarding valuable information alongside the noise. Relaxing the approximation to $k=2$ allows the reconstruction to capture the remainder of this underlying signal, as evidenced by the significant recovery of COMET and COMETKiwi scores, before the noise begins to dominate again at higher ranks. Ultimately, expanding the rank causes the performance to rapidly regress toward the MBR baseline, validating the hypothesis that consensus information is strictly a low-rank signal.

\paragraph{Finding 3: Denoising robustness varies by metric architecture.}
Finally, our experiments reveal that SVD denoising behaves fundamentally differently depending on the underlying utility function. When applied to MBR matrices constructed using surface-level metrics, separating the noise from the true information proves exceedingly difficult. For instance, when utilizing BLEU as the utility function, the $\bar{Z}_{other}$ scores stagnate even as the rank is expanded up to $k=3$. Similarly, chrF exhibits unstable behavior: it suffers massive degradation under a $k=1$ approximation, yields positive gains at $k=2$, but diminishes again by $k=3$. Conversely, neural metrics yields more robust denoising outcomes. Metrics like COMET and BLEURT display generally positive trends when retaining the top two components~($k=2$). Meanwhile, BERTScore demonstrates a distinct pattern, yielding notable mean improvements under both $k=1$ and $k=3$. We hypothesize that neural metrics, which rely on dense continuous embeddings, inherently encode a richer, low-rank consensus that SVD can effectively isolate. In contrast, character- and $n$-gram-level overlap metrics construct highly localized and entangled pairwise distributions. Consequently, forcing them into a strict low-rank approximation inadvertently discards critical discriminating information alongside the noise.

\paragraph{Finding 4: SVD-MBR generalizes across domains and tasks.}
Similar to \S\ref{sec:mbr_overfit}, we further investigate the generalizability of our method across domains and tasks, e.g., summarization, in Appendix~\ref{appendix:additional_exp}. Consistent with our primary findings, the supplementary results confirm that metric overfitting remains a universal issue. Applying SVD-MBR mitigates this overfitting effectively for neural metrics, demonstrating the broad compatibility of low-rank matrix approximations with dense continuous embeddings. On the other hand, these evaluations also reaffirm that applying low-rank denoising to surface-level metrics is highly unstable and struggles to produce robust, generalized impacts.

\paragraph{Summary.}
Our results demonstrate that \textbf{SVD-MBR is an effective method for mitigating metric overfitting in MBR decoding}. By strictly enforcing a low-rank approximation, SVD-MBR successfully decouples the true consensus from high-frequency metric noise, yielding significant gains in generalized generation quality ($\bar{Z}_{other}$). Crucially, retaining higher singular components simply reintroduces the noise responsible for overfitting, validating that true consensus is inherently a low-rank signal. Furthermore, the efficacy of the denoising method is fundamentally tied to the architecture of the utility function: while neural metrics, such as COMET, BLEURT, and BERTScore, encode a low-rank structure ideal for SVD decomposition, surface-level metrics, e.g., BLEU and chrF, construct highly entangled distributions where noise cannot be cleanly separated. Finally, our investigation across domains and tasks in Appendix~\ref{appendix:additional_exp} confirm that metric overfitting is a universal issue and SVD-MBR consistently solved this issue.

\section{Quality Analysis}

\subsection{Error Analysis}
\label{subsec:error_analysis}

To further investigate the mechanics of how SVD-MBR mitigates metric overfitting, we conducted an analysis connecting the intrinsic metric variance across the hypotheses ($\sigma^2$) to the evaluation score deltas between MBR and SVD-MBR 
($\Delta = \text{Score}_{\text{SVD-MBR}} - \text{Score}_{\text{MBR}}$). We report the Spearman correlation ($\rho$) between the variance of the evaluation metric scores across the hypothesis pool per source document. We measure this relationship across two dimensions:
\begin{itemize}
    \item \textbf{Dir. (Directional Change):} Calculates $\rho$ between $\sigma^2$ and $\Delta$. This measures whether higher inherent variance correlates with a net positive or negative score improvement against standard MBR.
    \item \textbf{Mag. (Magnitude Sensitivity):} Calculates $\rho$ between $\sigma^2$ and $\vert{}\Delta\vert{}$. This evaluates how strongly the variance dictates the score shift, regardless of its improvement.
\end{itemize}
As shown in Table~\ref{tab:corr_variance_analysis_ultra_compact_wmt22-ende}, SVD-MBR consistently increases the aggregate off-target scores~($\bar{Z}_{other}$) while simultaneously regularizing the inflated utility metric. However, this efficacy is heavily dependent on the specific utility function employed.

Crucially, the success of SVD in filtering the noise correlates with the intrinsic variance of the utility function. For example, COMET exhibits the highest variance and provides the clearest evidence of noise mitigation: it yields a notable positive gain in off-target metrics ($+0.05$) while restricting the utility metric's inflation. Similarly, BLEURT and chrF possess comparable variance levels, and both exhibit proportional increases in their $\bar{Z}_{other}$ scores~($|\mathcal{Y}|=256$). Interestingly, despite possessing the lowest intrinsic variance of the evaluated metrics, BERTScore still yields significant positive gains for the off-target average.
These variance-dependent characteristic generalize across domains, as detailed in Appendix~\ref{appendix:additional_exp}.

\begin{table}[ht!]
\centering
\resizebox{\columnwidth}{!}{
\begin{NiceTabular}{@{} lll c c c c @{}}
\toprule
\multirow{2}{*}{\textbf{Util. Function}} & \multirow{2}{*}{\textbf{$|\mathcal{Y}|$}} & \multirow{2}{*}{\textbf{Top-$k$}} & \multicolumn{2}{c}{\textbf{Util. Metric}} & \multicolumn{2}{c}{\makecell{\textbf{$\bar{Z}_{\text{other}}$} \\ ($\sigma^2=0.186$)}} \\
\cmidrule(lr){4-5} \cmidrule(lr){6-7}
 &  &  & \textit{Dir.} & \textit{Mag.} & \textit{Dir.} & \textit{Mag.} \\
\midrule
\multirow{9}{*}{\makecell{\textbf{BLEU} \\ ($\sigma^2=79.972$)}} & \multirow{3}{*}{4} & $k=1$ & \cellcolor{green!3} $+0.02$ & \cellcolor{blue!15} $0.10^{*}$ & \cellcolor{green!3} $+0.02$ & $-0.03$ \\
 &  & $k=2$ & \cellcolor{red!5} $-0.04$ & \cellcolor{blue!7} $0.05^{*}$ & $-0.01$ & \cellcolor{blue!7} $0.05^{*}$ \\
 &  & $k=3$ & \cellcolor{red!4} $-0.03$ & \cellcolor{blue!1} $0.01$ & \cellcolor{red!4} $-0.03$ & \cellcolor{blue!2} $0.02$ \\
\hhline{~------}
 & \multirow{3}{*}{32} & $k=1$ & \cellcolor{green!3} $+0.02$ & \cellcolor{blue!8} $0.06^{*}$ & \cellcolor{green!1} $+0.01$ & \cellcolor{blue!2} $0.02$ \\
 &  & $k=2$ & \cellcolor{red!4} $-0.03$ & \cellcolor{blue!5} $0.04$ & \cellcolor{red!1} $-0.01$ & \cellcolor{blue!12} $0.08^{*}$ \\
 &  & $k=3$ & \cellcolor{red!2} $-0.02$ & \cellcolor{blue!3} $0.02$ & \cellcolor{red!3} $-0.02$ & \cellcolor{blue!16} $0.11^{*}$ \\
\hhline{~------}
 & \multirow{3}{*}{256} & $k=1$ & \cellcolor{green!1} $+0.01$ & \cellcolor{blue!4} $0.03$ & $+0.00$ & \cellcolor{blue!2} $0.02$ \\
 &  & $k=2$ & \cellcolor{green!2} $+0.01$ & $-0.01$ & \cellcolor{red!3} $-0.02$ & \cellcolor{blue!8} $0.05^{*}$ \\
 &  & $k=3$ & \cellcolor{green!3} $+0.02$ & $-0.03$ & $-0.01$ & \cellcolor{blue!15} $0.11^{*}$ \\
\midrule
\multirow{9}{*}{\makecell{\textbf{chrF} \\ ($\sigma^2=70.045$)}} & \multirow{3}{*}{4} & $k=1$ & $+0.00$ & \cellcolor{blue!16} $0.11^{*}$ & $+0.01$ & \cellcolor{blue!11} $0.08^{*}$ \\
 &  & $k=2$ & \cellcolor{green!3} $+0.02$ & \cellcolor{blue!11} $0.08^{*}$ & \cellcolor{green!1} $+0.01$ & \cellcolor{blue!8} $0.06^{*}$ \\
 &  & $k=3$ & $-0.01$ & \cellcolor{blue!4} $0.03$ & \cellcolor{green!1} $+0.01$ & \cellcolor{blue!5} $0.03$ \\
\hhline{~------}
 & \multirow{3}{*}{32} & $k=1$ & $+0.00$ & \cellcolor{blue!16} $0.11^{*}$ & \cellcolor{green!1} $+0.01$ & \cellcolor{blue!12} $0.08^{*}$ \\
 &  & $k=2$ & \cellcolor{red!3} $-0.02$ & \cellcolor{blue!18} $0.12^{*}$ & \cellcolor{green!1} $+0.01$ & \cellcolor{blue!12} $0.09^{*}$ \\
 &  & $k=3$ & \cellcolor{red!3} $-0.02$ & \cellcolor{blue!11} $0.08^{*}$ & \cellcolor{red!2} $-0.02$ & \cellcolor{blue!10} $0.07^{*}$ \\
\hhline{~------}
 & \multirow{3}{*}{256} & $k=1$ & \cellcolor{green!6} $+0.04$ & \cellcolor{blue!14} $0.10^{*}$ & \cellcolor{green!9} $+0.06^{*}$ & \cellcolor{blue!16} $0.11^{*}$ \\
 &  & $k=2$ & \cellcolor{green!1} $+0.01$ & \cellcolor{blue!20} $0.14^{*}$ & \cellcolor{green!10} $+0.07^{*}$ & \cellcolor{blue!18} $0.12^{*}$ \\
 &  & $k=3$ & \cellcolor{red!2} $-0.02$ & \cellcolor{blue!14} $0.09^{*}$ & \cellcolor{green!5} $+0.04$ & \cellcolor{blue!11} $0.08^{*}$ \\
\midrule
\multirow{9}{*}{\makecell{\textbf{BLEURT} \\ ($\sigma^2=71.421$)}} & \multirow{3}{*}{4} & $k=1$ & \cellcolor{red!7} $-0.05^{*}$ & \cellcolor{blue!17} $0.12^{*}$ & $+0.00$ & \cellcolor{blue!19} $0.13^{*}$ \\
 &  & $k=2$ & \cellcolor{red!4} $-0.03$ & \cellcolor{blue!10} $0.07^{*}$ & $+0.00$ & \cellcolor{blue!13} $0.09^{*}$ \\
 &  & $k=3$ & \cellcolor{green!5} $+0.04$ & \cellcolor{blue!1} $0.01$ & \cellcolor{green!5} $+0.04$ & \cellcolor{blue!5} $0.03$ \\
\hhline{~------}
 & \multirow{3}{*}{32} & $k=1$ & \cellcolor{green!2} $+0.01$ & \cellcolor{blue!16} $0.11^{*}$ & \cellcolor{green!3} $+0.03$ & \cellcolor{blue!31} $0.21^{*}$ \\
 &  & $k=2$ & \cellcolor{green!5} $+0.03$ & \cellcolor{blue!15} $0.10^{*}$ & \cellcolor{green!2} $+0.02$ & \cellcolor{blue!20} $0.13^{*}$ \\
 &  & $k=3$ & \cellcolor{red!1} $-0.01$ & \cellcolor{blue!6} $0.05^{*}$ & \cellcolor{green!2} $+0.02$ & \cellcolor{blue!14} $0.10^{*}$ \\
\hhline{~------}
 & \multirow{3}{*}{256} & $k=1$ & \cellcolor{red!5} $-0.04$ & \cellcolor{blue!20} $0.14^{*}$ & \cellcolor{green!5} $+0.04$ & \cellcolor{blue!29} $0.20^{*}$ \\
 &  & $k=2$ & $+0.00$ & \cellcolor{blue!16} $0.11^{*}$ & \cellcolor{green!5} $+0.03$ & \cellcolor{blue!15} $0.10^{*}$ \\
 &  & $k=3$ & \cellcolor{red!1} $-0.01$ & \cellcolor{blue!8} $0.06^{*}$ & \cellcolor{green!3} $+0.02$ & \cellcolor{blue!14} $0.10^{*}$ \\
\midrule
\multirow{9}{*}{\makecell{\textbf{COMET} \\ ($\sigma^2=94.773$)}} & \multirow{3}{*}{4} & $k=1$ & \cellcolor{green!1} $+0.01$ & $-0.07$ & \cellcolor{green!7} $+0.05$ & \cellcolor{blue!29} $0.20^{*}$ \\
 &  & $k=2$ & \cellcolor{red!5} $-0.04$ & $-0.02$ & \cellcolor{green!7} $+0.05$ & \cellcolor{blue!25} $0.17^{*}$ \\
 &  & $k=3$ & \cellcolor{red!6} $-0.04$ & $-0.02$ & \cellcolor{green!6} $+0.04$ & \cellcolor{blue!22} $0.15^{*}$ \\
\hhline{~------}
 & \multirow{3}{*}{32} & $k=1$ & \cellcolor{green!14} $+0.10$ & $-0.06$ & \cellcolor{red!3} $-0.03$ & \cellcolor{blue!21} $0.14^{*}$ \\
 &  & $k=2$ & \cellcolor{green!1} $+0.01$ & \cellcolor{blue!1} $0.01$ & $-0.01$ & \cellcolor{blue!20} $0.14^{*}$ \\
 &  & $k=3$ & \cellcolor{green!1} $+0.01$ & \cellcolor{blue!1} $0.01$ & \cellcolor{red!2} $-0.02$ & \cellcolor{blue!18} $0.12$ \\
\hhline{~------}
 & \multirow{3}{*}{256} & $k=1$ & \cellcolor{green!2} $+0.01^{*}$ & $-0.11^{*}$ & \cellcolor{green!6} $+0.05^{*}$ & \cellcolor{blue!36} $0.25^{*}$ \\
 &  & $k=2$ & \cellcolor{red!7} $-0.05$ & $-0.05$ & \cellcolor{green!11} $+0.08^{*}$ & \cellcolor{blue!20} $0.14$ \\
 &  & $k=3$ & \cellcolor{red!10} $-0.07$ & $-0.05$ & \cellcolor{green!9} $+0.07$ & \cellcolor{blue!21} $0.15^{*}$ \\
\midrule
\multirow{9}{*}{\makecell{\textbf{BERTScore} \\ ($\sigma^2=12.698$)}} & \multirow{3}{*}{4} & $k=1$ & \cellcolor{green!2} $+0.02$ & \cellcolor{blue!13} $0.09^{*}$ & \cellcolor{green!3} $+0.02$ & \cellcolor{blue!12} $0.08^{*}$ \\
 &  & $k=2$ & \cellcolor{red!3} $-0.02$ & \cellcolor{blue!12} $0.08^{*}$ & \cellcolor{green!2} $+0.02$ & \cellcolor{blue!7} $0.05^{*}$ \\
 &  & $k=3$ & $+0.00$ & \cellcolor{blue!7} $0.05^{*}$ & $+0.00$ & $-0.00$ \\
\hhline{~------}
 & \multirow{3}{*}{32} & $k=1$ & \cellcolor{green!4} $+0.03$ & \cellcolor{blue!25} $0.17^{*}$ & \cellcolor{green!2} $+0.02$ & \cellcolor{blue!21} $0.14^{*}$ \\
 &  & $k=2$ & $+0.00$ & \cellcolor{blue!29} $0.20^{*}$ & \cellcolor{green!1} $+0.01$ & \cellcolor{blue!18} $0.12^{*}$ \\
 &  & $k=3$ & \cellcolor{green!1} $+0.01$ & \cellcolor{blue!22} $0.15^{*}$ & \cellcolor{green!3} $+0.02$ & \cellcolor{blue!13} $0.09^{*}$ \\
\hhline{~------}
 & \multirow{3}{*}{256} & $k=1$ & \cellcolor{green!5} $+0.03$ & \cellcolor{blue!24} $0.16^{*}$ & \cellcolor{green!8} $+0.06^{*}$ & \cellcolor{blue!16} $0.11^{*}$ \\
 &  & $k=2$ & \cellcolor{red!4} $-0.03$ & \cellcolor{blue!28} $0.19^{*}$ & \cellcolor{green!3} $+0.03$ & \cellcolor{blue!17} $0.12^{*}$ \\
 &  & $k=3$ & \cellcolor{green!2} $+0.02$ & \cellcolor{blue!14} $0.10^{*}$ & \cellcolor{green!6} $+0.05^{*}$ & \cellcolor{blue!4} $0.03$ \\
\bottomrule
\end{NiceTabular}
}
\caption{Spearman correlation ($\rho$) between hypotheses oracle variance and evaluation score differences. The Util. Metric is the metric used as utility function, while \textbf{$\bar{Z}_{\text{other}}$} is the mean z-score of all non-utility function evaluation metrics. The hypotheses variance ($\sigma^2$) is indicated beneath each corresponding utility function and in $\bar{Z}_{\text{other}}$. \textit{Dir.} measures directional improvement (\colorbox{green!15}{+} / \colorbox{red!15}{-}), and \colorbox{blue!15}{\textit{Mag.}} measures absolute score sensitivity. Significance ($p < 0.05$) is denoted by $^*$.}
\label{tab:corr_variance_analysis_ultra_compact_wmt22-ende}
\end{table}

Further analysis of each evaluation metric (Appendix~\ref{appendix:expand_result}) also shows that some utility functions distribute performance gains equally across all metrics, whereas others concentrate improvements within specific metrics. Moreover, consistent with our findings in \S\ref{sec:svd_mbr_result}, applying SVD when using BLEU as the utility function yields no improvement for $\bar{Z}_{other}$, despite BLEU demonstrating high variance. This further implies that while pairwise BLEU scores are noisy, their localized, $n$-gram-based noise cannot be cleanly decoupled from the true consensus using a low-rank approximation.

\begin{table*}[ht!]
\centering
\small
\begin{tabular}{@{} l c c @{\hspace{0.2\textwidth}} l c c @{}}
\toprule
\multicolumn{3}{@{}p{0.4\textwidth}@{}}{\textbf{Case 1: COMET as Util. Function}} & \multicolumn{3}{@{}p{0.4\textwidth}@{}}{\textbf{Case 2: BLEU as Util. Function}} \\
\midrule
\multicolumn{3}{@{}p{0.4\textwidth}@{}}{\textbf{Source:} My water heater is set pretty low.} & \multicolumn{3}{@{}p{0.4\textwidth}@{}}{\textbf{Source:} Do I have to pay any duties or customs?} \\
\addlinespace
\multicolumn{3}{@{}p{0.4\textwidth}@{}}{\textbf{Reference:} Mein Wassererhitzer ist ziemlich niedrig eingestellt.} & \multicolumn{3}{@{}p{0.4\textwidth}@{}}{\textbf{Reference:} Muss ich Zollgebühren zahlen?} \\
\midrule
\multicolumn{3}{@{}p{0.4\textwidth}@{}}{\textbf{MBR:}} & \multicolumn{3}{@{}p{0.4\textwidth}@{}}{\textbf{MBR:}} \\
\multicolumn{3}{@{}p{0.4\textwidth}@{}}{Meine Wassertemperatur ist ziemlich niedrig} & \multicolumn{3}{@{}p{0.4\textwidth}@{}}{Muss ich Gebühren oder Zölle bezahlen?} \\
\addlinespace
\multicolumn{3}{@{}p{0.4\textwidth}@{}}{\textbf{SVD-MBR:}} & \multicolumn{3}{@{}p{0.4\textwidth}@{}}{\textbf{SVD-MBR:}} \\
\multicolumn{3}{@{}p{0.4\textwidth}@{}}{Der Wasserheater ist ziemlich niedrig eingestellt.} & \multicolumn{3}{@{}p{0.48\textwidth}@{}}{Muss ich Zölle oder Zölle bezahlen?} \\
\midrule
\textbf{Metric} & \textbf{MBR} & \textbf{SVD-MBR} & \textbf{Metric} & \textbf{MBR} & \textbf{SVD-MBR} \\
\midrule
\textbf{COMET} (Util) & 87.5 & \textbf{95.7} (\textcolor{green}{$\uparrow$}) & \textbf{COMET} & \textbf{94.1} & 59.4 (\textcolor{red}{$\downarrow$}) \\
\textbf{BLEURT} & 38.2 & \textbf{68.3} (\textcolor{green}{$\uparrow$}) & \textbf{BLEURT} & \textbf{62.3} & 35.7 (\textcolor{red}{$\downarrow$}) \\
\textbf{BERTScore} & 77.4 & \textbf{86.2} (\textcolor{green}{$\uparrow$}) & \textbf{BERTScore} & \textbf{83.6} & 74.0 (\textcolor{red}{$\downarrow$}) \\
\textbf{chrF} & 52.1 & \textbf{76.2} (\textcolor{green}{$\uparrow$}) & \textbf{chrF} & \textbf{56.1} & 46.0 (\textcolor{red}{$\downarrow$}) \\
\textbf{BLEU} & 26.7 & \textbf{61.5} (\textcolor{green}{$\uparrow$}) & \textbf{BLEU} (Util) & 14.5 & 14.5 \\
\textbf{COMETKiwi} & 80.5 & \textbf{81.2} (\textcolor{green}{$\uparrow$}) & \textbf{COMETKiwi} & \textbf{87.5} & 67.6 (\textcolor{red}{$\downarrow$}) \\
\bottomrule
\end{tabular}
\caption{Sentence comparison of MBR and SVD-MBR ($k=1$, $|\mathcal{Y}|=256$). \textbf{Case 1} uses COMET as the utility function and \textbf{Case 2} uses BLEU as the utility function.}
\label{tab:qualitative_side_by_side_unified}
\end{table*}

\begin{figure*}[t]
    \centering
    \includegraphics[width=1.0\textwidth]{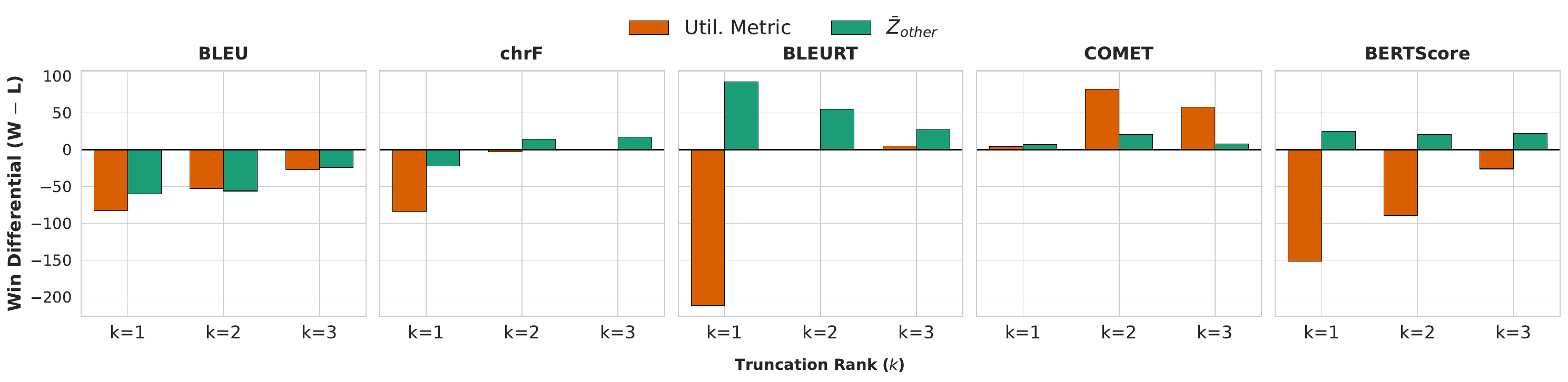}
    \caption{Sentence-level win difference (Wins $-$ Losses) of SVD-MBR compared to the MBR for $|\mathcal{Y}|=256$. The \colorbox{orange!15}{Util. Metric} is the metric used as utility function, while \colorbox{green!15}{$\bar{Z}_{other}$} is the mean z-score of all non-utility function evaluation metrics.}
    \label{fig:svd_mbr_win_diff}
\end{figure*}

\subsection{Qualitative Analysis}
\label{subsec:qualitative_analysis}

To further understand how SVD-MBR mitigates metric overfitting in practice, we conduct a qualitative analysis at the sentence level. By examining individual translations, we can observe exactly how SVD-MBR alters the final hypothesis selection. To this end, we highlight two specific scenarios in Table~\ref{tab:qualitative_side_by_side_unified}: a success case where SVD-MBR~($k=1$) outperforms MBR (using COMET as the utility function), and a failure case where it underperforms (using BLEU as the utility function). To contextualize these individual examples within the broader decoding behavior, we also present a Win Difference analysis comparing SVD-MBR against MBR for both the target utility metric and the aggregate off-target score~($\bar{Z}_{other}$), illustrated in Figure~\ref{fig:svd_mbr_win_diff}. 

Qualitative inspection reveals that when SVD-MBR diverges from MBR, the resulting differences in translation quality can be substantial. In the COMET success case, SVD-MBR yields improvements across all metrics, providing massive gains for metrics like BLEU and BLEURT. Notably, even while regularizing the overfitting behavior, the COMET score remains higher than that of the MBR baseline. Conversely, the failure case demonstrates a similar magnitude of divergence in the opposite direction, with MBR strictly outperforming the SVD-MBR selection across all metrics. 

The Win Difference analysis in Figure~\ref{fig:svd_mbr_win_diff} corroborates these sentence-level observations on a macro scale. For most utility functions, SVD-MBR consistently improves $\bar{Z}_{other}$ while appropriately regularizing the inflated target utility metric. Furthermore, optimizing for COMET proves uniquely robust, frequently yielding simultaneous wins for both the utility metric and off-target evaluations. However, further analysis (Appendix~\ref{appendix:expand_result}) demonstrates that SVD-MBR is a conservative regularizer, yielding gains only when the pairwise utility matrix exhibits exploitable noise. Consequently, utilizing surface-level metrics like BLEU (and certain configurations of chrF) fails to yield off-target improvements, reaffirming that not all metric spaces are equally amenable to low-rank regularization.

\section{Conclusion}

In this study, we first highlighted metric overfitting as a problem in Minimum Bayes Risk (MBR) decoding. We demonstrated that \textbf{MBR inflates the target utility metric while causing unoptimized metrics to stagnate or severely degrade}. Crucially, our evaluations show that this phenomenon is not isolated to a single utility function, but is a persistent issue across multiple MBR variants. 

\textbf{To mitigate this issue, we proposed SVD-MBR}, a novel decoding approach that applies Singular Value Decomposition (SVD) to the pairwise utility matrix. By reconstructing the matrix using only the top-$k$ components, we effectively decouple the underlying consensus from metric noise. Our quantitative results prove that a strict low-rank approximation, e.g., $k=1$ or $k=2$, successfully regularizes this overfitting issue, particularly for neural metrics like COMET and BLEURT. 

Moreover, our error analysis confirms that the efficacy of SVD denoising is strongly correlated with the intrinsic variance of the evaluator, solidifying its role as a robust regularizer. 
In practice, qualitative sentence-level analysis corroborates that SVD-MBR successfully redirects the selection to hypotheses that yield substantial gains across the entire evaluation suite.

Our supplementary experiments confirm that this metric overfitting issue is universal and SVD-MBR consistently mitigates it across different domains and metrics. Ultimately, this demonstrates that \textbf{\textit{surface-level metrics are mathematically resistant to low-rank denoising, and we advise practitioners, supported by our findings, to pair efficient, low-rank MBR approximations specifically with neural metrics for the best generation quality.}}

\section*{Limitations}

While SVD-MBR successfully mitigates metric overfitting and improves generalized translation quality, we identify several limitations to our current approach.

First, the application of SVD introduces additional computational complexity. MBR is already bottlenecked by its $\mathcal{O}(N \times M)$ pairwise evaluation cost. Decomposing the resulting matrix adds an $\mathcal{O}(\min(N^2M, NM^2))$ operation per sentence. However, our primary objective in this work is to establish SVD as a robust mechanism for outlier mitigation and metric regularization, rather than to optimize for decoding latency. We demonstrate that overfitting is a tendency that can be neutralized by our method. Consequently, we view our approach as a complementary component that can be combined with existing efficient MBR frameworks to address throughput constraints in future research.

Second, our experimental analysis relies on automated metrics. This is a deliberate choice based on our objective in investigating and mitigating the tendency of MBR decoding to overfit, rather than to pursue the incremental performance gains that requires human preference. Since our results demonstrate that SVD effectively reduces this overfitting and restores alignment across various metrics, it is sufficient for establishing the efficacy of our method. Future work may incorporate human evaluation, such as Multidimensional Quality Metrics (MQM), to assess the qualitative impact of this denoising on translation fluency and adequacy. However, this issue is also raised widely in other MBR decoding researches.

Finally, our experimental scope is constrained to $\epsilon$-sampling for candidate generation and SVD for denoising. Rather than expanding the generation methods or denoising algorithms, we prioritized an extensive sweep across a set of variables, including $k$-values, utility functions, datasets, and pseudo-reference counts, to comprehensively characterize the overfitting problem. We justify this design because our results consistently demonstrate that SVD-MBR successfully mitigates metric overfitting across this diverse variables, thereby establishing the robustness of our method. Future work may build upon this framework to explore whether alternative generation methods or denoising algorithms, such as NMF, provide complementary benefits.

\section*{Ethical Considerations}

This study fully complies with the ACL Ethics Policy and addresses all relevant items in the Responsible Research Checklist. All resources used in this work are publicly available and properly licensed, with no concerns regarding licensing. The study does not involve or produce any harmful content.
Although AI assistants were utilized for minor writing support, such as rephrasing and spell-checking, all original content was manually created by the authors.
Given these points, we affirm that this work raises no ethical concerns.

\section*{Acknowledgment}

We thank Kazuki Hayashi for reviewing the initial draft of this manuscript and providing valuable feedback on the structure of the paper from a fresh perspective. We also thank the anonymous reviewers for their constructive feedback and insightful suggestions that helped improve the final version of this paper. This work was partly supported by JSPS Kakenhi Grant Number JP24K02993, JP25K24369, and JP26H02537.

\bibliography{custom}

\appendix

\section{MBR vs Normalized MBR}
\label{appendix:mbr_normed_mbr}

In our methodology, we preprocess the pairwise utility matrix using z-score normalization. Crucially, this standardization is performed element-wise (globally) across the entire $N \times M$ matrix for each source document, rather than row-wise or column-wise. The choice of the normalization axis fundamentally impacts the MBR objective: (1) Row-wise normalization (per-hypothesis) forces every row to have a mean of 0, which destroys the MBR ranking by mathematically forcing all hypotheses to tie; (2) Column-wise normalization (per-reference) centers the data per column, which alters the voting power of the pseudo-references and shifts the objective toward a maximum-variance selection similar to Principal Component Analysis (PCA); (3) Global normalization utilizes a single scalar mean ($\mu$) and standard deviation ($\sigma$) across the entire matrix.

The element-wise transformation is calculated as:
\begin{equation}
    z_{i,j} = \frac{m_{i,j} - \mu}{\sigma}.
\end{equation}
When calculating the MBR row mean on this globally standardized matrix (without applying SVD), the decision rule simplifies to:
\begin{equation}
\label{eq:normed_mbr}
    \arg \max_i \left( \frac{\bar{m}_i - \mu}{\sigma} \right).
\end{equation}
Because $\mu$ and $\sigma$ are global constants for a given matrix, this transformation is perfectly monotonic. It perfectly preserves the relative expected utility and the relative distances between candidate scores, ensuring that the fundamental MBR consensus objective remains mathematically unchanged.

\begin{figure}[!t]
    \centering
    \includegraphics[width=1.0\linewidth]{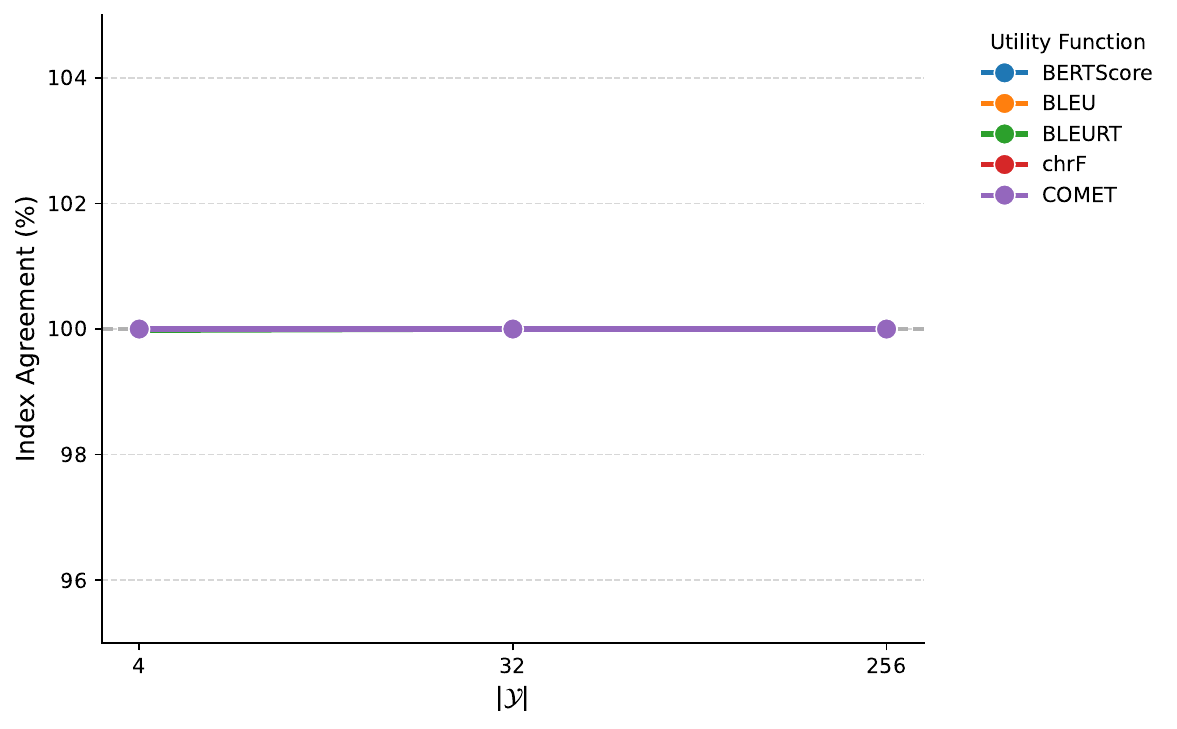}
    \caption{Selected sentence agreement between MBR and Normalized MBR. The grey line indicates 100\% threshold.}
    \label{fig:norm_mbr_agree}
\end{figure}

To empirically verify that global normalization does not alter the hypothesis selection, we compared the selected indices of the best candidates chosen by standard MBR against those chosen by Normalized MBR (Equation~\ref{eq:normed_mbr}, without SVD). We evaluated this on the WMT-22 En$\rightarrow$De dataset across varying reference pool sizes ($|\mathcal{Y}| \in \{4, 32, 256\}$). As visualized in Figure~\ref{fig:norm_mbr_agree}, the results demonstrate a 100\% index agreement between the two methods across all utility functions. This confirms that the step strictly standardizes the scale of the metric scores without altering the underlying MBR ranking prior to rank truncation.

\section{MBR Matrix is Low-Rank}
\label{appendix:mbr_low_rank}

We extracted and retained the generated singular values during the Singular Value Decomposition (SVD) of the pairwise utility matrix for each source document. To quantify the information captured by each singular value, we first average these values across all documents in the dataset. We then calculate the explained variance for each component index $k$ using the squared averaged singular values, normalized by the total sum of the squared values:
\begin{equation}
  V_k=\frac{\bar{\sigma}_k^2}{\sum_{i}\bar{\sigma}_i^2}  
\end{equation}
where $\bar{\sigma}$ represents the averaged singular values.

\begin{figure}[!t]
    \centering
    \includegraphics[width=1.0\linewidth]{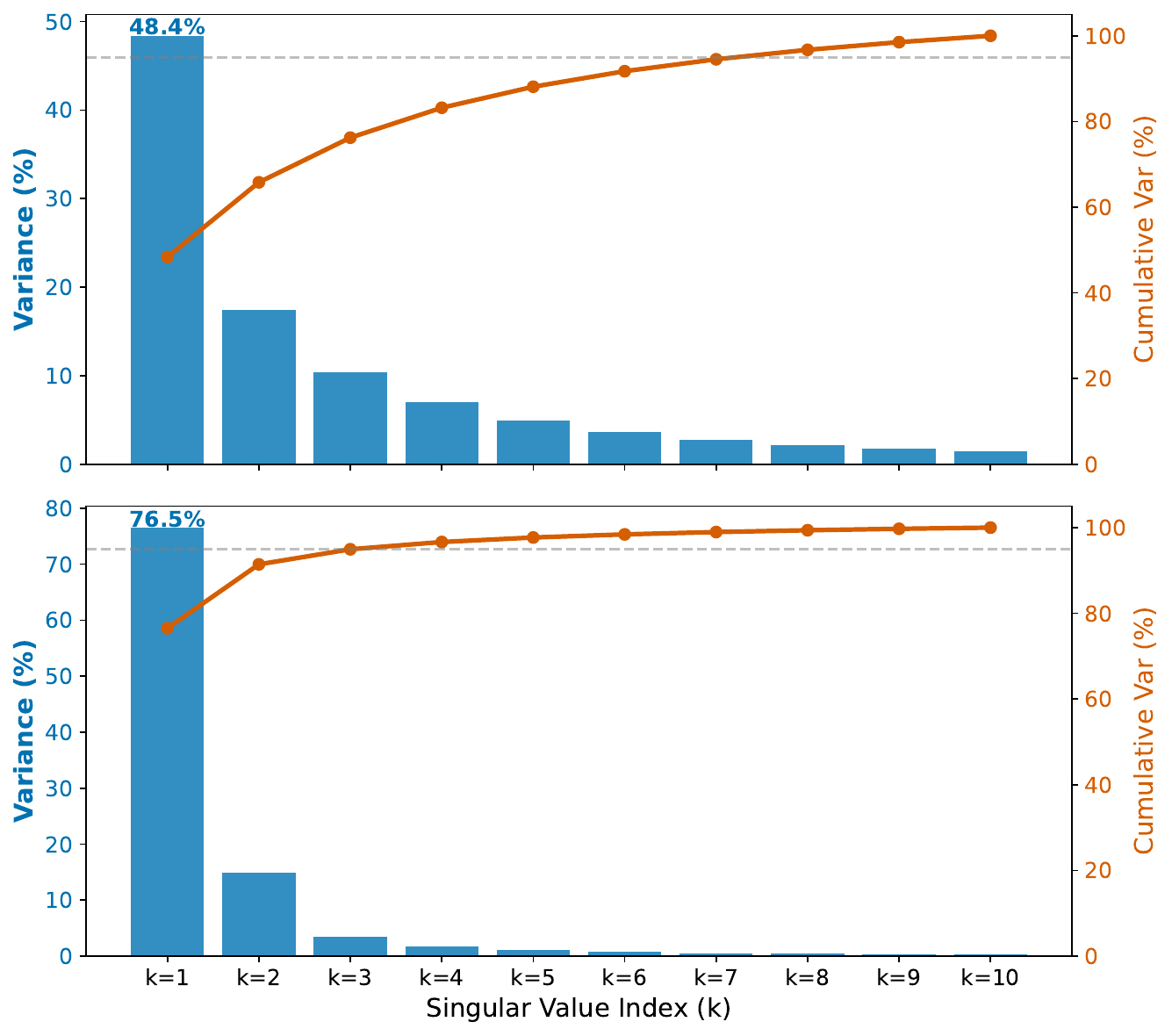}
    \caption{Cumulative variance and individual variance explained by the Top-10 singular values ($k$) for BLEU (top) and COMET (bottom) on WMT22 En$\rightarrow$De. The grey line indicates 95\% threshold.}
    \label{fig:singular_variance}
\end{figure}

For this analysis, we specifically examine matrices generated with a reference pool size of $\vert{}\mathcal{Y}\vert{}=256$. Figure \ref{fig:singular_variance} visualizes the explained variance for the BLEU and COMET utility functions on the WMT22-EnDe dataset. We selected these as representative examples of lexical and neural metrics, respectively; other evaluated metrics exhibit similar structural patterns.

As illustrated in Figure \ref{fig:singular_variance}, \textbf{the variance is heavily concentrated in the first singular value ($k=1$)}, which accounts for 48.4\% of the variance for BLEU and 76.5\% for COMET. This dominant initial peak is followed by a steep decrease at $k=2$, after which the explained variance diminishes gradually and flattens out.

This extreme concentration of variance empirically confirms that the \textbf{MBR pairwise matrix is inherently low-rank}. Consequently, we restrict our tested truncation ranks to small values ($k \in \{1, 2, 3\}$). Retaining components beyond $k=3$ captures negligible additional variance and expanding the rank beyond this threshold simply reintroduces the high-frequency metric noise responsible for overfitting, validating our design choice to aggressively truncate the matrix to isolate the true semantic consensus.

\section{Detailed Results}
\label{appendix:expand_result}

\subsection{Error Analysis}
\label{appendix:error_analysis}

\begin{table*}[ht!]
\centering
\resizebox{\textwidth}{!}{
\begin{NiceTabular}{@{} lll c c c c c c c c c c c c c c @{}}
\toprule
\multirow{2}{*}{\textbf{Util. Function}} & \multirow{2}{*}{\textbf{$|\mathcal{Y}|$}} & \multirow{2}{*}{\textbf{Top-$k$}} & \multicolumn{2}{c}{\makecell{\textbf{BLEU} \\ ($\sigma^2=79.972$)}} & \multicolumn{2}{c}{\makecell{\textbf{chrF} \\ ($\sigma^2=70.045$)}} & \multicolumn{2}{c}{\makecell{\textbf{BLEURT} \\ ($\sigma^2=71.421$)}} & \multicolumn{2}{c}{\makecell{\textbf{COMET} \\ ($\sigma^2=94.773$)}} & \multicolumn{2}{c}{\makecell{\textbf{BERTScore} \\ ($\sigma^2=12.698$)}} & \multicolumn{2}{c}{\makecell{\textbf{COMETKiwi} \\ ($\sigma^2=87.765$)}} & \multicolumn{2}{c}{\makecell{\textbf{$\bar{Z}_{\text{other}}$} \\ ($\sigma^2=0.186$)}} \\
\cmidrule(lr){4-5} \cmidrule(lr){6-7} \cmidrule(lr){8-9} \cmidrule(lr){10-11} \cmidrule(lr){12-13} \cmidrule(lr){14-15} \cmidrule(lr){16-17}
 &  &  & \textit{Dir.} & \textit{Mag.} & \textit{Dir.} & \textit{Mag.} & \textit{Dir.} & \textit{Mag.} & \textit{Dir.} & \textit{Mag.} & \textit{Dir.} & \textit{Mag.} & \textit{Dir.} & \textit{Mag.} & \textit{Dir.} & \textit{Mag.} \\
\midrule
\multirow{9}{*}{\textbf{BLEU}} & \multirow{3}{*}{4} & $k=1$ & \cellcolor{green!3} $+0.02$ & \cellcolor{blue!15} $0.10^{*}$ & \cellcolor{green!2} $+0.02$ & $-0.01$ & \cellcolor{green!2} $+0.02$ & \cellcolor{blue!1} $0.01$ & $+0.00$ & \cellcolor{blue!4} $0.03$ & \cellcolor{green!3} $+0.02$ & $-0.03$ & \cellcolor{red!2} $-0.01$ & $-0.01$ & \cellcolor{green!3} $+0.02$ & $-0.03$ \\
 &  & $k=2$ & \cellcolor{red!5} $-0.04$ & \cellcolor{blue!7} $0.05^{*}$ & \cellcolor{red!3} $-0.02$ & \cellcolor{blue!8} $0.06^{*}$ & \cellcolor{red!2} $-0.01$ & \cellcolor{blue!7} $0.05^{*}$ & \cellcolor{green!1} $+0.01$ & \cellcolor{blue!11} $0.08^{*}$ & \cellcolor{red!2} $-0.02$ & \cellcolor{blue!6} $0.05^{*}$ & $+0.01$ & \cellcolor{blue!9} $0.06^{*}$ & $-0.01$ & \cellcolor{blue!7} $0.05^{*}$ \\
 &  & $k=3$ & \cellcolor{red!4} $-0.03$ & \cellcolor{blue!1} $0.01$ & \cellcolor{red!6} $-0.05^{*}$ & \cellcolor{blue!5} $0.04$ & $-0.01$ & $-0.01$ & $-0.01$ & \cellcolor{blue!2} $0.01$ & \cellcolor{red!3} $-0.02$ & \cellcolor{blue!2} $0.02$ & \cellcolor{red!1} $-0.01$ & \cellcolor{blue!3} $0.03$ & \cellcolor{red!4} $-0.03$ & \cellcolor{blue!2} $0.02$ \\
\hhline{~----------------}
 & \multirow{3}{*}{32} & $k=1$ & \cellcolor{green!3} $+0.02$ & \cellcolor{blue!8} $0.06^{*}$ & \cellcolor{green!2} $+0.02$ & $0.01$ & $-0.01$ & \cellcolor{blue!4} $0.03$ & $-0.01$ & \cellcolor{blue!8} $0.06^{*}$ & \cellcolor{red!2} $-0.02$ & $-0.00$ & $+0.00$ & \cellcolor{blue!6} $0.04$ & \cellcolor{green!1} $+0.01$ & \cellcolor{blue!2} $0.02$ \\
 &  & $k=2$ & \cellcolor{red!4} $-0.03$ & \cellcolor{blue!5} $0.04$ & \cellcolor{red!1} $-0.01$ & \cellcolor{blue!6} $0.04^{*}$ & $+0.00$ & \cellcolor{blue!13} $0.09^{*}$ & $+0.00$ & \cellcolor{blue!16} $0.11^{*}$ & $+0.00$ & \cellcolor{blue!11} $0.08^{*}$ & \cellcolor{red!4} $-0.03$ & \cellcolor{blue!13} $0.09^{*}$ & \cellcolor{red!1} $-0.01$ & \cellcolor{blue!12} $0.08^{*}$ \\
 &  & $k=3$ & \cellcolor{red!2} $-0.02$ & \cellcolor{blue!3} $0.02$ & \cellcolor{red!3} $-0.02$ & \cellcolor{blue!7} $0.05^{*}$ & \cellcolor{red!5} $-0.03$ & \cellcolor{blue!14} $0.10^{*}$ & $+0.01$ & \cellcolor{blue!20} $0.14^{*}$ & $+0.00$ & \cellcolor{blue!14} $0.10^{*}$ & \cellcolor{red!3} $-0.02$ & \cellcolor{blue!19} $0.13^{*}$ & \cellcolor{red!3} $-0.02$ & \cellcolor{blue!16} $0.11^{*}$ \\
\hhline{~----------------}
 & \multirow{3}{*}{256} & $k=1$ & \cellcolor{green!1} $+0.01$ & \cellcolor{blue!4} $0.03$ & \cellcolor{red!3} $-0.02$ & $-0.02$ & $+0.00$ & \cellcolor{blue!5} $0.03$ & \cellcolor{green!1} $+0.01$ & \cellcolor{blue!8} $0.06^{*}$ & $+0.00$ & \cellcolor{blue!6} $0.04$ & \cellcolor{green!2} $+0.01$ & \cellcolor{blue!4} $0.03$ & $+0.00$ & \cellcolor{blue!2} $0.02$ \\
 &  & $k=2$ & \cellcolor{green!2} $+0.01$ & $-0.01$ & $+0.00$ & \cellcolor{blue!2} $0.02$ & $-0.01$ & \cellcolor{blue!8} $0.06^{*}$ & \cellcolor{red!1} $-0.01$ & \cellcolor{blue!8} $0.05^{*}$ & \cellcolor{green!3} $+0.02$ & \cellcolor{blue!10} $0.07^{*}$ & \cellcolor{red!3} $-0.02$ & \cellcolor{blue!4} $0.03$ & \cellcolor{red!3} $-0.02$ & \cellcolor{blue!8} $0.05^{*}$ \\
 &  & $k=3$ & \cellcolor{green!3} $+0.02$ & $-0.03$ & \cellcolor{red!2} $-0.01$ & \cellcolor{blue!8} $0.06^{*}$ & \cellcolor{red!4} $-0.03$ & \cellcolor{blue!14} $0.10^{*}$ & \cellcolor{red!5} $-0.04$ & \cellcolor{blue!13} $0.09^{*}$ & \cellcolor{green!2} $+0.02$ & \cellcolor{blue!15} $0.11^{*}$ & \cellcolor{red!1} $-0.01$ & \cellcolor{blue!12} $0.09^{*}$ & $-0.01$ & \cellcolor{blue!15} $0.11^{*}$ \\
\midrule
\multirow{9}{*}{\textbf{chrF}} & \multirow{3}{*}{4} & $k=1$ & \cellcolor{green!1} $+0.01$ & \cellcolor{blue!16} $0.11^{*}$ & $+0.00$ & \cellcolor{blue!16} $0.11^{*}$ & \cellcolor{red!7} $-0.05^{*}$ & \cellcolor{blue!16} $0.11^{*}$ & \cellcolor{red!4} $-0.03$ & \cellcolor{blue!13} $0.09^{*}$ & \cellcolor{red!2} $-0.02$ & \cellcolor{blue!13} $0.09^{*}$ & \cellcolor{red!2} $-0.02$ & \cellcolor{blue!8} $0.06^{*}$ & $+0.01$ & \cellcolor{blue!11} $0.08^{*}$ \\
 &  & $k=2$ & \cellcolor{red!1} $-0.01$ & \cellcolor{blue!7} $0.05^{*}$ & \cellcolor{green!3} $+0.02$ & \cellcolor{blue!11} $0.08^{*}$ & \cellcolor{red!2} $-0.02$ & \cellcolor{blue!11} $0.08^{*}$ & $+0.01$ & \cellcolor{blue!10} $0.07^{*}$ & $+0.00$ & \cellcolor{blue!10} $0.07^{*}$ & \cellcolor{green!1} $+0.01$ & \cellcolor{blue!10} $0.07^{*}$ & \cellcolor{green!1} $+0.01$ & \cellcolor{blue!8} $0.06^{*}$ \\
 &  & $k=3$ & \cellcolor{green!3} $+0.02$ & $-0.02$ & $-0.01$ & \cellcolor{blue!4} $0.03$ & \cellcolor{red!3} $-0.02$ & \cellcolor{blue!9} $0.06^{*}$ & \cellcolor{green!7} $+0.05^{*}$ & \cellcolor{blue!10} $0.07^{*}$ & $+0.00$ & \cellcolor{blue!4} $0.03$ & \cellcolor{green!1} $+0.01$ & \cellcolor{blue!11} $0.08^{*}$ & \cellcolor{green!1} $+0.01$ & \cellcolor{blue!5} $0.03$ \\
\hhline{~----------------}
 & \multirow{3}{*}{32} & $k=1$ & \cellcolor{red!1} $-0.01$ & \cellcolor{blue!9} $0.06^{*}$ & $+0.00$ & \cellcolor{blue!16} $0.11^{*}$ & \cellcolor{red!2} $-0.02$ & \cellcolor{blue!14} $0.10^{*}$ & $+0.00$ & \cellcolor{blue!14} $0.10^{*}$ & \cellcolor{red!5} $-0.04$ & \cellcolor{blue!16} $0.11^{*}$ & \cellcolor{red!2} $-0.01$ & \cellcolor{blue!14} $0.10^{*}$ & \cellcolor{green!1} $+0.01$ & \cellcolor{blue!12} $0.08^{*}$ \\
 &  & $k=2$ & \cellcolor{red!5} $-0.04$ & \cellcolor{blue!3} $0.02$ & \cellcolor{red!3} $-0.02$ & \cellcolor{blue!18} $0.12^{*}$ & \cellcolor{green!3} $+0.02$ & \cellcolor{blue!11} $0.08^{*}$ & \cellcolor{green!1} $+0.01$ & \cellcolor{blue!13} $0.09^{*}$ & \cellcolor{red!1} $-0.01$ & \cellcolor{blue!18} $0.13^{*}$ & $+0.00$ & \cellcolor{blue!17} $0.12^{*}$ & \cellcolor{green!1} $+0.01$ & \cellcolor{blue!12} $0.09^{*}$ \\
 &  & $k=3$ & \cellcolor{red!7} $-0.05^{*}$ & \cellcolor{blue!1} $0.01$ & \cellcolor{red!3} $-0.02$ & \cellcolor{blue!11} $0.08^{*}$ & \cellcolor{red!3} $-0.02$ & \cellcolor{blue!9} $0.06^{*}$ & \cellcolor{green!4} $+0.03$ & \cellcolor{blue!8} $0.06^{*}$ & \cellcolor{red!6} $-0.04^{*}$ & \cellcolor{blue!15} $0.10^{*}$ & \cellcolor{green!1} $+0.01$ & \cellcolor{blue!10} $0.07^{*}$ & \cellcolor{red!2} $-0.02$ & \cellcolor{blue!10} $0.07^{*}$ \\
\hhline{~----------------}
 & \multirow{3}{*}{256} & $k=1$ & \cellcolor{green!4} $+0.03$ & \cellcolor{blue!7} $0.05^{*}$ & \cellcolor{green!6} $+0.04$ & \cellcolor{blue!14} $0.10^{*}$ & \cellcolor{green!2} $+0.02$ & \cellcolor{blue!21} $0.14^{*}$ & \cellcolor{green!1} $+0.01$ & \cellcolor{blue!22} $0.15^{*}$ & \cellcolor{green!2} $+0.02$ & \cellcolor{blue!23} $0.15^{*}$ & \cellcolor{green!4} $+0.03$ & \cellcolor{blue!23} $0.15^{*}$ & \cellcolor{green!9} $+0.06^{*}$ & \cellcolor{blue!16} $0.11^{*}$ \\
 &  & $k=2$ & \cellcolor{red!1} $-0.01$ & \cellcolor{blue!6} $0.04^{*}$ & \cellcolor{green!1} $+0.01$ & \cellcolor{blue!20} $0.14^{*}$ & \cellcolor{green!7} $+0.05^{*}$ & \cellcolor{blue!22} $0.15^{*}$ & \cellcolor{green!4} $+0.03$ & \cellcolor{blue!19} $0.13^{*}$ & \cellcolor{green!6} $+0.04$ & \cellcolor{blue!21} $0.14^{*}$ & \cellcolor{green!9} $+0.06^{*}$ & \cellcolor{blue!21} $0.14^{*}$ & \cellcolor{green!10} $+0.07^{*}$ & \cellcolor{blue!18} $0.12^{*}$ \\
 &  & $k=3$ & $+0.00$ & \cellcolor{blue!6} $0.04^{*}$ & \cellcolor{red!2} $-0.02$ & \cellcolor{blue!14} $0.09^{*}$ & \cellcolor{green!2} $+0.02$ & \cellcolor{blue!14} $0.10^{*}$ & \cellcolor{green!2} $+0.02$ & \cellcolor{blue!14} $0.10^{*}$ & $+0.00$ & \cellcolor{blue!16} $0.11^{*}$ & \cellcolor{green!1} $+0.01$ & \cellcolor{blue!14} $0.10^{*}$ & \cellcolor{green!5} $+0.04$ & \cellcolor{blue!11} $0.08^{*}$ \\
\midrule
\multirow{9}{*}{\textbf{BLEURT}} & \multirow{3}{*}{4} & $k=1$ & \cellcolor{green!4} $+0.03$ & \cellcolor{blue!9} $0.07^{*}$ & \cellcolor{green!1} $+0.01$ & \cellcolor{blue!24} $0.16^{*}$ & \cellcolor{red!7} $-0.05^{*}$ & \cellcolor{blue!17} $0.12^{*}$ & \cellcolor{red!1} $-0.01$ & \cellcolor{blue!13} $0.09^{*}$ & $+0.00$ & \cellcolor{blue!24} $0.16^{*}$ & $+0.00$ & \cellcolor{blue!20} $0.14^{*}$ & $+0.00$ & \cellcolor{blue!19} $0.13^{*}$ \\
 &  & $k=2$ & \cellcolor{green!4} $+0.03$ & \cellcolor{blue!2} $0.02$ & \cellcolor{green!4} $+0.03$ & \cellcolor{blue!16} $0.11^{*}$ & \cellcolor{red!4} $-0.03$ & \cellcolor{blue!10} $0.07^{*}$ & \cellcolor{red!1} $-0.01$ & \cellcolor{blue!10} $0.07^{*}$ & \cellcolor{green!2} $+0.02$ & \cellcolor{blue!13} $0.09^{*}$ & \cellcolor{green!1} $+0.01$ & \cellcolor{blue!13} $0.09^{*}$ & $+0.00$ & \cellcolor{blue!13} $0.09^{*}$ \\
 &  & $k=3$ & \cellcolor{green!2} $+0.02$ & \cellcolor{blue!5} $0.04$ & \cellcolor{green!2} $+0.02$ & \cellcolor{blue!7} $0.05^{*}$ & \cellcolor{green!5} $+0.04$ & \cellcolor{blue!1} $0.01$ & \cellcolor{green!1} $+0.01$ & \cellcolor{blue!1} $0.01$ & \cellcolor{green!4} $+0.03$ & \cellcolor{blue!6} $0.05^{*}$ & \cellcolor{green!3} $+0.02$ & \cellcolor{blue!3} $0.03$ & \cellcolor{green!5} $+0.04$ & \cellcolor{blue!5} $0.03$ \\
\hhline{~----------------}
 & \multirow{3}{*}{32} & $k=1$ & \cellcolor{green!11} $+0.08^{*}$ & \cellcolor{blue!18} $0.12^{*}$ & \cellcolor{green!2} $+0.02$ & \cellcolor{blue!33} $0.22^{*}$ & \cellcolor{green!2} $+0.01$ & \cellcolor{blue!16} $0.11^{*}$ & \cellcolor{green!2} $+0.02$ & \cellcolor{blue!17} $0.12^{*}$ & \cellcolor{red!1} $-0.01$ & \cellcolor{blue!34} $0.23^{*}$ & $+0.00$ & \cellcolor{blue!24} $0.16^{*}$ & \cellcolor{green!3} $+0.03$ & \cellcolor{blue!31} $0.21^{*}$ \\
 &  & $k=2$ & \cellcolor{green!6} $+0.04$ & \cellcolor{blue!8} $0.06^{*}$ & \cellcolor{green!1} $+0.01$ & \cellcolor{blue!21} $0.14^{*}$ & \cellcolor{green!5} $+0.03$ & \cellcolor{blue!15} $0.10^{*}$ & \cellcolor{green!6} $+0.04^{*}$ & \cellcolor{blue!10} $0.07^{*}$ & $+0.00$ & \cellcolor{blue!24} $0.17^{*}$ & \cellcolor{green!4} $+0.03$ & \cellcolor{blue!13} $0.09^{*}$ & \cellcolor{green!2} $+0.02$ & \cellcolor{blue!20} $0.13^{*}$ \\
 &  & $k=3$ & \cellcolor{green!3} $+0.02$ & \cellcolor{blue!7} $0.05^{*}$ & \cellcolor{green!3} $+0.02$ & \cellcolor{blue!14} $0.10^{*}$ & \cellcolor{red!1} $-0.01$ & \cellcolor{blue!6} $0.05^{*}$ & \cellcolor{green!2} $+0.02$ & \cellcolor{blue!3} $0.03$ & \cellcolor{red!1} $-0.01$ & \cellcolor{blue!16} $0.11^{*}$ & $+0.00$ & \cellcolor{blue!8} $0.06^{*}$ & \cellcolor{green!2} $+0.02$ & \cellcolor{blue!14} $0.10^{*}$ \\
\hhline{~----------------}
 & \multirow{3}{*}{256} & $k=1$ & \cellcolor{green!9} $+0.06^{*}$ & \cellcolor{blue!14} $0.10^{*}$ & \cellcolor{green!6} $+0.04$ & \cellcolor{blue!33} $0.23^{*}$ & \cellcolor{red!5} $-0.04$ & \cellcolor{blue!20} $0.14^{*}$ & \cellcolor{green!7} $+0.05^{*}$ & \cellcolor{blue!18} $0.13^{*}$ & $+0.01$ & \cellcolor{blue!33} $0.22^{*}$ & \cellcolor{green!5} $+0.04$ & \cellcolor{blue!24} $0.16^{*}$ & \cellcolor{green!5} $+0.04$ & \cellcolor{blue!29} $0.20^{*}$ \\
 &  & $k=2$ & \cellcolor{green!4} $+0.03$ & \cellcolor{blue!5} $0.04$ & \cellcolor{green!2} $+0.01$ & \cellcolor{blue!19} $0.13^{*}$ & $+0.00$ & \cellcolor{blue!16} $0.11^{*}$ & \cellcolor{red!3} $-0.02$ & \cellcolor{blue!6} $0.05^{*}$ & \cellcolor{red!2} $-0.01$ & \cellcolor{blue!22} $0.15^{*}$ & \cellcolor{green!1} $+0.01$ & \cellcolor{blue!14} $0.10^{*}$ & \cellcolor{green!5} $+0.03$ & \cellcolor{blue!15} $0.10^{*}$ \\
 &  & $k=3$ & \cellcolor{green!4} $+0.03$ & \cellcolor{blue!8} $0.06^{*}$ & \cellcolor{green!2} $+0.02$ & \cellcolor{blue!13} $0.09^{*}$ & \cellcolor{red!1} $-0.01$ & \cellcolor{blue!8} $0.06^{*}$ & \cellcolor{green!1} $+0.01$ & \cellcolor{blue!4} $0.03$ & \cellcolor{red!2} $-0.02$ & \cellcolor{blue!16} $0.11^{*}$ & \cellcolor{green!4} $+0.03$ & \cellcolor{blue!8} $0.06^{*}$ & \cellcolor{green!3} $+0.02$ & \cellcolor{blue!14} $0.10^{*}$ \\
\midrule
\multirow{9}{*}{\textbf{COMET}} & \multirow{3}{*}{4} & $k=1$ & \cellcolor{green!14} $+0.10^{*}$ & \cellcolor{blue!51} $0.34^{*}$ & \cellcolor{green!6} $+0.04^{*}$ & \cellcolor{blue!38} $0.26^{*}$ & $-0.01$ & \cellcolor{blue!12} $0.08$ & \cellcolor{green!1} $+0.01$ & $-0.07$ & \cellcolor{green!4} $+0.03$ & \cellcolor{blue!28} $0.19^{*}$ & \cellcolor{green!1} $+0.01$ & $-0.02^{*}$ & \cellcolor{green!7} $+0.05$ & \cellcolor{blue!29} $0.20^{*}$ \\
 &  & $k=2$ & \cellcolor{green!18} $+0.12$ & \cellcolor{blue!36} $0.24^{*}$ & \cellcolor{green!6} $+0.05$ & \cellcolor{blue!30} $0.20^{*}$ & \cellcolor{red!2} $-0.02$ & \cellcolor{blue!11} $0.08$ & \cellcolor{red!5} $-0.04$ & $-0.02$ & \cellcolor{green!3} $+0.02$ & \cellcolor{blue!27} $0.18^{*}$ & \cellcolor{red!1} $-0.01$ & \cellcolor{blue!4} $0.03$ & \cellcolor{green!7} $+0.05$ & \cellcolor{blue!25} $0.17^{*}$ \\
 &  & $k=3$ & \cellcolor{green!15} $+0.10$ & \cellcolor{blue!35} $0.24^{*}$ & \cellcolor{green!7} $+0.05$ & \cellcolor{blue!29} $0.20^{*}$ & \cellcolor{red!1} $-0.01$ & \cellcolor{blue!9} $0.07$ & \cellcolor{red!6} $-0.04$ & $-0.02$ & \cellcolor{green!2} $+0.02$ & \cellcolor{blue!24} $0.16^{*}$ & \cellcolor{red!2} $-0.01$ & \cellcolor{blue!3} $0.03$ & \cellcolor{green!6} $+0.04$ & \cellcolor{blue!22} $0.15^{*}$ \\
\hhline{~----------------}
 & \multirow{3}{*}{32} & $k=1$ & \cellcolor{red!5} $-0.04$ & \cellcolor{blue!20} $0.14$ & \cellcolor{red!7} $-0.05$ & \cellcolor{blue!24} $0.16^{*}$ & \cellcolor{green!2} $+0.02$ & \cellcolor{blue!8} $0.06$ & \cellcolor{green!14} $+0.10$ & $-0.06$ & \cellcolor{red!5} $-0.03$ & \cellcolor{blue!22} $0.15^{*}$ & \cellcolor{green!9} $+0.07$ & $-0.02^{*}$ & \cellcolor{red!3} $-0.03$ & \cellcolor{blue!21} $0.14^{*}$ \\
 &  & $k=2$ & \cellcolor{red!3} $-0.02$ & \cellcolor{blue!19} $0.13$ & \cellcolor{red!4} $-0.03$ & \cellcolor{blue!19} $0.13$ & $+0.00$ & \cellcolor{blue!10} $0.07$ & \cellcolor{green!1} $+0.01$ & \cellcolor{blue!1} $0.01$ & $+0.00$ & \cellcolor{blue!19} $0.13^{*}$ & \cellcolor{green!5} $+0.04$ & \cellcolor{blue!4} $0.03$ & $-0.01$ & \cellcolor{blue!20} $0.14^{*}$ \\
 &  & $k=3$ & \cellcolor{red!4} $-0.03$ & \cellcolor{blue!19} $0.13$ & \cellcolor{red!5} $-0.04$ & \cellcolor{blue!17} $0.12$ & $+0.00$ & \cellcolor{blue!9} $0.06$ & \cellcolor{green!1} $+0.01$ & \cellcolor{blue!1} $0.01$ & \cellcolor{red!2} $-0.01$ & \cellcolor{blue!17} $0.11$ & \cellcolor{green!3} $+0.02$ & \cellcolor{blue!3} $0.02$ & \cellcolor{red!2} $-0.02$ & \cellcolor{blue!18} $0.12$ \\
\hhline{~----------------}
 & \multirow{3}{*}{256} & $k=1$ & \cellcolor{green!17} $+0.12$ & \cellcolor{blue!54} $0.36^{*}$ & \cellcolor{green!13} $+0.09$ & \cellcolor{blue!42} $0.28^{*}$ & $+0.00$ & \cellcolor{blue!15} $0.10^{*}$ & \cellcolor{green!2} $+0.01^{*}$ & $-0.11^{*}$ & \cellcolor{green!4} $+0.03^{*}$ & \cellcolor{blue!31} $0.21^{*}$ & \cellcolor{red!1} $-0.01^{*}$ & $-0.07$ & \cellcolor{green!6} $+0.05^{*}$ & \cellcolor{blue!36} $0.25^{*}$ \\
 &  & $k=2$ & \cellcolor{green!19} $+0.13^{*}$ & \cellcolor{blue!30} $0.20$ & \cellcolor{green!17} $+0.11^{*}$ & \cellcolor{blue!24} $0.16^{*}$ & $-0.01$ & \cellcolor{blue!6} $0.05$ & \cellcolor{red!7} $-0.05$ & $-0.05$ & \cellcolor{green!9} $+0.06^{*}$ & \cellcolor{blue!15} $0.10$ & \cellcolor{red!5} $-0.04^{*}$ & $-0.03$ & \cellcolor{green!11} $+0.08^{*}$ & \cellcolor{blue!20} $0.14$ \\
 &  & $k=3$ & \cellcolor{green!17} $+0.12$ & \cellcolor{blue!30} $0.21$ & \cellcolor{green!15} $+0.10$ & \cellcolor{blue!24} $0.16^{*}$ & \cellcolor{red!1} $-0.01$ & \cellcolor{blue!8} $0.05$ & \cellcolor{red!10} $-0.07$ & $-0.05$ & \cellcolor{green!8} $+0.06^{*}$ & \cellcolor{blue!16} $0.11^{*}$ & \cellcolor{red!7} $-0.05$ & $-0.03$ & \cellcolor{green!9} $+0.07$ & \cellcolor{blue!21} $0.15^{*}$ \\
\midrule
\multirow{9}{*}{\textbf{BERTScore}} & \multirow{3}{*}{4} & $k=1$ & $+0.00$ & \cellcolor{blue!15} $0.10^{*}$ & \cellcolor{green!6} $+0.05^{*}$ & \cellcolor{blue!9} $0.06^{*}$ & \cellcolor{green!3} $+0.02$ & \cellcolor{blue!14} $0.09^{*}$ & $+0.00$ & \cellcolor{blue!14} $0.10^{*}$ & \cellcolor{green!2} $+0.02$ & \cellcolor{blue!13} $0.09^{*}$ & $+0.01$ & \cellcolor{blue!15} $0.10^{*}$ & \cellcolor{green!3} $+0.02$ & \cellcolor{blue!12} $0.08^{*}$ \\
 &  & $k=2$ & \cellcolor{red!2} $-0.02$ & \cellcolor{blue!5} $0.04$ & \cellcolor{green!2} $+0.02$ & \cellcolor{blue!9} $0.06^{*}$ & \cellcolor{green!4} $+0.03$ & \cellcolor{blue!6} $0.04$ & \cellcolor{green!4} $+0.03$ & \cellcolor{blue!8} $0.06^{*}$ & \cellcolor{red!3} $-0.02$ & \cellcolor{blue!12} $0.08^{*}$ & \cellcolor{green!11} $+0.08^{*}$ & \cellcolor{blue!13} $0.09^{*}$ & \cellcolor{green!2} $+0.02$ & \cellcolor{blue!7} $0.05^{*}$ \\
 &  & $k=3$ & \cellcolor{red!1} $-0.01$ & $-0.04$ & \cellcolor{green!5} $+0.04$ & $0.00$ & $+0.00$ & \cellcolor{blue!5} $0.03$ & \cellcolor{green!4} $+0.03$ & \cellcolor{blue!3} $0.02$ & $+0.00$ & \cellcolor{blue!7} $0.05^{*}$ & \cellcolor{green!1} $+0.01$ & \cellcolor{blue!4} $0.03$ & $+0.00$ & $-0.00$ \\
\hhline{~----------------}
 & \multirow{3}{*}{32} & $k=1$ & \cellcolor{red!4} $-0.03$ & \cellcolor{blue!18} $0.12^{*}$ & \cellcolor{green!3} $+0.02$ & \cellcolor{blue!20} $0.14^{*}$ & \cellcolor{red!1} $-0.01$ & \cellcolor{blue!26} $0.18^{*}$ & \cellcolor{green!3} $+0.02$ & \cellcolor{blue!25} $0.17^{*}$ & \cellcolor{green!4} $+0.03$ & \cellcolor{blue!25} $0.17^{*}$ & \cellcolor{green!6} $+0.04$ & \cellcolor{blue!24} $0.16^{*}$ & \cellcolor{green!2} $+0.02$ & \cellcolor{blue!21} $0.14^{*}$ \\
 &  & $k=2$ & \cellcolor{red!3} $-0.02$ & \cellcolor{blue!8} $0.06^{*}$ & \cellcolor{green!3} $+0.03$ & \cellcolor{blue!20} $0.14^{*}$ & \cellcolor{red!4} $-0.03$ & \cellcolor{blue!20} $0.14^{*}$ & \cellcolor{green!2} $+0.01$ & \cellcolor{blue!17} $0.12^{*}$ & $+0.00$ & \cellcolor{blue!29} $0.20^{*}$ & \cellcolor{green!4} $+0.03$ & \cellcolor{blue!18} $0.13^{*}$ & \cellcolor{green!1} $+0.01$ & \cellcolor{blue!18} $0.12^{*}$ \\
 &  & $k=3$ & $-0.01$ & \cellcolor{blue!2} $0.02$ & \cellcolor{green!4} $+0.03$ & \cellcolor{blue!12} $0.09^{*}$ & \cellcolor{green!2} $+0.02$ & \cellcolor{blue!15} $0.10^{*}$ & \cellcolor{green!5} $+0.03$ & \cellcolor{blue!15} $0.10^{*}$ & \cellcolor{green!1} $+0.01$ & \cellcolor{blue!22} $0.15^{*}$ & \cellcolor{green!4} $+0.03$ & \cellcolor{blue!16} $0.11^{*}$ & \cellcolor{green!3} $+0.02$ & \cellcolor{blue!13} $0.09^{*}$ \\
\hhline{~----------------}
 & \multirow{3}{*}{256} & $k=1$ & \cellcolor{green!2} $+0.02$ & \cellcolor{blue!10} $0.07^{*}$ & \cellcolor{green!6} $+0.04$ & \cellcolor{blue!16} $0.11^{*}$ & \cellcolor{green!2} $+0.01$ & \cellcolor{blue!21} $0.15^{*}$ & \cellcolor{green!6} $+0.05^{*}$ & \cellcolor{blue!22} $0.15^{*}$ & \cellcolor{green!5} $+0.03$ & \cellcolor{blue!24} $0.16^{*}$ & \cellcolor{green!1} $+0.01$ & \cellcolor{blue!21} $0.14^{*}$ & \cellcolor{green!8} $+0.06^{*}$ & \cellcolor{blue!16} $0.11^{*}$ \\
 &  & $k=2$ & $+0.01$ & \cellcolor{blue!3} $0.02$ & $+0.00$ & \cellcolor{blue!18} $0.12^{*}$ & \cellcolor{green!1} $+0.01$ & \cellcolor{blue!20} $0.14^{*}$ & \cellcolor{green!3} $+0.02$ & \cellcolor{blue!19} $0.13^{*}$ & \cellcolor{red!4} $-0.03$ & \cellcolor{blue!28} $0.19^{*}$ & \cellcolor{green!3} $+0.02$ & \cellcolor{blue!19} $0.13^{*}$ & \cellcolor{green!3} $+0.03$ & \cellcolor{blue!17} $0.12^{*}$ \\
 &  & $k=3$ & \cellcolor{green!1} $+0.01$ & $-0.01$ & \cellcolor{green!4} $+0.03$ & \cellcolor{blue!3} $0.03$ & \cellcolor{green!2} $+0.02$ & \cellcolor{blue!5} $0.03$ & $+0.00$ & \cellcolor{blue!9} $0.06^{*}$ & \cellcolor{green!2} $+0.02$ & \cellcolor{blue!14} $0.10^{*}$ & \cellcolor{green!4} $+0.03$ & \cellcolor{blue!7} $0.05^{*}$ & \cellcolor{green!6} $+0.05^{*}$ & \cellcolor{blue!4} $0.03$ \\
\bottomrule
\end{NiceTabular}
}
\caption{Spearman correlation ($\rho$) between inherent hypotheses oracle variance and evaluation score differences across varying pseudo-reference pool sizes ($|\mathcal{Y}|$) on WMT22 En$\rightarrow$De. The hypothesis pool variance ($\sigma^2$) for each evaluation metric is displayed in the respective column headers. \textit{Dir.} (Directional Change) measures score improvements, highlighted as \colorbox{green!15}{positive} and \colorbox{red!15}{negative} impacts. \textit{Mag.} (Magnitude Sensitivity) measures absolute score variance, with \colorbox{blue!15}{blue} highlighting metric sensitivity. Statistical significance ($p < 0.05$) is denoted by an asterisk ($^*$).}
\label{tab:corr_variance_analysis_complete_wmt22-ende}
\end{table*}

While \S\ref{subsec:error_analysis} establishes the relationship between a utility function's inherent variance~($\sigma^2$) and the generalized performance of SVD-MBR~($\bar{Z}_{\text{other}}$), Table~\ref{tab:corr_variance_analysis_complete_wmt22-ende} allows for a granular inspection of how specific metric families interact under denoising. By tracking the individual directional shifts~(\textit{Dir.}) across each evaluation metric, two nuanced phenomena emerge.

The expanded data reveals that \textbf{utility functions distribute their denoising benefits in fundamentally different ways}. Certain metrics, when regularized, distribute their performance gains broadly and equally across evaluation metrics. For example, utilizing BLEURT as the utility function~($|\mathcal{Y}|=256, k=1$) yields balanced directional improvements across both lexical metrics~(BLEU $+0.06^*$, chrF $+0.04$) and neural metrics~(COMET $+0.05^*$, COMETKiwi $+0.04$). Conversely, other utility functions concentrate their improvements into specific metrics. When optimizing for COMET~($|\mathcal{Y}|=256, k=2$), the denoising does not yield uniform gains. BLEURT ~($-0.01$) is slightly degraded while driving massive improvements in BLEU~($+0.13^*$) and chrF~($+0.11^*$). Furthermore, this behavior exposes a strong intra-family synergy: because COMET and COMETKiwi share foundational architectural and semantic spaces, filtering the noise from the COMET pairwise matrix results in regularization to COMETKiwi scores~($-0.04^*$).

Although the main text groups surface-level metrics together due to their resistance to low-rank approximation, Table~\ref{tab:corr_variance_analysis_complete_wmt22-ende} highlights \textbf{a distinct divergence in behavior between BLEU and chrF}. When BLEU is employed as the utility function, SVD-MBR consistently fails to generate significant directional improvements across any other metric. Even at maximum reference size~($|\mathcal{Y}|=256$), the off-target directional shifts remain stagnant or near zero, reinforcing that BLEU’s strict, sparse $n$-gram matching completely lacks a compressible, low-rank semantic signal. However, chrF exhibits a uniquely different tendency. Likely due to its character-level $n$-gram smoothing, the chrF pairwise matrix retains a slightly more extractable continuous structure. Specifically, at $|\mathcal{Y}|=32$ and $k=1$, employing chrF as the utility function cascades positive directional improvements across the entirety of the off-target neural metric suite~(BLEURT $+0.05^*$, COMET $+0.01$, BERTScore $+0.03$, COMETKiwi $+0.00$). This indicates that while surface-level metrics are generally resistant to SVD, sub-word smoothing algorithms like chrF encode just enough structural consensus to benefit from denoising under optimal rank configurations.

\begin{table*}[ht!]
\centering
\resizebox{\textwidth}{!}{
\begin{NiceTabular}{@{} lll c c c c c c c c c c c c c c @{}}
\toprule
\multirow{2}{*}{\textbf{Util. Function}} & \multirow{2}{*}{\textbf{$|\mathcal{Y}|$}} & \multirow{2}{*}{\textbf{Top-$k$}} & \multicolumn{2}{c}{\textbf{BLEU}} & \multicolumn{2}{c}{\textbf{chrF}} & \multicolumn{2}{c}{\textbf{BLEURT}} & \multicolumn{2}{c}{\textbf{COMET}} & \multicolumn{2}{c}{\textbf{BERTScore}} & \multicolumn{2}{c}{\textbf{COMETKiwi}} & \multicolumn{2}{c}{\textbf{$\bar{Z}_{\text{other}}$}} \\
\cmidrule(lr){4-5} \cmidrule(lr){6-7} \cmidrule(lr){8-9} \cmidrule(lr){10-11} \cmidrule(lr){12-13} \cmidrule(lr){14-15} \cmidrule(lr){16-17}
 &  &  & \textit{Dir.} & \textit{Mag.} & \textit{Dir.} & \textit{Mag.} & \textit{Dir.} & \textit{Mag.} & \textit{Dir.} & \textit{Mag.} & \textit{Dir.} & \textit{Mag.} & \textit{Dir.} & \textit{Mag.} & \textit{Dir.} & \textit{Mag.} \\
\midrule
\multirow{9}{*}{\textbf{BLEU}} & \multirow{3}{*}{4} & $k=1$ & \cellcolor{red!3} $-0.02$ & \cellcolor{blue!37} $0.25^{*}$ & \cellcolor{red!4} $-0.03$ & \cellcolor{blue!34} $0.23^{*}$ & $-0.01$ & \cellcolor{blue!31} $0.21^{*}$ & \cellcolor{green!1} $+0.01$ & \cellcolor{blue!31} $0.21^{*}$ & \cellcolor{red!3} $-0.02$ & \cellcolor{blue!33} $0.22^{*}$ & \cellcolor{green!4} $+0.03$ & \cellcolor{blue!33} $0.22^{*}$ & $+0.01$ & \cellcolor{blue!31} $0.21^{*}$ \\
 &  & $k=2$ & $-0.01$ & \cellcolor{blue!34} $0.23^{*}$ & \cellcolor{red!1} $-0.01$ & \cellcolor{blue!33} $0.22^{*}$ & $+0.01$ & \cellcolor{blue!31} $0.21^{*}$ & $+0.00$ & \cellcolor{blue!32} $0.22^{*}$ & \cellcolor{red!3} $-0.02$ & \cellcolor{blue!32} $0.21^{*}$ & \cellcolor{red!1} $-0.01$ & \cellcolor{blue!33} $0.22^{*}$ & $+0.00$ & \cellcolor{blue!31} $0.21^{*}$ \\
 &  & $k=3$ & \cellcolor{red!7} $-0.05^{*}$ & \cellcolor{blue!30} $0.20^{*}$ & \cellcolor{red!2} $-0.02$ & \cellcolor{blue!29} $0.20^{*}$ & \cellcolor{red!4} $-0.03$ & \cellcolor{blue!29} $0.20^{*}$ & \cellcolor{red!3} $-0.03$ & \cellcolor{blue!28} $0.19^{*}$ & \cellcolor{red!5} $-0.04$ & \cellcolor{blue!28} $0.19^{*}$ & \cellcolor{red!1} $-0.01$ & \cellcolor{blue!29} $0.19^{*}$ & \cellcolor{red!6} $-0.04$ & \cellcolor{blue!28} $0.19^{*}$ \\
\hhline{~----------------}
 & \multirow{3}{*}{32} & $k=1$ & \cellcolor{green!4} $+0.03$ & \cellcolor{blue!31} $0.21^{*}$ & $+0.00$ & \cellcolor{blue!26} $0.18^{*}$ & $+0.01$ & \cellcolor{blue!24} $0.16^{*}$ & $+0.00$ & \cellcolor{blue!24} $0.16^{*}$ & $+0.01$ & \cellcolor{blue!24} $0.16^{*}$ & \cellcolor{red!2} $-0.02$ & \cellcolor{blue!27} $0.18^{*}$ & \cellcolor{red!2} $-0.01$ & \cellcolor{blue!24} $0.16^{*}$ \\
 &  & $k=2$ & \cellcolor{green!4} $+0.03$ & \cellcolor{blue!35} $0.24^{*}$ & $+0.00$ & \cellcolor{blue!33} $0.22^{*}$ & \cellcolor{green!1} $+0.01$ & \cellcolor{blue!32} $0.22^{*}$ & \cellcolor{red!3} $-0.02$ & \cellcolor{blue!33} $0.22^{*}$ & \cellcolor{red!2} $-0.02$ & \cellcolor{blue!30} $0.20^{*}$ & \cellcolor{red!3} $-0.02$ & \cellcolor{blue!34} $0.23^{*}$ & \cellcolor{red!2} $-0.02$ & \cellcolor{blue!30} $0.21^{*}$ \\
 &  & $k=3$ & \cellcolor{red!1} $-0.01$ & \cellcolor{blue!32} $0.22^{*}$ & $-0.01$ & \cellcolor{blue!31} $0.21^{*}$ & \cellcolor{red!3} $-0.02$ & \cellcolor{blue!29} $0.20^{*}$ & \cellcolor{red!6} $-0.04$ & \cellcolor{blue!31} $0.21^{*}$ & \cellcolor{red!4} $-0.03$ & \cellcolor{blue!26} $0.18^{*}$ & \cellcolor{red!12} $-0.09^{*}$ & \cellcolor{blue!31} $0.21^{*}$ & \cellcolor{red!10} $-0.07^{*}$ & \cellcolor{blue!26} $0.18^{*}$ \\
\hhline{~----------------}
 & \multirow{3}{*}{256} & $k=1$ & \cellcolor{red!7} $-0.05^{*}$ & \cellcolor{blue!23} $0.16^{*}$ & \cellcolor{red!2} $-0.02$ & \cellcolor{blue!22} $0.15^{*}$ & \cellcolor{red!3} $-0.02$ & \cellcolor{blue!21} $0.14^{*}$ & \cellcolor{red!3} $-0.03$ & \cellcolor{blue!22} $0.15^{*}$ & \cellcolor{green!1} $+0.01$ & \cellcolor{blue!19} $0.13^{*}$ & \cellcolor{red!4} $-0.03$ & \cellcolor{blue!21} $0.15^{*}$ & \cellcolor{red!3} $-0.02$ & \cellcolor{blue!19} $0.13^{*}$ \\
 &  & $k=2$ & \cellcolor{red!7} $-0.05^{*}$ & \cellcolor{blue!28} $0.19^{*}$ & \cellcolor{red!6} $-0.05^{*}$ & \cellcolor{blue!28} $0.19^{*}$ & \cellcolor{red!3} $-0.02$ & \cellcolor{blue!28} $0.19^{*}$ & \cellcolor{red!5} $-0.04$ & \cellcolor{blue!29} $0.20^{*}$ & \cellcolor{green!1} $+0.01$ & \cellcolor{blue!23} $0.15^{*}$ & \cellcolor{red!7} $-0.05^{*}$ & \cellcolor{blue!28} $0.19^{*}$ & \cellcolor{red!3} $-0.02$ & \cellcolor{blue!24} $0.16^{*}$ \\
 &  & $k=3$ & \cellcolor{red!3} $-0.03$ & \cellcolor{blue!23} $0.16^{*}$ & \cellcolor{red!1} $-0.01$ & \cellcolor{blue!24} $0.16^{*}$ & $+0.00$ & \cellcolor{blue!22} $0.15^{*}$ & \cellcolor{red!2} $-0.02$ & \cellcolor{blue!23} $0.15^{*}$ & $+0.00$ & \cellcolor{blue!17} $0.12^{*}$ & \cellcolor{red!1} $-0.01$ & \cellcolor{blue!22} $0.15^{*}$ & $+0.00$ & \cellcolor{blue!17} $0.12^{*}$ \\
\midrule
\multirow{9}{*}{\textbf{chrF}} & \multirow{3}{*}{4} & $k=1$ & \cellcolor{red!9} $-0.06^{*}$ & \cellcolor{blue!34} $0.23^{*}$ & \cellcolor{red!7} $-0.05^{*}$ & \cellcolor{blue!31} $0.21^{*}$ & \cellcolor{red!6} $-0.04$ & \cellcolor{blue!28} $0.19^{*}$ & \cellcolor{red!9} $-0.06^{*}$ & \cellcolor{blue!31} $0.21^{*}$ & \cellcolor{red!5} $-0.03$ & \cellcolor{blue!28} $0.19^{*}$ & \cellcolor{red!8} $-0.05^{*}$ & \cellcolor{blue!31} $0.21^{*}$ & \cellcolor{red!8} $-0.06^{*}$ & \cellcolor{blue!29} $0.20^{*}$ \\
 &  & $k=2$ & \cellcolor{green!1} $+0.01$ & \cellcolor{blue!36} $0.24^{*}$ & \cellcolor{red!5} $-0.04$ & \cellcolor{blue!35} $0.24^{*}$ & \cellcolor{red!6} $-0.05^{*}$ & \cellcolor{blue!35} $0.23^{*}$ & $+0.00$ & \cellcolor{blue!35} $0.24^{*}$ & $-0.01$ & \cellcolor{blue!34} $0.23^{*}$ & \cellcolor{red!5} $-0.03$ & \cellcolor{blue!35} $0.24^{*}$ & \cellcolor{red!1} $-0.01$ & \cellcolor{blue!34} $0.23^{*}$ \\
 &  & $k=3$ & \cellcolor{red!3} $-0.03$ & \cellcolor{blue!23} $0.16^{*}$ & \cellcolor{red!6} $-0.04$ & \cellcolor{blue!24} $0.16^{*}$ & \cellcolor{red!3} $-0.02$ & \cellcolor{blue!24} $0.16^{*}$ & \cellcolor{green!4} $+0.03$ & \cellcolor{blue!24} $0.16^{*}$ & \cellcolor{red!3} $-0.03$ & \cellcolor{blue!23} $0.15^{*}$ & $+0.00$ & \cellcolor{blue!24} $0.16^{*}$ & $+0.00$ & \cellcolor{blue!23} $0.16^{*}$ \\
\hhline{~----------------}
 & \multirow{3}{*}{32} & $k=1$ & \cellcolor{green!5} $+0.04$ & \cellcolor{blue!19} $0.13^{*}$ & \cellcolor{green!6} $+0.04$ & \cellcolor{blue!17} $0.12^{*}$ & \cellcolor{green!7} $+0.05^{*}$ & \cellcolor{blue!14} $0.10^{*}$ & $+0.01$ & \cellcolor{blue!16} $0.11^{*}$ & \cellcolor{green!4} $+0.03$ & \cellcolor{blue!12} $0.09^{*}$ & $+0.00$ & \cellcolor{blue!17} $0.12^{*}$ & \cellcolor{green!3} $+0.02$ & \cellcolor{blue!13} $0.09^{*}$ \\
 &  & $k=2$ & \cellcolor{red!3} $-0.03$ & \cellcolor{blue!21} $0.15^{*}$ & $+0.01$ & \cellcolor{blue!23} $0.16^{*}$ & \cellcolor{green!5} $+0.04$ & \cellcolor{blue!22} $0.15^{*}$ & \cellcolor{green!1} $+0.01$ & \cellcolor{blue!23} $0.16^{*}$ & \cellcolor{red!1} $-0.01$ & \cellcolor{blue!17} $0.12^{*}$ & \cellcolor{green!7} $+0.05^{*}$ & \cellcolor{blue!23} $0.16^{*}$ & \cellcolor{green!1} $+0.01$ & \cellcolor{blue!18} $0.12^{*}$ \\
 &  & $k=3$ & \cellcolor{green!1} $+0.01$ & \cellcolor{blue!16} $0.11^{*}$ & $+0.00$ & \cellcolor{blue!16} $0.11^{*}$ & \cellcolor{green!6} $+0.04^{*}$ & \cellcolor{blue!15} $0.10^{*}$ & $-0.01$ & \cellcolor{blue!15} $0.11^{*}$ & $+0.00$ & \cellcolor{blue!9} $0.06^{*}$ & \cellcolor{green!3} $+0.03$ & \cellcolor{blue!15} $0.11^{*}$ & \cellcolor{green!1} $+0.01$ & \cellcolor{blue!9} $0.06^{*}$ \\
\hhline{~----------------}
 & \multirow{3}{*}{256} & $k=1$ & \cellcolor{red!1} $-0.01$ & \cellcolor{blue!12} $0.09^{*}$ & \cellcolor{green!2} $+0.02$ & \cellcolor{blue!10} $0.07^{*}$ & $+0.00$ & \cellcolor{blue!9} $0.06^{*}$ & \cellcolor{red!4} $-0.03$ & \cellcolor{blue!13} $0.09^{*}$ & \cellcolor{red!2} $-0.01$ & \cellcolor{blue!6} $0.05^{*}$ & \cellcolor{red!2} $-0.02$ & \cellcolor{blue!13} $0.09^{*}$ & $+0.00$ & \cellcolor{blue!8} $0.06^{*}$ \\
 &  & $k=2$ & \cellcolor{red!2} $-0.02$ & \cellcolor{blue!20} $0.14^{*}$ & \cellcolor{green!1} $+0.01$ & \cellcolor{blue!19} $0.13^{*}$ & \cellcolor{red!1} $-0.01$ & \cellcolor{blue!18} $0.13^{*}$ & \cellcolor{red!5} $-0.04$ & \cellcolor{blue!20} $0.14^{*}$ & \cellcolor{red!5} $-0.03$ & \cellcolor{blue!15} $0.10^{*}$ & \cellcolor{red!6} $-0.04$ & \cellcolor{blue!20} $0.14^{*}$ & \cellcolor{red!3} $-0.02$ & \cellcolor{blue!15} $0.11^{*}$ \\
 &  & $k=3$ & $+0.00$ & \cellcolor{blue!15} $0.10^{*}$ & \cellcolor{green!3} $+0.02$ & \cellcolor{blue!13} $0.09^{*}$ & \cellcolor{red!1} $-0.01$ & \cellcolor{blue!12} $0.09^{*}$ & \cellcolor{red!1} $-0.01$ & \cellcolor{blue!13} $0.09^{*}$ & \cellcolor{green!2} $+0.01$ & \cellcolor{blue!7} $0.05^{*}$ & \cellcolor{red!3} $-0.02$ & \cellcolor{blue!14} $0.09^{*}$ & $+0.00$ & \cellcolor{blue!7} $0.05^{*}$ \\
\midrule
\multirow{9}{*}{\textbf{BLEURT}} & \multirow{3}{*}{4} & $k=1$ & \cellcolor{green!2} $+0.02$ & \cellcolor{blue!47} $0.32^{*}$ & \cellcolor{green!5} $+0.03$ & \cellcolor{blue!50} $0.34^{*}$ & \cellcolor{red!21} $-0.14^{*}$ & \cellcolor{blue!52} $0.35^{*}$ & \cellcolor{green!1} $+0.01$ & \cellcolor{blue!52} $0.35^{*}$ & $+0.00$ & \cellcolor{blue!50} $0.34^{*}$ & \cellcolor{green!1} $+0.01$ & \cellcolor{blue!51} $0.35^{*}$ & \cellcolor{red!1} $-0.01$ & \cellcolor{blue!52} $0.35^{*}$ \\
 &  & $k=2$ & \cellcolor{green!6} $+0.04$ & \cellcolor{blue!24} $0.16^{*}$ & \cellcolor{green!5} $+0.04$ & \cellcolor{blue!26} $0.18^{*}$ & \cellcolor{red!13} $-0.09^{*}$ & \cellcolor{blue!27} $0.19^{*}$ & \cellcolor{red!1} $-0.01$ & \cellcolor{blue!27} $0.19^{*}$ & $+0.00$ & \cellcolor{blue!27} $0.18^{*}$ & \cellcolor{green!1} $+0.01$ & \cellcolor{blue!27} $0.18^{*}$ & \cellcolor{red!2} $-0.01$ & \cellcolor{blue!27} $0.18^{*}$ \\
 &  & $k=3$ & \cellcolor{green!6} $+0.05^{*}$ & \cellcolor{blue!12} $0.08^{*}$ & \cellcolor{green!6} $+0.04$ & \cellcolor{blue!13} $0.09^{*}$ & \cellcolor{green!4} $+0.03$ & \cellcolor{blue!14} $0.09^{*}$ & \cellcolor{green!6} $+0.04$ & \cellcolor{blue!14} $0.09^{*}$ & \cellcolor{green!8} $+0.06^{*}$ & \cellcolor{blue!13} $0.09^{*}$ & \cellcolor{green!3} $+0.02$ & \cellcolor{blue!14} $0.10^{*}$ & \cellcolor{green!5} $+0.03$ & \cellcolor{blue!13} $0.09^{*}$ \\
\hhline{~----------------}
 & \multirow{3}{*}{32} & $k=1$ & $+0.00$ & \cellcolor{blue!35} $0.24^{*}$ & \cellcolor{red!1} $-0.01$ & \cellcolor{blue!41} $0.28^{*}$ & \cellcolor{red!20} $-0.14^{*}$ & \cellcolor{blue!43} $0.29^{*}$ & \cellcolor{green!3} $+0.02$ & \cellcolor{blue!41} $0.28^{*}$ & \cellcolor{red!3} $-0.03$ & \cellcolor{blue!40} $0.27^{*}$ & \cellcolor{green!5} $+0.03$ & \cellcolor{blue!40} $0.27^{*}$ & \cellcolor{green!2} $+0.02$ & \cellcolor{blue!39} $0.27^{*}$ \\
 &  & $k=2$ & \cellcolor{green!1} $+0.01$ & \cellcolor{blue!18} $0.12^{*}$ & \cellcolor{green!2} $+0.01$ & \cellcolor{blue!21} $0.15^{*}$ & \cellcolor{red!3} $-0.02$ & \cellcolor{blue!22} $0.15^{*}$ & $+0.00$ & \cellcolor{blue!22} $0.15^{*}$ & \cellcolor{green!1} $+0.01$ & \cellcolor{blue!20} $0.14^{*}$ & \cellcolor{green!11} $+0.08^{*}$ & \cellcolor{blue!21} $0.14^{*}$ & \cellcolor{green!6} $+0.04$ & \cellcolor{blue!20} $0.14^{*}$ \\
 &  & $k=3$ & \cellcolor{green!1} $+0.01$ & \cellcolor{blue!7} $0.05^{*}$ & \cellcolor{red!2} $-0.02$ & \cellcolor{blue!8} $0.06^{*}$ & $+0.00$ & \cellcolor{blue!9} $0.06^{*}$ & $+0.00$ & \cellcolor{blue!9} $0.06^{*}$ & $+0.00$ & \cellcolor{blue!6} $0.04$ & \cellcolor{green!2} $+0.02$ & \cellcolor{blue!9} $0.06^{*}$ & $+0.01$ & \cellcolor{blue!5} $0.04$ \\
\hhline{~----------------}
 & \multirow{3}{*}{256} & $k=1$ & \cellcolor{red!2} $-0.02$ & \cellcolor{blue!28} $0.19^{*}$ & \cellcolor{red!4} $-0.03$ & \cellcolor{blue!32} $0.21^{*}$ & \cellcolor{red!21} $-0.15^{*}$ & \cellcolor{blue!33} $0.23^{*}$ & \cellcolor{red!2} $-0.02$ & \cellcolor{blue!32} $0.22^{*}$ & \cellcolor{red!7} $-0.05^{*}$ & \cellcolor{blue!31} $0.21^{*}$ & \cellcolor{green!3} $+0.02$ & \cellcolor{blue!33} $0.22^{*}$ & \cellcolor{red!4} $-0.03$ & \cellcolor{blue!31} $0.21^{*}$ \\
 &  & $k=2$ & \cellcolor{green!5} $+0.03$ & \cellcolor{blue!11} $0.07^{*}$ & \cellcolor{red!1} $-0.01$ & \cellcolor{blue!12} $0.08^{*}$ & \cellcolor{red!1} $-0.01$ & \cellcolor{blue!13} $0.09^{*}$ & \cellcolor{green!3} $+0.02$ & \cellcolor{blue!12} $0.09^{*}$ & \cellcolor{green!2} $+0.02$ & \cellcolor{blue!10} $0.07^{*}$ & \cellcolor{green!5} $+0.04$ & \cellcolor{blue!12} $0.08^{*}$ & \cellcolor{green!1} $+0.01$ & \cellcolor{blue!10} $0.07^{*}$ \\
 &  & $k=3$ & $+0.00$ & \cellcolor{blue!2} $0.01$ & \cellcolor{red!3} $-0.03$ & \cellcolor{blue!1} $0.01$ & $+0.00$ & \cellcolor{blue!2} $0.02$ & $+0.00$ & \cellcolor{blue!2} $0.02$ & \cellcolor{green!2} $+0.01$ & $-0.02$ & $+0.00$ & \cellcolor{blue!2} $0.02$ & \cellcolor{red!2} $-0.02$ & $-0.02$ \\
\midrule
\multirow{9}{*}{\textbf{COMET}} & \multirow{3}{*}{4} & $k=1$ & \cellcolor{green!3} $+0.03$ & $-0.06^{*}$ & \cellcolor{green!3} $+0.02$ & $-0.08$ & \cellcolor{green!8} $+0.06$ & $-0.06$ & \cellcolor{green!17} $+0.12^{*}$ & $-0.02^{*}$ & \cellcolor{green!3} $+0.02$ & $-0.08$ & \cellcolor{green!16} $+0.11^{*}$ & $-0.02^{*}$ & \cellcolor{green!9} $+0.07$ & $-0.05^{*}$ \\
 &  & $k=2$ & \cellcolor{green!2} $+0.02$ & \cellcolor{blue!5} $0.04$ & $+0.00$ & \cellcolor{blue!4} $0.03$ & \cellcolor{green!2} $+0.01$ & \cellcolor{blue!5} $0.04$ & \cellcolor{green!7} $+0.05$ & \cellcolor{blue!11} $0.08^{*}$ & $+0.00$ & \cellcolor{blue!4} $0.03$ & \cellcolor{green!5} $+0.03$ & \cellcolor{blue!9} $0.07^{*}$ & \cellcolor{green!2} $+0.02$ & \cellcolor{blue!6} $0.04$ \\
 &  & $k=3$ & \cellcolor{green!3} $+0.02$ & \cellcolor{blue!2} $0.02$ & $+0.00$ & $0.00$ & \cellcolor{green!3} $+0.03$ & \cellcolor{blue!2} $0.02$ & \cellcolor{green!5} $+0.04$ & \cellcolor{blue!9} $0.06$ & $+0.00$ & $0.00$ & \cellcolor{green!5} $+0.04$ & \cellcolor{blue!7} $0.05$ & \cellcolor{green!3} $+0.03$ & \cellcolor{blue!3} $0.02$ \\
\hhline{~----------------}
 & \multirow{3}{*}{32} & $k=1$ & \cellcolor{red!4} $-0.03$ & \cellcolor{blue!21} $0.14^{*}$ & \cellcolor{red!4} $-0.03$ & \cellcolor{blue!22} $0.15^{*}$ & \cellcolor{red!9} $-0.06$ & \cellcolor{blue!23} $0.15^{*}$ & \cellcolor{red!22} $-0.15^{*}$ & \cellcolor{blue!28} $0.19^{*}$ & \cellcolor{red!4} $-0.03$ & \cellcolor{blue!21} $0.14^{*}$ & \cellcolor{red!10} $-0.07$ & \cellcolor{blue!25} $0.17^{*}$ & \cellcolor{red!9} $-0.06$ & \cellcolor{blue!22} $0.15^{*}$ \\
 &  & $k=2$ & \cellcolor{red!2} $-0.01$ & \cellcolor{blue!7} $0.05$ & $+0.00$ & \cellcolor{blue!8} $0.05$ & $+0.00$ & \cellcolor{blue!7} $0.05$ & \cellcolor{red!4} $-0.03$ & \cellcolor{blue!14} $0.10^{*}$ & \cellcolor{red!1} $-0.01$ & \cellcolor{blue!6} $0.04$ & \cellcolor{red!6} $-0.04$ & \cellcolor{blue!11} $0.08^{*}$ & \cellcolor{red!3} $-0.02$ & \cellcolor{blue!8} $0.06$ \\
 &  & $k=3$ & $-0.01$ & \cellcolor{blue!5} $0.03$ & $+0.01$ & \cellcolor{blue!5} $0.04$ & $-0.01$ & \cellcolor{blue!4} $0.03$ & \cellcolor{red!5} $-0.04$ & \cellcolor{blue!12} $0.08^{*}$ & $+0.00$ & \cellcolor{blue!3} $0.02$ & \cellcolor{red!6} $-0.05$ & \cellcolor{blue!9} $0.07^{*}$ & \cellcolor{red!2} $-0.02$ & \cellcolor{blue!5} $0.04$ \\
\hhline{~----------------}
 & \multirow{3}{*}{256} & $k=1$ & \cellcolor{green!6} $+0.04$ & $0.01^{*}$ & \cellcolor{green!3} $+0.03$ & $-0.01^{*}$ & \cellcolor{green!8} $+0.05$ & $-0.00$ & \cellcolor{green!19} $+0.13^{*}$ & \cellcolor{blue!6} $0.05^{*}$ & \cellcolor{green!1} $+0.01$ & $-0.00$ & \cellcolor{green!17} $+0.12^{*}$ & \cellcolor{blue!4} $0.03^{*}$ & \cellcolor{green!10} $+0.07^{*}$ & $0.01^{*}$ \\
 &  & $k=2$ & \cellcolor{green!4} $+0.03$ & \cellcolor{blue!16} $0.11^{*}$ & \cellcolor{green!5} $+0.04$ & \cellcolor{blue!13} $0.09^{*}$ & \cellcolor{green!4} $+0.03$ & \cellcolor{blue!10} $0.07$ & \cellcolor{green!15} $+0.11$ & \cellcolor{blue!19} $0.13^{*}$ & \cellcolor{green!2} $+0.02$ & \cellcolor{blue!11} $0.08^{*}$ & \cellcolor{green!16} $+0.11^{*}$ & \cellcolor{blue!17} $0.12^{*}$ & \cellcolor{green!11} $+0.07$ & \cellcolor{blue!15} $0.10^{*}$ \\
 &  & $k=3$ & \cellcolor{green!4} $+0.03$ & \cellcolor{blue!12} $0.08^{*}$ & \cellcolor{green!4} $+0.03$ & \cellcolor{blue!10} $0.07^{*}$ & \cellcolor{green!6} $+0.04$ & \cellcolor{blue!7} $0.05$ & \cellcolor{green!16} $+0.11$ & \cellcolor{blue!17} $0.11^{*}$ & \cellcolor{green!2} $+0.02$ & \cellcolor{blue!8} $0.06$ & \cellcolor{green!15} $+0.10$ & \cellcolor{blue!14} $0.10^{*}$ & \cellcolor{green!10} $+0.07$ & \cellcolor{blue!12} $0.09^{*}$ \\
\midrule
\multirow{9}{*}{\textbf{BERTScore}} & \multirow{3}{*}{4} & $k=1$ & \cellcolor{red!8} $-0.05^{*}$ & \cellcolor{blue!28} $0.19^{*}$ & \cellcolor{red!7} $-0.05^{*}$ & \cellcolor{blue!27} $0.19^{*}$ & \cellcolor{red!5} $-0.04$ & \cellcolor{blue!24} $0.17^{*}$ & \cellcolor{red!9} $-0.06^{*}$ & \cellcolor{blue!30} $0.20^{*}$ & \cellcolor{red!8} $-0.06^{*}$ & \cellcolor{blue!25} $0.17^{*}$ & \cellcolor{red!1} $-0.01$ & \cellcolor{blue!30} $0.21^{*}$ & \cellcolor{red!8} $-0.06^{*}$ & \cellcolor{blue!26} $0.18^{*}$ \\
 &  & $k=2$ & \cellcolor{red!5} $-0.04$ & \cellcolor{blue!27} $0.18^{*}$ & \cellcolor{red!2} $-0.01$ & \cellcolor{blue!31} $0.21^{*}$ & \cellcolor{green!2} $+0.01$ & \cellcolor{blue!30} $0.20^{*}$ & \cellcolor{green!3} $+0.02$ & \cellcolor{blue!30} $0.21^{*}$ & \cellcolor{red!7} $-0.05^{*}$ & \cellcolor{blue!26} $0.18^{*}$ & \cellcolor{green!3} $+0.02$ & \cellcolor{blue!31} $0.21^{*}$ & \cellcolor{green!1} $+0.01$ & \cellcolor{blue!29} $0.20^{*}$ \\
 &  & $k=3$ & \cellcolor{red!1} $-0.01$ & \cellcolor{blue!20} $0.14^{*}$ & \cellcolor{red!6} $-0.04$ & \cellcolor{blue!22} $0.15^{*}$ & \cellcolor{green!4} $+0.03$ & \cellcolor{blue!22} $0.15^{*}$ & \cellcolor{red!3} $-0.02$ & \cellcolor{blue!22} $0.15^{*}$ & \cellcolor{red!7} $-0.05^{*}$ & \cellcolor{blue!12} $0.08^{*}$ & \cellcolor{red!2} $-0.01$ & \cellcolor{blue!22} $0.15^{*}$ & \cellcolor{red!3} $-0.03$ & \cellcolor{blue!22} $0.15^{*}$ \\
\hhline{~----------------}
 & \multirow{3}{*}{32} & $k=1$ & \cellcolor{green!4} $+0.03$ & \cellcolor{blue!19} $0.13^{*}$ & \cellcolor{red!2} $-0.01$ & \cellcolor{blue!17} $0.12^{*}$ & $+0.00$ & \cellcolor{blue!18} $0.12^{*}$ & \cellcolor{green!1} $+0.01$ & \cellcolor{blue!20} $0.14^{*}$ & \cellcolor{red!2} $-0.01$ & \cellcolor{blue!15} $0.10^{*}$ & \cellcolor{red!4} $-0.03$ & \cellcolor{blue!20} $0.13^{*}$ & $+0.01$ & \cellcolor{blue!17} $0.12^{*}$ \\
 &  & $k=2$ & \cellcolor{green!3} $+0.03$ & \cellcolor{blue!36} $0.24^{*}$ & $+0.00$ & \cellcolor{blue!36} $0.24^{*}$ & \cellcolor{red!5} $-0.03$ & \cellcolor{blue!37} $0.25^{*}$ & \cellcolor{green!5} $+0.04$ & \cellcolor{blue!37} $0.25^{*}$ & \cellcolor{red!9} $-0.06^{*}$ & \cellcolor{blue!30} $0.20^{*}$ & $+0.01$ & \cellcolor{blue!36} $0.24^{*}$ & $+0.00$ & \cellcolor{blue!36} $0.24^{*}$ \\
 &  & $k=3$ & \cellcolor{green!3} $+0.02$ & \cellcolor{blue!16} $0.11^{*}$ & \cellcolor{green!3} $+0.02$ & \cellcolor{blue!16} $0.11^{*}$ & \cellcolor{red!1} $-0.01$ & \cellcolor{blue!17} $0.11^{*}$ & \cellcolor{green!7} $+0.05^{*}$ & \cellcolor{blue!17} $0.11^{*}$ & \cellcolor{red!1} $-0.01$ & \cellcolor{blue!6} $0.04$ & \cellcolor{green!6} $+0.04$ & \cellcolor{blue!17} $0.11^{*}$ & \cellcolor{green!4} $+0.03$ & \cellcolor{blue!16} $0.11^{*}$ \\
\hhline{~----------------}
 & \multirow{3}{*}{256} & $k=1$ & \cellcolor{red!1} $-0.01$ & \cellcolor{blue!14} $0.09^{*}$ & \cellcolor{red!7} $-0.05^{*}$ & \cellcolor{blue!7} $0.05^{*}$ & \cellcolor{red!2} $-0.02$ & \cellcolor{blue!9} $0.06^{*}$ & \cellcolor{red!3} $-0.03$ & \cellcolor{blue!11} $0.08^{*}$ & \cellcolor{red!4} $-0.03$ & \cellcolor{blue!5} $0.04$ & \cellcolor{green!2} $+0.02$ & \cellcolor{blue!11} $0.08^{*}$ & \cellcolor{red!4} $-0.03$ & \cellcolor{blue!8} $0.06^{*}$ \\
 &  & $k=2$ & \cellcolor{green!2} $+0.01$ & \cellcolor{blue!20} $0.14^{*}$ & \cellcolor{red!4} $-0.03$ & \cellcolor{blue!20} $0.13^{*}$ & \cellcolor{red!2} $-0.01$ & \cellcolor{blue!19} $0.13^{*}$ & \cellcolor{red!2} $-0.02$ & \cellcolor{blue!19} $0.13^{*}$ & \cellcolor{red!4} $-0.03$ & \cellcolor{blue!9} $0.06^{*}$ & \cellcolor{red!3} $-0.02$ & \cellcolor{blue!19} $0.13^{*}$ & \cellcolor{red!3} $-0.02$ & \cellcolor{blue!19} $0.13^{*}$ \\
 &  & $k=3$ & \cellcolor{green!3} $+0.02$ & \cellcolor{blue!10} $0.07^{*}$ & \cellcolor{red!3} $-0.03$ & \cellcolor{blue!6} $0.05^{*}$ & $-0.01$ & \cellcolor{blue!6} $0.04$ & \cellcolor{red!1} $-0.01$ & \cellcolor{blue!6} $0.04$ & \cellcolor{red!5} $-0.04$ & $-0.06^{*}$ & \cellcolor{red!4} $-0.03$ & \cellcolor{blue!5} $0.04$ & \cellcolor{red!5} $-0.04$ & \cellcolor{blue!6} $0.04$ \\
\bottomrule
\end{NiceTabular}
}
\caption{Spearman correlation ($\rho$) between matrix reconstruction errors and evaluation score differences across varying reference pool sizes ($|\mathcal{Y}|$) on WMT22 En$\rightarrow$De. \textit{Dir.} (Directional Change) measures score improvements, highlighted as \colorbox{green!15}{positive} and \colorbox{red!15}{negative} impacts. \textit{Mag.} (Magnitude Sensitivity) measures absolute score variance, with \colorbox{blue!15}{blue} highlighting metric sensitivity to matrix modifications. Statistical significance ($p < 0.05$) is denoted by an asterisk ($^*$).}
\label{tab:corr_mat_analysis_complete_wmt22-ende}
\end{table*}

To further investigate how SVD denoise the MBR pairwise matrix, we conducted a correlation analysis mirroring our previous variance study. However, rather than evaluating inherent variance, we calculated the correlation between the matrix reconstruction error and the evaluation score differences between SVD-MBR and MBR. Our primary objective is to mathematically determine whether the information discarded by the low-rank approximation represents noise or valuable signal. To quantify this error, we computed the Frobenius norm of the difference between the normalized pairwise utility matrix~(used in MBR) and the reconstructed low-rank matrix utilized by SVD-MBR.

Table~\ref{tab:corr_mat_analysis_complete_wmt22-ende} reveals that \textbf{the discarded information depends heavily on the metric used as utility function}. For neural metrics, specifically COMET and BLEURT, the information discarded by SVD is demonstrably noise; a higher reconstruction error correlates with positive gains across most off-target metrics, even in instances where the utility metric itself degrades. Conversely, BERTScore and chrF exhibit a more muted response, showing minimal generalized improvements~($\bar{Z}_{other}$) except at specific reference size~(e.g., $|\mathcal{Y}|=32$), where positive gains become more pronounced. Finally, BLEU stands out as the only metric where reconstruction error yields almost no improvement to off-target metrics, further confirming that truncating BLEU matrices destroys uncompressible signals rather than filtering noise.

Looking at the absolute magnitude of change~(\textit{Mag.}), there exist multiple instances of statistically significant relationship to matrix reconstruction error and evaluation scores. Across the vast majority of configurations, higher reconstruction errors correlate heavily with larger absolute differences in the metrics. This indicates that the \textbf{reconstruction error acts as a reliable indicator for the scale of the SVD's intervention: when the matrix has a high reconstruction error, SVD-MBR aggressively alters hypothesis selection, driving substantial changes in the overall evaluation scores}. Conversely, when the reconstruction error is minimal, SVD-MBR's behavior closely mirrors standard MBR, resulting in only marginal shifts in the final scores.

\begin{table*}[ht!]
\centering
\resizebox{\textwidth}{!}{
\begin{tabular}{llllllllll}
\toprule
\textbf{Util. Function} & \textbf{$|\mathcal{Y}|$} & \textbf{Top-$k$} &  \textbf{BLEU} & \textbf{chrF} & \textbf{BLEURT} & \textbf{COMET} & \textbf{BERTScore} & \textbf{COMETKiwi} & \textbf{$\bar{Z}_{\text{other}}$} \\
\midrule
\multirow[t]{9}{*}{BLEU} & \multirow[t]{3}{*}{4} & $k=1$ & \cellcolor{red!15} 419 / 1169 / 449 & \cellcolor{green!15} 493 / 1053 / 491 & \cellcolor{red!15} 480 / 1050 / 507 & \cellcolor{red!15} 486 / 1048 / 503 & \cellcolor{red!15} 492 / 1043 / 502 & \cellcolor{red!15} 475 / 1050 / 512 & \cellcolor{red!15} 478 / 1041 / 518 \\
 &  & $k=2$ & \cellcolor{red!15} 223 / 1543 / 271 & \cellcolor{red!15} 259 / 1482 / 296 & \cellcolor{red!15} 260 / 1479 / 298 & \cellcolor{red!15} 256 / 1480 / 301 & \cellcolor{red!15} 255 / 1475 / 307 & \cellcolor{red!15} 257 / 1481 / 299 & \cellcolor{red!15} 250 / 1472 / 315 \\
 &  & $k=3$ & \cellcolor{red!15} 71 / 1858 / 108 & 109 / 1819 / 109 & \cellcolor{red!15} 91 / 1817 / 129 & \cellcolor{red!15} 99 / 1817 / 121 & \cellcolor{red!15} 95 / 1814 / 128 & \cellcolor{red!15} 99 / 1817 / 121 & \cellcolor{red!15} 98 / 1811 / 128 \\
\cline{2-10}
 & \multirow[t]{3}{*}{32} & $k=1$ & \cellcolor{green!15} 381 / 1280 / 376 & \cellcolor{red!15} 434 / 1155 / 448 & \cellcolor{red!15} 438 / 1147 / 452 & \cellcolor{red!15} 423 / 1147 / 467 & \cellcolor{green!15} 455 / 1135 / 447 & \cellcolor{red!15} 399 / 1151 / 487 & \cellcolor{red!15} 452 / 1118 / 467 \\
 &  & $k=2$ & \cellcolor{red!15} 259 / 1507 / 271 & \cellcolor{red!15} 301 / 1412 / 324 & \cellcolor{red!15} 300 / 1403 / 334 & \cellcolor{red!15} 295 / 1404 / 338 & \cellcolor{red!15} 312 / 1394 / 331 & \cellcolor{red!15} 282 / 1409 / 346 & \cellcolor{red!15} 313 / 1375 / 349 \\
 &  & $k=3$ & \cellcolor{red!15} 175 / 1669 / 193 & \cellcolor{red!15} 207 / 1605 / 225 & \cellcolor{red!15} 198 / 1597 / 242 & \cellcolor{red!15} 192 / 1598 / 247 & \cellcolor{red!15} 217 / 1586 / 234 & \cellcolor{red!15} 188 / 1601 / 248 & \cellcolor{red!15} 220 / 1564 / 253 \\
\cline{2-10}
 & \multirow[t]{3}{*}{256} & $k=1$ & \cellcolor{red!15} 199 / 1556 / 282 & \cellcolor{red!15} 269 / 1458 / 310 & \cellcolor{red!15} 279 / 1451 / 307 & \cellcolor{red!15} 268 / 1449 / 320 & \cellcolor{green!15} 303 / 1435 / 299 & \cellcolor{red!15} 276 / 1454 / 307 & \cellcolor{red!15} 276 / 1425 / 336 \\
 &  & $k=2$ & \cellcolor{red!15} 120 / 1744 / 173 & \cellcolor{red!15} 145 / 1676 / 216 & \cellcolor{red!15} 169 / 1670 / 198 & \cellcolor{red!15} 153 / 1668 / 216 & \cellcolor{red!15} 189 / 1653 / 195 & \cellcolor{red!15} 168 / 1670 / 199 & \cellcolor{red!15} 175 / 1631 / 231 \\
 &  & $k=3$ & \cellcolor{red!15} 81 / 1848 / 108 & \cellcolor{red!15} 106 / 1788 / 143 & \cellcolor{red!15} 120 / 1778 / 139 & \cellcolor{red!15} 119 / 1778 / 140 & \cellcolor{red!15} 137 / 1758 / 142 & \cellcolor{red!15} 119 / 1781 / 137 & \cellcolor{red!15} 135 / 1743 / 159 \\
\cline{1-10} \cline{2-10}
\multirow[t]{9}{*}{chrF} & \multirow[t]{3}{*}{4} & $k=1$ & \cellcolor{red!15} 434 / 1119 / 484 & \cellcolor{red!15} 443 / 998 / 596 & \cellcolor{red!15} 464 / 993 / 580 & \cellcolor{red!15} 500 / 992 / 545 & \cellcolor{red!15} 487 / 992 / 558 & \cellcolor{red!15} 496 / 995 / 546 & \cellcolor{red!15} 476 / 990 / 571 \\
 &  & $k=2$ & \cellcolor{green!15} 185 / 1672 / 180 & \cellcolor{red!15} 190 / 1625 / 222 & \cellcolor{red!15} 182 / 1622 / 233 & \cellcolor{red!15} 204 / 1621 / 212 & \cellcolor{red!15} 195 / 1618 / 224 & \cellcolor{red!15} 183 / 1622 / 232 & \cellcolor{red!15} 201 / 1615 / 221 \\
 &  & $k=3$ & \cellcolor{red!15} 49 / 1903 / 85 & \cellcolor{red!15} 55 / 1879 / 103 & \cellcolor{red!15} 71 / 1879 / 87 & \cellcolor{green!15} 80 / 1878 / 79 & \cellcolor{red!15} 60 / 1877 / 100 & \cellcolor{red!15} 70 / 1878 / 89 & \cellcolor{red!15} 73 / 1874 / 90 \\
\cline{2-10}
 & \multirow[t]{3}{*}{32} & $k=1$ & \cellcolor{green!15} 444 / 1256 / 337 & \cellcolor{red!15} 405 / 1127 / 505 & \cellcolor{red!15} 445 / 1118 / 474 & \cellcolor{red!15} 450 / 1120 / 467 & \cellcolor{red!15} 450 / 1110 / 477 & \cellcolor{green!15} 460 / 1125 / 452 & \cellcolor{red!15} 459 / 1096 / 482 \\
 &  & $k=2$ & \cellcolor{red!15} 168 / 1676 / 193 & \cellcolor{red!15} 212 / 1583 / 242 & \cellcolor{green!15} 254 / 1572 / 211 & \cellcolor{green!15} 245 / 1572 / 220 & \cellcolor{red!15} 227 / 1559 / 251 & \cellcolor{green!15} 253 / 1574 / 210 & \cellcolor{green!15} 254 / 1542 / 241 \\
 &  & $k=3$ & \cellcolor{red!15} 113 / 1797 / 127 & \cellcolor{red!15} 139 / 1741 / 157 & \cellcolor{red!15} 148 / 1734 / 155 & \cellcolor{red!15} 150 / 1735 / 152 & \cellcolor{red!15} 137 / 1721 / 179 & \cellcolor{red!15} 149 / 1735 / 153 & \cellcolor{red!15} 154 / 1703 / 180 \\
\cline{2-10}
 & \multirow[t]{3}{*}{256} & $k=1$ & \cellcolor{green!15} 378 / 1365 / 294 & \cellcolor{red!15} 343 / 1267 / 427 & \cellcolor{red!15} 357 / 1260 / 420 & \cellcolor{red!15} 368 / 1263 / 406 & \cellcolor{red!15} 383 / 1258 / 396 & \cellcolor{red!15} 359 / 1268 / 410 & \cellcolor{red!15} 384 / 1247 / 406 \\
 &  & $k=2$ & \cellcolor{red!15} 191 / 1653 / 193 & \cellcolor{red!15} 226 / 1582 / 229 & \cellcolor{green!15} 239 / 1578 / 220 & \cellcolor{green!15} 242 / 1578 / 217 & \cellcolor{red!15} 229 / 1569 / 239 & \cellcolor{green!15} 247 / 1584 / 206 & \cellcolor{green!15} 247 / 1557 / 233 \\
 &  & $k=3$ & \cellcolor{green!15} 124 / 1792 / 121 & \cellcolor{red!15} 147 / 1742 / 148 & \cellcolor{green!15} 160 / 1736 / 141 & \cellcolor{green!15} 159 / 1737 / 141 & \cellcolor{red!15} 156 / 1724 / 157 & \cellcolor{green!15} 152 / 1740 / 145 & \cellcolor{green!15} 172 / 1710 / 155 \\
\cline{1-10} \cline{2-10}
\multirow[t]{9}{*}{BLEURT} & \multirow[t]{3}{*}{4} & $k=1$ & \cellcolor{green!15} 277 / 1511 / 249 & \cellcolor{green!15} 316 / 1448 / 273 & \cellcolor{red!15} 187 / 1440 / 410 & \cellcolor{red!15} 288 / 1441 / 308 & \cellcolor{red!15} 298 / 1439 / 300 & \cellcolor{red!15} 289 / 1443 / 305 & \cellcolor{green!15} 306 / 1439 / 292 \\
 &  & $k=2$ & \cellcolor{green!15} 98 / 1859 / 80 & \cellcolor{green!15} 120 / 1832 / 85 & \cellcolor{red!15} 84 / 1829 / 124 & \cellcolor{red!15} 97 / 1830 / 110 & \cellcolor{green!15} 110 / 1828 / 99 & \cellcolor{red!15} 100 / 1831 / 106 & \cellcolor{green!15} 107 / 1827 / 103 \\
 &  & $k=3$ & \cellcolor{green!15} 42 / 1968 / 27 & \cellcolor{green!15} 53 / 1955 / 29 & \cellcolor{green!15} 51 / 1954 / 32 & \cellcolor{green!15} 44 / 1954 / 39 & \cellcolor{green!15} 53 / 1953 / 31 & \cellcolor{green!15} 44 / 1955 / 38 & \cellcolor{green!15} 48 / 1952 / 37 \\
\cline{2-10}
 & \multirow[t]{3}{*}{32} & $k=1$ & \cellcolor{green!15} 249 / 1591 / 197 & \cellcolor{green!15} 282 / 1517 / 238 & \cellcolor{red!15} 145 / 1511 / 381 & \cellcolor{green!15} 284 / 1510 / 243 & \cellcolor{green!15} 285 / 1501 / 251 & \cellcolor{green!15} 268 / 1511 / 258 & \cellcolor{green!15} 300 / 1496 / 241 \\
 &  & $k=2$ & \cellcolor{green!15} 100 / 1856 / 81 & \cellcolor{green!15} 120 / 1823 / 94 & \cellcolor{red!15} 99 / 1823 / 115 & \cellcolor{green!15} 124 / 1824 / 89 & \cellcolor{green!15} 128 / 1814 / 95 & \cellcolor{green!15} 133 / 1825 / 79 & \cellcolor{green!15} 142 / 1808 / 87 \\
 &  & $k=3$ & \cellcolor{green!15} 59 / 1920 / 58 & \cellcolor{red!15} 65 / 1906 / 66 & \cellcolor{red!15} 65 / 1905 / 67 & \cellcolor{green!15} 70 / 1905 / 62 & \cellcolor{green!15} 75 / 1897 / 65 & \cellcolor{green!15} 69 / 1905 / 63 & \cellcolor{green!15} 81 / 1891 / 65 \\
\cline{2-10}
 & \multirow[t]{3}{*}{256} & $k=1$ & \cellcolor{green!15} 273 / 1571 / 193 & \cellcolor{green!15} 325 / 1486 / 226 & \cellcolor{red!15} 173 / 1480 / 384 & \cellcolor{green!15} 302 / 1481 / 254 & \cellcolor{green!15} 321 / 1480 / 236 & \cellcolor{green!15} 305 / 1483 / 249 & \cellcolor{green!15} 327 / 1475 / 235 \\
 &  & $k=2$ & \cellcolor{green!15} 99 / 1860 / 78 & \cellcolor{green!15} 115 / 1834 / 88 & \cellcolor{green!15} 102 / 1834 / 101 & \cellcolor{green!15} 114 / 1834 / 89 & \cellcolor{green!15} 123 / 1826 / 88 & \cellcolor{green!15} 123 / 1835 / 79 & \cellcolor{green!15} 137 / 1818 / 82 \\
 &  & $k=3$ & \cellcolor{green!15} 63 / 1930 / 44 & \cellcolor{green!15} 70 / 1916 / 51 & \cellcolor{green!15} 63 / 1916 / 58 & \cellcolor{green!15} 63 / 1917 / 57 & \cellcolor{green!15} 73 / 1909 / 55 & \cellcolor{green!15} 66 / 1916 / 55 & \cellcolor{green!15} 82 / 1900 / 55 \\
\cline{1-10} \cline{2-10}
\multirow[t]{9}{*}{COMET} & \multirow[t]{3}{*}{4} & $k=1$ & \cellcolor{green!15} 163 / 1723 / 151 & \cellcolor{red!15} 170 / 1687 / 180 & \cellcolor{red!15} 165 / 1686 / 186 & \cellcolor{red!15} 167 / 1687 / 183 & \cellcolor{red!15} 162 / 1682 / 193 & \cellcolor{red!15} 166 / 1687 / 184 & \cellcolor{red!15} 164 / 1679 / 194 \\
 &  & $k=2$ & \cellcolor{green!15} 62 / 1920 / 55 & \cellcolor{red!15} 65 / 1900 / 72 & \cellcolor{green!15} 69 / 1900 / 68 & \cellcolor{green!15} 73 / 1902 / 62 & \cellcolor{green!15} 72 / 1898 / 67 & \cellcolor{green!15} 68 / 1902 / 67 & \cellcolor{green!15} 76 / 1895 / 66 \\
 &  & $k=3$ & \cellcolor{red!15} 27 / 1981 / 29 & \cellcolor{green!15} 31 / 1977 / 29 & \cellcolor{red!15} 27 / 1976 / 34 & \cellcolor{green!15} 32 / 1977 / 28 & \cellcolor{green!15} 33 / 1974 / 30 & \cellcolor{green!15} 36 / 1976 / 25 & \cellcolor{green!15} 35 / 1971 / 31 \\
\cline{2-10}
 & \multirow[t]{3}{*}{32} & $k=1$ & \cellcolor{green!15} 149 / 1783 / 105 & \cellcolor{green!15} 159 / 1746 / 132 & \cellcolor{green!15} 156 / 1741 / 140 & \cellcolor{red!15} 139 / 1744 / 154 & \cellcolor{green!15} 160 / 1734 / 143 & \cellcolor{green!15} 162 / 1742 / 133 & \cellcolor{green!15} 160 / 1725 / 152 \\
 &  & $k=2$ & \cellcolor{green!15} 87 / 1883 / 67 & \cellcolor{green!15} 89 / 1871 / 77 & \cellcolor{green!15} 100 / 1871 / 66 & \cellcolor{green!15} 107 / 1872 / 58 & \cellcolor{green!15} 95 / 1863 / 79 & \cellcolor{green!15} 106 / 1870 / 61 & \cellcolor{green!15} 107 / 1850 / 80 \\
 &  & $k=3$ & \cellcolor{green!15} 49 / 1947 / 41 & \cellcolor{green!15} 58 / 1939 / 40 & \cellcolor{green!15} 57 / 1939 / 41 & \cellcolor{green!15} 57 / 1939 / 41 & \cellcolor{green!15} 61 / 1931 / 45 & \cellcolor{green!15} 62 / 1939 / 36 & \cellcolor{green!15} 68 / 1921 / 48 \\
\cline{2-10}
 & \multirow[t]{3}{*}{256} & $k=1$ & \cellcolor{green!15} 117 / 1810 / 110 & \cellcolor{green!15} 129 / 1784 / 124 & \cellcolor{green!15} 132 / 1780 / 125 & \cellcolor{green!15} 129 / 1783 / 125 & \cellcolor{green!15} 143 / 1776 / 118 & \cellcolor{green!15} 129 / 1788 / 120 & \cellcolor{green!15} 138 / 1768 / 131 \\
 &  & $k=2$ & \cellcolor{green!15} 82 / 1879 / 76 & \cellcolor{green!15} 91 / 1861 / 85 & \cellcolor{green!15} 103 / 1861 / 73 & \cellcolor{green!15} 129 / 1861 / 47 & \cellcolor{green!15} 95 / 1854 / 88 & \cellcolor{green!15} 106 / 1865 / 66 & \cellcolor{green!15} 107 / 1844 / 86 \\
 &  & $k=3$ & \cellcolor{red!15} 48 / 1933 / 56 & \cellcolor{red!15} 59 / 1918 / 60 & \cellcolor{green!15} 67 / 1919 / 51 & \cellcolor{green!15} 87 / 1921 / 29 & \cellcolor{red!15} 61 / 1911 / 65 & \cellcolor{green!15} 66 / 1921 / 50 & \cellcolor{green!15} 73 / 1899 / 65 \\
\cline{1-10} \cline{2-10}
\multirow[t]{9}{*}{BERTScore} & \multirow[t]{3}{*}{4} & $k=1$ & \cellcolor{red!15} 399 / 1151 / 487 & \cellcolor{red!15} 460 / 1026 / 551 & \cellcolor{red!15} 499 / 1017 / 521 & 508 / 1021 / 508 & \cellcolor{red!15} 420 / 1006 / 611 & \cellcolor{green!15} 516 / 1025 / 496 & \cellcolor{red!15} 482 / 1016 / 539 \\
 &  & $k=2$ & \cellcolor{red!15} 164 / 1691 / 182 & \cellcolor{red!15} 201 / 1609 / 227 & \cellcolor{red!15} 205 / 1607 / 225 & \cellcolor{green!15} 223 / 1608 / 206 & \cellcolor{red!15} 179 / 1596 / 262 & \cellcolor{green!15} 224 / 1611 / 202 & \cellcolor{red!15} 214 / 1607 / 216 \\
 &  & $k=3$ & \cellcolor{red!15} 56 / 1909 / 72 & \cellcolor{red!15} 67 / 1876 / 94 & \cellcolor{green!15} 82 / 1874 / 81 & \cellcolor{green!15} 82 / 1874 / 81 & \cellcolor{red!15} 70 / 1862 / 105 & \cellcolor{green!15} 82 / 1875 / 80 & \cellcolor{red!15} 79 / 1874 / 84 \\
\cline{2-10}
 & \multirow[t]{3}{*}{32} & $k=1$ & \cellcolor{green!15} 401 / 1238 / 398 & \cellcolor{red!15} 442 / 1083 / 512 & \cellcolor{red!15} 456 / 1079 / 502 & \cellcolor{green!15} 497 / 1077 / 463 & \cellcolor{red!15} 399 / 1052 / 586 & \cellcolor{red!15} 476 / 1084 / 477 & \cellcolor{green!15} 485 / 1074 / 478 \\
 &  & $k=2$ & \cellcolor{green!15} 196 / 1656 / 185 & \cellcolor{red!15} 226 / 1572 / 239 & \cellcolor{red!15} 213 / 1566 / 258 & \cellcolor{green!15} 252 / 1567 / 218 & \cellcolor{red!15} 209 / 1547 / 281 & \cellcolor{green!15} 237 / 1575 / 225 & \cellcolor{red!15} 233 / 1566 / 238 \\
 &  & $k=3$ & \cellcolor{green!15} 106 / 1842 / 89 & 125 / 1787 / 125 & \cellcolor{green!15} 126 / 1786 / 125 & \cellcolor{green!15} 138 / 1785 / 114 & \cellcolor{red!15} 117 / 1764 / 156 & \cellcolor{green!15} 142 / 1787 / 108 & \cellcolor{green!15} 143 / 1785 / 109 \\
\cline{2-10}
 & \multirow[t]{3}{*}{256} & $k=1$ & \cellcolor{green!15} 339 / 1360 / 338 & \cellcolor{red!15} 390 / 1215 / 432 & \cellcolor{red!15} 412 / 1203 / 422 & \cellcolor{green!15} 440 / 1208 / 389 & \cellcolor{red!15} 353 / 1180 / 504 & \cellcolor{green!15} 425 / 1209 / 403 & \cellcolor{green!15} 430 / 1202 / 405 \\
 &  & $k=2$ & \cellcolor{red!15} 122 / 1784 / 131 & \cellcolor{red!15} 149 / 1719 / 169 & \cellcolor{red!15} 158 / 1709 / 170 & \cellcolor{green!15} 186 / 1711 / 140 & \cellcolor{red!15} 133 / 1682 / 222 & \cellcolor{green!15} 179 / 1713 / 145 & \cellcolor{green!15} 175 / 1708 / 154 \\
 &  & $k=3$ & \cellcolor{green!15} 78 / 1900 / 59 & \cellcolor{green!15} 89 / 1860 / 88 & \cellcolor{green!15} 92 / 1857 / 88 & \cellcolor{green!15} 109 / 1857 / 71 & \cellcolor{red!15} 92 / 1827 / 118 & \cellcolor{green!15} 99 / 1861 / 77 & \cellcolor{green!15} 101 / 1857 / 79 \\
\cline{1-10} \cline{2-10}
\bottomrule
\end{tabular}

}
\caption{Sentence-level comparison between SVD-MBR and MBR on the WMT22 En$\rightarrow$De. Cells are formatted as \textbf{W / T / L}, indicating the number of sentences where SVD-MBR achieved a higher score (Win), an identical score (Tie), or a lower score (Loss). The rightmost column, $\bar{Z}_{\text{other}}$, aggregates the net win rate across all off-target evaluation metrics. \colorbox{green!15}{Positive} and \colorbox{red!15}{Negative} win rate for SVD-MBR is indicated by the color of the cells.}
\label{tab:svd_wtl_full_wmt22_ende}
\end{table*}

\subsection{Qualitative Analysis}

In \S\ref{subsec:qualitative_analysis}, our analysis focused on the net win-loss differential specifically at the highest reference size~($|\mathcal{Y}|=256$). To provide a comprehensive view of the hypothesis selection dynamics, Table~\ref{tab:svd_wtl_full_wmt22_ende} expands this analysis to include the full Win/Tie/Loss (W/T/L) distributions across all evaluated utility functions, pseudo-reference sizes, and truncation ranks~($k$).

The complete W/T/L distributions reveal a critical characteristic regarding SVD-MBR: it is highly conservative in its interventions. Across most configurations, the vast majority of sentence-level comparison result in a "Tie," indicating that SVD-MBR and MBR ultimately select the exact same hypothesis or hypotheses with identical metric scores. For example, when using COMET as utility function, $|\mathcal{Y}|=256$, and  $k=1$; the methods agreed on approximately 1,780 sentences out of 2,037 sentences on average across all metrics. This high agreement is desirable; it demonstrates that \textbf{SVD-MBR operates strictly as a regularization mechanism, preserving the consensus-finding strengths of MBR while only intervening when the pairwise matrix exhibits exploitable noise}.

When SVD-MBR does override the baseline decision~(resulting in a Win or a Loss), the frequency of these interventions is generally symmetrical for most metrics. The sheer number of Wins and Losses remains relatively close, suggesting that while \textbf{SVD-MBR trades specific local translations, it generally maintains a stable baseline of quality}. However, notable exceptions occur where metric overfitting is most severe. Specifically, when utilizing neural metrics like BLEURT as utility function, $|\mathcal{Y}|=256$, and $k=1$; a pronounced asymmetry emerges. For the BLEURT configuration, SVD-MBR achieved 327 wins against only 235 losses on the aggregate off-target evaluation ($\bar{Z}_{other}$). This shows that with metrics that contain noise, SVD-MBR systematically removes it, resulting in a substantial surplus of "Wins" as the algorithm consistently selects  superior hypotheses over utility function-biased hypotheses.

\begin{table*}[ht!]
\centering
\resizebox{\textwidth}{!}{
\begin{tabular}{p{0.2\linewidth} p{0.15\linewidth} l l l l l l l l}
\toprule
 &  &  & \textbf{BLEU} & \textbf{chrF} & \textbf{BLEURT} & \textbf{COMET} & \textbf{BERTScore} & \textbf{COMETKiwi} & \textbf{$\bar{Z}_{\text{other}}$} \\
\textbf{Decoding Method} & \textbf{Util. Function} & \textbf{$|\mathcal{Y}|$} &  &  &  &  &  &  &  \\
\midrule
MAP$\epsilon$ & N/A & - & 26.284 & 51.041 & 61.281 & 78.400 & 79.477 & 75.565 & 0.313 \\
\cline{1-10} \cline{2-10}
\multirow[t]{15}{*}{MBR} & \multirow[t]{3}{*}{BLEU} & 4 & \cellcolor{red!17} 24.491 (-1.793) & \cellcolor{red!21} 49.240 (-1.801) & \cellcolor{red!24} 59.936 (-1.346) & \cellcolor{red!11} 77.427 (-0.973) & \cellcolor{red!13} 79.056 (-0.420) & \cellcolor{red!24} 74.187 (-1.378) & \cellcolor{red!22} -0.663 (-0.976) \\
 &  & 32 & 26.332 (+0.048) & \cellcolor{red!4} 50.648 (-0.393) & \cellcolor{red!10} 60.723 (-0.558) & \cellcolor{red!1} 78.231 (-0.169) & 79.492 (+0.015) & \cellcolor{red!8} 75.094 (-0.471) & \cellcolor{red!8} -0.048 (-0.361) \\
 &  & 256 & \cellcolor{green!3} 26.590 (+0.306) & 50.983 (-0.058) & \cellcolor{red!7} 60.865 (-0.416) & 78.396 (-0.004) & \cellcolor{green!4} 79.623 (+0.146) & \cellcolor{red!5} 75.265 (-0.300) & \cellcolor{red!5} 0.090 (-0.223) \\
\cline{2-10}
 & \multirow[t]{3}{*}{chrF} & 4 & \cellcolor{red!16} 24.586 (-1.698) & \cellcolor{red!5} 50.614 (-0.427) & \cellcolor{red!14} 60.505 (-0.776) & \cellcolor{red!9} 77.573 (-0.827) & \cellcolor{red!9} 79.170 (-0.307) & \cellcolor{red!17} 74.577 (-0.988) & \cellcolor{red!16} -0.395 (-0.708) \\
 &  & 32 & 26.359 (+0.075) & \cellcolor{green!14} 52.263 (+1.223) & \cellcolor{green!6} 61.624 (+0.343) & \cellcolor{green!5} 78.861 (+0.461) & \cellcolor{green!13} 79.885 (+0.408) & \cellcolor{green!5} 75.869 (+0.304) & \cellcolor{green!3} 0.481 (+0.168) \\
 &  & 256 & \cellcolor{green!4} 26.726 (+0.442) & \cellcolor{green!18} 52.620 (+1.580) & \cellcolor{green!9} 61.807 (+0.526) & \cellcolor{green!6} 78.921 (+0.521) & \cellcolor{green!15} 79.961 (+0.484) & \cellcolor{green!6} 75.915 (+0.350) & \cellcolor{green!6} 0.586 (+0.273) \\
\cline{2-10}
 & \multirow[t]{3}{*}{BLEURT} & 4 & \cellcolor{red!24} 23.838 (-2.446) & \cellcolor{red!20} 49.318 (-1.723) & \cellcolor{green!21} 62.451 (+1.170) & \cellcolor{green!2} 78.596 (+0.196) & \cellcolor{red!4} 79.345 (-0.131) & 75.603 (+0.038) & \cellcolor{red!12} -0.213 (-0.527) \\
 &  & 32 & \cellcolor{red!9} 25.343 (-0.941) & \cellcolor{red!4} 50.677 (-0.364) & \cellcolor{green!41} 63.578 (+2.297) & \cellcolor{green!10} 79.295 (+0.894) & \cellcolor{green!7} 79.725 (+0.249) & \cellcolor{green!12} 76.252 (+0.687) & \cellcolor{green!2} 0.426 (+0.113) \\
 &  & 256 & \cellcolor{red!7} 25.543 (-0.741) & \cellcolor{red!2} 50.802 (-0.239) & \cellcolor{green!46} 63.851 (+2.570) & \cellcolor{green!11} 79.422 (+1.021) & \cellcolor{green!12} 79.873 (+0.396) & \cellcolor{green!15} 76.420 (+0.855) & \cellcolor{green!5} 0.542 (+0.229) \\
\cline{2-10}
 & \multirow[t]{3}{*}{COMET} & 4 & \cellcolor{red!16} 24.643 (-1.641) & \cellcolor{red!13} 49.911 (-1.130) & \cellcolor{green!8} 61.730 (+0.449) & \cellcolor{green!38} 81.730 (+3.330) & \cellcolor{green!25} 80.274 (+0.797) & \cellcolor{green!33} 77.467 (+1.902) & \cellcolor{green!5} 0.543 (+0.230) \\
 &  & 32 & \cellcolor{red!8} 25.396 (-0.888) & \cellcolor{red!3} 50.738 (-0.303) & \cellcolor{green!18} 62.324 (+1.042) & \cellcolor{green!47} 82.505 (+4.104) & \cellcolor{green!36} 80.625 (+1.149) & \cellcolor{green!45} 78.099 (+2.534) & \cellcolor{green!16} 1.006 (+0.693) \\
 &  & 256 & \cellcolor{red!4} 25.846 (-0.438) & 51.086 (+0.045) & \cellcolor{green!21} 62.449 (+1.168) & \cellcolor{green!48} 82.599 (+4.198) & \cellcolor{green!36} 80.635 (+1.159) & \cellcolor{green!45} 78.118 (+2.553) & \cellcolor{green!19} 1.135 (+0.822) \\
\cline{2-10}
 & \multirow[t]{3}{*}{BERTScore} & 4 & \cellcolor{red!23} 23.910 (-2.374) & \cellcolor{red!22} 49.154 (-1.887) & \cellcolor{red!20} 60.145 (-1.136) & \cellcolor{red!5} 77.929 (-0.471) & \cellcolor{green!6} 79.689 (+0.212) & \cellcolor{red!12} 74.879 (-0.686) & \cellcolor{red!18} -0.483 (-0.796) \\
 &  & 32 & \cellcolor{red!8} 25.407 (-0.877) & \cellcolor{red!6} 50.479 (-0.562) & \cellcolor{red!4} 61.017 (-0.264) & \cellcolor{green!3} 78.741 (+0.340) & \cellcolor{green!29} 80.389 (+0.913) & \cellcolor{green!3} 75.752 (+0.187) & \cellcolor{red!2} 0.219 (-0.094) \\
 &  & 256 & \cellcolor{red!7} 25.576 (-0.708) & \cellcolor{red!3} 50.782 (-0.259) & 61.251 (-0.030) & \cellcolor{green!5} 78.892 (+0.491) & \cellcolor{green!34} 80.554 (+1.077) & \cellcolor{green!4} 75.812 (+0.247) & 0.340 (+0.027) \\
\cline{1-10} \cline{2-10}
\multirow[t]{15}{*}{Probabilistic MBR} & \multirow[t]{3}{*}{BLEU} & 4 & \cellcolor{red!29} 23.280 (-3.004) & \cellcolor{red!33} 48.254 (-2.787) & \cellcolor{red!39} 59.120 (-2.161) & \cellcolor{red!24} 76.333 (-2.068) & \cellcolor{red!38} 78.285 (-1.192) & \cellcolor{red!43} 73.118 (-2.448) & \cellcolor{red!38} -1.360 (-1.673) \\
 &  & 32 & \cellcolor{red!14} 24.849 (-1.435) & \cellcolor{red!19} 49.428 (-1.613) & \cellcolor{red!26} 59.834 (-1.447) & \cellcolor{red!15} 77.072 (-1.328) & \cellcolor{red!21} 78.817 (-0.659) & \cellcolor{red!29} 73.892 (-1.674) & \cellcolor{red!25} -0.788 (-1.101) \\
 &  & 256 & \cellcolor{red!6} 25.636 (-0.648) & \cellcolor{red!9} 50.253 (-0.788) & \cellcolor{red!14} 60.478 (-0.803) & \cellcolor{red!8} 77.708 (-0.692) & \cellcolor{red!6} 79.271 (-0.205) & \cellcolor{red!19} 74.447 (-1.118) & \cellcolor{red!15} -0.333 (-0.646) \\
\cline{2-10}
 & \multirow[t]{3}{*}{chrF} & 4 & \cellcolor{red!35} 22.695 (-3.589) & \cellcolor{red!31} 48.425 (-2.615) & \cellcolor{red!39} 59.085 (-2.196) & \cellcolor{red!25} 76.187 (-2.214) & \cellcolor{red!42} 78.145 (-1.332) & \cellcolor{red!45} 73.020 (-2.545) & \cellcolor{red!40} -1.449 (-1.762) \\
 &  & 32 & \cellcolor{red!26} 23.656 (-2.628) & \cellcolor{red!15} 49.743 (-1.298) & \cellcolor{red!27} 59.770 (-1.511) & \cellcolor{red!15} 77.025 (-1.375) & \cellcolor{red!25} 78.689 (-0.787) & \cellcolor{red!29} 73.895 (-1.670) & \cellcolor{red!27} -0.887 (-1.200) \\
 &  & 256 & \cellcolor{red!20} 24.266 (-2.018) & \cellcolor{red!6} 50.506 (-0.535) & \cellcolor{red!17} 60.301 (-0.980) & \cellcolor{red!11} 77.442 (-0.958) & \cellcolor{red!16} 78.954 (-0.522) & \cellcolor{red!23} 74.220 (-1.345) & \cellcolor{red!20} -0.580 (-0.893) \\
\cline{2-10}
 & \multirow[t]{3}{*}{BLEURT} & 4 & \cellcolor{red!50} 21.255 (-5.029) & \cellcolor{red!49} 46.917 (-4.124) & \cellcolor{red!35} 59.314 (-1.967) & \cellcolor{red!24} 76.254 (-2.146) & \cellcolor{red!50} 77.910 (-1.566) & \cellcolor{red!43} 73.119 (-2.447) & \cellcolor{red!49} -1.819 (-2.132) \\
 &  & 32 & \cellcolor{red!24} 23.781 (-2.503) & \cellcolor{red!21} 49.260 (-1.781) & \cellcolor{green!7} 61.716 (+0.434) & \cellcolor{red!2} 78.200 (-0.200) & \cellcolor{red!8} 79.214 (-0.262) & \cellcolor{red!8} 75.081 (-0.484) & \cellcolor{red!15} -0.374 (-0.687) \\
 &  & 256 & \cellcolor{red!9} 25.323 (-0.961) & \cellcolor{red!5} 50.612 (-0.429) & \cellcolor{green!40} 63.497 (+2.216) & \cellcolor{green!8} 79.143 (+0.743) & \cellcolor{green!9} 79.774 (+0.298) & \cellcolor{green!11} 76.221 (+0.655) & \cellcolor{green!2} 0.406 (+0.093) \\
\cline{2-10}
 & \multirow[t]{3}{*}{COMET} & 4 & \cellcolor{red!46} 21.605 (-4.679) & \cellcolor{red!46} 47.101 (-3.939) & \cellcolor{red!41} 59.000 (-2.281) & \cellcolor{red!15} 77.077 (-1.324) & \cellcolor{red!43} 78.110 (-1.366) & \cellcolor{red!36} 73.515 (-2.050) & \cellcolor{red!46} -1.670 (-1.984) \\
 &  & 32 & \cellcolor{red!36} 22.646 (-3.639) & \cellcolor{red!33} 48.262 (-2.779) & \cellcolor{red!19} 60.229 (-1.052) & \cellcolor{green!11} 79.408 (+1.008) & \cellcolor{red!10} 79.148 (-0.328) & 75.548 (-0.018) & \cellcolor{red!23} -0.682 (-0.995) \\
 &  & 256 & \cellcolor{red!4} 25.857 (-0.427) & 51.061 (+0.021) & \cellcolor{green!20} 62.423 (+1.141) & \cellcolor{green!44} 82.242 (+3.841) & \cellcolor{green!34} 80.568 (+1.091) & \cellcolor{green!42} 77.958 (+2.393) & \cellcolor{green!18} 1.092 (+0.779) \\
\cline{2-10}
 & \multirow[t]{3}{*}{BERTScore} & 4 & \cellcolor{red!47} 21.477 (-4.807) & \cellcolor{red!50} 46.839 (-4.202) & \cellcolor{red!50} 58.535 (-2.746) & \cellcolor{red!28} 75.970 (-2.430) & \cellcolor{red!44} 78.088 (-1.388) & \cellcolor{red!50} 72.760 (-2.806) & \cellcolor{red!50} -1.838 (-2.151) \\
 &  & 32 & \cellcolor{red!40} 22.241 (-4.043) & \cellcolor{red!39} 47.718 (-3.322) & \cellcolor{red!36} 59.301 (-1.981) & \cellcolor{red!18} 76.824 (-1.576) & \cellcolor{red!22} 78.782 (-0.695) & \cellcolor{red!30} 73.852 (-1.713) & \cellcolor{red!36} -1.273 (-1.586) \\
 &  & 256 & \cellcolor{red!9} 25.359 (-0.925) & \cellcolor{red!8} 50.363 (-0.678) & \cellcolor{red!2} 61.143 (-0.138) & \cellcolor{green!4} 78.764 (+0.364) & \cellcolor{green!27} 80.330 (+0.854) & \cellcolor{green!2} 75.720 (+0.155) & \cellcolor{red!2} 0.213 (-0.100) \\
\cline{1-10} \cline{2-10}
\multirow[t]{5}{*}{Model-based MBR} & BLEU & 256 & \cellcolor{green!1} 26.468 (+0.184) & 50.997 (-0.044) & \cellcolor{red!8} 60.809 (-0.473) & 78.372 (-0.028) & \cellcolor{green!7} 79.701 (+0.225) & \cellcolor{red!6} 75.211 (-0.354) & \cellcolor{red!5} 0.093 (-0.220) \\
\cline{2-10}
 & chrF & 256 & 26.236 (-0.048) & \cellcolor{green!19} 52.665 (+1.624) & \cellcolor{green!6} 61.659 (+0.377) & \cellcolor{green!3} 78.723 (+0.323) & \cellcolor{green!14} 79.937 (+0.461) & \cellcolor{green!3} 75.765 (+0.200) & \cellcolor{green!3} 0.453 (+0.140) \\
\cline{2-10}
 & BLEURT & 256 & \cellcolor{red!12} 24.993 (-1.291) & \cellcolor{red!6} 50.481 (-0.560) & \cellcolor{green!49} 64.023 (+2.742) & \cellcolor{green!11} 79.380 (+0.979) & \cellcolor{green!11} 79.826 (+0.349) & \cellcolor{green!15} 76.455 (+0.890) & \cellcolor{green!2} 0.414 (+0.101) \\
\cline{2-10}
 & COMET & 256 & \cellcolor{red!9} 25.334 (-0.950) & \cellcolor{red!1} 50.928 (-0.113) & \cellcolor{green!21} 62.471 (+1.190) & \cellcolor{green!50} 82.704 (+4.303) & \cellcolor{green!37} 80.661 (+1.184) & \cellcolor{green!46} 78.186 (+2.621) & \cellcolor{green!17} 1.065 (+0.751) \\
\cline{2-10}
 & BERTScore & 256 & \cellcolor{red!11} 25.119 (-1.165) & \cellcolor{red!4} 50.652 (-0.388) & \cellcolor{red!1} 61.194 (-0.087) & \cellcolor{green!5} 78.912 (+0.512) & \cellcolor{green!37} 80.643 (+1.167) & \cellcolor{green!5} 75.891 (+0.326) & \cellcolor{red!1} 0.267 (-0.046) \\
\cline{1-10} \cline{2-10}
\textit{Oracle} & N/A & - & \textit{45.472} & \textit{65.771} & \textit{70.205} & \textit{85.631} & \textit{85.226} & \textit{81.385} & \textit{4.001} \\
\cline{1-10} \cline{2-10}
\bottomrule
\end{tabular}

}
\caption{Performance of decoding variants on the WMT22 De$\rightarrow$En ($|\mathcal{H}|=256$). Absolute differences (in parentheses) are relative to the \texttt{MAP$\epsilon$} baseline. Color gradation indicates the relative magnitude of \colorbox{green!15}{improvement} or \colorbox{red!15}{degradation}.}
\label{tab:mbr_overfitting_wmt22_deen}
\end{table*}

\begin{table*}[ht!]
\centering
\resizebox{\textwidth}{!}{
\begin{tabular}{llllllllll}
\toprule
\textbf{Decoding Method} & \textbf{Util. Function} & \textbf{$|\mathcal{Y}|$} & \textbf{BLEU} & \textbf{chrF} & \textbf{BLEURT} & \textbf{COMET} & \textbf{BERTScore} & \textbf{COMETKiwi} & \textbf{$\bar{Z}_{\text{other}}$} \\
\midrule
\multirow[t]{15}{*}{MBR} & \multirow[t]{3}{*}{BLEU} & 4 & 24.491 & 49.240 & 59.936 & 77.427 & 79.056 & 74.187 & -1.449 \\
 &  & 32 & 26.332 & 50.648 & 60.723 & 78.231 & 79.492 & 75.094 & -0.576 \\
 &  & 256 & 26.590 & 50.983 & 60.865 & 78.396 & 79.623 & 75.265 & -0.377 \\
\cline{2-10}
 & \multirow[t]{3}{*}{chrF} & 4 & 24.586 & 50.614 & 60.505 & 77.573 & 79.170 & 74.577 & -1.097 \\
 &  & 32 & 26.359 & 52.263 & 61.624 & 78.861 & 79.885 & 75.869 & 0.164 \\
 &  & 256 & 26.726 & 52.620 & 61.807 & 78.921 & 79.961 & 75.915 & 0.324 \\
\cline{2-10}
 & \multirow[t]{3}{*}{BLEURT} & 4 & 23.838 & 49.318 & 62.451 & 78.596 & 79.345 & 75.603 & -0.965 \\
 &  & 32 & 25.343 & 50.677 & 63.578 & 79.295 & 79.725 & 76.252 & 0.005 \\
 &  & 256 & 25.543 & 50.802 & 63.851 & 79.422 & 79.873 & 76.420 & 0.177 \\
\cline{2-10}
 & \multirow[t]{3}{*}{COMET} & 4 & 24.643 & 49.911 & 61.730 & 81.730 & 80.274 & 77.467 & 0.133 \\
 &  & 32 & 25.396 & 50.738 & 62.324 & 82.505 & 80.625 & 78.099 & 0.832 \\
 &  & 256 & 25.846 & 51.086 & 62.449 & 82.599 & 80.635 & 78.118 & 1.036 \\
\cline{2-10}
 & \multirow[t]{3}{*}{BERTScore} & 4 & 23.910 & 49.154 & 60.145 & 77.929 & 79.689 & 74.879 & -1.253 \\
 &  & 32 & 25.407 & 50.479 & 61.017 & 78.741 & 80.389 & 75.752 & -0.224 \\
 &  & 256 & 25.576 & 50.782 & 61.251 & 78.892 & 80.554 & 75.812 & -0.049 \\
\cline{1-10} \cline{2-10}
\multirow[t]{15}{*}{SVD-MBR ($k=1$)} & \multirow[t]{3}{*}{BLEU} & 4 & \cellcolor{green!10} 24.577 (+0.09) & \cellcolor{green!8} 49.359 (+0.12) & \cellcolor{red!2} 59.899 (-0.04) & \cellcolor{red!8} 77.332 (-0.10) & \cellcolor{red!8} 79.027 (-0.03) & \cellcolor{green!2} 74.214 (+0.03) & -1.449 (-0.000) \\
 &  & 32 & \cellcolor{green!10} 26.417 (+0.08) & \cellcolor{green!10} 50.794 (+0.15) & \cellcolor{green!2} 60.765 (+0.04) & \cellcolor{red!6} 78.153 (-0.08) & \cellcolor{green!12} 79.532 (+0.04) & \cellcolor{red!6} 75.024 (-0.07) & \cellcolor{green!4} -0.543 (+0.032) \\
 &  & 256 & \cellcolor{green!16} 26.730 (+0.14) & \cellcolor{green!6} 51.079 (+0.10) & \cellcolor{green!10} 61.016 (+0.15)$^*$ & \cellcolor{red!3} 78.353 (-0.04) & \cellcolor{green!10} 79.659 (+0.04) & \cellcolor{red!3} 75.227 (-0.04) & \cellcolor{green!6} -0.327 (+0.050) \\
\cline{2-10}
 & \multirow[t]{3}{*}{chrF} & 4 & \cellcolor{red!10} 24.497 (-0.09) & \cellcolor{red!32} 50.172 (-0.44) & \cellcolor{red!6} 60.414 (-0.09) & \cellcolor{green!10} 77.694 (+0.12) & \cellcolor{red!10} 79.133 (-0.04) & 74.575 (-0.00) & \cellcolor{red!4} -1.133 (-0.036) \\
 &  & 32 & \cellcolor{red!33} 26.085 (-0.27) & \cellcolor{red!50} 51.576 (-0.69) & \cellcolor{red!25} 61.255 (-0.37) & \cellcolor{red!25} 78.559 (-0.30) & \cellcolor{red!36} 79.761 (-0.12) & \cellcolor{red!24} 75.617 (-0.25) & \cellcolor{red!35} -0.095 (-0.259) \\
 &  & 256 & \cellcolor{red!4} 26.693 (-0.03) & \cellcolor{red!46} 51.977 (-0.64) & \cellcolor{red!22} 61.479 (-0.33) & \cellcolor{red!11} 78.789 (-0.13) & \cellcolor{red!26} 79.871 (-0.09) & \cellcolor{red!15} 75.752 (-0.16) & \cellcolor{red!20} 0.176 (-0.148) \\
\cline{2-10}
 & \multirow[t]{3}{*}{BLEURT} & 4 & \cellcolor{green!10} 23.925 (+0.09) & \cellcolor{green!1} 49.333 (+0.02) & \cellcolor{red!49} 61.724 (-0.73) & \cellcolor{red!11} 78.457 (-0.14) & \cellcolor{red!5} 79.328 (-0.02) & \cellcolor{red!21} 75.375 (-0.23) & \cellcolor{red!5} -1.006 (-0.041) \\
 &  & 32 & \cellcolor{green!3} 25.372 (+0.03) & \cellcolor{red!5} 50.601 (-0.08) & \cellcolor{red!49} 62.844 (-0.73) & \cellcolor{green!1} 79.310 (+0.01) & \cellcolor{green!26} 79.815 (+0.09)$^*$ & \cellcolor{green!5} 76.311 (+0.06) & \cellcolor{green!5} 0.041 (+0.037) \\
 &  & 256 & \cellcolor{green!16} 25.676 (+0.13) & \cellcolor{green!6} 50.891 (+0.09) & \cellcolor{red!50} 63.113 (-0.74) & \cellcolor{red!8} 79.318 (-0.10) & \cellcolor{green!12} 79.916 (+0.04) & 76.415 (-0.00) & \cellcolor{green!7} 0.229 (+0.051) \\
\cline{2-10}
 & \multirow[t]{3}{*}{COMET} & 4 & \cellcolor{red!35} 24.350 (-0.29) & \cellcolor{red!19} 49.642 (-0.27) & \cellcolor{red!14} 61.518 (-0.21) & \cellcolor{red!50} 81.140 (-0.59) & \cellcolor{red!48} 80.111 (-0.16) & \cellcolor{red!26} 77.190 (-0.28) & \cellcolor{red!37} -0.139 (-0.272) \\
 &  & 32 & \cellcolor{green!7} 25.461 (+0.06) & \cellcolor{green!10} 50.876 (+0.14) & \cellcolor{red!8} 62.193 (-0.13) & \cellcolor{red!32} 82.127 (-0.38) & \cellcolor{red!4} 80.610 (-0.02) & \cellcolor{red!15} 77.933 (-0.17) & \cellcolor{red!2} 0.816 (-0.015) \\
 &  & 256 & \cellcolor{green!1} 25.861 (+0.01) & 51.099 (+0.01) & \cellcolor{red!5} 62.361 (-0.09) & \cellcolor{red!27} 82.274 (-0.32) & 80.638 (+0.00) & \cellcolor{red!7} 78.043 (-0.08) & \cellcolor{red!3} 1.014 (-0.022) \\
\cline{2-10}
 & \multirow[t]{3}{*}{BERTScore} & 4 & \cellcolor{green!14} 24.033 (+0.12) & \cellcolor{red!3} 49.103 (-0.05) & \cellcolor{green!11} 60.319 (+0.17)$^*$ & 77.925 (-0.00) & \cellcolor{red!50} 79.522 (-0.17) & \cellcolor{green!3} 74.920 (+0.04) & \cellcolor{green!7} -1.198 (+0.055) \\
 &  & 32 & \cellcolor{green!42} 25.760 (+0.35)$^*$ & \cellcolor{green!28} 50.869 (+0.39)$^*$ & \cellcolor{green!19} 61.310 (+0.29)$^*$ & \cellcolor{green!37} 79.178 (+0.44)$^*$ & \cellcolor{green!15} 80.440 (+0.05) & \cellcolor{green!50} 76.274 (+0.52)$^*$ & \cellcolor{green!50} 0.137 (+0.361)$^*$ \\
 &  & 256 & \cellcolor{green!50} 25.990 (+0.41)$^*$ & \cellcolor{green!21} 51.081 (+0.30)$^*$ & \cellcolor{green!17} 61.505 (+0.25)$^*$ & \cellcolor{green!35} 79.311 (+0.42)$^*$ & \cellcolor{red!4} 80.538 (-0.02) & \cellcolor{green!48} 76.320 (+0.51)$^*$ & \cellcolor{green!47} 0.294 (+0.343)$^*$ \\
\cline{1-10} \cline{2-10}
\multirow[t]{15}{*}{SVD-MBR ($k=2$)} & \multirow[t]{3}{*}{BLEU} & 4 & \cellcolor{red!41} 24.143 (-0.35) & \cellcolor{red!26} 48.881 (-0.36) & \cellcolor{red!5} 59.849 (-0.09) & \cellcolor{red!7} 77.340 (-0.09) & \cellcolor{red!22} 78.980 (-0.08) & \cellcolor{red!5} 74.134 (-0.05) & \cellcolor{red!19} -1.591 (-0.143) \\
 &  & 32 & \cellcolor{red!19} 26.173 (-0.16) & \cellcolor{red!6} 50.565 (-0.08) & \cellcolor{green!2} 60.765 (+0.04) & \cellcolor{red!2} 78.200 (-0.03) & \cellcolor{green!3} 79.503 (+0.01) & \cellcolor{red!3} 75.056 (-0.04) & \cellcolor{red!2} -0.592 (-0.016) \\
 &  & 256 & \cellcolor{red!15} 26.462 (-0.13) & \cellcolor{red!4} 50.918 (-0.07) & \cellcolor{green!3} 60.920 (+0.06) & 78.399 (+0.00) & \cellcolor{green!3} 79.634 (+0.01) & \cellcolor{green!3} 75.303 (+0.04) & \cellcolor{green!1} -0.370 (+0.008) \\
\cline{2-10}
 & \multirow[t]{3}{*}{chrF} & 4 & \cellcolor{red!7} 24.527 (-0.06) & \cellcolor{red!9} 50.481 (-0.13) & \cellcolor{green!2} 60.547 (+0.04) & \cellcolor{green!2} 77.602 (+0.03) & \cellcolor{red!2} 79.161 (-0.01) & \cellcolor{green!3} 74.617 (+0.04) & -1.095 (+0.002) \\
 &  & 32 & \cellcolor{red!8} 26.293 (-0.07) & \cellcolor{red!8} 52.146 (-0.12) & \cellcolor{red!6} 61.524 (-0.10) & \cellcolor{red!7} 78.774 (-0.09) & \cellcolor{red!19} 79.819 (-0.07) & \cellcolor{red!6} 75.797 (-0.07) & \cellcolor{red!11} 0.082 (-0.082) \\
 &  & 256 & \cellcolor{green!10} 26.817 (+0.09) & \cellcolor{red!3} 52.577 (-0.04) & \cellcolor{red!2} 61.777 (-0.03) & \cellcolor{red!9} 78.810 (-0.11) & \cellcolor{red!4} 79.945 (-0.02) & \cellcolor{red!9} 75.821 (-0.09) & \cellcolor{red!3} 0.302 (-0.022) \\
\cline{2-10}
 & \multirow[t]{3}{*}{BLEURT} & 4 & \cellcolor{red!12} 23.737 (-0.10) & 49.326 (+0.01) & \cellcolor{red!11} 62.288 (-0.16) & \cellcolor{red!6} 78.522 (-0.07) & \cellcolor{red!23} 79.267 (-0.08) & \cellcolor{red!3} 75.570 (-0.03) & \cellcolor{red!9} -1.031 (-0.066) \\
 &  & 32 & \cellcolor{green!11} 25.436 (+0.09) & \cellcolor{green!6} 50.761 (+0.08) & \cellcolor{red!3} 63.524 (-0.05) & 79.303 (+0.01) & \cellcolor{green!7} 79.749 (+0.02) & \cellcolor{green!9} 76.355 (+0.10)$^*$ & \cellcolor{green!9} 0.071 (+0.066) \\
 &  & 256 & \cellcolor{green!4} 25.579 (+0.04) & \cellcolor{green!6} 50.892 (+0.09) & \cellcolor{red!3} 63.796 (-0.06) & 79.420 (-0.00) & \cellcolor{green!2} 79.882 (+0.01) & \cellcolor{green!8} 76.505 (+0.09) & \cellcolor{green!6} 0.223 (+0.045) \\
\cline{2-10}
 & \multirow[t]{3}{*}{COMET} & 4 & \cellcolor{red!7} 24.579 (-0.06) & 49.904 (-0.01) & \cellcolor{red!2} 61.693 (-0.04) & \cellcolor{red!7} 81.637 (-0.09) & \cellcolor{red!3} 80.262 (-0.01) & \cellcolor{red!6} 77.396 (-0.07) & \cellcolor{red!5} 0.094 (-0.040) \\
 &  & 32 & \cellcolor{green!14} 25.518 (+0.12) & \cellcolor{green!9} 50.862 (+0.12) & \cellcolor{green!2} 62.360 (+0.04) & \cellcolor{green!1} 82.527 (+0.02) & \cellcolor{green!21} 80.699 (+0.07)$^*$ & \cellcolor{green!7} 78.175 (+0.08)$^*$ & \cellcolor{green!14} 0.933 (+0.102)$^*$ \\
 &  & 256 & \cellcolor{red!3} 25.817 (-0.03) & \cellcolor{green!5} 51.156 (+0.07) & \cellcolor{green!5} 62.529 (+0.08)$^*$ & \cellcolor{green!4} 82.651 (+0.05)$^*$ & \cellcolor{green!31} 80.742 (+0.11)$^*$ & \cellcolor{green!8} 78.211 (+0.09)$^*$ & \cellcolor{green!11} 1.116 (+0.081)$^*$ \\
\cline{2-10}
 & \multirow[t]{3}{*}{BERTScore} & 4 & \cellcolor{green!9} 23.991 (+0.08) & \cellcolor{red!5} 49.081 (-0.07) & \cellcolor{green!3} 60.197 (+0.05) & \cellcolor{green!4} 77.981 (+0.05) & \cellcolor{red!18} 79.626 (-0.06) & \cellcolor{green!3} 74.914 (+0.04) & \cellcolor{green!3} -1.228 (+0.024) \\
 &  & 32 & \cellcolor{green!7} 25.472 (+0.07) & \cellcolor{green!3} 50.525 (+0.05) & \cellcolor{green!3} 61.062 (+0.05) & \cellcolor{green!1} 78.757 (+0.02) & \cellcolor{red!8} 80.359 (-0.03) & \cellcolor{green!11} 75.870 (+0.12)$^*$ & \cellcolor{green!7} -0.169 (+0.055) \\
 &  & 256 & \cellcolor{green!31} 25.839 (+0.26)$^*$ & \cellcolor{green!10} 50.928 (+0.15)$^*$ & \cellcolor{green!3} 61.308 (+0.06) & \cellcolor{green!15} 79.072 (+0.18)$^*$ & \cellcolor{red!2} 80.545 (-0.01) & \cellcolor{green!18} 76.004 (+0.19)$^*$ & \cellcolor{green!21} 0.107 (+0.156)$^*$ \\
\cline{1-10} \cline{2-10}
\multirow[t]{15}{*}{SVD-MBR ($k=3$)} & \multirow[t]{3}{*}{BLEU} & 4 & \cellcolor{red!30} 24.237 (-0.25) & \cellcolor{red!13} 49.056 (-0.18) & \cellcolor{red!8} 59.806 (-0.13) & \cellcolor{red!12} 77.279 (-0.15) & \cellcolor{red!11} 79.016 (-0.04) & \cellcolor{red!9} 74.083 (-0.10) & \cellcolor{red!15} -1.564 (-0.115) \\
 &  & 32 & \cellcolor{red!33} 26.054 (-0.28) & \cellcolor{red!8} 50.526 (-0.12) & \cellcolor{red!4} 60.663 (-0.06) & \cellcolor{red!3} 78.195 (-0.04) & \cellcolor{red!8} 79.462 (-0.03) & \cellcolor{green!4} 75.138 (+0.04) & \cellcolor{red!6} -0.622 (-0.046) \\
 &  & 256 & \cellcolor{red!8} 26.519 (-0.07) & \cellcolor{red!1} 50.962 (-0.02) & \cellcolor{red!1} 60.845 (-0.02) & \cellcolor{green!1} 78.412 (+0.02) & \cellcolor{red!6} 79.602 (-0.02) & 75.274 (+0.01) & \cellcolor{red!1} -0.390 (-0.013) \\
\cline{2-10}
 & \multirow[t]{3}{*}{chrF} & 4 & \cellcolor{red!13} 24.474 (-0.11) & \cellcolor{red!7} 50.517 (-0.10) & \cellcolor{red!2} 60.472 (-0.03) & \cellcolor{red!3} 77.528 (-0.04) & \cellcolor{red!10} 79.135 (-0.03) & \cellcolor{red!1} 74.565 (-0.01) & \cellcolor{red!7} -1.149 (-0.052) \\
 &  & 32 & 26.361 (+0.00) & 52.259 (-0.00) & 61.613 (-0.01) & \cellcolor{red!1} 78.847 (-0.01) & \cellcolor{green!4} 79.898 (+0.01) & \cellcolor{green!6} 75.939 (+0.07) & \cellcolor{green!1} 0.178 (+0.014) \\
 &  & 256 & \cellcolor{red!3} 26.693 (-0.03) & \cellcolor{green!1} 52.645 (+0.02) & \cellcolor{red!2} 61.766 (-0.04) & \cellcolor{red!9} 78.810 (-0.11) & \cellcolor{red!8} 79.931 (-0.03) & \cellcolor{red!6} 75.847 (-0.07) & \cellcolor{red!7} 0.272 (-0.052) \\
\cline{2-10}
 & \multirow[t]{3}{*}{BLEURT} & 4 & \cellcolor{green!7} 23.900 (+0.06) & \cellcolor{green!5} 49.392 (+0.07) & 62.448 (-0.00) & 78.588 (-0.01) & \cellcolor{red!3} 79.334 (-0.01) & \cellcolor{red!4} 75.552 (-0.05) & \cellcolor{green!2} -0.950 (+0.015) \\
 &  & 32 & \cellcolor{green!13} 25.455 (+0.11)$^*$ & \cellcolor{green!6} 50.768 (+0.09) & \cellcolor{green!2} 63.616 (+0.04) & \cellcolor{green!5} 79.363 (+0.07)$^*$ & \cellcolor{green!14} 79.772 (+0.05)$^*$ & \cellcolor{green!7} 76.327 (+0.08)$^*$ & \cellcolor{green!11} 0.088 (+0.084)$^*$ \\
 &  & 256 & \cellcolor{red!1} 25.526 (-0.02) & \cellcolor{green!3} 50.847 (+0.05) & \cellcolor{red!3} 63.801 (-0.05) & \cellcolor{green!2} 79.447 (+0.03) & \cellcolor{red!1} 79.868 (-0.00) & \cellcolor{green!3} 76.461 (+0.04) & \cellcolor{green!2} 0.192 (+0.015) \\
\cline{2-10}
 & \multirow[t]{3}{*}{COMET} & 4 & \cellcolor{green!10} 24.731 (+0.09) & \cellcolor{green!7} 50.007 (+0.10)$^*$ & 61.740 (+0.01) & \cellcolor{red!2} 81.705 (-0.03) & 80.274 (+0.00) & \cellcolor{red!2} 77.437 (-0.03) & \cellcolor{green!5} 0.170 (+0.037) \\
 &  & 32 & \cellcolor{green!2} 25.416 (+0.02) & \cellcolor{green!4} 50.805 (+0.07) & \cellcolor{green!3} 62.377 (+0.05) & \cellcolor{green!3} 82.545 (+0.04) & \cellcolor{green!15} 80.678 (+0.05)$^*$ & \cellcolor{green!6} 78.170 (+0.07)$^*$ & \cellcolor{green!8} 0.893 (+0.061)$^*$ \\
 &  & 256 & 25.840 (-0.01) & \cellcolor{green!2} 51.114 (+0.03) & \cellcolor{green!2} 62.481 (+0.03) & \cellcolor{green!4} 82.650 (+0.05)$^*$ & \cellcolor{green!15} 80.687 (+0.05)$^*$ & \cellcolor{green!5} 78.180 (+0.06)$^*$ & \cellcolor{green!5} 1.077 (+0.041)$^*$ \\
\cline{2-10}
 & \multirow[t]{3}{*}{BERTScore} & 4 & \cellcolor{green!11} 24.007 (+0.10)$^*$ & \cellcolor{green!3} 49.195 (+0.04) & \cellcolor{green!1} 60.166 (+0.02) & 77.920 (-0.01) & \cellcolor{red!1} 79.683 (-0.01) & \cellcolor{red!3} 74.838 (-0.04) & \cellcolor{green!3} -1.227 (+0.026) \\
 &  & 32 & \cellcolor{green!5} 25.456 (+0.05) & \cellcolor{green!3} 50.533 (+0.05) & \cellcolor{green!2} 61.058 (+0.04) & \cellcolor{green!5} 78.805 (+0.06) & \cellcolor{red!2} 80.381 (-0.01) & \cellcolor{green!12} 75.883 (+0.13)$^*$ & \cellcolor{green!8} -0.163 (+0.061)$^*$ \\
 &  & 256 & \cellcolor{green!3} 25.605 (+0.03) & \cellcolor{green!6} 50.875 (+0.09) & 61.253 (+0.00) & \cellcolor{green!3} 78.935 (+0.04) & 80.552 (-0.00) & \cellcolor{green!11} 75.931 (+0.12)$^*$ & \cellcolor{green!7} 0.003 (+0.053)$^*$ \\
\cline{1-10} \cline{2-10}
\textit{Oracle} & N/A & - & \textit{45.472} & \textit{65.771} & \textit{70.205} & \textit{85.631} & \textit{85.226} & \textit{81.385} & \textit{5.528} \\
\cline{1-10} \cline{2-10}
\bottomrule
\end{tabular}

}
\caption{Performance of MBR vs. SVD-MBR on WMT22 De$\rightarrow$En ($|\mathcal{H}|=256$). $\bar{Z}_{\text{other}}$ represents the mean z-score of all non-utility function evaluation metrics. Values in parentheses are absolute differences against MBR, with colors indicating relative \colorbox{green!15}{improvement} or \colorbox{red!15}{degradation}. Statistical significance ($p < 0.05$) is marked with $^*$.}
\label{tab:svd_mbr_results_wmt22_deen}
\end{table*}

\begin{table*}[ht!]
\centering
\resizebox{\textwidth}{!}{
\begin{NiceTabular}{@{} lll c c c c c c c c c c c c c c @{}}
\toprule
\multirow{2}{*}{\textbf{Util. Function}} & \multirow{2}{*}{\textbf{$|\mathcal{Y}|$}} & \multirow{2}{*}{\textbf{Top-$k$}} & \multicolumn{2}{c}{\makecell{\textbf{BLEU} \\ ($\sigma^2=96.972$)}} & \multicolumn{2}{c}{\makecell{\textbf{chrF} \\ ($\sigma^2=76.595$)}} & \multicolumn{2}{c}{\makecell{\textbf{BLEURT} \\ ($\sigma^2=34.042$)}} & \multicolumn{2}{c}{\makecell{\textbf{COMET} \\ ($\sigma^2=39.038$)}} & \multicolumn{2}{c}{\makecell{\textbf{BERTScore} \\ ($\sigma^2=16.240$)}} & \multicolumn{2}{c}{\makecell{\textbf{COMETKiwi} \\ ($\sigma^2=44.653$)}} & \multicolumn{2}{c}{\makecell{\textbf{$\bar{Z}_{\text{other}}$} \\ ($\sigma^2=0.174$)}} \\
\cmidrule(lr){4-5} \cmidrule(lr){6-7} \cmidrule(lr){8-9} \cmidrule(lr){10-11} \cmidrule(lr){12-13} \cmidrule(lr){14-15} \cmidrule(lr){16-17}
 &  &  & \textit{Dir.} & \textit{Mag.} & \textit{Dir.} & \textit{Mag.} & \textit{Dir.} & \textit{Mag.} & \textit{Dir.} & \textit{Mag.} & \textit{Dir.} & \textit{Mag.} & \textit{Dir.} & \textit{Mag.} & \textit{Dir.} & \textit{Mag.} \\
\midrule
\multirow{9}{*}{\textbf{BLEU}} & \multirow{3}{*}{4} & $k=1$ & \cellcolor{green!4} $+0.03$ & \cellcolor{blue!11} $0.08^{*}$ & \cellcolor{green!3} $+0.02$ & \cellcolor{blue!2} $0.01$ & \cellcolor{green!2} $+0.02$ & \cellcolor{blue!1} $0.01$ & $+0.00$ & \cellcolor{blue!10} $0.07^{*}$ & \cellcolor{green!10} $+0.07^{*}$ & \cellcolor{blue!2} $0.02$ & \cellcolor{green!4} $+0.03$ & \cellcolor{blue!13} $0.09^{*}$ & \cellcolor{green!6} $+0.04$ & \cellcolor{blue!2} $0.02$ \\
 &  & $k=2$ & \cellcolor{red!4} $-0.03$ & \cellcolor{blue!1} $0.01$ & \cellcolor{red!1} $-0.01$ & \cellcolor{blue!3} $0.03$ & \cellcolor{red!2} $-0.01$ & \cellcolor{blue!2} $0.02$ & \cellcolor{green!1} $+0.01$ & \cellcolor{blue!6} $0.04^{*}$ & \cellcolor{green!8} $+0.06^{*}$ & \cellcolor{blue!3} $0.02$ & \cellcolor{green!1} $+0.01$ & \cellcolor{blue!8} $0.06^{*}$ & \cellcolor{red!2} $-0.01$ & \cellcolor{blue!2} $0.02$ \\
 &  & $k=3$ & \cellcolor{red!6} $-0.04^{*}$ & $-0.02$ & \cellcolor{red!9} $-0.06^{*}$ & $0.00$ & $-0.01$ & $0.00$ & \cellcolor{red!3} $-0.02$ & \cellcolor{blue!6} $0.04$ & \cellcolor{green!2} $+0.01$ & \cellcolor{blue!2} $0.01$ & \cellcolor{green!2} $+0.02$ & \cellcolor{blue!6} $0.04$ & \cellcolor{red!3} $-0.02$ & \cellcolor{blue!2} $0.01$ \\
\hhline{~----------------}
 & \multirow{3}{*}{32} & $k=1$ & $+0.00$ & $0.01$ & \cellcolor{green!2} $+0.01$ & $-0.06^{*}$ & \cellcolor{green!9} $+0.06^{*}$ & \cellcolor{blue!1} $0.01$ & \cellcolor{green!4} $+0.03$ & \cellcolor{blue!11} $0.08^{*}$ & \cellcolor{green!4} $+0.03$ & $0.01$ & $+0.00$ & \cellcolor{blue!19} $0.13^{*}$ & \cellcolor{green!8} $+0.06^{*}$ & \cellcolor{blue!2} $0.02$ \\
 &  & $k=2$ & \cellcolor{red!5} $-0.04$ & \cellcolor{blue!6} $0.04$ & \cellcolor{red!3} $-0.02$ & \cellcolor{blue!5} $0.04$ & \cellcolor{green!3} $+0.02$ & \cellcolor{blue!7} $0.05^{*}$ & \cellcolor{green!5} $+0.03$ & \cellcolor{blue!16} $0.11^{*}$ & \cellcolor{green!2} $+0.02$ & \cellcolor{blue!12} $0.08^{*}$ & \cellcolor{red!1} $-0.01$ & \cellcolor{blue!19} $0.13^{*}$ & $+0.00$ & \cellcolor{blue!11} $0.08^{*}$ \\
 &  & $k=3$ & \cellcolor{red!4} $-0.03$ & $-0.03$ & \cellcolor{red!3} $-0.02$ & \cellcolor{blue!1} $0.01$ & $+0.01$ & \cellcolor{blue!7} $0.05^{*}$ & \cellcolor{green!3} $+0.03$ & \cellcolor{blue!18} $0.12^{*}$ & \cellcolor{green!1} $+0.01$ & \cellcolor{blue!14} $0.10^{*}$ & $+0.00$ & \cellcolor{blue!21} $0.14^{*}$ & $+0.00$ & \cellcolor{blue!10} $0.07^{*}$ \\
\hhline{~----------------}
 & \multirow{3}{*}{256} & $k=1$ & \cellcolor{green!5} $+0.04$ & \cellcolor{blue!1} $0.01$ & \cellcolor{green!7} $+0.05^{*}$ & $-0.01$ & $+0.00$ & \cellcolor{blue!3} $0.02$ & \cellcolor{red!2} $-0.02$ & \cellcolor{blue!11} $0.07^{*}$ & \cellcolor{green!1} $+0.01$ & \cellcolor{blue!8} $0.06^{*}$ & $+0.01$ & \cellcolor{blue!11} $0.08^{*}$ & $+0.00$ & \cellcolor{blue!6} $0.04$ \\
 &  & $k=2$ & \cellcolor{red!4} $-0.03$ & $-0.03$ & $+0.00$ & \cellcolor{blue!1} $0.01$ & \cellcolor{green!3} $+0.02$ & \cellcolor{blue!5} $0.03$ & \cellcolor{red!1} $-0.01$ & \cellcolor{blue!12} $0.09^{*}$ & $+0.00$ & \cellcolor{blue!9} $0.06^{*}$ & $+0.00$ & \cellcolor{blue!11} $0.08^{*}$ & $+0.00$ & \cellcolor{blue!8} $0.06^{*}$ \\
 &  & $k=3$ & \cellcolor{green!2} $+0.02$ & $-0.01$ & \cellcolor{red!2} $-0.02$ & \cellcolor{blue!7} $0.05^{*}$ & \cellcolor{red!3} $-0.02$ & \cellcolor{blue!5} $0.04$ & \cellcolor{red!1} $-0.01$ & \cellcolor{blue!8} $0.05^{*}$ & \cellcolor{red!3} $-0.02$ & \cellcolor{blue!12} $0.08^{*}$ & \cellcolor{green!1} $+0.01$ & \cellcolor{blue!9} $0.06^{*}$ & \cellcolor{red!1} $-0.01$ & \cellcolor{blue!10} $0.07^{*}$ \\
\midrule
\multirow{9}{*}{\textbf{chrF}} & \multirow{3}{*}{4} & $k=1$ & \cellcolor{green!3} $+0.02$ & \cellcolor{blue!14} $0.10^{*}$ & \cellcolor{green!1} $+0.01$ & \cellcolor{blue!12} $0.09^{*}$ & \cellcolor{green!1} $+0.01$ & \cellcolor{blue!13} $0.09^{*}$ & \cellcolor{green!4} $+0.03$ & \cellcolor{blue!20} $0.13^{*}$ & \cellcolor{green!6} $+0.04$ & \cellcolor{blue!15} $0.10^{*}$ & $+0.00$ & \cellcolor{blue!27} $0.18^{*}$ & \cellcolor{green!7} $+0.05^{*}$ & \cellcolor{blue!13} $0.09^{*}$ \\
 &  & $k=2$ & \cellcolor{red!3} $-0.02$ & \cellcolor{blue!3} $0.03$ & \cellcolor{red!2} $-0.02$ & \cellcolor{blue!7} $0.05^{*}$ & \cellcolor{green!6} $+0.04$ & \cellcolor{blue!9} $0.06^{*}$ & \cellcolor{green!6} $+0.05^{*}$ & \cellcolor{blue!10} $0.07^{*}$ & \cellcolor{green!1} $+0.01$ & \cellcolor{blue!9} $0.06^{*}$ & $+0.00$ & \cellcolor{blue!14} $0.10^{*}$ & \cellcolor{green!5} $+0.04$ & \cellcolor{blue!8} $0.06^{*}$ \\
 &  & $k=3$ & $+0.00$ & $-0.01$ & \cellcolor{red!7} $-0.05^{*}$ & $-0.00$ & $+0.00$ & \cellcolor{blue!6} $0.04$ & $+0.00$ & \cellcolor{blue!7} $0.05^{*}$ & \cellcolor{green!2} $+0.02$ & \cellcolor{blue!3} $0.02$ & \cellcolor{red!4} $-0.03$ & \cellcolor{blue!9} $0.06^{*}$ & $+0.00$ & \cellcolor{blue!4} $0.03$ \\
\hhline{~----------------}
 & \multirow{3}{*}{32} & $k=1$ & \cellcolor{green!3} $+0.03$ & \cellcolor{blue!11} $0.07^{*}$ & \cellcolor{green!8} $+0.06^{*}$ & \cellcolor{blue!10} $0.07^{*}$ & \cellcolor{red!2} $-0.02$ & \cellcolor{blue!17} $0.12^{*}$ & \cellcolor{green!2} $+0.02$ & \cellcolor{blue!23} $0.16^{*}$ & \cellcolor{red!1} $-0.01$ & \cellcolor{blue!20} $0.14^{*}$ & \cellcolor{red!1} $-0.01$ & \cellcolor{blue!24} $0.16^{*}$ & $+0.00$ & \cellcolor{blue!16} $0.11^{*}$ \\
 &  & $k=2$ & \cellcolor{red!3} $-0.02$ & $-0.01$ & \cellcolor{red!1} $-0.01$ & \cellcolor{blue!7} $0.05^{*}$ & \cellcolor{red!2} $-0.01$ & \cellcolor{blue!9} $0.06^{*}$ & \cellcolor{green!3} $+0.02$ & \cellcolor{blue!17} $0.12^{*}$ & $+0.00$ & \cellcolor{blue!15} $0.11^{*}$ & $+0.00$ & \cellcolor{blue!20} $0.14^{*}$ & \cellcolor{green!2} $+0.01$ & \cellcolor{blue!12} $0.08^{*}$ \\
 &  & $k=3$ & \cellcolor{green!2} $+0.02$ & \cellcolor{blue!1} $0.01$ & $+0.00$ & \cellcolor{blue!10} $0.07^{*}$ & \cellcolor{green!4} $+0.03$ & \cellcolor{blue!13} $0.09^{*}$ & \cellcolor{green!4} $+0.03$ & \cellcolor{blue!20} $0.14^{*}$ & \cellcolor{red!1} $-0.01$ & \cellcolor{blue!17} $0.12^{*}$ & $+0.00$ & \cellcolor{blue!18} $0.12^{*}$ & \cellcolor{green!4} $+0.03$ & \cellcolor{blue!14} $0.10^{*}$ \\
\hhline{~----------------}
 & \multirow{3}{*}{256} & $k=1$ & \cellcolor{red!1} $-0.01$ & \cellcolor{blue!7} $0.05^{*}$ & \cellcolor{green!2} $+0.02$ & \cellcolor{blue!12} $0.08^{*}$ & $+0.00$ & \cellcolor{blue!17} $0.12^{*}$ & \cellcolor{red!2} $-0.01$ & \cellcolor{blue!26} $0.17^{*}$ & \cellcolor{red!7} $-0.05^{*}$ & \cellcolor{blue!20} $0.14^{*}$ & \cellcolor{red!6} $-0.04^{*}$ & \cellcolor{blue!27} $0.18^{*}$ & \cellcolor{red!2} $-0.02$ & \cellcolor{blue!17} $0.12^{*}$ \\
 &  & $k=2$ & \cellcolor{green!1} $+0.01$ & \cellcolor{blue!2} $0.02$ & $+0.00$ & \cellcolor{blue!13} $0.09^{*}$ & \cellcolor{green!2} $+0.02$ & \cellcolor{blue!17} $0.11^{*}$ & \cellcolor{red!1} $-0.01$ & \cellcolor{blue!21} $0.14^{*}$ & \cellcolor{green!1} $+0.01$ & \cellcolor{blue!16} $0.11^{*}$ & $+0.00$ & \cellcolor{blue!21} $0.14^{*}$ & $+0.00$ & \cellcolor{blue!16} $0.11^{*}$ \\
 &  & $k=3$ & $+0.00$ & $-0.01$ & $+0.00$ & \cellcolor{blue!7} $0.05^{*}$ & $+0.00$ & \cellcolor{blue!12} $0.08^{*}$ & \cellcolor{green!3} $+0.02$ & \cellcolor{blue!16} $0.11^{*}$ & $+0.00$ & \cellcolor{blue!15} $0.11^{*}$ & \cellcolor{green!2} $+0.02$ & \cellcolor{blue!15} $0.10^{*}$ & $+0.00$ & \cellcolor{blue!13} $0.09^{*}$ \\
\midrule
\multirow{9}{*}{\textbf{BLEURT}} & \multirow{3}{*}{4} & $k=1$ & $+0.00$ & \cellcolor{blue!8} $0.06^{*}$ & \cellcolor{green!3} $+0.02$ & \cellcolor{blue!11} $0.08^{*}$ & $+0.00$ & \cellcolor{blue!12} $0.08^{*}$ & \cellcolor{red!3} $-0.02$ & \cellcolor{blue!16} $0.11^{*}$ & \cellcolor{green!5} $+0.04$ & \cellcolor{blue!15} $0.11^{*}$ & \cellcolor{green!3} $+0.02$ & \cellcolor{blue!18} $0.13^{*}$ & \cellcolor{green!3} $+0.02$ & \cellcolor{blue!12} $0.08^{*}$ \\
 &  & $k=2$ & \cellcolor{green!1} $+0.01$ & \cellcolor{blue!4} $0.03$ & \cellcolor{green!6} $+0.04$ & \cellcolor{blue!9} $0.06^{*}$ & \cellcolor{red!5} $-0.04$ & \cellcolor{blue!4} $0.03$ & \cellcolor{red!10} $-0.07^{*}$ & \cellcolor{blue!10} $0.07^{*}$ & \cellcolor{red!2} $-0.02$ & \cellcolor{blue!7} $0.05^{*}$ & \cellcolor{red!4} $-0.03$ & \cellcolor{blue!10} $0.07^{*}$ & $+0.00$ & \cellcolor{blue!8} $0.06^{*}$ \\
 &  & $k=3$ & \cellcolor{green!3} $+0.02$ & $0.00$ & \cellcolor{green!6} $+0.04$ & \cellcolor{blue!7} $0.05^{*}$ & \cellcolor{red!4} $-0.03$ & \cellcolor{blue!6} $0.04$ & \cellcolor{red!2} $-0.01$ & \cellcolor{blue!10} $0.07^{*}$ & \cellcolor{red!2} $-0.02$ & \cellcolor{blue!6} $0.04$ & \cellcolor{red!3} $-0.02$ & \cellcolor{blue!8} $0.06^{*}$ & \cellcolor{green!1} $+0.01$ & \cellcolor{blue!8} $0.06^{*}$ \\
\hhline{~----------------}
 & \multirow{3}{*}{32} & $k=1$ & \cellcolor{green!1} $+0.01$ & \cellcolor{blue!7} $0.05^{*}$ & \cellcolor{green!4} $+0.03$ & \cellcolor{blue!12} $0.09^{*}$ & $+0.01$ & \cellcolor{blue!15} $0.11^{*}$ & \cellcolor{green!8} $+0.05^{*}$ & \cellcolor{blue!23} $0.15^{*}$ & \cellcolor{green!2} $+0.01$ & \cellcolor{blue!19} $0.13^{*}$ & \cellcolor{green!11} $+0.08^{*}$ & \cellcolor{blue!23} $0.16^{*}$ & \cellcolor{green!7} $+0.05^{*}$ & \cellcolor{blue!15} $0.10^{*}$ \\
 &  & $k=2$ & $-0.01$ & \cellcolor{blue!4} $0.03$ & \cellcolor{green!7} $+0.05^{*}$ & \cellcolor{blue!12} $0.08^{*}$ & \cellcolor{green!3} $+0.02$ & \cellcolor{blue!13} $0.09^{*}$ & \cellcolor{green!4} $+0.03$ & \cellcolor{blue!16} $0.11^{*}$ & \cellcolor{green!2} $+0.02$ & \cellcolor{blue!13} $0.09^{*}$ & \cellcolor{green!6} $+0.04$ & \cellcolor{blue!17} $0.12^{*}$ & \cellcolor{green!1} $+0.01$ & \cellcolor{blue!14} $0.09^{*}$ \\
 &  & $k=3$ & \cellcolor{green!1} $+0.01$ & \cellcolor{blue!1} $0.01$ & \cellcolor{green!2} $+0.01$ & \cellcolor{blue!5} $0.03$ & \cellcolor{red!2} $-0.02$ & \cellcolor{blue!8} $0.05^{*}$ & \cellcolor{green!3} $+0.02$ & \cellcolor{blue!9} $0.06^{*}$ & $+0.00$ & \cellcolor{blue!13} $0.09^{*}$ & \cellcolor{green!3} $+0.02$ & \cellcolor{blue!10} $0.07^{*}$ & \cellcolor{green!1} $+0.01$ & \cellcolor{blue!9} $0.06^{*}$ \\
\hhline{~----------------}
 & \multirow{3}{*}{256} & $k=1$ & \cellcolor{red!1} $-0.01$ & \cellcolor{blue!12} $0.08^{*}$ & \cellcolor{green!2} $+0.01$ & \cellcolor{blue!19} $0.13^{*}$ & \cellcolor{green!1} $+0.01$ & \cellcolor{blue!23} $0.16^{*}$ & \cellcolor{green!6} $+0.04$ & \cellcolor{blue!29} $0.20^{*}$ & \cellcolor{red!1} $-0.01$ & \cellcolor{blue!27} $0.18^{*}$ & \cellcolor{green!5} $+0.04$ & \cellcolor{blue!29} $0.20^{*}$ & \cellcolor{green!3} $+0.02$ & \cellcolor{blue!25} $0.17^{*}$ \\
 &  & $k=2$ & $+0.00$ & \cellcolor{blue!3} $0.03$ & $+0.00$ & \cellcolor{blue!14} $0.10^{*}$ & \cellcolor{green!3} $+0.03$ & \cellcolor{blue!15} $0.11^{*}$ & \cellcolor{green!3} $+0.02$ & \cellcolor{blue!20} $0.14^{*}$ & \cellcolor{red!6} $-0.04$ & \cellcolor{blue!19} $0.13^{*}$ & \cellcolor{red!2} $-0.02$ & \cellcolor{blue!20} $0.14^{*}$ & \cellcolor{red!2} $-0.02$ & \cellcolor{blue!17} $0.12^{*}$ \\
 &  & $k=3$ & \cellcolor{green!1} $+0.01$ & $-0.02$ & \cellcolor{green!1} $+0.01$ & \cellcolor{blue!6} $0.05^{*}$ & \cellcolor{green!3} $+0.02$ & \cellcolor{blue!4} $0.03$ & \cellcolor{green!7} $+0.05^{*}$ & \cellcolor{blue!12} $0.09^{*}$ & \cellcolor{red!4} $-0.03$ & \cellcolor{blue!15} $0.10^{*}$ & \cellcolor{red!2} $-0.02$ & \cellcolor{blue!13} $0.09^{*}$ & $+0.00$ & \cellcolor{blue!11} $0.08^{*}$ \\
\midrule
\multirow{9}{*}{\textbf{COMET}} & \multirow{3}{*}{4} & $k=1$ & \cellcolor{red!3} $-0.02$ & \cellcolor{blue!19} $0.13^{*}$ & $+0.00$ & \cellcolor{blue!20} $0.14^{*}$ & $-0.01$ & \cellcolor{blue!15} $0.10^{*}$ & \cellcolor{red!1} $-0.01$ & \cellcolor{blue!17} $0.11^{*}$ & \cellcolor{red!1} $-0.01$ & \cellcolor{blue!22} $0.15^{*}$ & \cellcolor{red!2} $-0.01$ & \cellcolor{blue!12} $0.09^{*}$ & $+0.00$ & \cellcolor{blue!20} $0.14^{*}$ \\
 &  & $k=2$ & \cellcolor{red!1} $-0.01$ & \cellcolor{blue!1} $0.01$ & \cellcolor{green!3} $+0.02$ & $0.01$ & \cellcolor{red!1} $-0.01$ & \cellcolor{blue!3} $0.02$ & \cellcolor{red!3} $-0.02$ & \cellcolor{blue!7} $0.05^{*}$ & $+0.00$ & \cellcolor{blue!9} $0.06^{*}$ & \cellcolor{red!1} $-0.01$ & \cellcolor{blue!8} $0.05^{*}$ & \cellcolor{red!2} $-0.02$ & \cellcolor{blue!4} $0.03$ \\
 &  & $k=3$ & \cellcolor{green!1} $+0.01$ & $-0.01$ & \cellcolor{green!1} $+0.01$ & $-0.03$ & \cellcolor{red!4} $-0.03$ & $0.00$ & \cellcolor{red!4} $-0.03$ & $0.00$ & $+0.00$ & \cellcolor{blue!4} $0.03$ & \cellcolor{red!1} $-0.01$ & $0.00$ & $+0.00$ & $0.01$ \\
\hhline{~----------------}
 & \multirow{3}{*}{32} & $k=1$ & $+0.00$ & \cellcolor{blue!8} $0.06^{*}$ & \cellcolor{green!2} $+0.02$ & \cellcolor{blue!12} $0.08^{*}$ & \cellcolor{green!2} $+0.02$ & \cellcolor{blue!12} $0.08^{*}$ & $-0.01$ & \cellcolor{blue!12} $0.08^{*}$ & \cellcolor{red!2} $-0.02$ & \cellcolor{blue!11} $0.07^{*}$ & \cellcolor{red!7} $-0.05^{*}$ & \cellcolor{blue!11} $0.08^{*}$ & \cellcolor{red!1} $-0.01$ & \cellcolor{blue!11} $0.08^{*}$ \\
 &  & $k=2$ & \cellcolor{red!2} $-0.02$ & \cellcolor{blue!2} $0.01$ & \cellcolor{green!3} $+0.02$ & \cellcolor{blue!1} $0.01$ & $+0.00$ & $-0.01$ & \cellcolor{green!3} $+0.02$ & \cellcolor{blue!2} $0.01$ & \cellcolor{green!2} $+0.02$ & $0.01$ & \cellcolor{red!2} $-0.01$ & \cellcolor{blue!1} $0.01$ & \cellcolor{green!1} $+0.01$ & $0.00$ \\
 &  & $k=3$ & \cellcolor{red!4} $-0.03$ & $-0.02$ & \cellcolor{red!1} $-0.01$ & $-0.01$ & \cellcolor{green!5} $+0.03$ & $-0.02$ & \cellcolor{green!5} $+0.04$ & \cellcolor{blue!4} $0.03$ & \cellcolor{red!1} $-0.01$ & $0.00$ & \cellcolor{red!3} $-0.02$ & \cellcolor{blue!5} $0.04$ & \cellcolor{green!1} $+0.01$ & $-0.00$ \\
\hhline{~----------------}
 & \multirow{3}{*}{256} & $k=1$ & \cellcolor{green!3} $+0.03$ & \cellcolor{blue!7} $0.05^{*}$ & \cellcolor{green!3} $+0.02$ & \cellcolor{blue!13} $0.09^{*}$ & \cellcolor{green!4} $+0.03$ & \cellcolor{blue!12} $0.09^{*}$ & \cellcolor{green!4} $+0.03$ & \cellcolor{blue!14} $0.09^{*}$ & \cellcolor{green!1} $+0.01$ & \cellcolor{blue!17} $0.11^{*}$ & \cellcolor{red!3} $-0.03$ & \cellcolor{blue!15} $0.10^{*}$ & \cellcolor{green!2} $+0.02$ & \cellcolor{blue!16} $0.11^{*}$ \\
 &  & $k=2$ & \cellcolor{red!2} $-0.02$ & \cellcolor{blue!6} $0.04$ & $+0.00$ & \cellcolor{blue!7} $0.05^{*}$ & \cellcolor{green!5} $+0.03$ & \cellcolor{blue!2} $0.02$ & \cellcolor{green!6} $+0.04^{*}$ & \cellcolor{blue!4} $0.03$ & \cellcolor{green!6} $+0.04$ & \cellcolor{blue!6} $0.05^{*}$ & \cellcolor{red!4} $-0.03$ & \cellcolor{blue!3} $0.02$ & \cellcolor{green!4} $+0.03$ & \cellcolor{blue!7} $0.05^{*}$ \\
 &  & $k=3$ & $+0.00$ & \cellcolor{blue!3} $0.02$ & \cellcolor{green!1} $+0.01$ & \cellcolor{blue!6} $0.04^{*}$ & \cellcolor{green!4} $+0.03$ & $-0.00$ & \cellcolor{green!12} $+0.08^{*}$ & \cellcolor{blue!7} $0.05^{*}$ & $+0.01$ & \cellcolor{blue!6} $0.04$ & \cellcolor{green!1} $+0.01$ & \cellcolor{blue!6} $0.04$ & \cellcolor{green!4} $+0.03$ & \cellcolor{blue!6} $0.04^{*}$ \\
\midrule
\multirow{9}{*}{\textbf{BERTScore}} & \multirow{3}{*}{4} & $k=1$ & \cellcolor{green!1} $+0.01$ & \cellcolor{blue!21} $0.15^{*}$ & $+0.00$ & \cellcolor{blue!16} $0.11^{*}$ & \cellcolor{green!2} $+0.02$ & \cellcolor{blue!14} $0.10^{*}$ & \cellcolor{green!1} $+0.01$ & \cellcolor{blue!21} $0.14^{*}$ & \cellcolor{green!2} $+0.02$ & \cellcolor{blue!16} $0.11^{*}$ & \cellcolor{green!7} $+0.05^{*}$ & \cellcolor{blue!23} $0.15^{*}$ & \cellcolor{green!6} $+0.04$ & \cellcolor{blue!17} $0.11^{*}$ \\
 &  & $k=2$ & \cellcolor{green!2} $+0.02$ & \cellcolor{blue!2} $0.02$ & \cellcolor{red!1} $-0.01$ & \cellcolor{blue!6} $0.05^{*}$ & \cellcolor{green!1} $+0.01$ & \cellcolor{blue!8} $0.06^{*}$ & \cellcolor{green!2} $+0.01$ & \cellcolor{blue!14} $0.10^{*}$ & $+0.00$ & \cellcolor{blue!15} $0.11^{*}$ & $+0.00$ & \cellcolor{blue!14} $0.10^{*}$ & \cellcolor{green!4} $+0.03$ & \cellcolor{blue!11} $0.07^{*}$ \\
 &  & $k=3$ & $-0.01$ & $-0.01$ & \cellcolor{green!1} $+0.01$ & \cellcolor{blue!1} $0.01$ & $+0.00$ & \cellcolor{blue!2} $0.02$ & \cellcolor{green!5} $+0.04$ & \cellcolor{blue!8} $0.06^{*}$ & \cellcolor{green!2} $+0.02$ & \cellcolor{blue!7} $0.05^{*}$ & \cellcolor{green!1} $+0.01$ & \cellcolor{blue!9} $0.06^{*}$ & \cellcolor{green!7} $+0.05^{*}$ & \cellcolor{blue!5} $0.04$ \\
\hhline{~----------------}
 & \multirow{3}{*}{32} & $k=1$ & \cellcolor{green!1} $+0.01$ & \cellcolor{blue!12} $0.08^{*}$ & \cellcolor{green!4} $+0.03$ & \cellcolor{blue!9} $0.06^{*}$ & $+0.00$ & \cellcolor{blue!13} $0.09^{*}$ & \cellcolor{green!4} $+0.03$ & \cellcolor{blue!20} $0.14^{*}$ & \cellcolor{green!5} $+0.04$ & \cellcolor{blue!20} $0.14^{*}$ & \cellcolor{green!4} $+0.03$ & \cellcolor{blue!25} $0.17^{*}$ & \cellcolor{green!1} $+0.01$ & \cellcolor{blue!12} $0.08^{*}$ \\
 &  & $k=2$ & \cellcolor{green!2} $+0.01$ & \cellcolor{blue!2} $0.01$ & \cellcolor{green!5} $+0.04$ & \cellcolor{blue!6} $0.04$ & \cellcolor{green!2} $+0.01$ & \cellcolor{blue!14} $0.09^{*}$ & \cellcolor{green!3} $+0.03$ & \cellcolor{blue!21} $0.14^{*}$ & \cellcolor{green!2} $+0.01$ & \cellcolor{blue!20} $0.13^{*}$ & \cellcolor{green!2} $+0.01$ & \cellcolor{blue!18} $0.12^{*}$ & \cellcolor{green!1} $+0.01$ & \cellcolor{blue!13} $0.09^{*}$ \\
 &  & $k=3$ & \cellcolor{green!6} $+0.04$ & $-0.03$ & \cellcolor{green!7} $+0.05^{*}$ & \cellcolor{blue!3} $0.02$ & \cellcolor{green!4} $+0.03$ & \cellcolor{blue!11} $0.08^{*}$ & \cellcolor{green!2} $+0.01$ & \cellcolor{blue!17} $0.11^{*}$ & \cellcolor{red!2} $-0.01$ & \cellcolor{blue!19} $0.13^{*}$ & \cellcolor{green!3} $+0.02$ & \cellcolor{blue!15} $0.11^{*}$ & \cellcolor{green!3} $+0.02$ & \cellcolor{blue!11} $0.08^{*}$ \\
\hhline{~----------------}
 & \multirow{3}{*}{256} & $k=1$ & \cellcolor{green!5} $+0.04$ & \cellcolor{blue!15} $0.10^{*}$ & \cellcolor{green!4} $+0.03$ & \cellcolor{blue!15} $0.10^{*}$ & \cellcolor{green!1} $+0.01$ & \cellcolor{blue!21} $0.14^{*}$ & \cellcolor{green!7} $+0.05^{*}$ & \cellcolor{blue!29} $0.20^{*}$ & \cellcolor{green!6} $+0.05^{*}$ & \cellcolor{blue!29} $0.19^{*}$ & \cellcolor{green!8} $+0.06^{*}$ & \cellcolor{blue!26} $0.18^{*}$ & \cellcolor{green!5} $+0.03$ & \cellcolor{blue!20} $0.14^{*}$ \\
 &  & $k=2$ & \cellcolor{green!4} $+0.03$ & \cellcolor{blue!6} $0.05^{*}$ & $+0.01$ & \cellcolor{blue!14} $0.09^{*}$ & \cellcolor{green!6} $+0.04$ & \cellcolor{blue!16} $0.11^{*}$ & \cellcolor{green!6} $+0.04$ & \cellcolor{blue!21} $0.14^{*}$ & \cellcolor{green!2} $+0.02$ & \cellcolor{blue!21} $0.15^{*}$ & \cellcolor{green!5} $+0.04$ & \cellcolor{blue!20} $0.14^{*}$ & \cellcolor{green!3} $+0.02$ & \cellcolor{blue!19} $0.13^{*}$ \\
 &  & $k=3$ & \cellcolor{green!3} $+0.03$ & \cellcolor{blue!3} $0.02$ & $+0.01$ & \cellcolor{blue!6} $0.04^{*}$ & $+0.00$ & \cellcolor{blue!10} $0.07^{*}$ & \cellcolor{green!1} $+0.01$ & \cellcolor{blue!15} $0.10^{*}$ & \cellcolor{green!1} $+0.01$ & \cellcolor{blue!14} $0.10^{*}$ & \cellcolor{green!4} $+0.03$ & \cellcolor{blue!12} $0.08^{*}$ & $+0.01$ & \cellcolor{blue!13} $0.09^{*}$ \\
\bottomrule
\end{NiceTabular}
}
\caption{Spearman correlation ($\rho$) between inherent hypotheses oracle variance and evaluation score differences across varying pseudo-reference pool sizes ($|\mathcal{Y}|$) on WMT22 De$\rightarrow$En. The hypothesis pool variance ($\sigma^2$) for each evaluation metric is displayed in the respective column headers. \textit{Dir.} (Directional Change) measures score improvements, highlighted as \colorbox{green!15}{positive} and \colorbox{red!15}{negative} impacts. \textit{Mag.} (Magnitude Sensitivity) measures absolute score variance, with \colorbox{blue!15}{blue} highlighting metric sensitivity. Statistical significance ($p < 0.05$) is denoted by an asterisk ($^*$).}
\label{tab:corr_variance_analysis_complete_wmt22-deen}
\end{table*}

\begin{table*}[ht!]
\centering
\resizebox{\textwidth}{!}{
\begin{NiceTabular}{@{} lll c c c c c c c c c c c c c c @{}}
\toprule
\multirow{2}{*}{\textbf{Util. Function}} & \multirow{2}{*}{\textbf{$|\mathcal{Y}|$}} & \multirow{2}{*}{\textbf{Top-$k$}} & \multicolumn{2}{c}{\textbf{BLEU}} & \multicolumn{2}{c}{\textbf{chrF}} & \multicolumn{2}{c}{\textbf{BLEURT}} & \multicolumn{2}{c}{\textbf{COMET}} & \multicolumn{2}{c}{\textbf{BERTScore}} & \multicolumn{2}{c}{\textbf{COMETKiwi}} & \multicolumn{2}{c}{\textbf{$\bar{Z}_{\text{other}}$}} \\
\cmidrule(lr){4-5} \cmidrule(lr){6-7} \cmidrule(lr){8-9} \cmidrule(lr){10-11} \cmidrule(lr){12-13} \cmidrule(lr){14-15} \cmidrule(lr){16-17}
 &  &  & \textit{Dir.} & \textit{Mag.} & \textit{Dir.} & \textit{Mag.} & \textit{Dir.} & \textit{Mag.} & \textit{Dir.} & \textit{Mag.} & \textit{Dir.} & \textit{Mag.} & \textit{Dir.} & \textit{Mag.} & \textit{Dir.} & \textit{Mag.} \\
\midrule
\multirow{9}{*}{\textbf{BLEU}} & \multirow{3}{*}{4} & $k=1$ & \cellcolor{red!9} $-0.06^{*}$ & \cellcolor{blue!41} $0.28^{*}$ & \cellcolor{red!2} $-0.02$ & \cellcolor{blue!41} $0.27^{*}$ & \cellcolor{red!2} $-0.01$ & \cellcolor{blue!39} $0.26^{*}$ & \cellcolor{red!2} $-0.02$ & \cellcolor{blue!41} $0.28^{*}$ & \cellcolor{red!4} $-0.03$ & \cellcolor{blue!42} $0.28^{*}$ & \cellcolor{red!5} $-0.04$ & \cellcolor{blue!42} $0.28^{*}$ & \cellcolor{red!4} $-0.03$ & \cellcolor{blue!40} $0.27^{*}$ \\
 &  & $k=2$ & \cellcolor{red!5} $-0.03$ & \cellcolor{blue!35} $0.23^{*}$ & \cellcolor{red!5} $-0.03$ & \cellcolor{blue!34} $0.23^{*}$ & \cellcolor{red!2} $-0.01$ & \cellcolor{blue!34} $0.23^{*}$ & \cellcolor{red!3} $-0.02$ & \cellcolor{blue!35} $0.24^{*}$ & \cellcolor{red!3} $-0.02$ & \cellcolor{blue!34} $0.23^{*}$ & \cellcolor{red!8} $-0.06^{*}$ & \cellcolor{blue!35} $0.24^{*}$ & \cellcolor{red!3} $-0.03$ & \cellcolor{blue!34} $0.23^{*}$ \\
 &  & $k=3$ & \cellcolor{red!4} $-0.03$ & \cellcolor{blue!30} $0.20^{*}$ & \cellcolor{red!1} $-0.01$ & \cellcolor{blue!28} $0.19^{*}$ & \cellcolor{red!4} $-0.03$ & \cellcolor{blue!28} $0.19^{*}$ & \cellcolor{red!6} $-0.04$ & \cellcolor{blue!29} $0.19^{*}$ & \cellcolor{red!1} $-0.01$ & \cellcolor{blue!28} $0.19^{*}$ & \cellcolor{red!4} $-0.03$ & \cellcolor{blue!28} $0.19^{*}$ & \cellcolor{red!6} $-0.04$ & \cellcolor{blue!28} $0.19^{*}$ \\
\hhline{~----------------}
 & \multirow{3}{*}{32} & $k=1$ & \cellcolor{green!4} $+0.03$ & \cellcolor{blue!36} $0.24^{*}$ & \cellcolor{green!4} $+0.03$ & \cellcolor{blue!35} $0.24^{*}$ & \cellcolor{red!3} $-0.02$ & \cellcolor{blue!30} $0.20^{*}$ & \cellcolor{green!4} $+0.03$ & \cellcolor{blue!33} $0.22^{*}$ & $+0.01$ & \cellcolor{blue!32} $0.22^{*}$ & $+0.00$ & \cellcolor{blue!33} $0.22^{*}$ & \cellcolor{green!1} $+0.01$ & \cellcolor{blue!31} $0.21^{*}$ \\
 &  & $k=2$ & \cellcolor{red!3} $-0.03$ & \cellcolor{blue!36} $0.24^{*}$ & \cellcolor{red!1} $-0.01$ & \cellcolor{blue!37} $0.25^{*}$ & \cellcolor{red!1} $-0.01$ & \cellcolor{blue!33} $0.22^{*}$ & \cellcolor{red!1} $-0.01$ & \cellcolor{blue!37} $0.25^{*}$ & \cellcolor{green!4} $+0.03$ & \cellcolor{blue!35} $0.23^{*}$ & $+0.00$ & \cellcolor{blue!36} $0.24^{*}$ & $+0.00$ & \cellcolor{blue!33} $0.22^{*}$ \\
 &  & $k=3$ & \cellcolor{red!8} $-0.05^{*}$ & \cellcolor{blue!37} $0.25^{*}$ & \cellcolor{red!1} $-0.01$ & \cellcolor{blue!39} $0.26^{*}$ & \cellcolor{red!6} $-0.04$ & \cellcolor{blue!34} $0.23^{*}$ & \cellcolor{green!2} $+0.02$ & \cellcolor{blue!37} $0.25^{*}$ & \cellcolor{green!1} $+0.01$ & \cellcolor{blue!34} $0.23^{*}$ & \cellcolor{green!1} $+0.01$ & \cellcolor{blue!37} $0.25^{*}$ & $+0.00$ & \cellcolor{blue!32} $0.22^{*}$ \\
\hhline{~----------------}
 & \multirow{3}{*}{256} & $k=1$ & \cellcolor{red!5} $-0.04$ & \cellcolor{blue!33} $0.23^{*}$ & $+0.01$ & \cellcolor{blue!32} $0.22^{*}$ & \cellcolor{green!2} $+0.02$ & \cellcolor{blue!32} $0.22^{*}$ & \cellcolor{red!3} $-0.03$ & \cellcolor{blue!33} $0.23^{*}$ & $+0.00$ & \cellcolor{blue!31} $0.21^{*}$ & \cellcolor{red!2} $-0.02$ & \cellcolor{blue!33} $0.22^{*}$ & $+0.00$ & \cellcolor{blue!30} $0.20^{*}$ \\
 &  & $k=2$ & $+0.00$ & \cellcolor{blue!33} $0.22^{*}$ & \cellcolor{red!1} $-0.01$ & \cellcolor{blue!32} $0.22^{*}$ & \cellcolor{green!4} $+0.03$ & \cellcolor{blue!30} $0.21^{*}$ & \cellcolor{red!2} $-0.02$ & \cellcolor{blue!31} $0.21^{*}$ & \cellcolor{green!1} $+0.01$ & \cellcolor{blue!28} $0.19^{*}$ & $+0.01$ & \cellcolor{blue!32} $0.22^{*}$ & \cellcolor{green!1} $+0.01$ & \cellcolor{blue!27} $0.18^{*}$ \\
 &  & $k=3$ & \cellcolor{red!2} $-0.02$ & \cellcolor{blue!26} $0.17^{*}$ & $+0.00$ & \cellcolor{blue!25} $0.17^{*}$ & \cellcolor{red!2} $-0.02$ & \cellcolor{blue!20} $0.14^{*}$ & \cellcolor{red!3} $-0.02$ & \cellcolor{blue!23} $0.16^{*}$ & \cellcolor{red!6} $-0.04$ & \cellcolor{blue!17} $0.12^{*}$ & $+0.00$ & \cellcolor{blue!23} $0.16^{*}$ & \cellcolor{red!5} $-0.04$ & \cellcolor{blue!16} $0.11^{*}$ \\
\midrule
\multirow{9}{*}{\textbf{chrF}} & \multirow{3}{*}{4} & $k=1$ & \cellcolor{red!10} $-0.07^{*}$ & \cellcolor{blue!34} $0.23^{*}$ & \cellcolor{red!7} $-0.05^{*}$ & \cellcolor{blue!34} $0.23^{*}$ & \cellcolor{red!4} $-0.03$ & \cellcolor{blue!31} $0.21^{*}$ & \cellcolor{red!4} $-0.03$ & \cellcolor{blue!34} $0.23^{*}$ & \cellcolor{red!5} $-0.04$ & \cellcolor{blue!33} $0.22^{*}$ & \cellcolor{red!2} $-0.02$ & \cellcolor{blue!36} $0.24^{*}$ & \cellcolor{red!8} $-0.06^{*}$ & \cellcolor{blue!32} $0.22^{*}$ \\
 &  & $k=2$ & \cellcolor{green!3} $+0.02$ & \cellcolor{blue!36} $0.24^{*}$ & \cellcolor{green!1} $+0.01$ & \cellcolor{blue!35} $0.24^{*}$ & \cellcolor{green!1} $+0.01$ & \cellcolor{blue!34} $0.23^{*}$ & \cellcolor{green!4} $+0.03$ & \cellcolor{blue!35} $0.23^{*}$ & \cellcolor{green!6} $+0.04$ & \cellcolor{blue!35} $0.24^{*}$ & \cellcolor{red!1} $-0.01$ & \cellcolor{blue!35} $0.24^{*}$ & \cellcolor{green!3} $+0.02$ & \cellcolor{blue!34} $0.23^{*}$ \\
 &  & $k=3$ & $-0.01$ & \cellcolor{blue!27} $0.18^{*}$ & \cellcolor{green!3} $+0.02$ & \cellcolor{blue!27} $0.19^{*}$ & \cellcolor{green!4} $+0.03$ & \cellcolor{blue!27} $0.18^{*}$ & $+0.00$ & \cellcolor{blue!27} $0.18^{*}$ & $+0.00$ & \cellcolor{blue!27} $0.18^{*}$ & \cellcolor{green!1} $+0.01$ & \cellcolor{blue!27} $0.18^{*}$ & $+0.00$ & \cellcolor{blue!27} $0.19^{*}$ \\
\hhline{~----------------}
 & \multirow{3}{*}{32} & $k=1$ & \cellcolor{red!2} $-0.02$ & \cellcolor{blue!26} $0.18^{*}$ & \cellcolor{red!7} $-0.05^{*}$ & \cellcolor{blue!26} $0.18^{*}$ & \cellcolor{green!1} $+0.01$ & \cellcolor{blue!25} $0.17^{*}$ & $+0.00$ & \cellcolor{blue!25} $0.17^{*}$ & $+0.00$ & \cellcolor{blue!24} $0.16^{*}$ & $+0.01$ & \cellcolor{blue!27} $0.18^{*}$ & $+0.00$ & \cellcolor{blue!23} $0.16^{*}$ \\
 &  & $k=2$ & $-0.01$ & \cellcolor{blue!27} $0.18^{*}$ & \cellcolor{red!3} $-0.02$ & \cellcolor{blue!31} $0.21^{*}$ & $+0.00$ & \cellcolor{blue!29} $0.20^{*}$ & $+0.00$ & \cellcolor{blue!31} $0.21^{*}$ & \cellcolor{green!1} $+0.01$ & \cellcolor{blue!29} $0.20^{*}$ & $-0.01$ & \cellcolor{blue!32} $0.21^{*}$ & \cellcolor{red!1} $-0.01$ & \cellcolor{blue!28} $0.19^{*}$ \\
 &  & $k=3$ & $+0.00$ & \cellcolor{blue!21} $0.14^{*}$ & $+0.00$ & \cellcolor{blue!22} $0.15^{*}$ & \cellcolor{red!2} $-0.02$ & \cellcolor{blue!19} $0.13^{*}$ & $+0.00$ & \cellcolor{blue!21} $0.14^{*}$ & \cellcolor{green!1} $+0.01$ & \cellcolor{blue!18} $0.12^{*}$ & $+0.00$ & \cellcolor{blue!23} $0.16^{*}$ & \cellcolor{red!3} $-0.02$ & \cellcolor{blue!16} $0.11^{*}$ \\
\hhline{~----------------}
 & \multirow{3}{*}{256} & $k=1$ & \cellcolor{green!4} $+0.03$ & \cellcolor{blue!19} $0.13^{*}$ & \cellcolor{red!3} $-0.03$ & \cellcolor{blue!17} $0.12^{*}$ & $+0.00$ & \cellcolor{blue!14} $0.10^{*}$ & $+0.00$ & \cellcolor{blue!18} $0.12^{*}$ & \cellcolor{green!3} $+0.02$ & \cellcolor{blue!14} $0.09^{*}$ & \cellcolor{green!2} $+0.02$ & \cellcolor{blue!18} $0.13^{*}$ & $+0.00$ & \cellcolor{blue!14} $0.10^{*}$ \\
 &  & $k=2$ & $+0.00$ & \cellcolor{blue!23} $0.16^{*}$ & \cellcolor{red!3} $-0.02$ & \cellcolor{blue!24} $0.16^{*}$ & \cellcolor{red!2} $-0.02$ & \cellcolor{blue!23} $0.15^{*}$ & \cellcolor{red!7} $-0.05^{*}$ & \cellcolor{blue!24} $0.16^{*}$ & \cellcolor{red!1} $-0.01$ & \cellcolor{blue!20} $0.14^{*}$ & \cellcolor{red!2} $-0.02$ & \cellcolor{blue!23} $0.16^{*}$ & \cellcolor{red!6} $-0.04$ & \cellcolor{blue!20} $0.14^{*}$ \\
 &  & $k=3$ & \cellcolor{green!2} $+0.01$ & \cellcolor{blue!17} $0.12^{*}$ & \cellcolor{green!2} $+0.01$ & \cellcolor{blue!18} $0.13^{*}$ & $+0.00$ & \cellcolor{blue!16} $0.11^{*}$ & \cellcolor{green!1} $+0.01$ & \cellcolor{blue!16} $0.11^{*}$ & \cellcolor{green!1} $+0.01$ & \cellcolor{blue!13} $0.09^{*}$ & \cellcolor{red!3} $-0.02$ & \cellcolor{blue!17} $0.12^{*}$ & \cellcolor{red!2} $-0.01$ & \cellcolor{blue!13} $0.09^{*}$ \\
\midrule
\multirow{9}{*}{\textbf{BLEURT}} & \multirow{3}{*}{4} & $k=1$ & \cellcolor{green!3} $+0.02$ & \cellcolor{blue!41} $0.27^{*}$ & $+0.00$ & \cellcolor{blue!43} $0.29^{*}$ & \cellcolor{red!16} $-0.11^{*}$ & \cellcolor{blue!45} $0.30^{*}$ & \cellcolor{red!5} $-0.04$ & \cellcolor{blue!44} $0.30^{*}$ & \cellcolor{red!2} $-0.02$ & \cellcolor{blue!43} $0.29^{*}$ & \cellcolor{red!3} $-0.02$ & \cellcolor{blue!44} $0.30^{*}$ & \cellcolor{red!3} $-0.02$ & \cellcolor{blue!43} $0.29^{*}$ \\
 &  & $k=2$ & \cellcolor{green!2} $+0.02$ & \cellcolor{blue!29} $0.20^{*}$ & \cellcolor{green!2} $+0.02$ & \cellcolor{blue!30} $0.21^{*}$ & \cellcolor{red!8} $-0.06^{*}$ & \cellcolor{blue!31} $0.21^{*}$ & \cellcolor{red!5} $-0.04$ & \cellcolor{blue!31} $0.21^{*}$ & \cellcolor{red!2} $-0.02$ & \cellcolor{blue!30} $0.20^{*}$ & \cellcolor{red!5} $-0.04$ & \cellcolor{blue!31} $0.21^{*}$ & \cellcolor{red!1} $-0.01$ & \cellcolor{blue!30} $0.20^{*}$ \\
 &  & $k=3$ & $+0.00$ & \cellcolor{blue!17} $0.12^{*}$ & \cellcolor{green!2} $+0.02$ & \cellcolor{blue!15} $0.11^{*}$ & $+0.00$ & \cellcolor{blue!15} $0.10^{*}$ & \cellcolor{green!3} $+0.02$ & \cellcolor{blue!15} $0.10^{*}$ & \cellcolor{red!6} $-0.04$ & \cellcolor{blue!14} $0.10^{*}$ & \cellcolor{red!5} $-0.04$ & \cellcolor{blue!15} $0.10^{*}$ & $+0.00$ & \cellcolor{blue!14} $0.10^{*}$ \\
\hhline{~----------------}
 & \multirow{3}{*}{32} & $k=1$ & \cellcolor{green!1} $+0.01$ & \cellcolor{blue!22} $0.15^{*}$ & \cellcolor{green!3} $+0.02$ & \cellcolor{blue!25} $0.17^{*}$ & \cellcolor{red!6} $-0.05^{*}$ & \cellcolor{blue!26} $0.18^{*}$ & \cellcolor{green!3} $+0.02$ & \cellcolor{blue!26} $0.18^{*}$ & \cellcolor{green!6} $+0.05^{*}$ & \cellcolor{blue!26} $0.18^{*}$ & \cellcolor{green!5} $+0.03$ & \cellcolor{blue!28} $0.19^{*}$ & \cellcolor{green!1} $+0.01$ & \cellcolor{blue!26} $0.17^{*}$ \\
 &  & $k=2$ & \cellcolor{green!5} $+0.04$ & \cellcolor{blue!24} $0.16^{*}$ & \cellcolor{green!4} $+0.03$ & \cellcolor{blue!25} $0.17^{*}$ & \cellcolor{red!6} $-0.04$ & \cellcolor{blue!25} $0.17^{*}$ & \cellcolor{green!1} $+0.01$ & \cellcolor{blue!26} $0.17^{*}$ & \cellcolor{green!1} $+0.01$ & \cellcolor{blue!24} $0.16^{*}$ & \cellcolor{red!1} $-0.01$ & \cellcolor{blue!26} $0.18^{*}$ & \cellcolor{green!1} $+0.01$ & \cellcolor{blue!23} $0.16^{*}$ \\
 &  & $k=3$ & $+0.00$ & \cellcolor{blue!17} $0.12^{*}$ & \cellcolor{green!1} $+0.01$ & \cellcolor{blue!16} $0.11^{*}$ & \cellcolor{green!3} $+0.03$ & \cellcolor{blue!14} $0.10^{*}$ & \cellcolor{green!2} $+0.02$ & \cellcolor{blue!16} $0.11^{*}$ & \cellcolor{green!5} $+0.04$ & \cellcolor{blue!13} $0.09^{*}$ & \cellcolor{green!3} $+0.02$ & \cellcolor{blue!17} $0.11^{*}$ & \cellcolor{green!3} $+0.02$ & \cellcolor{blue!12} $0.08^{*}$ \\
\hhline{~----------------}
 & \multirow{3}{*}{256} & $k=1$ & \cellcolor{green!3} $+0.02$ & \cellcolor{blue!6} $0.05^{*}$ & \cellcolor{green!4} $+0.03$ & \cellcolor{blue!7} $0.05^{*}$ & $-0.01$ & \cellcolor{blue!7} $0.05^{*}$ & \cellcolor{green!1} $+0.01$ & \cellcolor{blue!8} $0.06^{*}$ & \cellcolor{green!1} $+0.01$ & \cellcolor{blue!6} $0.04$ & \cellcolor{red!2} $-0.02$ & \cellcolor{blue!9} $0.06^{*}$ & \cellcolor{green!3} $+0.02$ & \cellcolor{blue!5} $0.04$ \\
 &  & $k=2$ & $+0.00$ & \cellcolor{blue!13} $0.09^{*}$ & \cellcolor{green!3} $+0.02$ & \cellcolor{blue!14} $0.10^{*}$ & \cellcolor{red!2} $-0.02$ & \cellcolor{blue!13} $0.09^{*}$ & \cellcolor{green!2} $+0.02$ & \cellcolor{blue!14} $0.10^{*}$ & \cellcolor{green!1} $+0.01$ & \cellcolor{blue!12} $0.08^{*}$ & \cellcolor{red!6} $-0.04$ & \cellcolor{blue!16} $0.11^{*}$ & $+0.00$ & \cellcolor{blue!12} $0.08^{*}$ \\
 &  & $k=3$ & \cellcolor{red!4} $-0.03$ & \cellcolor{blue!10} $0.07^{*}$ & \cellcolor{red!5} $-0.04$ & \cellcolor{blue!8} $0.06^{*}$ & \cellcolor{red!4} $-0.03$ & \cellcolor{blue!6} $0.05^{*}$ & \cellcolor{red!7} $-0.05^{*}$ & \cellcolor{blue!8} $0.05^{*}$ & $+0.00$ & \cellcolor{blue!5} $0.04$ & $+0.00$ & \cellcolor{blue!9} $0.06^{*}$ & \cellcolor{red!3} $-0.02$ & \cellcolor{blue!4} $0.03$ \\
\midrule
\multirow{9}{*}{\textbf{COMET}} & \multirow{3}{*}{4} & $k=1$ & \cellcolor{red!2} $-0.02$ & \cellcolor{blue!45} $0.30^{*}$ & \cellcolor{red!5} $-0.03$ & \cellcolor{blue!49} $0.33^{*}$ & \cellcolor{red!11} $-0.08^{*}$ & \cellcolor{blue!50} $0.34^{*}$ & \cellcolor{red!24} $-0.16^{*}$ & \cellcolor{blue!51} $0.35^{*}$ & \cellcolor{red!8} $-0.06^{*}$ & \cellcolor{blue!51} $0.34^{*}$ & \cellcolor{red!10} $-0.07^{*}$ & \cellcolor{blue!52} $0.35^{*}$ & \cellcolor{red!10} $-0.07^{*}$ & \cellcolor{blue!50} $0.33^{*}$ \\
 &  & $k=2$ & \cellcolor{red!6} $-0.04$ & \cellcolor{blue!26} $0.18^{*}$ & $-0.01$ & \cellcolor{blue!29} $0.20^{*}$ & \cellcolor{red!11} $-0.08^{*}$ & \cellcolor{blue!29} $0.20^{*}$ & \cellcolor{red!10} $-0.07^{*}$ & \cellcolor{blue!30} $0.20^{*}$ & \cellcolor{red!5} $-0.04$ & \cellcolor{blue!29} $0.20^{*}$ & \cellcolor{red!7} $-0.05^{*}$ & \cellcolor{blue!29} $0.20^{*}$ & \cellcolor{red!6} $-0.05^{*}$ & \cellcolor{blue!29} $0.20^{*}$ \\
 &  & $k=3$ & \cellcolor{green!5} $+0.04$ & \cellcolor{blue!20} $0.14^{*}$ & \cellcolor{green!9} $+0.06^{*}$ & \cellcolor{blue!20} $0.14^{*}$ & $+0.00$ & \cellcolor{blue!20} $0.13^{*}$ & \cellcolor{red!3} $-0.02$ & \cellcolor{blue!21} $0.14^{*}$ & $+0.00$ & \cellcolor{blue!19} $0.13^{*}$ & \cellcolor{green!1} $+0.01$ & \cellcolor{blue!20} $0.14^{*}$ & $+0.00$ & \cellcolor{blue!19} $0.13^{*}$ \\
\hhline{~----------------}
 & \multirow{3}{*}{32} & $k=1$ & $+0.00$ & \cellcolor{blue!30} $0.20^{*}$ & \cellcolor{green!2} $+0.02$ & \cellcolor{blue!31} $0.21^{*}$ & \cellcolor{red!3} $-0.02$ & \cellcolor{blue!33} $0.22^{*}$ & \cellcolor{red!12} $-0.08^{*}$ & \cellcolor{blue!33} $0.22^{*}$ & \cellcolor{red!1} $-0.01$ & \cellcolor{blue!31} $0.21^{*}$ & \cellcolor{red!1} $-0.01$ & \cellcolor{blue!33} $0.22^{*}$ & \cellcolor{red!3} $-0.02$ & \cellcolor{blue!31} $0.21^{*}$ \\
 &  & $k=2$ & \cellcolor{green!3} $+0.02$ & \cellcolor{blue!19} $0.13^{*}$ & \cellcolor{green!2} $+0.01$ & \cellcolor{blue!21} $0.14^{*}$ & $-0.01$ & \cellcolor{blue!20} $0.14^{*}$ & $+0.00$ & \cellcolor{blue!21} $0.15^{*}$ & \cellcolor{green!4} $+0.03$ & \cellcolor{blue!20} $0.14^{*}$ & \cellcolor{green!7} $+0.05^{*}$ & \cellcolor{blue!21} $0.14^{*}$ & \cellcolor{green!3} $+0.02$ & \cellcolor{blue!20} $0.13^{*}$ \\
 &  & $k=3$ & \cellcolor{green!5} $+0.04$ & \cellcolor{blue!16} $0.11^{*}$ & \cellcolor{green!6} $+0.04$ & \cellcolor{blue!17} $0.11^{*}$ & \cellcolor{green!3} $+0.02$ & \cellcolor{blue!17} $0.12^{*}$ & \cellcolor{green!6} $+0.04$ & \cellcolor{blue!17} $0.12^{*}$ & \cellcolor{green!3} $+0.03$ & \cellcolor{blue!12} $0.09^{*}$ & \cellcolor{green!4} $+0.03$ & \cellcolor{blue!18} $0.12^{*}$ & \cellcolor{green!2} $+0.02$ & \cellcolor{blue!13} $0.09^{*}$ \\
\hhline{~----------------}
 & \multirow{3}{*}{256} & $k=1$ & \cellcolor{red!4} $-0.03$ & \cellcolor{blue!25} $0.17^{*}$ & $-0.01$ & \cellcolor{blue!26} $0.18^{*}$ & \cellcolor{red!6} $-0.04^{*}$ & \cellcolor{blue!27} $0.18^{*}$ & \cellcolor{red!15} $-0.10^{*}$ & \cellcolor{blue!29} $0.20^{*}$ & $-0.01$ & \cellcolor{blue!27} $0.18^{*}$ & $+0.00$ & \cellcolor{blue!29} $0.20^{*}$ & \cellcolor{red!3} $-0.02$ & \cellcolor{blue!26} $0.18^{*}$ \\
 &  & $k=2$ & \cellcolor{red!5} $-0.04$ & \cellcolor{blue!15} $0.10^{*}$ & \cellcolor{red!3} $-0.02$ & \cellcolor{blue!16} $0.11^{*}$ & \cellcolor{red!1} $-0.01$ & \cellcolor{blue!14} $0.10^{*}$ & \cellcolor{red!1} $-0.01$ & \cellcolor{blue!17} $0.12^{*}$ & \cellcolor{green!8} $+0.06^{*}$ & \cellcolor{blue!15} $0.11^{*}$ & \cellcolor{green!4} $+0.03$ & \cellcolor{blue!18} $0.12^{*}$ & $+0.00$ & \cellcolor{blue!14} $0.10^{*}$ \\
 &  & $k=3$ & \cellcolor{red!1} $-0.01$ & \cellcolor{blue!8} $0.05^{*}$ & $+0.00$ & \cellcolor{blue!9} $0.06^{*}$ & $+0.01$ & \cellcolor{blue!5} $0.04$ & $-0.01$ & \cellcolor{blue!9} $0.06^{*}$ & \cellcolor{green!3} $+0.02$ & \cellcolor{blue!5} $0.04$ & \cellcolor{green!2} $+0.02$ & \cellcolor{blue!10} $0.07^{*}$ & \cellcolor{red!1} $-0.01$ & \cellcolor{blue!3} $0.02$ \\
\midrule
\multirow{9}{*}{\textbf{BERTScore}} & \multirow{3}{*}{4} & $k=1$ & \cellcolor{red!2} $-0.02$ & \cellcolor{blue!28} $0.19^{*}$ & \cellcolor{red!2} $-0.01$ & \cellcolor{blue!30} $0.20^{*}$ & $+0.00$ & \cellcolor{blue!30} $0.20^{*}$ & $+0.01$ & \cellcolor{blue!36} $0.24^{*}$ & \cellcolor{red!5} $-0.04$ & \cellcolor{blue!32} $0.22^{*}$ & $+0.00$ & \cellcolor{blue!38} $0.26^{*}$ & \cellcolor{red!2} $-0.01$ & \cellcolor{blue!29} $0.20^{*}$ \\
 &  & $k=2$ & \cellcolor{red!6} $-0.05^{*}$ & \cellcolor{blue!32} $0.21^{*}$ & \cellcolor{red!8} $-0.06^{*}$ & \cellcolor{blue!33} $0.22^{*}$ & \cellcolor{red!3} $-0.02$ & \cellcolor{blue!30} $0.21^{*}$ & \cellcolor{red!4} $-0.03$ & \cellcolor{blue!33} $0.22^{*}$ & \cellcolor{red!12} $-0.09^{*}$ & \cellcolor{blue!30} $0.20^{*}$ & $+0.00$ & \cellcolor{blue!34} $0.23^{*}$ & \cellcolor{red!6} $-0.05^{*}$ & \cellcolor{blue!30} $0.20^{*}$ \\
 &  & $k=3$ & \cellcolor{green!3} $+0.02$ & \cellcolor{blue!22} $0.15^{*}$ & $+0.01$ & \cellcolor{blue!23} $0.16^{*}$ & $+0.00$ & \cellcolor{blue!15} $0.10^{*}$ & \cellcolor{red!4} $-0.03$ & \cellcolor{blue!21} $0.14^{*}$ & \cellcolor{red!1} $-0.01$ & \cellcolor{blue!9} $0.06^{*}$ & \cellcolor{red!8} $-0.06^{*}$ & \cellcolor{blue!23} $0.15^{*}$ & \cellcolor{red!5} $-0.04$ & \cellcolor{blue!13} $0.09^{*}$ \\
\hhline{~----------------}
 & \multirow{3}{*}{32} & $k=1$ & $+0.00$ & \cellcolor{blue!13} $0.09^{*}$ & \cellcolor{red!1} $-0.01$ & \cellcolor{blue!13} $0.09^{*}$ & \cellcolor{red!2} $-0.02$ & \cellcolor{blue!14} $0.10^{*}$ & \cellcolor{green!1} $+0.01$ & \cellcolor{blue!20} $0.13^{*}$ & \cellcolor{green!5} $+0.04$ & \cellcolor{blue!16} $0.11^{*}$ & $+0.00$ & \cellcolor{blue!21} $0.14^{*}$ & \cellcolor{red!1} $-0.01$ & \cellcolor{blue!13} $0.09^{*}$ \\
 &  & $k=2$ & \cellcolor{green!1} $+0.01$ & \cellcolor{blue!25} $0.17^{*}$ & $-0.01$ & \cellcolor{blue!27} $0.18^{*}$ & \cellcolor{green!2} $+0.02$ & \cellcolor{blue!24} $0.17^{*}$ & \cellcolor{red!1} $-0.01$ & \cellcolor{blue!28} $0.19^{*}$ & \cellcolor{red!1} $-0.01$ & \cellcolor{blue!24} $0.16^{*}$ & $+0.01$ & \cellcolor{blue!29} $0.20^{*}$ & $+0.00$ & \cellcolor{blue!24} $0.16^{*}$ \\
 &  & $k=3$ & \cellcolor{red!8} $-0.06^{*}$ & \cellcolor{blue!15} $0.10^{*}$ & \cellcolor{red!5} $-0.04$ & \cellcolor{blue!16} $0.11^{*}$ & \cellcolor{green!2} $+0.02$ & \cellcolor{blue!10} $0.07^{*}$ & \cellcolor{red!1} $-0.01$ & \cellcolor{blue!15} $0.11^{*}$ & $+0.00$ & \cellcolor{blue!4} $0.03$ & \cellcolor{green!1} $+0.01$ & \cellcolor{blue!16} $0.11^{*}$ & \cellcolor{green!1} $+0.01$ & \cellcolor{blue!9} $0.07^{*}$ \\
\hhline{~----------------}
 & \multirow{3}{*}{256} & $k=1$ & \cellcolor{red!2} $-0.02$ & \cellcolor{blue!16} $0.11^{*}$ & \cellcolor{red!2} $-0.02$ & \cellcolor{blue!13} $0.09^{*}$ & \cellcolor{red!4} $-0.03$ & \cellcolor{blue!14} $0.10^{*}$ & \cellcolor{green!4} $+0.03$ & \cellcolor{blue!20} $0.13^{*}$ & \cellcolor{green!6} $+0.04$ & \cellcolor{blue!16} $0.11^{*}$ & \cellcolor{green!4} $+0.03$ & \cellcolor{blue!22} $0.15^{*}$ & \cellcolor{green!1} $+0.01$ & \cellcolor{blue!14} $0.09^{*}$ \\
 &  & $k=2$ & \cellcolor{green!2} $+0.02$ & \cellcolor{blue!24} $0.16^{*}$ & $+0.00$ & \cellcolor{blue!25} $0.17^{*}$ & \cellcolor{red!4} $-0.03$ & \cellcolor{blue!19} $0.13^{*}$ & \cellcolor{red!1} $-0.01$ & \cellcolor{blue!26} $0.18^{*}$ & \cellcolor{red!3} $-0.02$ & \cellcolor{blue!18} $0.12^{*}$ & \cellcolor{green!4} $+0.03$ & \cellcolor{blue!27} $0.18^{*}$ & \cellcolor{green!1} $+0.01$ & \cellcolor{blue!19} $0.13^{*}$ \\
 &  & $k=3$ & \cellcolor{red!3} $-0.02$ & \cellcolor{blue!12} $0.08^{*}$ & \cellcolor{red!3} $-0.02$ & \cellcolor{blue!12} $0.08^{*}$ & \cellcolor{red!5} $-0.04$ & \cellcolor{blue!2} $0.02$ & \cellcolor{red!1} $-0.01$ & \cellcolor{blue!10} $0.07^{*}$ & \cellcolor{red!2} $-0.02$ & $-0.00$ & \cellcolor{green!2} $+0.02$ & \cellcolor{blue!12} $0.09^{*}$ & \cellcolor{red!4} $-0.03$ & \cellcolor{blue!1} $0.01$ \\
\bottomrule
\end{NiceTabular}
}
\caption{Spearman correlation ($\rho$) between matrix reconstruction errors and evaluation score differences across varying reference pool sizes ($|\mathcal{Y}|$) on WMT22 De$\rightarrow$En. \textit{Dir.} (Directional Change) measures score improvements, highlighted as \colorbox{green!15}{positive} and \colorbox{red!15}{negative} impacts. \textit{Mag.} (Magnitude Sensitivity) measures absolute score variance, with \colorbox{blue!15}{blue} highlighting metric sensitivity to matrix modifications. Statistical significance ($p < 0.05$) is denoted by an asterisk ($^*$).}
\label{tab:corr_mat_analysis_complete_wmt22-deen}
\end{table*}

\begin{table*}[ht!]
\centering
\resizebox{\textwidth}{!}{
\begin{tabular}{llllllllll}
\toprule
\textbf{Util. Function} & \textbf{$|\mathcal{Y}|$} & \textbf{Top-$k$} & \textbf{BLEU} & \textbf{chrF} & \textbf{BLEURT} & \textbf{COMET} & \textbf{BERTScore} & \textbf{COMETKiwi} & \textbf{$\bar{Z}_{\text{other}}$} \\
\midrule
\multirow[t]{9}{*}{BLEU} & \multirow[t]{3}{*}{4} & $k=1$ & \cellcolor{red!15} 440 / 1076 / 468 & \cellcolor{red!15} 491 / 992 / 501 & \cellcolor{green!15} 500 / 989 / 495 & \cellcolor{red!15} 477 / 993 / 514 & \cellcolor{red!15} 467 / 987 / 530 & \cellcolor{red!15} 467 / 995 / 522 & \cellcolor{red!15} 486 / 987 / 511 \\
 &  & $k=2$ & \cellcolor{red!15} 210 / 1507 / 267 & \cellcolor{red!15} 233 / 1446 / 305 & \cellcolor{red!15} 245 / 1447 / 292 & \cellcolor{red!15} 252 / 1448 / 284 & \cellcolor{red!15} 241 / 1444 / 299 & \cellcolor{red!15} 244 / 1451 / 289 & \cellcolor{red!15} 241 / 1444 / 299 \\
 &  & $k=3$ & \cellcolor{red!15} 97 / 1769 / 118 & \cellcolor{red!15} 114 / 1732 / 138 & \cellcolor{red!15} 111 / 1729 / 144 & \cellcolor{red!15} 97 / 1733 / 154 & \cellcolor{red!15} 117 / 1729 / 138 & \cellcolor{red!15} 111 / 1731 / 142 & \cellcolor{red!15} 105 / 1729 / 150 \\
\cline{2-10}
 & \multirow[t]{3}{*}{32} & $k=1$ & \cellcolor{green!15} 350 / 1294 / 340 & \cellcolor{green!15} 412 / 1174 / 398 & \cellcolor{red!15} 399 / 1160 / 425 & \cellcolor{green!15} 426 / 1167 / 391 & \cellcolor{red!15} 410 / 1153 / 421 & \cellcolor{red!15} 391 / 1174 / 419 & \cellcolor{green!15} 426 / 1133 / 425 \\
 &  & $k=2$ & \cellcolor{red!15} 233 / 1472 / 279 & \cellcolor{red!15} 282 / 1391 / 311 & \cellcolor{red!15} 297 / 1366 / 321 & \cellcolor{red!15} 280 / 1385 / 319 & \cellcolor{red!15} 308 / 1366 / 310 & \cellcolor{red!15} 280 / 1391 / 313 & \cellcolor{red!15} 310 / 1339 / 335 \\
 &  & $k=3$ & \cellcolor{red!15} 168 / 1614 / 202 & \cellcolor{red!15} 208 / 1556 / 220 & \cellcolor{green!15} 233 / 1525 / 226 & \cellcolor{red!15} 215 / 1544 / 225 & \cellcolor{red!15} 225 / 1533 / 226 & \cellcolor{red!15} 202 / 1549 / 233 & \cellcolor{red!15} 236 / 1503 / 245 \\
\cline{2-10}
 & \multirow[t]{3}{*}{256} & $k=1$ & \cellcolor{red!15} 226 / 1498 / 260 & \cellcolor{green!15} 277 / 1433 / 274 & \cellcolor{green!15} 311 / 1425 / 248 & \cellcolor{red!15} 267 / 1429 / 288 & \cellcolor{red!15} 275 / 1419 / 290 & \cellcolor{green!15} 276 / 1433 / 275 & \cellcolor{green!15} 294 / 1404 / 286 \\
 &  & $k=2$ & \cellcolor{red!15} 137 / 1682 / 165 & \cellcolor{red!15} 172 / 1628 / 184 & \cellcolor{green!15} 193 / 1613 / 178 & \cellcolor{red!15} 169 / 1617 / 198 & \cellcolor{red!15} 175 / 1611 / 198 & \cellcolor{green!15} 178 / 1629 / 177 & \cellcolor{red!15} 192 / 1587 / 205 \\
 &  & $k=3$ & \cellcolor{red!15} 93 / 1782 / 109 & \cellcolor{green!15} 129 / 1739 / 116 & \cellcolor{green!15} 137 / 1718 / 129 & \cellcolor{green!15} 130 / 1730 / 124 & \cellcolor{red!15} 122 / 1716 / 146 & \cellcolor{green!15} 131 / 1732 / 121 & \cellcolor{green!15} 150 / 1685 / 149 \\
\cline{1-10} \cline{2-10}
\multirow[t]{9}{*}{chrF} & \multirow[t]{3}{*}{4} & $k=1$ & \cellcolor{red!15} 420 / 1100 / 464 & \cellcolor{red!15} 437 / 1014 / 533 & \cellcolor{red!15} 453 / 1012 / 519 & \cellcolor{red!15} 476 / 1014 / 494 & \cellcolor{red!15} 455 / 1009 / 520 & \cellcolor{red!15} 471 / 1019 / 494 & \cellcolor{red!15} 451 / 1009 / 524 \\
 &  & $k=2$ & \cellcolor{green!15} 215 / 1572 / 197 & \cellcolor{red!15} 224 / 1525 / 235 & \cellcolor{red!15} 224 / 1522 / 238 & \cellcolor{green!15} 248 / 1521 / 215 & \cellcolor{green!15} 233 / 1520 / 231 & \cellcolor{green!15} 233 / 1528 / 223 & \cellcolor{green!15} 239 / 1519 / 226 \\
 &  & $k=3$ & \cellcolor{red!15} 80 / 1821 / 83 & \cellcolor{red!15} 93 / 1796 / 95 & \cellcolor{red!15} 85 / 1793 / 106 & \cellcolor{red!15} 92 / 1794 / 98 & \cellcolor{red!15} 89 / 1792 / 103 & \cellcolor{red!15} 92 / 1797 / 95 & \cellcolor{red!15} 92 / 1791 / 101 \\
\cline{2-10}
 & \multirow[t]{3}{*}{32} & $k=1$ & \cellcolor{red!15} 362 / 1249 / 373 & \cellcolor{red!15} 355 / 1146 / 483 & \cellcolor{red!15} 376 / 1132 / 476 & \cellcolor{red!15} 384 / 1146 / 454 & \cellcolor{red!15} 407 / 1128 / 449 & \cellcolor{red!15} 392 / 1151 / 441 & \cellcolor{red!15} 392 / 1115 / 477 \\
 &  & $k=2$ & \cellcolor{red!15} 192 / 1578 / 214 & \cellcolor{red!15} 216 / 1511 / 257 & \cellcolor{red!15} 232 / 1493 / 259 & \cellcolor{red!15} 224 / 1503 / 257 & \cellcolor{red!15} 228 / 1493 / 263 & \cellcolor{green!15} 237 / 1513 / 234 & \cellcolor{red!15} 232 / 1471 / 281 \\
 &  & $k=3$ & \cellcolor{red!15} 122 / 1724 / 138 & \cellcolor{red!15} 139 / 1686 / 159 & \cellcolor{red!15} 156 / 1664 / 164 & \cellcolor{green!15} 164 / 1674 / 146 & \cellcolor{green!15} 166 / 1666 / 152 & \cellcolor{green!15} 157 / 1686 / 141 & \cellcolor{green!15} 175 / 1638 / 171 \\
\cline{2-10}
 & \multirow[t]{3}{*}{256} & $k=1$ & \cellcolor{green!15} 324 / 1375 / 285 & \cellcolor{red!15} 281 / 1295 / 408 & \cellcolor{red!15} 306 / 1282 / 396 & \cellcolor{red!15} 304 / 1292 / 388 & \cellcolor{red!15} 337 / 1279 / 368 & \cellcolor{red!15} 299 / 1295 / 390 & \cellcolor{red!15} 320 / 1267 / 397 \\
 &  & $k=2$ & \cellcolor{green!15} 200 / 1590 / 194 & \cellcolor{red!15} 208 / 1532 / 244 & \cellcolor{red!15} 229 / 1512 / 243 & \cellcolor{red!15} 225 / 1518 / 241 & \cellcolor{red!15} 229 / 1509 / 246 & \cellcolor{red!15} 219 / 1527 / 238 & \cellcolor{red!15} 236 / 1490 / 258 \\
 &  & $k=3$ & \cellcolor{red!15} 122 / 1720 / 142 & \cellcolor{red!15} 134 / 1680 / 170 & \cellcolor{red!15} 150 / 1663 / 171 & \cellcolor{red!15} 151 / 1668 / 165 & \cellcolor{red!15} 151 / 1659 / 174 & \cellcolor{red!15} 145 / 1675 / 164 & \cellcolor{red!15} 158 / 1643 / 183 \\
\cline{1-10} \cline{2-10}
\multirow[t]{9}{*}{BLEURT} & \multirow[t]{3}{*}{4} & $k=1$ & \cellcolor{red!15} 312 / 1352 / 320 & \cellcolor{red!15} 338 / 1288 / 358 & \cellcolor{red!15} 234 / 1286 / 464 & \cellcolor{red!15} 325 / 1286 / 373 & \cellcolor{green!15} 352 / 1281 / 351 & \cellcolor{red!15} 310 / 1289 / 385 & \cellcolor{red!15} 333 / 1279 / 372 \\
 &  & $k=2$ & \cellcolor{red!15} 115 / 1738 / 131 & \cellcolor{green!15} 144 / 1710 / 130 & \cellcolor{red!15} 109 / 1707 / 168 & \cellcolor{red!15} 134 / 1710 / 140 & \cellcolor{red!15} 127 / 1706 / 151 & \cellcolor{red!15} 123 / 1709 / 152 & \cellcolor{red!15} 130 / 1704 / 150 \\
 &  & $k=3$ & \cellcolor{red!15} 40 / 1900 / 44 & \cellcolor{green!15} 59 / 1884 / 41 & \cellcolor{green!15} 52 / 1882 / 50 & \cellcolor{green!15} 61 / 1882 / 41 & \cellcolor{green!15} 52 / 1881 / 51 & \cellcolor{red!15} 49 / 1884 / 51 & \cellcolor{green!15} 59 / 1880 / 45 \\
\cline{2-10}
 & \multirow[t]{3}{*}{32} & $k=1$ & \cellcolor{green!15} 332 / 1338 / 314 & \cellcolor{red!15} 349 / 1268 / 367 & \cellcolor{red!15} 226 / 1252 / 506 & \cellcolor{red!15} 335 / 1260 / 389 & \cellcolor{green!15} 377 / 1250 / 357 & \cellcolor{red!15} 357 / 1266 / 361 & \cellcolor{green!15} 371 / 1244 / 369 \\
 &  & $k=2$ & \cellcolor{green!15} 137 / 1730 / 117 & \cellcolor{green!15} 155 / 1693 / 136 & \cellcolor{red!15} 145 / 1674 / 165 & \cellcolor{red!15} 141 / 1687 / 156 & \cellcolor{red!15} 147 / 1675 / 162 & \cellcolor{green!15} 155 / 1689 / 140 & \cellcolor{green!15} 162 / 1665 / 157 \\
 &  & $k=3$ & \cellcolor{green!15} 75 / 1847 / 62 & \cellcolor{green!15} 86 / 1827 / 71 & \cellcolor{red!15} 84 / 1809 / 91 & \cellcolor{green!15} 84 / 1822 / 78 & \cellcolor{green!15} 93 / 1809 / 82 & \cellcolor{green!15} 91 / 1825 / 68 & \cellcolor{green!15} 104 / 1796 / 84 \\
\cline{2-10}
 & \multirow[t]{3}{*}{256} & $k=1$ & \cellcolor{green!15} 362 / 1351 / 271 & \cellcolor{green!15} 369 / 1261 / 354 & \cellcolor{red!15} 232 / 1248 / 504 & \cellcolor{red!15} 350 / 1255 / 379 & \cellcolor{green!15} 375 / 1242 / 367 & \cellcolor{red!15} 358 / 1262 / 364 & \cellcolor{green!15} 384 / 1237 / 363 \\
 &  & $k=2$ & \cellcolor{green!15} 157 / 1697 / 130 & \cellcolor{green!15} 176 / 1641 / 167 & \cellcolor{red!15} 170 / 1627 / 187 & \cellcolor{red!15} 169 / 1634 / 181 & \cellcolor{red!15} 167 / 1626 / 191 & \cellcolor{green!15} 183 / 1638 / 163 & \cellcolor{red!15} 177 / 1622 / 185 \\
 &  & $k=3$ & \cellcolor{green!15} 83 / 1823 / 78 & \cellcolor{green!15} 102 / 1793 / 89 & \cellcolor{red!15} 101 / 1780 / 103 & \cellcolor{green!15} 103 / 1788 / 93 & \cellcolor{red!15} 97 / 1781 / 106 & \cellcolor{green!15} 100 / 1792 / 92 & \cellcolor{red!15} 103 / 1771 / 110 \\
\cline{1-10} \cline{2-10}
\multirow[t]{9}{*}{COMET} & \multirow[t]{3}{*}{4} & $k=1$ & \cellcolor{red!15} 220 / 1496 / 268 & \cellcolor{red!15} 247 / 1451 / 286 & \cellcolor{red!15} 238 / 1443 / 303 & \cellcolor{red!15} 191 / 1445 / 348 & \cellcolor{red!15} 241 / 1443 / 300 & \cellcolor{red!15} 241 / 1448 / 295 & \cellcolor{red!15} 235 / 1440 / 309 \\
 &  & $k=2$ & \cellcolor{red!15} 79 / 1803 / 102 & \cellcolor{red!15} 96 / 1783 / 105 & \cellcolor{red!15} 95 / 1778 / 111 & \cellcolor{red!15} 90 / 1780 / 114 & \cellcolor{red!15} 101 / 1778 / 105 & \cellcolor{red!15} 94 / 1781 / 109 & \cellcolor{red!15} 101 / 1775 / 108 \\
 &  & $k=3$ & \cellcolor{green!15} 38 / 1913 / 33 & \cellcolor{green!15} 46 / 1906 / 32 & \cellcolor{green!15} 43 / 1903 / 38 & \cellcolor{red!15} 34 / 1904 / 46 & \cellcolor{green!15} 41 / 1903 / 40 & \cellcolor{green!15} 41 / 1906 / 37 & \cellcolor{green!15} 43 / 1899 / 42 \\
\cline{2-10}
 & \multirow[t]{3}{*}{32} & $k=1$ & \cellcolor{green!15} 215 / 1574 / 195 & \cellcolor{green!15} 235 / 1524 / 225 & \cellcolor{red!15} 211 / 1514 / 259 & \cellcolor{red!15} 163 / 1523 / 298 & \cellcolor{green!15} 254 / 1514 / 216 & 227 / 1530 / 227 & \cellcolor{red!15} 242 / 1499 / 243 \\
 &  & $k=2$ & \cellcolor{green!15} 107 / 1794 / 83 & \cellcolor{green!15} 112 / 1766 / 106 & \cellcolor{green!15} 120 / 1758 / 106 & \cellcolor{red!15} 109 / 1765 / 110 & \cellcolor{green!15} 135 / 1761 / 88 & \cellcolor{green!15} 125 / 1769 / 90 & \cellcolor{green!15} 134 / 1742 / 108 \\
 &  & $k=3$ & \cellcolor{green!15} 72 / 1855 / 57 & \cellcolor{green!15} 71 / 1849 / 64 & \cellcolor{green!15} 80 / 1841 / 63 & \cellcolor{green!15} 80 / 1847 / 57 & \cellcolor{green!15} 85 / 1842 / 57 & \cellcolor{green!15} 79 / 1852 / 53 & \cellcolor{green!15} 88 / 1819 / 77 \\
\cline{2-10}
 & \multirow[t]{3}{*}{256} & $k=1$ & \cellcolor{green!15} 193 / 1613 / 178 & \cellcolor{green!15} 215 / 1556 / 213 & \cellcolor{red!15} 218 / 1542 / 224 & \cellcolor{red!15} 169 / 1550 / 265 & \cellcolor{green!15} 241 / 1548 / 195 & \cellcolor{red!15} 213 / 1557 / 214 & \cellcolor{green!15} 239 / 1533 / 212 \\
 &  & $k=2$ & \cellcolor{red!15} 83 / 1808 / 93 & \cellcolor{red!15} 104 / 1772 / 108 & \cellcolor{green!15} 128 / 1760 / 96 & \cellcolor{green!15} 123 / 1767 / 94 & \cellcolor{green!15} 135 / 1762 / 87 & \cellcolor{green!15} 127 / 1773 / 84 & \cellcolor{green!15} 140 / 1746 / 98 \\
 &  & $k=3$ & \cellcolor{red!15} 52 / 1876 / 56 & \cellcolor{red!15} 56 / 1861 / 67 & \cellcolor{green!15} 74 / 1849 / 61 & \cellcolor{green!15} 67 / 1861 / 56 & \cellcolor{green!15} 78 / 1857 / 49 & \cellcolor{green!15} 69 / 1864 / 51 & \cellcolor{green!15} 87 / 1835 / 62 \\
\cline{1-10} \cline{2-10}
\multirow[t]{9}{*}{BERTScore} & \multirow[t]{3}{*}{4} & $k=1$ & \cellcolor{green!15} 453 / 1081 / 450 & \cellcolor{red!15} 474 / 984 / 526 & \cellcolor{green!15} 533 / 958 / 493 & \cellcolor{red!15} 501 / 973 / 510 & \cellcolor{red!15} 471 / 954 / 559 & \cellcolor{green!15} 500 / 988 / 496 & \cellcolor{green!15} 515 / 955 / 514 \\
 &  & $k=2$ & \cellcolor{green!15} 173 / 1640 / 171 & \cellcolor{red!15} 171 / 1603 / 210 & \cellcolor{green!15} 204 / 1579 / 201 & \cellcolor{green!15} 200 / 1595 / 189 & \cellcolor{red!15} 177 / 1583 / 224 & \cellcolor{red!15} 187 / 1604 / 193 & \cellcolor{green!15} 209 / 1575 / 200 \\
 &  & $k=3$ & \cellcolor{green!15} 63 / 1884 / 37 & \cellcolor{green!15} 56 / 1874 / 54 & \cellcolor{green!15} 69 / 1848 / 67 & 58 / 1868 / 58 & \cellcolor{red!15} 69 / 1845 / 70 & \cellcolor{red!15} 44 / 1873 / 67 & \cellcolor{red!15} 70 / 1843 / 71 \\
\cline{2-10}
 & \multirow[t]{3}{*}{32} & $k=1$ & \cellcolor{green!15} 463 / 1087 / 434 & \cellcolor{green!15} 551 / 974 / 459 & \cellcolor{green!15} 575 / 941 / 468 & \cellcolor{green!15} 603 / 968 / 413 & 524 / 936 / 524 & \cellcolor{green!15} 597 / 979 / 408 & \cellcolor{green!15} 606 / 929 / 449 \\
 &  & $k=2$ & \cellcolor{red!15} 185 / 1611 / 188 & \cellcolor{red!15} 206 / 1567 / 211 & \cellcolor{green!15} 240 / 1528 / 216 & \cellcolor{green!15} 218 / 1562 / 204 & \cellcolor{red!15} 225 / 1524 / 235 & \cellcolor{green!15} 239 / 1566 / 179 & \cellcolor{green!15} 233 / 1522 / 229 \\
 &  & $k=3$ & \cellcolor{red!15} 93 / 1781 / 110 & \cellcolor{red!15} 113 / 1755 / 116 & \cellcolor{green!15} 152 / 1700 / 132 & \cellcolor{green!15} 130 / 1744 / 110 & \cellcolor{red!15} 117 / 1708 / 159 & \cellcolor{green!15} 136 / 1752 / 96 & \cellcolor{green!15} 149 / 1695 / 140 \\
\cline{2-10}
 & \multirow[t]{3}{*}{256} & $k=1$ & \cellcolor{green!15} 420 / 1159 / 405 & \cellcolor{green!15} 491 / 1046 / 447 & \cellcolor{green!15} 542 / 1009 / 433 & \cellcolor{green!15} 571 / 1032 / 381 & \cellcolor{green!15} 507 / 1005 / 472 & \cellcolor{green!15} 566 / 1047 / 371 & \cellcolor{green!15} 576 / 995 / 413 \\
 &  & $k=2$ & \cellcolor{green!15} 162 / 1670 / 152 & \cellcolor{green!15} 193 / 1620 / 171 & \cellcolor{green!15} 210 / 1574 / 200 & \cellcolor{green!15} 228 / 1605 / 151 & \cellcolor{green!15} 208 / 1587 / 189 & \cellcolor{green!15} 223 / 1618 / 143 & \cellcolor{green!15} 238 / 1565 / 181 \\
 &  & $k=3$ & \cellcolor{red!15} 78 / 1810 / 96 & \cellcolor{green!15} 108 / 1780 / 96 & \cellcolor{green!15} 126 / 1735 / 123 & \cellcolor{green!15} 115 / 1763 / 106 & \cellcolor{green!15} 125 / 1745 / 114 & \cellcolor{green!15} 127 / 1777 / 80 & \cellcolor{green!15} 141 / 1724 / 119 \\
\cline{1-10} \cline{2-10}
\bottomrule
\end{tabular}

}
\caption{Sentence-level comparison between SVD-MBR and MBR on the WMT22 De$\rightarrow$En. Cells are formatted as \textbf{W / T / L}, indicating the number of sentences where SVD-MBR achieved a higher score (Win), an identical score (Tie), or a lower score (Loss). The rightmost column, $\bar{Z}_{\text{other}}$, aggregates the net win rate across all off-target evaluation metrics. \colorbox{green!15}{Positive} and \colorbox{red!15}{Negative} win rate for SVD-MBR is indicated by the color of the cells.}
\label{tab:svd_wtl_full_wmt22_deen}
\end{table*}

\begin{table*}[ht!]
\centering
\small
\begin{tabular}{@{} l c c @{\hspace{0.2\textwidth}} l c c @{}}
\toprule
\multicolumn{3}{@{}p{0.4\textwidth}@{}}{\textbf{Case 1: COMET as Util. Function}} & \multicolumn{3}{@{}p{0.4\textwidth}@{}}{\textbf{Case 2: BLEU as Util. Function}} \\
\midrule
\multicolumn{3}{@{}p{0.4\textwidth}@{}}{\textbf{Source:} Weder ServusTV noch ORF tun das.} & \multicolumn{3}{@{}p{0.4\textwidth}@{}}{\textbf{Source:} geht das bei ihnen?} \\
\addlinespace
\multicolumn{3}{@{}p{0.4\textwidth}@{}}{\textbf{Reference:} Neither ServusTV nor ORF do this.} & \multicolumn{3}{@{}p{0.4\textwidth}@{}}{\textbf{Reference:} does that work for you?} \\
\midrule
\multicolumn{3}{@{}p{0.4\textwidth}@{}}{\textbf{MBR:}} & \multicolumn{3}{@{}p{0.4\textwidth}@{}}{\textbf{MBR:}} \\
\multicolumn{3}{@{}p{0.4\textwidth}@{}}{Neither TV nor ARF do that.} & \multicolumn{3}{@{}p{0.4\textwidth}@{}}{Does that happen to them?} \\
\addlinespace
\multicolumn{3}{@{}p{0.4\textwidth}@{}}{\textbf{SVD-MBR:}} & \multicolumn{3}{@{}p{0.4\textwidth}@{}}{\textbf{SVD-MBR:}} \\
\multicolumn{3}{@{}p{0.4\textwidth}@{}}{Neither Servus TV nor ORF do that.} & \multicolumn{3}{@{}p{0.48\textwidth}@{}}{Is it going to them?} \\
\midrule
\textbf{Metric} & \textbf{MBR} & \textbf{SVD-MBR} & \textbf{Metric} & \textbf{MBR} & \textbf{SVD-MBR} \\
\midrule
\textbf{BLEU} & 9.3 & \textbf{23.4} (\textcolor{green}{$\uparrow$}) & \textbf{BLEU} (Util) & \textbf{9.7} & 8.1 (\textcolor{red}{$\downarrow$}) \\
\textbf{chrF} & 42.6 & \textbf{88.8} (\textcolor{green}{$\uparrow$}) & \textbf{chrF} & \textbf{27.6} & 8.2 (\textcolor{red}{$\downarrow$}) \\
\textbf{BLEURT} & 63.6 & \textbf{83.6} (\textcolor{green}{$\uparrow$}) & \textbf{BLEURT} & \textbf{61.7} & 43.6 (\textcolor{red}{$\downarrow$}) \\
\textbf{COMET} (Util) & 91.8 & \textbf{96.5} (\textcolor{green}{$\uparrow$}) & \textbf{COMET} & \textbf{70.2} & 49.8 (\textcolor{red}{$\downarrow$}) \\
\textbf{BERTScore} & 82.2 & \textbf{86.8} (\textcolor{green}{$\uparrow$}) & \textbf{BERTScore} & \textbf{68.3} & 59.2 (\textcolor{red}{$\downarrow$}) \\
\textbf{COMETKiwi} & 60.6 & \textbf{85.1} (\textcolor{green}{$\uparrow$}) & \textbf{COMETKiwi} & \textbf{77.8} & 50.0 (\textcolor{red}{$\downarrow$}) \\
\bottomrule
\end{tabular}
\caption{Sentence comparison of MBR and SVD-MBR ($k=1$) using $|\mathcal{Y}|=256$. \textbf{Case 1} demonstrates the comparison using COMET as the utility function and \textbf{Case 2} demonstrates the comparison using BLEU as the utility function.}
\label{tab:qualitative_side_by_side_unified_wmt22-deen}
\end{table*}

\section{Additional Experiments}
\label{appendix:additional_exp}

To further examine whether overfitting in MBR and the success of our SVD-MBR approach in mitigating the overfitting is not an occurance of a single dataset, we expand our experiment using two datasets, the WMT-22 De$\rightarrow$En and XSum~\cite{narayan-etal-2018-dont}. 

\subsection{WMT22 De$\rightarrow$En}

Table~\ref{tab:mbr_overfitting_wmt22_deen} presents the overfitting results for the WMT-22 De$\rightarrow$En dataset. \textbf{Consistent with our findings on the En$\rightarrow$De direction, we observe persistent metric overfitting across all evaluated utility functions}. To assess the generalizability of our mitigation strategy, we applied SVD-MBR to this dataset, with the comparative performance detailed in Table~\ref{tab:svd_mbr_results_wmt22_deen}.

Interestingly, SVD-MBR yields mixed results on surface-level metrics for this language pair. When optimizing for BLEU at a high reference count~($|\mathcal{Y}|=256$) with a rank-1 approximation~($k=1$), we observe a significant increase in the off-target BLEURT score. However, optimizing for chrF yields no generalized improvement under any configuration. This instability further solidifies our hypothesis that surface-level metrics lack a robust, low-rank structural consensus, making them highly unreliable candidates for SVD denoising. 

Conversely, neural metrics exhibit the same consistent, positive denoising behavior observed previously. \textbf{One notable divergence from the En$\rightarrow$De dataset is the exceptional performance of BERTScore on De$\rightarrow$En}. When utilized as the utility function, BERTScore demonstrates significant improvements across all off-target metrics even under the most aggressive rank truncation~($k=1$), further reinforcing the compatibility between dense semantic embeddings and low-rank matrix approximation.

We further conducted a Spearman correlation analysis for the WMT22 De$\rightarrow$En dataset, following the methodology established in Appendix~\ref{appendix:error_analysis}. Tables~\ref{tab:corr_variance_analysis_complete_wmt22-deen} and \ref{tab:corr_mat_analysis_complete_wmt22-deen} present the correlations of the hypothesis oracle variance and the matrix reconstruction error against the score differences, respectively. Interestingly, \textbf{the intrinsic variance of the metrics diverges notably from the En$\rightarrow$De dataset}. Specifically, the variances for COMET~($\sigma^2=39.038$) and BLEURT~($\sigma^2=34.042$) are substantially lower, whereas BERTScore~($\sigma^2=16.240$) exhibits slightly increased variance. This shift theoretically explains why BERTScore emerges as the utility function with the most consistent off-target improvements for this specific language pair. Furthermore, despite their reduced variance, optimizing for both COMET and BLEURT still yields predominantly positive performance gains. Conversely, the correlations for surface-level metrics like BLEU and chrF remain unstable, mirroring the poor generalizability observed in the baseline performance comparisons.

Analyzing the matrix reconstruction error against the score differences~(Table~\ref{tab:corr_mat_analysis_complete_wmt22-deen}) yields a nuanced result. While the correlation magnitude~(\textit{Mag.}) shows a highly significant positive relationship with the aggregated off-target metrics~($\bar{Z}_{\text{other}}$), the directional correlation~(\textit{Dir.}) is frequently negative. This indicates a strong inverse relationship: higher matrix reconstruction errors reliably correlate with negative performance deltas~(score degradation). Consequently, \textbf{for the De$\rightarrow$En task, forcing an inaccurate low-rank approximation on the matrix directly harms the final selection quality}.

Furthermore, Table~\ref{tab:svd_wtl_full_wmt22_deen} presents the complete sentence-level Win/Tie/Loss~(WTL) distribution for the WMT22 De$\rightarrow$En dataset. Consistent with our primary findings on the En$\rightarrow$De task, the vast majority of SVD-MBR selections result in a Tie, reaffirming that \textbf{SVD-MBR acts as a safe, non-destructive regularizer regardless of the language direction}. Reflecting the aggregate performance results from Table~\ref{tab:svd_mbr_results_wmt22_deen}, the most substantial and consistent net win rates across the off-target metrics~($\bar{Z}_{\text{other}}$) occur when optimizing for BERTScore, followed closely by COMET and BLEURT. Conversely, the surface-level metric chrF rarely yields a net positive win rate, failing to improve generalized quality except in specific edge cases~(e.g., $|\mathcal{Y}|=4, k=2$ and $|\mathcal{Y}|=32, k=3$). Interestingly, BLEU demonstrates a slightly more favorable distribution than chrF on this dataset, achieving positive net wins under several configurations, particularly at the largest reference pool size~($|\mathcal{Y}|=256$).

These quantitative patterns are supported by the qualitative examples in Table~\ref{tab:qualitative_side_by_side_unified_wmt22-deen}, which align with our En$\rightarrow$De results. When SVD-MBR utilizes COMET as the utility function, we \textbf{observe consistent improvements across all metrics}, including COMET itself. In contrast, using BLEU as the utility function results in performance degradation across all metrics, underscoring the metric's failure to provide a meaningful consensus signal in this denoising paradigm.

\begin{table*}[ht!]
\centering
\resizebox{\textwidth}{!}{
\begin{tabular}{p{0.2\linewidth} p{0.15\linewidth} l l l l l l}
\toprule
\textbf{Decoding Method} & \textbf{Util. Function} & \textbf{$|\mathcal{Y}|$} & \textbf{BERTScore} & \textbf{ROUGE-1} & \textbf{ROUGE-2} & \textbf{ROUGE-L} & \textbf{$\bar{Z}_{\text{other}}$} \\
\midrule
MAP$\epsilon$ & N/A & - & 75.631 & 43.643 & 21.490 & 36.184 & -0.093 \\
\cline{1-8} \cline{2-8}
\multirow[t]{12}{*}{MBR} & \multirow[t]{3}{*}{BERTScore} & 4 & \cellcolor{green!21} 76.450 (+0.819) & \cellcolor{red!10} 42.777 (-0.866) & \cellcolor{red!17} 19.882 (-1.609) & \cellcolor{red!11} 35.218 (-0.966) & \cellcolor{red!11} -0.641 (-0.549) \\
 &  & 32 & \cellcolor{green!45} 77.345 (+1.714) & \cellcolor{green!5} 44.057 (+0.414) & \cellcolor{red!4} 21.110 (-0.380) & \cellcolor{green!4} 36.600 (+0.416) & \cellcolor{green!4} 0.123 (+0.216) \\
 &  & 256 & \cellcolor{green!50} 77.501 (+1.870) & \cellcolor{green!8} 44.376 (+0.733) & \cellcolor{red!1} 21.378 (-0.112) & \cellcolor{green!8} 36.937 (+0.754) & \cellcolor{green!8} 0.304 (+0.397) \\
\cline{2-8}
 & \multirow[t]{3}{*}{ROUGE-1} & 4 & \cellcolor{red!2} 75.533 (-0.098) & \cellcolor{green!2} 43.863 (+0.220) & \cellcolor{red!14} 20.121 (-1.369) & \cellcolor{red!9} 35.398 (-0.785) & \cellcolor{red!9} -0.535 (-0.443) \\
 &  & 32 & \cellcolor{green!13} 76.138 (+0.507) & \cellcolor{green!30} 46.114 (+2.471) & \cellcolor{green!2} 21.680 (+0.190) & \cellcolor{green!10} 37.093 (+0.909) & \cellcolor{green!9} 0.379 (+0.472) \\
 &  & 256 & \cellcolor{green!17} 76.293 (+0.662) & \cellcolor{green!36} 46.608 (+2.965) & \cellcolor{green!6} 22.126 (+0.636) & \cellcolor{green!16} 37.594 (+1.411) & \cellcolor{green!15} 0.637 (+0.730) \\
\cline{2-8}
 & \multirow[t]{3}{*}{ROUGE-2} & 4 & \cellcolor{red!4} 75.447 (-0.183) & \cellcolor{red!9} 42.894 (-0.749) & \cellcolor{red!11} 20.435 (-1.056) & \cellcolor{red!10} 35.299 (-0.885) & \cellcolor{red!11} -0.624 (-0.531) \\
 &  & 32 & \cellcolor{green!14} 76.170 (+0.539) & \cellcolor{green!15} 44.885 (+1.242) & \cellcolor{green!12} 22.654 (+1.163) & \cellcolor{green!14} 37.367 (+1.183) & \cellcolor{green!11} 0.459 (+0.552) \\
 &  & 256 & \cellcolor{green!17} 76.285 (+0.655) & \cellcolor{green!19} 45.246 (+1.603) & \cellcolor{green!16} 23.005 (+1.515) & \cellcolor{green!18} 37.746 (+1.562) & \cellcolor{green!15} 0.650 (+0.743) \\
\cline{2-8}
 & \multirow[t]{3}{*}{\texttt{ROUGE}} & 4 & \cellcolor{red!2} 75.545 (-0.086) & \cellcolor{red!4} 43.310 (-0.333) & \cellcolor{red!12} 20.373 (-1.117) & \cellcolor{red!6} 35.655 (-0.529) & \cellcolor{red!6} -0.418 (-0.325) \\
 &  & 32 & \cellcolor{green!17} 76.299 (+0.668) & \cellcolor{green!21} 45.445 (+1.802) & \cellcolor{green!8} 22.297 (+0.807) & \cellcolor{green!18} 37.722 (+1.539) & \cellcolor{green!16} 0.721 (+0.814) \\
 &  & 256 & \cellcolor{green!21} 76.423 (+0.792) & \cellcolor{green!27} 45.860 (+2.217) & \cellcolor{green!13} 22.711 (+1.220) & \cellcolor{green!23} 38.142 (+1.958) & \cellcolor{green!21} 0.944 (+1.037) \\
\cline{1-8} \cline{2-8}
\multirow[t]{12}{*}{Probabilistic MBR} & \multirow[t]{3}{*}{BERTScore} & 4 & \cellcolor{red!32} 74.431 (-1.200) & \cellcolor{red!50} 39.545 (-4.098) & \cellcolor{red!50} 16.882 (-4.608) & \cellcolor{red!50} 31.959 (-4.225) & \cellcolor{red!50} -2.503 (-2.411) \\
 &  & 32 & \cellcolor{red!4} 75.468 (-0.162) & \cellcolor{red!33} 40.893 (-2.750) & \cellcolor{red!37} 18.038 (-3.452) & \cellcolor{red!34} 33.243 (-2.941) & \cellcolor{red!34} -1.761 (-1.669) \\
 &  & 256 & \cellcolor{green!42} 77.215 (+1.585) & \cellcolor{green!4} 43.993 (+0.350) & \cellcolor{red!5} 20.985 (-0.505) & \cellcolor{green!4} 36.522 (+0.339) & \cellcolor{green!3} 0.070 (+0.163) \\
\cline{2-8}
 & \multirow[t]{3}{*}{ROUGE-1} & 4 & \cellcolor{red!23} 74.743 (-0.887) & \cellcolor{red!25} 41.543 (-2.100) & \cellcolor{red!33} 18.393 (-3.097) & \cellcolor{red!32} 33.442 (-2.742) & \cellcolor{red!31} -1.615 (-1.522) \\
 &  & 32 & \cellcolor{red!3} 75.496 (-0.135) & 43.720 (+0.077) & \cellcolor{red!15} 20.035 (-1.455) & \cellcolor{red!10} 35.328 (-0.856) & \cellcolor{red!10} -0.583 (-0.490) \\
 &  & 256 & \cellcolor{green!14} 76.154 (+0.524) & \cellcolor{green!29} 46.041 (+2.398) & \cellcolor{green!3} 21.814 (+0.323) & \cellcolor{green!12} 37.237 (+1.053) & \cellcolor{green!11} 0.443 (+0.535) \\
\cline{2-8}
 & \multirow[t]{3}{*}{ROUGE-2} & 4 & \cellcolor{red!19} 74.883 (-0.748) & \cellcolor{red!26} 41.468 (-2.175) & \cellcolor{red!26} 19.037 (-2.453) & \cellcolor{red!27} 33.856 (-2.328) & \cellcolor{red!27} -1.411 (-1.319) \\
 &  & 32 & \cellcolor{red!1} 75.572 (-0.059) & \cellcolor{red!5} 43.195 (-0.448) & \cellcolor{red!6} 20.876 (-0.615) & \cellcolor{red!5} 35.741 (-0.443) & \cellcolor{red!6} -0.427 (-0.335) \\
 &  & 256 & \cellcolor{green!11} 76.065 (+0.435) & \cellcolor{green!11} 44.622 (+0.978) & \cellcolor{green!10} 22.474 (+0.984) & \cellcolor{green!11} 37.146 (+0.962) & \cellcolor{green!8} 0.323 (+0.416) \\
\cline{2-8}
 & \multirow[t]{3}{*}{\texttt{ROUGE}} & 4 & \cellcolor{red!19} 74.896 (-0.734) & \cellcolor{red!26} 41.505 (-2.138) & \cellcolor{red!28} 18.839 (-2.651) & \cellcolor{red!26} 33.959 (-2.224) & \cellcolor{red!26} -1.362 (-1.269) \\
 &  & 32 & 75.600 (-0.030) & \cellcolor{red!4} 43.265 (-0.378) & \cellcolor{red!12} 20.359 (-1.131) & \cellcolor{red!6} 35.642 (-0.541) & \cellcolor{red!6} -0.410 (-0.317) \\
 &  & 256 & \cellcolor{green!16} 76.231 (+0.600) & \cellcolor{green!19} 45.278 (+1.635) & \cellcolor{green!7} 22.202 (+0.712) & \cellcolor{green!16} 37.593 (+1.410) & \cellcolor{green!15} 0.643 (+0.736) \\
\cline{1-8} \cline{2-8}
\multirow[t]{4}{*}{Model-based MBR} & BERTScore & 256 & \cellcolor{green!49} 77.483 (+1.853) & \cellcolor{green!9} 44.444 (+0.801) & 21.473 (-0.018) & \cellcolor{green!9} 37.024 (+0.840) & \cellcolor{green!9} 0.353 (+0.446) \\
\cline{2-8}
 & ROUGE-1 & 256 & \cellcolor{green!18} 76.307 (+0.677) & \cellcolor{green!35} 46.551 (+2.908) & \cellcolor{green!6} 22.129 (+0.639) & \cellcolor{green!16} 37.604 (+1.421) & \cellcolor{green!15} 0.646 (+0.738) \\
\cline{2-8}
 & ROUGE-2 & 256 & \cellcolor{green!17} 76.281 (+0.650) & \cellcolor{green!18} 45.189 (+1.545) & \cellcolor{green!16} 22.973 (+1.483) & \cellcolor{green!17} 37.688 (+1.504) & \cellcolor{green!14} 0.626 (+0.719) \\
\cline{2-8}
 & \texttt{ROUGE} & 256 & \cellcolor{green!21} 76.429 (+0.798) & \cellcolor{green!26} 45.794 (+2.151) & \cellcolor{green!13} 22.698 (+1.207) & \cellcolor{green!22} 38.106 (+1.922) & \cellcolor{green!21} 0.930 (+1.022) \\
\cline{1-8} \cline{2-8}
\textit{Oracle} & N/A & - & \textit{83.108} & \textit{63.274} & \textit{42.572} & \textit{57.403} & \textit{9.210} \\
\cline{1-8} \cline{2-8}
\bottomrule
\end{tabular}

}
\caption{Performance of decoding variants on the XSum ($|\mathcal{H}|=256$). Absolute differences (in parentheses) are relative to the \texttt{MAP$\epsilon$} baseline. Color gradation indicates the relative magnitude of \colorbox{green!15}{improvement} or \colorbox{red!15}{degradation}.}
\label{tab:mbr_overfitting_xsum}
\end{table*}

\begin{table*}[ht!]
\centering
\resizebox{\textwidth}{!}{
\begin{tabular}{llllllll}
\toprule
\textbf{Decoding Method} & \textbf{Util. Function} & \textbf{$|\mathcal{Y}|$} & \textbf{BERTScore} & \textbf{ROUGE-1} & \textbf{ROUGE-2} & \textbf{ROUGE-L} & \textbf{$\bar{Z}_{\text{other}}$} \\
\midrule
\multirow[t]{12}{*}{MBR} & \multirow[t]{3}{*}{BERTScore} & 4 & 76.450 & 42.777 & 19.882 & 35.218 & -1.288 \\
 &  & 32 & 77.345 & 44.057 & 21.110 & 36.600 & -0.129 \\
 &  & 256 & 77.501 & 44.376 & 21.378 & 36.937 & 0.147 \\
\cline{2-8}
 & \multirow[t]{3}{*}{ROUGE-1} & 4 & 75.533 & 43.863 & 20.121 & 35.398 & -1.112 \\
 &  & 32 & 76.138 & 46.114 & 21.680 & 37.093 & 0.228 \\
 &  & 256 & 76.293 & 46.608 & 22.126 & 37.594 & 0.607 \\
\cline{2-8}
 & \multirow[t]{3}{*}{ROUGE-2} & 4 & 75.447 & 42.894 & 20.435 & 35.299 & -1.190 \\
 &  & 32 & 76.170 & 44.885 & 22.654 & 37.367 & 0.395 \\
 &  & 256 & 76.285 & 45.246 & 23.005 & 37.746 & 0.675 \\
\cline{2-8}
 & \multirow[t]{3}{*}{\texttt{ROUGE}} & 4 & 75.545 & 43.310 & 20.373 & 35.655 & -0.912 \\
 &  & 32 & 76.299 & 45.445 & 22.297 & 37.722 & 0.764 \\
 &  & 256 & 76.423 & 45.860 & 22.711 & 38.142 & 1.094 \\
\cline{1-8} \cline{2-8}
\multirow[t]{12}{*}{SVD-MBR ($k=1$)} & \multirow[t]{3}{*}{BERTScore} & 4 & \cellcolor{red!50} 75.921 (-0.53) & \cellcolor{red!26} 42.152 (-0.62) & \cellcolor{red!45} 19.458 (-0.42) & \cellcolor{red!44} 34.721 (-0.50) & \cellcolor{red!47} -1.745 (-0.457) \\
 &  & 32 & \cellcolor{red!38} 76.939 (-0.41) & \cellcolor{red!8} 43.848 (-0.21) & \cellcolor{green!8} 21.185 (+0.08) & \cellcolor{red!2} 36.573 (-0.03) & \cellcolor{red!4} -0.171 (-0.042) \\
 &  & 256 & \cellcolor{red!36} 77.118 (-0.38) & \cellcolor{red!6} 44.220 (-0.16) & \cellcolor{green!21} 21.575 (+0.20)$^*$ & \cellcolor{green!1} 36.951 (+0.01) & \cellcolor{green!2} 0.169 (+0.022) \\
\cline{2-8}
 & \multirow[t]{3}{*}{ROUGE-1} & 4 & \cellcolor{red!22} 75.294 (-0.24) & \cellcolor{red!50} 42.689 (-1.17) & \cellcolor{red!42} 19.731 (-0.39) & \cellcolor{red!48} 34.855 (-0.54) & \cellcolor{red!43} -1.533 (-0.421) \\
 &  & 32 & \cellcolor{red!2} 76.116 (-0.02) & \cellcolor{red!42} 45.108 (-1.01) & \cellcolor{green!7} 21.746 (+0.07) & \cellcolor{red!8} 37.002 (-0.09) & \cellcolor{red!2} 0.208 (-0.020) \\
 &  & 256 & \cellcolor{red!3} 76.259 (-0.03) & \cellcolor{red!45} 45.544 (-1.06) & 22.129 (+0.00) & \cellcolor{red!17} 37.393 (-0.20) & \cellcolor{red!8} 0.526 (-0.080) \\
\cline{2-8}
 & \multirow[t]{3}{*}{ROUGE-2} & 4 & \cellcolor{red!12} 75.319 (-0.13) & \cellcolor{red!24} 42.328 (-0.57) & \cellcolor{red!40} 20.057 (-0.38) & \cellcolor{red!31} 34.948 (-0.35) & \cellcolor{red!34} -1.524 (-0.334) \\
 &  & 32 & \cellcolor{red!11} 76.052 (-0.12) & \cellcolor{red!27} 44.237 (-0.65) & \cellcolor{red!47} 22.210 (-0.44) & \cellcolor{red!34} 36.979 (-0.39) & \cellcolor{red!37} 0.032 (-0.363) \\
 &  & 256 & \cellcolor{red!10} 76.171 (-0.11) & \cellcolor{red!27} 44.605 (-0.64) & \cellcolor{red!47} 22.568 (-0.44) & \cellcolor{red!33} 37.373 (-0.37) & \cellcolor{red!36} 0.321 (-0.354) \\
\cline{2-8}
 & \multirow[t]{3}{*}{\texttt{ROUGE}} & 4 & \cellcolor{red!15} 75.376 (-0.17) & \cellcolor{red!37} 42.436 (-0.87) & \cellcolor{red!44} 19.960 (-0.41) & \cellcolor{red!49} 35.093 (-0.56) & \cellcolor{red!49} -1.388 (-0.476) \\
 &  & 32 & \cellcolor{red!12} 76.161 (-0.14) & \cellcolor{red!40} 44.504 (-0.94) & \cellcolor{red!39} 21.931 (-0.37) & \cellcolor{red!47} 37.191 (-0.53) & \cellcolor{red!47} 0.305 (-0.459) \\
 &  & 256 & \cellcolor{red!13} 76.280 (-0.14) & \cellcolor{red!43} 44.847 (-1.01) & \cellcolor{red!39} 22.348 (-0.36) & \cellcolor{red!50} 37.579 (-0.56) & \cellcolor{red!50} 0.612 (-0.482) \\
\cline{1-8} \cline{2-8}
\multirow[t]{12}{*}{SVD-MBR ($k=2$)} & \multirow[t]{3}{*}{BERTScore} & 4 & \cellcolor{red!16} 76.274 (-0.18) & \cellcolor{red!7} 42.600 (-0.18) & \cellcolor{red!16} 19.725 (-0.16) & \cellcolor{red!14} 35.052 (-0.17) & \cellcolor{red!15} -1.437 (-0.148) \\
 &  & 32 & \cellcolor{red!16} 77.173 (-0.17) & \cellcolor{red!5} 43.920 (-0.14) & \cellcolor{red!5} 21.062 (-0.05) & \cellcolor{red!9} 36.491 (-0.11) & \cellcolor{red!8} -0.215 (-0.086) \\
 &  & 256 & \cellcolor{red!8} 77.407 (-0.09) & \cellcolor{green!1} 44.400 (+0.02) & \cellcolor{green!14} 21.513 (+0.13)$^*$ & \cellcolor{green!5} 37.004 (+0.07) & \cellcolor{green!7} 0.215 (+0.069)$^*$ \\
\cline{2-8}
 & \multirow[t]{3}{*}{ROUGE-1} & 4 & \cellcolor{red!9} 75.428 (-0.10) & \cellcolor{red!22} 43.340 (-0.52) & \cellcolor{red!29} 19.848 (-0.27) & \cellcolor{red!28} 35.083 (-0.32) & \cellcolor{red!24} -1.352 (-0.240) \\
 &  & 32 & \cellcolor{red!3} 76.097 (-0.04) & \cellcolor{red!26} 45.483 (-0.63) & \cellcolor{red!7} 21.606 (-0.07) & \cellcolor{red!16} 36.902 (-0.19) & \cellcolor{red!10} 0.123 (-0.104) \\
 &  & 256 & \cellcolor{green!2} 76.319 (+0.03) & \cellcolor{red!17} 46.201 (-0.41) & \cellcolor{green!9} 22.210 (+0.08) & \cellcolor{red!1} 37.574 (-0.02) & \cellcolor{green!3} 0.641 (+0.034) \\
\cline{2-8}
 & \multirow[t]{3}{*}{ROUGE-2} & 4 & \cellcolor{red!10} 75.335 (-0.11) & \cellcolor{red!18} 42.454 (-0.44) & \cellcolor{red!43} 20.032 (-0.40) & \cellcolor{red!28} 34.973 (-0.33) & \cellcolor{red!29} -1.473 (-0.283) \\
 &  & 32 & \cellcolor{red!11} 76.047 (-0.12) & \cellcolor{red!24} 44.320 (-0.56) & \cellcolor{red!50} 22.190 (-0.46) & \cellcolor{red!36} 36.962 (-0.41) & \cellcolor{red!36} 0.047 (-0.348) \\
 &  & 256 & \cellcolor{red!4} 76.241 (-0.04) & \cellcolor{red!14} 44.915 (-0.33) & \cellcolor{red!20} 22.816 (-0.19) & \cellcolor{red!16} 37.564 (-0.18) & \cellcolor{red!17} 0.504 (-0.172) \\
\cline{2-8}
 & \multirow[t]{3}{*}{\texttt{ROUGE}} & 4 & \cellcolor{red!12} 75.416 (-0.13) & \cellcolor{red!25} 42.712 (-0.60) & \cellcolor{red!43} 19.972 (-0.40) & \cellcolor{red!45} 35.139 (-0.52) & \cellcolor{red!40} -1.301 (-0.389) \\
 &  & 32 & \cellcolor{red!13} 76.160 (-0.14) & \cellcolor{red!29} 44.763 (-0.68) & \cellcolor{red!34} 21.975 (-0.32) & \cellcolor{red!42} 37.242 (-0.48) & \cellcolor{red!39} 0.380 (-0.384) \\
 &  & 256 & \cellcolor{red!2} 76.396 (-0.03) & \cellcolor{red!19} 45.405 (-0.46) & \cellcolor{red!7} 22.645 (-0.07) & \cellcolor{red!18} 37.938 (-0.20) & \cellcolor{red!17} 0.926 (-0.167) \\
\cline{1-8} \cline{2-8}
\multirow[t]{12}{*}{SVD-MBR ($k=3$)} & \multirow[t]{3}{*}{BERTScore} & 4 & \cellcolor{red!3} 76.411 (-0.04) & \cellcolor{red!1} 42.735 (-0.04) & \cellcolor{red!1} 19.863 (-0.02) & \cellcolor{red!5} 35.157 (-0.06) & \cellcolor{red!3} -1.324 (-0.036) \\
 &  & 32 & \cellcolor{red!3} 77.303 (-0.04) & 44.050 (-0.01) & 21.114 (+0.00) & 36.605 (+0.01) & -0.128 (+0.001) \\
 &  & 256 & \cellcolor{red!3} 77.467 (-0.03) & 44.387 (+0.01) & \cellcolor{green!6} 21.442 (+0.06)$^*$ & \cellcolor{green!1} 36.958 (+0.02) & \cellcolor{green!3} 0.176 (+0.029) \\
\cline{2-8}
 & \multirow[t]{3}{*}{ROUGE-1} & 4 & \cellcolor{red!1} 75.516 (-0.02) & \cellcolor{red!3} 43.781 (-0.08) & \cellcolor{red!4} 20.081 (-0.04) & \cellcolor{red!2} 35.365 (-0.03) & \cellcolor{red!3} -1.144 (-0.032) \\
 &  & 32 & 76.143 (+0.01) & \cellcolor{red!11} 45.850 (-0.26) & 21.675 (-0.00) & \cellcolor{red!2} 37.065 (-0.03) & 0.220 (-0.007) \\
 &  & 256 & \cellcolor{green!3} 76.325 (+0.03)$^*$ & \cellcolor{red!6} 46.450 (-0.16) & \cellcolor{green!10} 22.224 (+0.10)$^*$ & \cellcolor{green!3} 37.639 (+0.04) & \cellcolor{green!6} 0.668 (+0.061)$^*$ \\
\cline{2-8}
 & \multirow[t]{3}{*}{ROUGE-2} & 4 & \cellcolor{red!5} 75.387 (-0.06) & \cellcolor{red!7} 42.729 (-0.17) & \cellcolor{red!18} 20.267 (-0.17) & \cellcolor{red!14} 35.137 (-0.16) & \cellcolor{red!13} -1.319 (-0.129) \\
 &  & 32 & \cellcolor{red!7} 76.092 (-0.08) & \cellcolor{red!16} 44.491 (-0.39) & \cellcolor{red!34} 22.338 (-0.32) & \cellcolor{red!27} 37.062 (-0.30) & \cellcolor{red!25} 0.150 (-0.246) \\
 &  & 256 & \cellcolor{red!1} 76.274 (-0.01) & \cellcolor{red!7} 45.068 (-0.18) & \cellcolor{red!10} 22.903 (-0.10) & \cellcolor{red!8} 37.645 (-0.10) & \cellcolor{red!8} 0.589 (-0.086) \\
\cline{2-8}
 & \multirow[t]{3}{*}{\texttt{ROUGE}} & 4 & \cellcolor{red!4} 75.498 (-0.05) & \cellcolor{red!5} 43.171 (-0.14) & \cellcolor{red!11} 20.268 (-0.10) & \cellcolor{red!13} 35.504 (-0.15) & \cellcolor{red!11} -1.019 (-0.107) \\
 &  & 32 & \cellcolor{red!6} 76.233 (-0.07) & \cellcolor{red!17} 45.039 (-0.41) & \cellcolor{red!19} 22.117 (-0.18) & \cellcolor{red!24} 37.443 (-0.28) & \cellcolor{red!22} 0.547 (-0.217) \\
 &  & 256 & 76.425 (+0.00) & \cellcolor{red!9} 45.643 (-0.22) & 22.711 (+0.00) & \cellcolor{red!7} 38.061 (-0.08) & \cellcolor{red!6} 1.031 (-0.063) \\
\cline{1-8} \cline{2-8}
\textit{Oracle} & N/A & - & \textit{83.108} & \textit{63.274} & \textit{42.572} & \textit{57.403} & \textit{13.320} \\
\cline{1-8} \cline{2-8}
\bottomrule
\end{tabular}

}
\caption{Performance of MBR vs. SVD-MBR on XSum ($|\mathcal{H}|=256$). $\bar{Z}_{\text{other}}$ represents the mean z-score of all non-utility function evaluation metrics. Values in parentheses are absolute differences against MBR, with colors indicating relative \colorbox{green!15}{improvement} or \colorbox{red!15}{degradation}. Statistical significance ($p < 0.05$) is marked with $^*$.}
\label{tab:svd_mbr_results_xsum}
\end{table*}

\begin{table*}[ht!]
\centering
\resizebox{\textwidth}{!}{
\begin{NiceTabular}{@{} lll c c c c c c c c c c @{}}
\toprule
\multirow{2}{*}{\textbf{Util. Function}} & \multirow{2}{*}{\textbf{$|\mathcal{Y}|$}} & \multirow{2}{*}{\textbf{Top-$k$}} & \multicolumn{2}{c}{\makecell{\textbf{BERTScore} \\ ($\sigma^2=13.815$)}} & \multicolumn{2}{c}{\makecell{\textbf{ROUGE-1} \\ ($\sigma^2=78.896$)}} & \multicolumn{2}{c}{\makecell{\textbf{ROUGE-2} \\ ($\sigma^2=72.118$)}} & \multicolumn{2}{c}{\makecell{\textbf{ROUGE-L} \\ ($\sigma^2=83.027$)}} & \multicolumn{2}{c}{\makecell{\textbf{$\bar{Z}_{\text{other}}$} \\ ($\sigma^2=0.302$)}} \\
\cmidrule(lr){4-5} \cmidrule(lr){6-7} \cmidrule(lr){8-9} \cmidrule(lr){10-11} \cmidrule(lr){12-13}
  &   &   & \textit{Dir.} & \textit{Mag.} & \textit{Dir.} & \textit{Mag.} & \textit{Dir.} & \textit{Mag.} & \textit{Dir.} & \textit{Mag.} & \textit{Dir.} & \textit{Mag.} \\
\midrule
\multirow{9}{*}{\textbf{BERTScore}} & \multirow{3}{*}{4} & $k=1$ & \cellcolor{green!4} $+0.03^{*}$ & \cellcolor{blue!9} $0.07^{*}$ & \cellcolor{red!2} $-0.02$ & \cellcolor{blue!10} $0.07^{*}$ & \cellcolor{red!1} $-0.01$ & \cellcolor{blue!25} $0.17^{*}$ & \cellcolor{red!1} $-0.01$ & \cellcolor{blue!13} $0.09^{*}$ & \cellcolor{red!1} $-0.01$ & \cellcolor{blue!14} $0.10^{*}$ \\
  &   & $k=2$ & \cellcolor{green!1} $+0.01$ & \cellcolor{blue!4} $0.03^{*}$ & \cellcolor{red!3} $-0.02^{*}$ & \cellcolor{blue!4} $0.03^{*}$ & \cellcolor{red!1} $-0.01$ & \cellcolor{blue!10} $0.07^{*}$ & \cellcolor{red!2} $-0.01$ & \cellcolor{blue!4} $0.03^{*}$ & \cellcolor{red!3} $-0.02^{*}$ & \cellcolor{blue!4} $0.03^{*}$ \\
  &   & $k=3$ & $+0.00$ & \cellcolor{blue!1} $0.01$ & \cellcolor{red!1} $-0.01$ & \cellcolor{blue!1} $0.01$ & \cellcolor{green!1} $+0.01$ & \cellcolor{blue!6} $0.04^{*}$ & \cellcolor{red!1} $-0.01$ & \cellcolor{blue!3} $0.02^{*}$ & \cellcolor{red!1} $-0.01$ & \cellcolor{blue!2} $0.02$ \\
\hhline{~------------}
  & \multirow{3}{*}{32} & $k=1$ & \cellcolor{green!13} $+0.09^{*}$ & \cellcolor{blue!12} $0.09^{*}$ & \cellcolor{green!3} $+0.03^{*}$ & \cellcolor{blue!13} $0.09^{*}$ & \cellcolor{green!3} $+0.02^{*}$ & \cellcolor{blue!28} $0.19^{*}$ & \cellcolor{green!5} $+0.03^{*}$ & \cellcolor{blue!17} $0.12^{*}$ & \cellcolor{green!4} $+0.03^{*}$ & \cellcolor{blue!18} $0.12^{*}$ \\
  &   & $k=2$ & \cellcolor{green!6} $+0.04^{*}$ & \cellcolor{blue!10} $0.07^{*}$ & \cellcolor{green!1} $+0.01$ & \cellcolor{blue!7} $0.05^{*}$ & $+0.00$ & \cellcolor{blue!13} $0.09^{*}$ & \cellcolor{green!2} $+0.02^{*}$ & \cellcolor{blue!8} $0.05^{*}$ & \cellcolor{green!2} $+0.02$ & \cellcolor{blue!8} $0.05^{*}$ \\
  &   & $k=3$ & \cellcolor{green!5} $+0.04^{*}$ & \cellcolor{blue!4} $0.03^{*}$ & \cellcolor{green!1} $+0.01$ & \cellcolor{blue!2} $0.02$ & $+0.00$ & \cellcolor{blue!5} $0.04^{*}$ & $+0.01$ & \cellcolor{blue!3} $0.02^{*}$ & \cellcolor{green!1} $+0.01$ & \cellcolor{blue!2} $0.02$ \\
\hhline{~------------}
  & \multirow{3}{*}{256} & $k=1$ & \cellcolor{green!13} $+0.09^{*}$ & \cellcolor{blue!15} $0.10^{*}$ & \cellcolor{green!5} $+0.04^{*}$ & \cellcolor{blue!15} $0.11^{*}$ & \cellcolor{green!5} $+0.04^{*}$ & \cellcolor{blue!26} $0.17^{*}$ & \cellcolor{green!5} $+0.04^{*}$ & \cellcolor{blue!16} $0.11^{*}$ & \cellcolor{green!6} $+0.04^{*}$ & \cellcolor{blue!17} $0.12^{*}$ \\
  &   & $k=2$ & \cellcolor{green!10} $+0.07^{*}$ & \cellcolor{blue!9} $0.06^{*}$ & \cellcolor{green!2} $+0.02^{*}$ & \cellcolor{blue!7} $0.05^{*}$ & \cellcolor{green!5} $+0.04^{*}$ & \cellcolor{blue!9} $0.06^{*}$ & \cellcolor{green!5} $+0.04^{*}$ & \cellcolor{blue!3} $0.03^{*}$ & \cellcolor{green!4} $+0.03^{*}$ & \cellcolor{blue!4} $0.03^{*}$ \\
  &   & $k=3$ & \cellcolor{green!5} $+0.04^{*}$ & \cellcolor{blue!2} $0.02$ & \cellcolor{green!1} $+0.01$ & \cellcolor{blue!3} $0.02^{*}$ & \cellcolor{green!2} $+0.01$ & \cellcolor{blue!6} $0.04^{*}$ & \cellcolor{green!2} $+0.01$ & \cellcolor{blue!2} $0.02$ & \cellcolor{green!2} $+0.02$ & \cellcolor{blue!2} $0.02$ \\
\midrule
\multirow{9}{*}{\textbf{ROUGE-1}} & \multirow{3}{*}{4} & $k=1$ & \cellcolor{green!1} $+0.01$ & \cellcolor{blue!9} $0.07^{*}$ & \cellcolor{green!2} $+0.02$ & \cellcolor{blue!10} $0.07^{*}$ & \cellcolor{red!1} $-0.01$ & \cellcolor{blue!27} $0.19^{*}$ & \cellcolor{green!2} $+0.02$ & \cellcolor{blue!13} $0.09^{*}$ & \cellcolor{green!2} $+0.02$ & \cellcolor{blue!15} $0.10^{*}$ \\
  &   & $k=2$ & \cellcolor{red!1} $-0.01$ & \cellcolor{blue!3} $0.02^{*}$ & $+0.00$ & \cellcolor{blue!7} $0.05^{*}$ & \cellcolor{red!3} $-0.03^{*}$ & \cellcolor{blue!11} $0.08^{*}$ & \cellcolor{red!1} $-0.01$ & \cellcolor{blue!4} $0.03^{*}$ & $-0.01$ & \cellcolor{blue!5} $0.03^{*}$ \\
  &   & $k=3$ & \cellcolor{red!1} $-0.01$ & \cellcolor{blue!1} $0.01$ & \cellcolor{green!2} $+0.02$ & \cellcolor{blue!2} $0.02$ & \cellcolor{red!1} $-0.01$ & \cellcolor{blue!5} $0.04^{*}$ & \cellcolor{green!1} $+0.01$ & $0.01$ & $+0.00$ & \cellcolor{blue!1} $0.01$ \\
\hhline{~------------}
  & \multirow{3}{*}{32} & $k=1$ & \cellcolor{green!4} $+0.03^{*}$ & \cellcolor{blue!14} $0.10^{*}$ & \cellcolor{green!8} $+0.06^{*}$ & \cellcolor{blue!12} $0.09^{*}$ & \cellcolor{green!4} $+0.03^{*}$ & \cellcolor{blue!26} $0.18^{*}$ & \cellcolor{green!6} $+0.04^{*}$ & \cellcolor{blue!17} $0.11^{*}$ & \cellcolor{green!6} $+0.05^{*}$ & \cellcolor{blue!18} $0.12^{*}$ \\
  &   & $k=2$ & \cellcolor{green!1} $+0.01$ & \cellcolor{blue!10} $0.07^{*}$ & \cellcolor{green!4} $+0.03^{*}$ & \cellcolor{blue!9} $0.06^{*}$ & $+0.00$ & \cellcolor{blue!13} $0.09^{*}$ & $+0.00$ & \cellcolor{blue!6} $0.04^{*}$ & $+0.00$ & \cellcolor{blue!7} $0.05^{*}$ \\
  &   & $k=3$ & \cellcolor{green!2} $+0.02$ & \cellcolor{blue!2} $0.02$ & \cellcolor{green!3} $+0.02^{*}$ & \cellcolor{blue!3} $0.02^{*}$ & $+0.00$ & \cellcolor{blue!5} $0.04^{*}$ & $+0.00$ & $0.00$ & $+0.00$ & $0.00$ \\
\hhline{~------------}
  & \multirow{3}{*}{256} & $k=1$ & \cellcolor{green!5} $+0.04^{*}$ & \cellcolor{blue!16} $0.11^{*}$ & \cellcolor{green!9} $+0.07^{*}$ & \cellcolor{blue!12} $0.08^{*}$ & \cellcolor{green!4} $+0.03^{*}$ & \cellcolor{blue!22} $0.15^{*}$ & \cellcolor{green!7} $+0.05^{*}$ & \cellcolor{blue!12} $0.09^{*}$ & \cellcolor{green!7} $+0.05^{*}$ & \cellcolor{blue!14} $0.10^{*}$ \\
  &   & $k=2$ & \cellcolor{green!5} $+0.03^{*}$ & \cellcolor{blue!9} $0.06^{*}$ & \cellcolor{green!8} $+0.06^{*}$ & \cellcolor{blue!8} $0.05^{*}$ & \cellcolor{green!4} $+0.03^{*}$ & \cellcolor{blue!9} $0.06^{*}$ & \cellcolor{green!5} $+0.04^{*}$ & \cellcolor{blue!4} $0.03^{*}$ & \cellcolor{green!5} $+0.04^{*}$ & \cellcolor{blue!5} $0.03^{*}$ \\
  &   & $k=3$ & \cellcolor{green!2} $+0.02^{*}$ & \cellcolor{blue!5} $0.03^{*}$ & \cellcolor{green!6} $+0.05^{*}$ & \cellcolor{blue!4} $0.03^{*}$ & \cellcolor{green!4} $+0.03^{*}$ & \cellcolor{blue!5} $0.04^{*}$ & \cellcolor{green!4} $+0.03^{*}$ & \cellcolor{blue!1} $0.01$ & \cellcolor{green!3} $+0.03^{*}$ & \cellcolor{blue!1} $0.01$ \\
\midrule
\multirow{9}{*}{\textbf{ROUGE-2}} & \multirow{3}{*}{4} & $k=1$ & $+0.00$ & \cellcolor{blue!6} $0.04^{*}$ & \cellcolor{red!2} $-0.01$ & \cellcolor{blue!9} $0.06^{*}$ & $+0.00$ & \cellcolor{blue!22} $0.15^{*}$ & $+0.00$ & \cellcolor{blue!12} $0.08^{*}$ & $+0.00$ & \cellcolor{blue!10} $0.07^{*}$ \\
  &   & $k=2$ & \cellcolor{red!2} $-0.02$ & \cellcolor{blue!4} $0.03^{*}$ & \cellcolor{red!3} $-0.03^{*}$ & \cellcolor{blue!6} $0.04^{*}$ & \cellcolor{red!4} $-0.03^{*}$ & \cellcolor{blue!11} $0.07^{*}$ & \cellcolor{red!3} $-0.02^{*}$ & \cellcolor{blue!4} $0.03^{*}$ & \cellcolor{red!2} $-0.02$ & \cellcolor{blue!4} $0.03^{*}$ \\
  &   & $k=3$ & \cellcolor{red!1} $-0.01$ & $-0.00$ & $+0.00$ & \cellcolor{blue!1} $0.01$ & \cellcolor{red!1} $-0.01$ & \cellcolor{blue!2} $0.02^{*}$ & $+0.00$ & $-0.00$ & $-0.01$ & $0.00$ \\
\hhline{~------------}
  & \multirow{3}{*}{32} & $k=1$ & $+0.01$ & \cellcolor{blue!7} $0.05^{*}$ & \cellcolor{green!2} $+0.02$ & \cellcolor{blue!9} $0.06^{*}$ & $+0.00$ & \cellcolor{blue!21} $0.14^{*}$ & \cellcolor{green!5} $+0.03^{*}$ & \cellcolor{blue!10} $0.07^{*}$ & \cellcolor{green!5} $+0.04^{*}$ & \cellcolor{blue!9} $0.07^{*}$ \\
  &   & $k=2$ & $+0.00$ & \cellcolor{blue!7} $0.05^{*}$ & $+0.01$ & \cellcolor{blue!7} $0.05^{*}$ & \cellcolor{red!1} $-0.01$ & \cellcolor{blue!14} $0.10^{*}$ & $+0.00$ & \cellcolor{blue!6} $0.05^{*}$ & \cellcolor{green!1} $+0.01$ & \cellcolor{blue!7} $0.05^{*}$ \\
  &   & $k=3$ & \cellcolor{red!1} $-0.01$ & \cellcolor{blue!4} $0.03^{*}$ & $+0.00$ & \cellcolor{blue!3} $0.02^{*}$ & $+0.00$ & \cellcolor{blue!4} $0.03^{*}$ & \cellcolor{green!1} $+0.01$ & $-0.01$ & \cellcolor{green!1} $+0.01$ & \cellcolor{blue!1} $0.01$ \\
\hhline{~------------}
  & \multirow{3}{*}{256} & $k=1$ & \cellcolor{green!1} $+0.01$ & \cellcolor{blue!9} $0.06^{*}$ & \cellcolor{green!3} $+0.02^{*}$ & \cellcolor{blue!7} $0.05^{*}$ & $+0.00$ & \cellcolor{blue!14} $0.09^{*}$ & $+0.00$ & \cellcolor{blue!5} $0.04^{*}$ & \cellcolor{green!3} $+0.02^{*}$ & \cellcolor{blue!6} $0.05^{*}$ \\
  &   & $k=2$ & \cellcolor{green!1} $+0.01$ & \cellcolor{blue!4} $0.03^{*}$ & \cellcolor{green!1} $+0.01$ & \cellcolor{blue!2} $0.02^{*}$ & \cellcolor{green!1} $+0.01$ & \cellcolor{blue!7} $0.05^{*}$ & $+0.00$ & $0.00$ & \cellcolor{green!2} $+0.02$ & \cellcolor{blue!1} $0.01$ \\
  &   & $k=3$ & \cellcolor{green!1} $+0.01$ & \cellcolor{blue!2} $0.02$ & \cellcolor{green!2} $+0.02$ & \cellcolor{blue!2} $0.01$ & \cellcolor{green!1} $+0.01$ & \cellcolor{blue!1} $0.01$ & \cellcolor{green!1} $+0.01$ & $-0.02^{*}$ & \cellcolor{green!1} $+0.01$ & $-0.01$ \\
\midrule
\multirow{9}{*}{\textbf{\texttt{ROUGE}}} & \multirow{3}{*}{4} & $k=1$ & \cellcolor{green!1} $+0.01$ & \cellcolor{blue!8} $0.05^{*}$ & \cellcolor{green!1} $+0.01$ & \cellcolor{blue!10} $0.07^{*}$ & $+0.01$ & \cellcolor{blue!23} $0.16^{*}$ & \cellcolor{green!2} $+0.01$ & \cellcolor{blue!11} $0.08^{*}$ & \cellcolor{green!2} $+0.02^{*}$ & \cellcolor{blue!12} $0.08^{*}$ \\
  &   & $k=2$ & \cellcolor{red!1} $-0.01$ & \cellcolor{blue!2} $0.02$ & \cellcolor{red!1} $-0.01$ & \cellcolor{blue!5} $0.03^{*}$ & $-0.01$ & \cellcolor{blue!10} $0.07^{*}$ & \cellcolor{red!2} $-0.01$ & \cellcolor{blue!4} $0.03^{*}$ & $+0.00$ & \cellcolor{blue!4} $0.03^{*}$ \\
  &   & $k=3$ & $+0.00$ & \cellcolor{blue!2} $0.01$ & \cellcolor{green!1} $+0.01$ & \cellcolor{blue!3} $0.02^{*}$ & \cellcolor{red!1} $-0.01$ & \cellcolor{blue!3} $0.02^{*}$ & $+0.00$ & \cellcolor{blue!1} $0.01$ & $+0.00$ & \cellcolor{blue!1} $0.01$ \\
\hhline{~------------}
  & \multirow{3}{*}{32} & $k=1$ & \cellcolor{green!2} $+0.02^{*}$ & \cellcolor{blue!11} $0.08^{*}$ & \cellcolor{green!3} $+0.02^{*}$ & \cellcolor{blue!12} $0.08^{*}$ & $-0.01$ & \cellcolor{blue!23} $0.16^{*}$ & \cellcolor{green!2} $+0.02$ & \cellcolor{blue!13} $0.09^{*}$ & \cellcolor{green!4} $+0.03^{*}$ & \cellcolor{blue!14} $0.10^{*}$ \\
  &   & $k=2$ & $+0.00$ & \cellcolor{blue!6} $0.05^{*}$ & \cellcolor{green!2} $+0.01$ & \cellcolor{blue!7} $0.05^{*}$ & \cellcolor{red!2} $-0.01$ & \cellcolor{blue!13} $0.09^{*}$ & \cellcolor{green!1} $+0.01$ & \cellcolor{blue!5} $0.03^{*}$ & \cellcolor{green!2} $+0.02^{*}$ & \cellcolor{blue!6} $0.04^{*}$ \\
  &   & $k=3$ & \cellcolor{green!2} $+0.02$ & \cellcolor{blue!3} $0.02^{*}$ & \cellcolor{green!1} $+0.01$ & \cellcolor{blue!4} $0.03^{*}$ & $+0.00$ & \cellcolor{blue!5} $0.04^{*}$ & $+0.00$ & $0.00$ & \cellcolor{green!2} $+0.01$ & $0.00$ \\
\hhline{~------------}
  & \multirow{3}{*}{256} & $k=1$ & $+0.00$ & \cellcolor{blue!12} $0.08^{*}$ & \cellcolor{green!3} $+0.02^{*}$ & \cellcolor{blue!10} $0.07^{*}$ & \cellcolor{red!1} $-0.01$ & \cellcolor{blue!16} $0.11^{*}$ & \cellcolor{green!2} $+0.01$ & \cellcolor{blue!9} $0.06^{*}$ & \cellcolor{green!3} $+0.03^{*}$ & \cellcolor{blue!9} $0.06^{*}$ \\
  &   & $k=2$ & \cellcolor{green!3} $+0.02^{*}$ & \cellcolor{blue!8} $0.06^{*}$ & \cellcolor{green!4} $+0.03^{*}$ & \cellcolor{blue!5} $0.04^{*}$ & $+0.01$ & \cellcolor{blue!6} $0.04^{*}$ & \cellcolor{green!4} $+0.03^{*}$ & \cellcolor{blue!1} $0.01$ & \cellcolor{green!6} $+0.04^{*}$ & \cellcolor{blue!2} $0.02$ \\
  &   & $k=3$ & \cellcolor{green!1} $+0.01$ & \cellcolor{blue!7} $0.05^{*}$ & \cellcolor{green!3} $+0.03^{*}$ & \cellcolor{blue!4} $0.03^{*}$ & $+0.00$ & \cellcolor{blue!3} $0.02^{*}$ & \cellcolor{green!2} $+0.02^{*}$ & $-0.01$ & \cellcolor{green!3} $+0.02^{*}$ & $0.00$ \\
\bottomrule
\end{NiceTabular}
}
\caption{Spearman correlation ($\rho$) between inherent hypotheses oracle variance and evaluation score differences across varying pseudo-reference pool sizes ($|\mathcal{Y}|$) on XSum. The hypothesis pool variance ($\sigma^2$) for each evaluation metric is displayed in the respective column headers. \textit{Dir.} (Directional Change) measures score improvements, highlighted as \colorbox{green!15}{positive} and \colorbox{red!15}{negative} impacts. \textit{Mag.} (Magnitude Sensitivity) measures absolute score variance, with \colorbox{blue!15}{blue} highlighting metric sensitivity. Statistical significance ($p < 0.05$) is denoted by an asterisk ($^*$).}
\label{tab:corr_variance_analysis_complete_xsum}
\end{table*}

\begin{table*}[ht!]
\centering
\resizebox{\textwidth}{!}{
\begin{NiceTabular}{@{} lll c c c c c c c c c c @{}}
\toprule
\multirow{2}{*}{\textbf{Utility}} & \multirow{2}{*}{\textbf{$|\mathcal{Y}|$}} & \multirow{2}{*}{\textbf{Top-$k$}} & \multicolumn{2}{c}{\textbf{BERTScore}} & \multicolumn{2}{c}{\textbf{ROUGE-1}} & \multicolumn{2}{c}{\textbf{ROUGE-2}} & \multicolumn{2}{c}{\textbf{ROUGE-L}} & \multicolumn{2}{c}{\textbf{$\bar{Z}_{\text{other}}$}} \\
\cmidrule(lr){4-5} \cmidrule(lr){6-7} \cmidrule(lr){8-9} \cmidrule(lr){10-11} \cmidrule(lr){12-13}
 &  &  & \textit{Dir.} & \textit{Mag.} & \textit{Dir.} & \textit{Mag.} & \textit{Dir.} & \textit{Mag.} & \textit{Dir.} & \textit{Mag.} & \textit{Dir.} & \textit{Mag.} \\
\midrule
\multirow{9}{*}{\textbf{BERTScore}} & \multirow{3}{*}{4} & $k=1$ & \cellcolor{red!7} $-0.05^{*}$ & \cellcolor{blue!32} $0.22^{*}$ & \cellcolor{red!3} $-0.03^{*}$ & \cellcolor{blue!29} $0.20^{*}$ & \cellcolor{red!4} $-0.03^{*}$ & \cellcolor{blue!26} $0.18^{*}$ & \cellcolor{red!5} $-0.04^{*}$ & \cellcolor{blue!30} $0.20^{*}$ & \cellcolor{red!5} $-0.04^{*}$ & \cellcolor{blue!29} $0.20^{*}$ \\
 &  & $k=2$ & \cellcolor{red!8} $-0.06^{*}$ & \cellcolor{blue!33} $0.23^{*}$ & \cellcolor{red!1} $-0.01$ & \cellcolor{blue!34} $0.23^{*}$ & \cellcolor{red!2} $-0.02^{*}$ & \cellcolor{blue!31} $0.21^{*}$ & \cellcolor{red!3} $-0.02^{*}$ & \cellcolor{blue!34} $0.23^{*}$ & \cellcolor{red!2} $-0.02^{*}$ & \cellcolor{blue!34} $0.23^{*}$ \\
 &  & $k=3$ & \cellcolor{red!1} $-0.01$ & \cellcolor{blue!19} $0.13^{*}$ & $-0.01$ & \cellcolor{blue!23} $0.16^{*}$ & \cellcolor{red!1} $-0.01$ & \cellcolor{blue!22} $0.15^{*}$ & \cellcolor{red!1} $-0.01$ & \cellcolor{blue!23} $0.16^{*}$ & \cellcolor{red!1} $-0.01$ & \cellcolor{blue!23} $0.16^{*}$ \\
\hhline{~------------}
 & \multirow{3}{*}{32} & $k=1$ & \cellcolor{green!3} $+0.03^{*}$ & \cellcolor{blue!10} $0.07^{*}$ & \cellcolor{green!2} $+0.02$ & \cellcolor{blue!9} $0.07^{*}$ & $+0.01$ & \cellcolor{blue!7} $0.05^{*}$ & $+0.00$ & \cellcolor{blue!9} $0.06^{*}$ & \cellcolor{green!1} $+0.01$ & \cellcolor{blue!8} $0.06^{*}$ \\
 &  & $k=2$ & \cellcolor{red!6} $-0.05^{*}$ & \cellcolor{blue!29} $0.19^{*}$ & \cellcolor{red!2} $-0.01$ & \cellcolor{blue!30} $0.21^{*}$ & \cellcolor{red!2} $-0.02$ & \cellcolor{blue!28} $0.19^{*}$ & \cellcolor{red!1} $-0.01$ & \cellcolor{blue!30} $0.20^{*}$ & \cellcolor{red!2} $-0.02$ & \cellcolor{blue!30} $0.20^{*}$ \\
 &  & $k=3$ & \cellcolor{red!2} $-0.01$ & \cellcolor{blue!9} $0.06^{*}$ & $+0.00$ & \cellcolor{blue!14} $0.09^{*}$ & \cellcolor{red!1} $-0.01$ & \cellcolor{blue!13} $0.09^{*}$ & $-0.01$ & \cellcolor{blue!13} $0.09^{*}$ & $-0.01$ & \cellcolor{blue!14} $0.09^{*}$ \\
\hhline{~------------}
 & \multirow{3}{*}{256} & $k=1$ & \cellcolor{green!1} $+0.01$ & \cellcolor{blue!9} $0.06^{*}$ & $+0.00$ & \cellcolor{blue!9} $0.06^{*}$ & \cellcolor{red!2} $-0.02$ & \cellcolor{blue!6} $0.04^{*}$ & \cellcolor{red!2} $-0.02$ & \cellcolor{blue!9} $0.07^{*}$ & \cellcolor{red!1} $-0.01$ & \cellcolor{blue!8} $0.06^{*}$ \\
 &  & $k=2$ & \cellcolor{red!3} $-0.02^{*}$ & \cellcolor{blue!21} $0.14^{*}$ & $+0.00$ & \cellcolor{blue!24} $0.16^{*}$ & $+0.00$ & \cellcolor{blue!21} $0.15^{*}$ & \cellcolor{red!1} $-0.01$ & \cellcolor{blue!24} $0.16^{*}$ & $-0.01$ & \cellcolor{blue!23} $0.16^{*}$ \\
 &  & $k=3$ & \cellcolor{red!1} $-0.01$ & \cellcolor{blue!8} $0.06^{*}$ & $+0.00$ & \cellcolor{blue!14} $0.10^{*}$ & $+0.00$ & \cellcolor{blue!13} $0.09^{*}$ & \cellcolor{red!1} $-0.01$ & \cellcolor{blue!14} $0.10^{*}$ & $-0.01$ & \cellcolor{blue!14} $0.10^{*}$ \\
\midrule
\multirow{9}{*}{\textbf{ROUGE-1}} & \multirow{3}{*}{4} & $k=1$ & \cellcolor{red!5} $-0.04^{*}$ & \cellcolor{blue!27} $0.18^{*}$ & \cellcolor{red!8} $-0.06^{*}$ & \cellcolor{blue!28} $0.19^{*}$ & \cellcolor{red!4} $-0.03^{*}$ & \cellcolor{blue!22} $0.15^{*}$ & \cellcolor{red!6} $-0.04^{*}$ & \cellcolor{blue!26} $0.17^{*}$ & \cellcolor{red!7} $-0.05^{*}$ & \cellcolor{blue!24} $0.16^{*}$ \\
 &  & $k=2$ & \cellcolor{red!2} $-0.02$ & \cellcolor{blue!33} $0.22^{*}$ & \cellcolor{red!4} $-0.03^{*}$ & \cellcolor{blue!33} $0.22^{*}$ & \cellcolor{red!1} $-0.01$ & \cellcolor{blue!31} $0.21^{*}$ & $+0.00$ & \cellcolor{blue!32} $0.22^{*}$ & \cellcolor{red!2} $-0.02$ & \cellcolor{blue!32} $0.21^{*}$ \\
 &  & $k=3$ & \cellcolor{red!1} $-0.01$ & \cellcolor{blue!26} $0.18^{*}$ & \cellcolor{red!1} $-0.01$ & \cellcolor{blue!28} $0.19^{*}$ & \cellcolor{green!2} $+0.02$ & \cellcolor{blue!25} $0.17^{*}$ & \cellcolor{green!1} $+0.01$ & \cellcolor{blue!27} $0.18^{*}$ & \cellcolor{green!1} $+0.01$ & \cellcolor{blue!26} $0.18^{*}$ \\
\hhline{~------------}
 & \multirow{3}{*}{32} & $k=1$ & \cellcolor{green!1} $+0.01$ & \cellcolor{blue!16} $0.11^{*}$ & \cellcolor{red!1} $-0.01$ & \cellcolor{blue!18} $0.12^{*}$ & $+0.00$ & \cellcolor{blue!11} $0.07^{*}$ & $+0.00$ & \cellcolor{blue!15} $0.10^{*}$ & $+0.00$ & \cellcolor{blue!13} $0.09^{*}$ \\
 &  & $k=2$ & $+0.00$ & \cellcolor{blue!25} $0.17^{*}$ & \cellcolor{red!2} $-0.02$ & \cellcolor{blue!26} $0.18^{*}$ & $+0.00$ & \cellcolor{blue!21} $0.14^{*}$ & \cellcolor{green!1} $+0.01$ & \cellcolor{blue!24} $0.16^{*}$ & \cellcolor{green!1} $+0.01$ & \cellcolor{blue!23} $0.16^{*}$ \\
 &  & $k=3$ & \cellcolor{green!3} $+0.02^{*}$ & \cellcolor{blue!21} $0.14^{*}$ & $+0.00$ & \cellcolor{blue!24} $0.16^{*}$ & \cellcolor{green!1} $+0.01$ & \cellcolor{blue!19} $0.13^{*}$ & \cellcolor{green!2} $+0.02^{*}$ & \cellcolor{blue!22} $0.15^{*}$ & \cellcolor{green!2} $+0.02$ & \cellcolor{blue!20} $0.14^{*}$ \\
\hhline{~------------}
 & \multirow{3}{*}{256} & $k=1$ & \cellcolor{green!1} $+0.01$ & \cellcolor{blue!17} $0.11^{*}$ & \cellcolor{red!4} $-0.03^{*}$ & \cellcolor{blue!17} $0.11^{*}$ & \cellcolor{red!3} $-0.02^{*}$ & \cellcolor{blue!8} $0.06^{*}$ & \cellcolor{red!2} $-0.02$ & \cellcolor{blue!15} $0.10^{*}$ & \cellcolor{red!1} $-0.01$ & \cellcolor{blue!12} $0.08^{*}$ \\
 &  & $k=2$ & $+0.00$ & \cellcolor{blue!21} $0.14^{*}$ & \cellcolor{red!5} $-0.04^{*}$ & \cellcolor{blue!22} $0.15^{*}$ & $+0.00$ & \cellcolor{blue!17} $0.11^{*}$ & \cellcolor{red!2} $-0.02$ & \cellcolor{blue!20} $0.14^{*}$ & $-0.01$ & \cellcolor{blue!19} $0.13^{*}$ \\
 &  & $k=3$ & $-0.01$ & \cellcolor{blue!13} $0.09^{*}$ & \cellcolor{red!3} $-0.02^{*}$ & \cellcolor{blue!15} $0.11^{*}$ & \cellcolor{red!1} $-0.01$ & \cellcolor{blue!11} $0.08^{*}$ & \cellcolor{red!1} $-0.01$ & \cellcolor{blue!14} $0.10^{*}$ & \cellcolor{red!1} $-0.01$ & \cellcolor{blue!12} $0.08^{*}$ \\
\midrule
\multirow{9}{*}{\textbf{ROUGE-2}} & \multirow{3}{*}{4} & $k=1$ & \cellcolor{red!3} $-0.02^{*}$ & \cellcolor{blue!31} $0.21^{*}$ & \cellcolor{red!4} $-0.03^{*}$ & \cellcolor{blue!30} $0.20^{*}$ & \cellcolor{red!4} $-0.03^{*}$ & \cellcolor{blue!29} $0.20^{*}$ & \cellcolor{red!4} $-0.03^{*}$ & \cellcolor{blue!31} $0.21^{*}$ & \cellcolor{red!4} $-0.03^{*}$ & \cellcolor{blue!30} $0.21^{*}$ \\
 &  & $k=2$ & $+0.00$ & \cellcolor{blue!36} $0.25^{*}$ & \cellcolor{red!1} $-0.01$ & \cellcolor{blue!35} $0.24^{*}$ & \cellcolor{red!2} $-0.02$ & \cellcolor{blue!33} $0.23^{*}$ & \cellcolor{red!1} $-0.01$ & \cellcolor{blue!35} $0.24^{*}$ & \cellcolor{red!1} $-0.01$ & \cellcolor{blue!35} $0.24^{*}$ \\
 &  & $k=3$ & $-0.01$ & \cellcolor{blue!33} $0.23^{*}$ & \cellcolor{red!3} $-0.02^{*}$ & \cellcolor{blue!34} $0.23^{*}$ & \cellcolor{red!3} $-0.02^{*}$ & \cellcolor{blue!32} $0.22^{*}$ & $+0.00$ & \cellcolor{blue!34} $0.23^{*}$ & \cellcolor{red!1} $-0.01$ & \cellcolor{blue!33} $0.22^{*}$ \\
\hhline{~------------}
 & \multirow{3}{*}{32} & $k=1$ & $+0.00$ & \cellcolor{blue!25} $0.17^{*}$ & \cellcolor{red!2} $-0.01$ & \cellcolor{blue!26} $0.18^{*}$ & $+0.00$ & \cellcolor{blue!21} $0.14^{*}$ & $+0.00$ & \cellcolor{blue!25} $0.17^{*}$ & $+0.00$ & \cellcolor{blue!24} $0.16^{*}$ \\
 &  & $k=2$ & $+0.00$ & \cellcolor{blue!30} $0.21^{*}$ & \cellcolor{red!4} $-0.03^{*}$ & \cellcolor{blue!31} $0.21^{*}$ & \cellcolor{red!3} $-0.02^{*}$ & \cellcolor{blue!26} $0.18^{*}$ & \cellcolor{red!1} $-0.01$ & \cellcolor{blue!31} $0.21^{*}$ & \cellcolor{red!2} $-0.02$ & \cellcolor{blue!29} $0.20^{*}$ \\
 &  & $k=3$ & \cellcolor{red!2} $-0.02$ & \cellcolor{blue!35} $0.24^{*}$ & \cellcolor{red!5} $-0.03^{*}$ & \cellcolor{blue!38} $0.26^{*}$ & \cellcolor{red!3} $-0.02^{*}$ & \cellcolor{blue!33} $0.22^{*}$ & \cellcolor{red!1} $-0.01$ & \cellcolor{blue!38} $0.25^{*}$ & \cellcolor{red!3} $-0.02^{*}$ & \cellcolor{blue!35} $0.24^{*}$ \\
\hhline{~------------}
 & \multirow{3}{*}{256} & $k=1$ & \cellcolor{green!2} $+0.01$ & \cellcolor{blue!25} $0.17^{*}$ & $-0.01$ & \cellcolor{blue!26} $0.18^{*}$ & \cellcolor{red!1} $-0.01$ & \cellcolor{blue!19} $0.13^{*}$ & \cellcolor{green!1} $+0.01$ & \cellcolor{blue!26} $0.17^{*}$ & \cellcolor{green!1} $+0.01$ & \cellcolor{blue!23} $0.16^{*}$ \\
 &  & $k=2$ & \cellcolor{green!1} $+0.01$ & \cellcolor{blue!27} $0.18^{*}$ & \cellcolor{red!1} $-0.01$ & \cellcolor{blue!28} $0.19^{*}$ & $+0.00$ & \cellcolor{blue!21} $0.14^{*}$ & $+0.00$ & \cellcolor{blue!27} $0.19^{*}$ & $+0.00$ & \cellcolor{blue!25} $0.17^{*}$ \\
 &  & $k=3$ & \cellcolor{green!2} $+0.02$ & \cellcolor{blue!29} $0.19^{*}$ & \cellcolor{red!2} $-0.02$ & \cellcolor{blue!32} $0.21^{*}$ & \cellcolor{red!1} $-0.01$ & \cellcolor{blue!25} $0.17^{*}$ & \cellcolor{red!1} $-0.01$ & \cellcolor{blue!31} $0.21^{*}$ & $+0.00$ & \cellcolor{blue!28} $0.19^{*}$ \\
\midrule
\multirow{9}{*}{\textbf{\texttt{ROUGE}}} & \multirow{3}{*}{4} & $k=1$ & \cellcolor{red!2} $-0.02^{*}$ & \cellcolor{blue!30} $0.20^{*}$ & \cellcolor{red!5} $-0.03^{*}$ & \cellcolor{blue!30} $0.20^{*}$ & \cellcolor{red!4} $-0.03^{*}$ & \cellcolor{blue!24} $0.16^{*}$ & \cellcolor{red!4} $-0.03^{*}$ & \cellcolor{blue!29} $0.20^{*}$ & \cellcolor{red!5} $-0.04^{*}$ & \cellcolor{blue!28} $0.19^{*}$ \\
 &  & $k=2$ & \cellcolor{red!3} $-0.02^{*}$ & \cellcolor{blue!36} $0.24^{*}$ & \cellcolor{red!6} $-0.04^{*}$ & \cellcolor{blue!36} $0.24^{*}$ & \cellcolor{red!2} $-0.02$ & \cellcolor{blue!33} $0.23^{*}$ & \cellcolor{red!4} $-0.03^{*}$ & \cellcolor{blue!36} $0.24^{*}$ & \cellcolor{red!5} $-0.04^{*}$ & \cellcolor{blue!35} $0.24^{*}$ \\
 &  & $k=3$ & $+0.00$ & \cellcolor{blue!31} $0.21^{*}$ & \cellcolor{red!4} $-0.03^{*}$ & \cellcolor{blue!31} $0.21^{*}$ & $+0.00$ & \cellcolor{blue!30} $0.20^{*}$ & \cellcolor{red!4} $-0.03^{*}$ & \cellcolor{blue!32} $0.21^{*}$ & \cellcolor{red!3} $-0.02^{*}$ & \cellcolor{blue!30} $0.21^{*}$ \\
\hhline{~------------}
 & \multirow{3}{*}{32} & $k=1$ & $+0.00$ & \cellcolor{blue!25} $0.17^{*}$ & $+0.00$ & \cellcolor{blue!25} $0.17^{*}$ & $+0.00$ & \cellcolor{blue!21} $0.14^{*}$ & $+0.00$ & \cellcolor{blue!26} $0.18^{*}$ & $+0.00$ & \cellcolor{blue!23} $0.16^{*}$ \\
 &  & $k=2$ & \cellcolor{red!1} $-0.01$ & \cellcolor{blue!32} $0.22^{*}$ & \cellcolor{red!5} $-0.03^{*}$ & \cellcolor{blue!33} $0.22^{*}$ & \cellcolor{red!1} $-0.01$ & \cellcolor{blue!26} $0.18^{*}$ & \cellcolor{red!3} $-0.02^{*}$ & \cellcolor{blue!32} $0.21^{*}$ & \cellcolor{red!4} $-0.03^{*}$ & \cellcolor{blue!30} $0.20^{*}$ \\
 &  & $k=3$ & \cellcolor{red!1} $-0.01$ & \cellcolor{blue!29} $0.20^{*}$ & \cellcolor{red!4} $-0.03^{*}$ & \cellcolor{blue!31} $0.21^{*}$ & \cellcolor{red!3} $-0.02^{*}$ & \cellcolor{blue!26} $0.17^{*}$ & \cellcolor{red!2} $-0.02$ & \cellcolor{blue!31} $0.21^{*}$ & \cellcolor{red!3} $-0.03^{*}$ & \cellcolor{blue!28} $0.19^{*}$ \\
\hhline{~------------}
 & \multirow{3}{*}{256} & $k=1$ & $+0.00$ & \cellcolor{blue!27} $0.19^{*}$ & \cellcolor{red!5} $-0.04^{*}$ & \cellcolor{blue!27} $0.18^{*}$ & $-0.01$ & \cellcolor{blue!20} $0.14^{*}$ & \cellcolor{red!2} $-0.02$ & \cellcolor{blue!27} $0.18^{*}$ & \cellcolor{red!3} $-0.02^{*}$ & \cellcolor{blue!24} $0.17^{*}$ \\
 &  & $k=2$ & $+0.00$ & \cellcolor{blue!28} $0.19^{*}$ & \cellcolor{red!5} $-0.03^{*}$ & \cellcolor{blue!29} $0.20^{*}$ & $+0.00$ & \cellcolor{blue!23} $0.16^{*}$ & $-0.01$ & \cellcolor{blue!28} $0.19^{*}$ & \cellcolor{red!2} $-0.02$ & \cellcolor{blue!26} $0.18^{*}$ \\
 &  & $k=3$ & \cellcolor{green!1} $+0.01$ & \cellcolor{blue!27} $0.18^{*}$ & \cellcolor{red!3} $-0.02^{*}$ & \cellcolor{blue!29} $0.20^{*}$ & \cellcolor{green!1} $+0.01$ & \cellcolor{blue!23} $0.16^{*}$ & $+0.00$ & \cellcolor{blue!28} $0.19^{*}$ & $+0.00$ & \cellcolor{blue!25} $0.17^{*}$ \\
\bottomrule
\end{NiceTabular}
}
\caption{Spearman correlation ($\rho$) between matrix reconstruction errors and evaluation score differences across varying reference pool sizes ($|\mathcal{Y}|$) on XSum. \textit{Dir.} (Directional Change) measures score improvements, highlighted as \colorbox{green!15}{positive} and \colorbox{red!15}{negative} impacts. \textit{Mag.} (Magnitude Sensitivity) measures absolute score variance, with \colorbox{blue!15}{blue} highlighting metric sensitivity to matrix modifications. Statistical significance ($p < 0.05$) is denoted by an asterisk ($^*$).}
\label{tab:corr_mat_analysis_complete_xsum}
\end{table*}

\begin{table*}[ht!]
\centering
\resizebox{\textwidth}{!}{
\begin{tabular}{llllllll}
\toprule
\textbf{Util. Function} & \textbf{$|\mathcal{Y}|$} & \textbf{Top-$k$} & \textbf{BERTScore} & \textbf{ROUGE-1} & \textbf{ROUGE-2} & \textbf{ROUGE-L} & \textbf{$\bar{Z}_{\text{other}}$} \\
\midrule
\multirow[t]{9}{*}{BERTScore} & \multirow[t]{3}{*}{4} & $k=1$ & \cellcolor{red!15} 2333 / 4992 / 4008 & \cellcolor{red!15} 2724 / 5314 / 3295 & \cellcolor{red!15} 2583 / 5729 / 3021 & \cellcolor{red!15} 2785 / 5324 / 3224 & \cellcolor{red!15} 2823 / 5169 / 3341 \\
 &  & $k=2$ & \cellcolor{red!15} 1038 / 8666 / 1629 & \cellcolor{red!15} 1164 / 8823 / 1346 & \cellcolor{red!15} 1059 / 9004 / 1270 & \cellcolor{red!15} 1153 / 8830 / 1350 & \cellcolor{red!15} 1167 / 8761 / 1405 \\
 &  & $k=3$ & \cellcolor{red!15} 422 / 10368 / 543 & \cellcolor{red!15} 421 / 10452 / 460 & \cellcolor{red!15} 393 / 10515 / 425 & \cellcolor{red!15} 400 / 10451 / 482 & \cellcolor{red!15} 429 / 10426 / 478 \\
\cline{2-8}
 & \multirow[t]{3}{*}{32} & $k=1$ & \cellcolor{red!15} 2485 / 4693 / 4155 & \cellcolor{red!15} 3035 / 5117 / 3181 & \cellcolor{green!15} 2941 / 5560 / 2832 & \cellcolor{red!15} 3069 / 5162 / 3102 & \cellcolor{red!15} 3123 / 4969 / 3241 \\
 &  & $k=2$ & \cellcolor{red!15} 1278 / 8140 / 1915 & \cellcolor{red!15} 1424 / 8335 / 1574 & \cellcolor{red!15} 1380 / 8518 / 1435 & \cellcolor{red!15} 1452 / 8358 / 1523 & \cellcolor{red!15} 1465 / 8271 / 1597 \\
 &  & $k=3$ & \cellcolor{red!15} 741 / 9638 / 954 & \cellcolor{red!15} 757 / 9784 / 792 & \cellcolor{green!15} 745 / 9868 / 720 & \cellcolor{red!15} 761 / 9795 / 777 & \cellcolor{red!15} 783 / 9742 / 808 \\
\cline{2-8}
 & \multirow[t]{3}{*}{256} & $k=1$ & \cellcolor{red!15} 2335 / 5108 / 3890 & \cellcolor{red!15} 2842 / 5546 / 2945 & \cellcolor{green!15} 2788 / 5976 / 2569 & \cellcolor{green!15} 2873 / 5590 / 2870 & \cellcolor{red!15} 2946 / 5411 / 2976 \\
 &  & $k=2$ & \cellcolor{red!15} 1174 / 8587 / 1572 & \cellcolor{green!15} 1298 / 8772 / 1263 & \cellcolor{green!15} 1255 / 8960 / 1118 & \cellcolor{green!15} 1325 / 8793 / 1215 & \cellcolor{green!15} 1361 / 8720 / 1252 \\
 &  & $k=3$ & \cellcolor{red!15} 702 / 9791 / 840 & \cellcolor{green!15} 717 / 9921 / 695 & \cellcolor{green!15} 683 / 10027 / 623 & \cellcolor{green!15} 714 / 9932 / 687 & \cellcolor{green!15} 738 / 9896 / 699 \\
\cline{1-8} \cline{2-8}
\multirow[t]{9}{*}{ROUGE-1} & \multirow[t]{3}{*}{4} & $k=1$ & \cellcolor{red!15} 3199 / 4293 / 3841 & \cellcolor{red!15} 2803 / 4578 / 3952 & \cellcolor{red!15} 2990 / 5048 / 3295 & \cellcolor{red!15} 3126 / 4559 / 3648 & \cellcolor{red!15} 3193 / 4291 / 3849 \\
 &  & $k=2$ & \cellcolor{red!15} 1653 / 7680 / 2000 & \cellcolor{red!15} 1470 / 7862 / 2001 & \cellcolor{red!15} 1493 / 8128 / 1712 & \cellcolor{red!15} 1576 / 7857 / 1900 & \cellcolor{red!15} 1626 / 7678 / 2029 \\
 &  & $k=3$ & \cellcolor{red!15} 680 / 9875 / 778 & \cellcolor{red!15} 644 / 9979 / 710 & \cellcolor{red!15} 589 / 10094 / 650 & \cellcolor{red!15} 646 / 9975 / 712 & \cellcolor{red!15} 691 / 9873 / 769 \\
\cline{2-8}
 & \multirow[t]{3}{*}{32} & $k=1$ & \cellcolor{red!15} 3546 / 4158 / 3629 & \cellcolor{red!15} 2825 / 4535 / 3973 & \cellcolor{green!15} 3320 / 4993 / 3020 & \cellcolor{red!15} 3385 / 4561 / 3387 & \cellcolor{red!15} 3557 / 4142 / 3634 \\
 &  & $k=2$ & \cellcolor{red!15} 2234 / 6688 / 2411 & \cellcolor{red!15} 1851 / 6902 / 2580 & \cellcolor{red!15} 2027 / 7218 / 2088 & \cellcolor{red!15} 2110 / 6907 / 2316 & \cellcolor{red!15} 2234 / 6662 / 2437 \\
 &  & $k=3$ & \cellcolor{red!15} 1440 / 8446 / 1447 & \cellcolor{red!15} 1180 / 8624 / 1529 & \cellcolor{green!15} 1264 / 8811 / 1258 & \cellcolor{red!15} 1334 / 8599 / 1400 & \cellcolor{red!15} 1424 / 8422 / 1487 \\
\cline{2-8}
 & \multirow[t]{3}{*}{256} & $k=1$ & \cellcolor{red!15} 3334 / 4586 / 3413 & \cellcolor{red!15} 2612 / 4953 / 3768 & \cellcolor{green!15} 3132 / 5378 / 2823 & \cellcolor{red!15} 3112 / 4977 / 3244 & \cellcolor{red!15} 3283 / 4575 / 3475 \\
 &  & $k=2$ & \cellcolor{green!15} 2050 / 7270 / 2013 & \cellcolor{red!15} 1654 / 7513 / 2166 & \cellcolor{green!15} 1881 / 7787 / 1665 & \cellcolor{green!15} 1913 / 7523 / 1897 & \cellcolor{green!15} 2045 / 7255 / 2033 \\
 &  & $k=3$ & \cellcolor{green!15} 1268 / 8878 / 1187 & \cellcolor{red!15} 1060 / 9027 / 1246 & \cellcolor{green!15} 1146 / 9200 / 987 & \cellcolor{green!15} 1208 / 9030 / 1095 & \cellcolor{green!15} 1277 / 8859 / 1197 \\
\cline{1-8} \cline{2-8}
\multirow[t]{9}{*}{ROUGE-2} & \multirow[t]{3}{*}{4} & $k=1$ & \cellcolor{red!15} 2981 / 5035 / 3317 & \cellcolor{red!15} 2746 / 5257 / 3330 & \cellcolor{red!15} 2616 / 5676 / 3041 & \cellcolor{red!15} 2792 / 5299 / 3242 & \cellcolor{red!15} 2918 / 5034 / 3381 \\
 &  & $k=2$ & \cellcolor{red!15} 1856 / 7442 / 2035 & \cellcolor{red!15} 1682 / 7625 / 2026 & \cellcolor{red!15} 1520 / 7882 / 1931 & \cellcolor{red!15} 1704 / 7633 / 1996 & \cellcolor{red!15} 1776 / 7441 / 2116 \\
 &  & $k=3$ & \cellcolor{red!15} 804 / 9650 / 879 & \cellcolor{red!15} 720 / 9725 / 888 & \cellcolor{red!15} 648 / 9854 / 831 & \cellcolor{red!15} 738 / 9730 / 865 & \cellcolor{red!15} 751 / 9648 / 934 \\
\cline{2-8}
 & \multirow[t]{3}{*}{32} & $k=1$ & \cellcolor{red!15} 2884 / 5146 / 3303 & \cellcolor{red!15} 2536 / 5557 / 3240 & \cellcolor{red!15} 2485 / 5997 / 2851 & \cellcolor{red!15} 2720 / 5563 / 3050 & \cellcolor{red!15} 2796 / 5121 / 3416 \\
 &  & $k=2$ & \cellcolor{red!15} 2256 / 6491 / 2586 & \cellcolor{red!15} 1997 / 6766 / 2570 & \cellcolor{red!15} 1917 / 7103 / 2313 & \cellcolor{red!15} 2109 / 6778 / 2446 & \cellcolor{red!15} 2181 / 6453 / 2699 \\
 &  & $k=3$ & \cellcolor{red!15} 1622 / 7880 / 1831 & \cellcolor{red!15} 1407 / 8107 / 1819 & \cellcolor{red!15} 1335 / 8376 / 1622 & \cellcolor{red!15} 1481 / 8104 / 1748 & \cellcolor{red!15} 1568 / 7835 / 1930 \\
\cline{2-8}
 & \multirow[t]{3}{*}{256} & $k=1$ & \cellcolor{red!15} 2513 / 6009 / 2811 & \cellcolor{red!15} 2137 / 6369 / 2827 & \cellcolor{red!15} 2185 / 6740 / 2408 & \cellcolor{red!15} 2354 / 6366 / 2613 & \cellcolor{red!15} 2403 / 5986 / 2944 \\
 &  & $k=2$ & \cellcolor{red!15} 1851 / 7562 / 1920 & \cellcolor{red!15} 1550 / 7842 / 1941 & \cellcolor{red!15} 1548 / 8152 / 1633 & \cellcolor{red!15} 1678 / 7831 / 1824 & \cellcolor{red!15} 1774 / 7529 / 2030 \\
 &  & $k=3$ & \cellcolor{green!15} 1364 / 8625 / 1344 & \cellcolor{red!15} 1142 / 8819 / 1372 & \cellcolor{red!15} 1109 / 9062 / 1162 & \cellcolor{red!15} 1198 / 8828 / 1307 & \cellcolor{red!15} 1313 / 8588 / 1432 \\
\cline{1-8} \cline{2-8}
\multirow[t]{9}{*}{\texttt{ROUGE}} & \multirow[t]{3}{*}{4} & $k=1$ & \cellcolor{red!15} 3076 / 4827 / 3430 & \cellcolor{red!15} 2654 / 5083 / 3596 & \cellcolor{red!15} 2662 / 5539 / 3132 & \cellcolor{red!15} 2786 / 5106 / 3441 & \cellcolor{red!15} 2873 / 4827 / 3633 \\
 &  & $k=2$ & \cellcolor{red!15} 1717 / 7562 / 2054 & \cellcolor{red!15} 1486 / 7728 / 2119 & \cellcolor{red!15} 1482 / 8002 / 1849 & \cellcolor{red!15} 1535 / 7739 / 2059 & \cellcolor{red!15} 1579 / 7559 / 2195 \\
 &  & $k=3$ & \cellcolor{red!15} 753 / 9740 / 840 & \cellcolor{red!15} 658 / 9829 / 846 & \cellcolor{red!15} 657 / 9950 / 726 & \cellcolor{red!15} 666 / 9830 / 837 & \cellcolor{red!15} 705 / 9736 / 892 \\
\cline{2-8}
 & \multirow[t]{3}{*}{32} & $k=1$ & \cellcolor{red!15} 3078 / 4761 / 3494 & \cellcolor{red!15} 2561 / 5151 / 3621 & \cellcolor{red!15} 2720 / 5622 / 2991 & \cellcolor{red!15} 2780 / 5188 / 3365 & \cellcolor{red!15} 2892 / 4735 / 3706 \\
 &  & $k=2$ & \cellcolor{red!15} 2210 / 6525 / 2598 & \cellcolor{red!15} 1887 / 6810 / 2636 & \cellcolor{red!15} 1986 / 7156 / 2191 & \cellcolor{red!15} 2046 / 6812 / 2475 & \cellcolor{red!15} 2114 / 6496 / 2723 \\
 &  & $k=3$ & \cellcolor{red!15} 1473 / 8204 / 1656 & \cellcolor{red!15} 1210 / 8396 / 1727 & \cellcolor{red!15} 1255 / 8660 / 1418 & \cellcolor{red!15} 1352 / 8406 / 1575 & \cellcolor{red!15} 1388 / 8171 / 1774 \\
\cline{2-8}
 & \multirow[t]{3}{*}{256} & $k=1$ & \cellcolor{red!15} 2772 / 5387 / 3174 & \cellcolor{red!15} 2237 / 5768 / 3328 & \cellcolor{red!15} 2476 / 6226 / 2631 & \cellcolor{red!15} 2494 / 5800 / 3039 & \cellcolor{red!15} 2601 / 5367 / 3365 \\
 &  & $k=2$ & \cellcolor{red!15} 1930 / 7391 / 2012 & \cellcolor{red!15} 1591 / 7665 / 2077 & \cellcolor{green!15} 1719 / 7962 / 1652 & \cellcolor{red!15} 1772 / 7690 / 1871 & \cellcolor{red!15} 1881 / 7365 / 2087 \\
 &  & $k=3$ & \cellcolor{red!15} 1289 / 8731 / 1313 & \cellcolor{red!15} 1112 / 8901 / 1320 & \cellcolor{green!15} 1153 / 9108 / 1072 & \cellcolor{red!15} 1188 / 8908 / 1237 & \cellcolor{red!15} 1284 / 8695 / 1354 \\
\cline{1-8} \cline{2-8}
\bottomrule
\end{tabular}

}
\caption{Sentence-level comparison between SVD-MBR and MBR on the XSum. Cells are formatted as \textbf{W / T / L}, indicating the number of sentences where SVD-MBR achieved a higher score (Win), an identical score (Tie), or a lower score (Loss). The rightmost column, $\bar{Z}_{\text{other}}$, aggregates the net win rate across all off-target evaluation metrics. \colorbox{green!15}{Positive} and \colorbox{red!15}{Negative} win rate for SVD-MBR is indicated by the color of the cells.}
\label{tab:svd_wtl_full_xsum}
\end{table*}

\begin{table*}[ht!]
\centering
\small
\setlength{\tabcolsep}{3.5pt}
\begin{tabular}{@{} l c c @{\hspace{0.2\textwidth}} l c c @{}}
\toprule
\multicolumn{3}{@{}p{0.4\textwidth}@{}}{\textbf{Case 1: BERTScore as Util. Function}} & \multicolumn{3}{@{}p{0.4\textwidth}@{}}{\textbf{Case 2: ROUGE-1 as Util. Function}} \\
\midrule
\multicolumn{3}{@{}p{0.4\textwidth}@{}}{\textbf{Source:} Charlie Walker's deflected free-kick put the hosts ahead, before Giorgio Rasulo's shot doubled the lead.Grimsby then took a 3-2 lead with goals from Padraig Amond, Omar Bogle and Nathan Arnold, before Rhys Browne fired in an equaliser for Aldershot.But Jon Nolan's cross went over Shots keeper Phil Smith and into the net to earn the Mariners all three points.The result keeps Paul Hurst's Grimsby side third in the table, seven points above sixth-placed Braintree.Aldershot Town boss Barry Smith told BBC Surrey:Media playback is not supported on this device"I am just disappointed with the goals we conceded, but I thought the first-half the players showed that commitment and desire. The passing football we like to do and like to play made us score two great goals."In the second half we still played some good stuff and the boys still worked hard, but lapses of concentration in our box cost us."The players know who they are picking up in the box. Phil Smith has held his hand up for two goals and feels he could've done better, which is credit to him, but it's lapses of concentration."Grimsby Town manager Paul Hurst told BBC Radio Humberside:Media playback is not supported on this device"First-half going in at 2-0 was very concerning because I asked them some questions and they were quiet."To drag themselves back into the game and get in front and then have the disappointment of going 3-3, but clearly there was some luck in the end with the winning goal from Jon Nolan."We've certainly earned some luck, we haven't had it recently and I'm hoping that's a change in that fortune. What a fantastic three points and reward for the fans that were not happy at half-time as we all weren't."} & \multicolumn{3}{@{}p{0.4\textwidth}@{}}{\textbf{Source:} The Loyalist Community Council (LCC) has created the flag to commemorate the centenary of the World War One battle.It says the flag is supported by three paramilitary groups, the Ulster Defence Association (UDA), the Ulster Volunteer Force (UVF) and the Red Hand Commando."We hope this is the only flag that is flown along arterial routes alongside the Union flag and the Ulster flag," Winston Irvine of the LCC said."I think this flag will hopefully reduce the amount of perceived paramilitary flags."Mr Irvine said he did not expect nationalists "to be screaming in support of the flag from the rooftops".But he added: "I do hope people see this is an attempt to lessen the impact of flags and to command a wider support for the respectful and dignified commemoration to mark the anniversary of the Somme."The Somme means something to everyone, regardless of your community background, both in Northern Ireland and in the Republic of Ireland, given the scale and severity of that battle on the people of this island."Members of the 36th Ulster Division were among 100,000 Allied soldiers who fought at the Somme in 1916.A private benefactor has paid for the purchase of the flags.The group has also drawn up a set of protocols for the flying of flags.Mr Irvine said people were "fed up with torn and tattered flags flying from lamp posts and buildings".The protocols give guidance on the period of time flags should be flown, suggesting a three-month spell starting in June.Assistant Chief Constable Stephen Martin of the Police Service of Northern Ireland said the flying of flags in public places "provokes a range of strong responses and very different viewpoints".He added: "We hope [the LCC's new flag and its protocols] can be a positive development in improving the overall context in which flags are flown."} \\
\addlinespace
\multicolumn{3}{@{}p{0.4\textwidth}@{}}{\textbf{Reference:} Grimsby fought back from two goals down to beat Aldershot and boost their National League play-off hopes.} & \multicolumn{3}{@{}p{0.4\textwidth}@{}}{\textbf{Reference:} A new flag to mark the Battle of the Somme has been unfurled in Belfast.} \\
\midrule
\multicolumn{3}{@{}p{0.4\textwidth}@{}}{\textbf{MBR:}} & \multicolumn{3}{@{}p{0.4\textwidth}@{}}{\textbf{MBR:}} \\
\multicolumn{3}{@{}p{0.4\textwidth}@{}}{Grimsby secured their third National League win of the season with a late winner at Aldershot.} & \multicolumn{3}{@{}p{0.4\textwidth}@{}}{A Loyalist community group in Belfast has unveiled a new flag to mark the Battle of the Somme.} \\
\addlinespace
\multicolumn{3}{@{}p{0.4\textwidth}@{}}{\textbf{SVD-MBR:}} & \multicolumn{3}{@{}p{0.4\textwidth}@{}}{\textbf{SVD-MBR:}} \\
\multicolumn{3}{@{}p{0.4\textwidth}@{}}{Grimsby came from two goals down to beat Aldershot and increase their National League play-off hopes.} & \multicolumn{3}{@{}p{0.48\textwidth}@{}}{A Loyalist community group has launched a "Somme flag" to "reduce the number of perceived paramilitary flags" in Northern Ireland.} \\
\midrule
\textbf{Metric} & \textbf{MBR} & \textbf{SVD-MBR} & \textbf{Metric} & \textbf{MBR} & \textbf{SVD-MBR} \\
\midrule
\textbf{BERTScore} (Util) & 78.1 & \textbf{93.8} (\textcolor{green}{$\uparrow$}) & \textbf{BERTScore} & \textbf{82.4} & 63.0 (\textcolor{red}{$\downarrow$}) \\
\textbf{ROUGE-1} & 29.4 & \textbf{85.7} (\textcolor{green}{$\uparrow$}) & \textbf{ROUGE-1} (Util) & \textbf{78.8} & 45.7 (\textcolor{red}{$\downarrow$}) \\
\textbf{ROUGE-2} & 6.2 & \textbf{72.7} (\textcolor{green}{$\uparrow$}) & \textbf{ROUGE-2} & \textbf{64.5} & 6.1 (\textcolor{red}{$\downarrow$}) \\
\textbf{ROUGE-L} & 23.5 & \textbf{85.7} (\textcolor{green}{$\uparrow$}) & \textbf{ROUGE-L} & \textbf{60.6} & 34.3 (\textcolor{red}{$\downarrow$}) \\
\bottomrule
\end{tabular}
\caption{Sentence comparison of MBR and SVD-MBR ($k=1$) using $|\mathcal{Y}|=256$. \textbf{Case 1} demonstrates the comparison using BERTScore as the utility function and \textbf{Case 2} demonstrates the comparison using ROUGE-1 as the utility function.}
\label{tab:qualitative_side_by_side_unified_xsum}
\end{table*}

\subsection{XSum}

To evaluate the generalizability of our findings, we extend our analysis to a different text generation domain: abstractive summarization. For this experiment, we utilize the XSum dataset and adapt our experimental setup accordingly, employing a BART-Large model fine-tuned on XSum for hypothesis generation. To capture both dense semantic similarity and surface-level lexical overlap, we evaluate performance using two metric families: BERTScore and ROUGE~\cite{lin-2004-rouge} (specifically ROUGE-1, ROUGE-2, and ROUGE-L). As MBR utility functions, we test BERTScore, ROUGE-1, ROUGE-2, and an unweighted macro-average of the evaluated ROUGE variants, which we denote as \texttt{ROUGE}.

Table~\ref{tab:mbr_overfitting_xsum} presents the overfitting results for the XSum dataset. \textbf{The observed trends closely mirror our translation findings: candidate hypotheses consistently overfit to the specifically designated utility metric}. For instance, in standard MBR~($|\mathcal{Y}|=256$), optimizing for ROUGE-1 disproportionately inflates the ROUGE-1 evaluation score~($+2.965$), while off-target metrics like BERTScore experience only marginal collateral gains~($+0.662$). Conversely, optimizing for BERTScore actively suppresses or degrades specific lexical overlaps, as evidenced by the degradation in ROUGE-2~($-0.112$). Interestingly, when the aggregated \texttt{ROUGE} metric is employed as the utility function, this narrow metric exploitation is noticeably mitigated, yielding a more balanced and harmonious improvement across all individual ROUGE variants.

Table~\ref{tab:svd_mbr_results_xsum} compares the performance of SVD-MBR against standard MBR on the XSum dataset. Consistent with the results from the translation tasks, employing BERTScore as the utility function yields significant performance increases, particularly at $|\mathcal{Y}|=256$ and $k=2$. Interestingly, SVD-MBR also achieves a significant increase in aggregated off-target performance~($\bar{Z}_{\text{other}}$) when utilizing ROUGE-1 at $|\mathcal{Y}|=256$ and $k=3$. This further demonstrates that our method can mitigate overfitting even when applied to surface-level metrics, although its efficacy remains more unstable compared to neural metrics.

The Spearman correlations between the performance deltas and both the inherent hypothesis oracle variance and the matrix reconstruction error are presented in Tables~\ref{tab:corr_variance_analysis_complete_xsum} and \ref{tab:corr_mat_analysis_complete_xsum}, respectively. Surprisingly, the variance correlation analysis reveals that for both BERTScore and the ROUGE variants, nearly all configurations achieve a significant positive correlation with the $\bar{Z}_{\text{other}}$ score, regardless of the overall intrinsic variance of each metric. This indicates that \textbf{SVD-MBR successfully improves outputs in high-variance scenarios while maintaining parity with standard MBR when variance is low}. Conversely, the correlation between matrix reconstruction error and evaluation score differences shows a negative trajectory for $\bar{Z}_{\text{other}}$. This suggests that although higher inherent variance leads to better denoising performance by SVD-MBR, this improvement does not directly mathematically translate from larger matrix reconstruction errors in this specific domain.

Furthermore, our sentence-level Win/Tie/Loss (WTL) comparison in Table~\ref{tab:svd_wtl_full_xsum} mirrors the aggregate performance scores observed in Table~\ref{tab:svd_mbr_results_xsum}. The data demonstrates that our method is highly effective when utilizing BERTScore and ROUGE-1 as utility functions, but yields less consistent improvements across the remaining metrics. Finally, the qualitative examples in Table~\ref{tab:qualitative_side_by_side_unified_xsum}, selected using the BERTScore and ROUGE-1 configurations, highlight that sentence-level generations can differ substantially. For example, in the success case utilizing BERTScore, the performance gains across all off-target metrics are exceptionally high. Conversely, the failure case utilizing ROUGE-1 illustrates a severe degradation, marked by significant decreases in both ROUGE-2 and ROUGE-L scores.

\end{document}